\documentclass[runningheads]{llncs}
\usepackage{graphicx}

\usepackage{comment_eccv}
\usepackage{amsmath,amssymb} 

\usepackage[accsupp]{axessibility}  

\usepackage{epsfig}
\usepackage[normalem]{ulem}
\usepackage{enumitem}
\usepackage{graphbox}
\usepackage{multirow} 
\usepackage{multicol} 
\usepackage{comment}
\usepackage{makecell} 
\usepackage{array}
\usepackage[font=small]{caption}
\usepackage{subcaption}
\usepackage{dsfont} 
\usepackage[dvipsnames]{xcolor}
\usepackage{colortbl}
\usepackage{diagbox} 
\usepackage{scrextend} 
\usepackage{balance} 
\usepackage{url}
\usepackage{wrapfig}
\usepackage[bottom]{footmisc}
\usepackage{fancyhdr}

\def\eg{\emph{e.g.}} 
\def\ie{\emph{i.e.}}

\def\etal{\emph{et al.}}

\newcommand{\ours}{SenFuNet}

\newcommand{\boldparagraph}[1]{\vspace{0.1em}\noindent{\bf #1}}

\newcommand{\1}[1]{\mathds{1}_{\{#1\}}}
\newcommand{\bplus}{$\boldsymbol{+}$}    

\DeclareMathOperator{\sign}{sign}

\usepackage{scrextend} 
\definecolor{gray}{rgb}{0.95,0.95,0.95}
\definecolor{mycol}{rgb}{0.90,0.95,1.0}

\fancypagestyle{firstpage}{
  \fancyhf{}
  \fancyfoot[L]{\footnotesize Corresponding author: \texttt{esandstroem@ee.ethz.ch}}%
}

\newif\ifeccv
\eccvtrue

\begin{document}
\pagestyle{headings}
\mainmatter
\def\ECCVSubNumber{1860}  

\title{Learning Online Multi-Sensor Depth Fusion} 

%
\author{Erik Sandström$^{1}$
\and
Martin R. Oswald$^{1,2}$
\and
Suryansh Kumar$^{1}$
\and
Silvan Weder$^{1}$
\and
Fisher Yu$^{1}$
\and
Cristian Sminchisescu$^{3,5}$
\and
Luc Van Gool$^{1,4}$
}

\authorrunning{E. Sandström \etal}
\institute{$^{1}$ETH Zürich, $^{2}$University of Amsterdam, $^{3}$Lund University, \\$^{4}$KU Leuven, $^{5}$Google Research}
\maketitle
\thispagestyle{firstpage}
\begin{abstract}
    Many hand-held or mixed reality devices are used with a single sensor for 3D reconstruction, although they often comprise multiple sensors. 
    Multi-sensor depth fusion is able to substantially improve the robustness and accuracy of 3D reconstruction methods, but existing techniques are not robust enough to handle sensors which operate with diverse value ranges as well as noise and outlier statistics. 
    To this end, we introduce SenFuNet, a depth fusion approach that learns sensor-specific noise and outlier statistics and combines the data streams of depth frames from different sensors in an online fashion. 
    Our method fuses multi-sensor depth streams regardless of time synchronization and calibration and generalizes well with little training data.
    We conduct experiments with various sensor combinations on the real-world CoRBS and Scene3D datasets, as well as the Replica dataset.
    Experiments demonstrate that our fusion strategy outperforms traditional and recent online depth fusion
    approaches. In addition, the combination of multiple sensors yields more robust outlier handling and more precise surface reconstruction than the use of a single sensor. The source code and data are available at \url{https://github.com/tfy14esa/SenFuNet}.
\end{abstract}
\section{Introduction}

\ifeccv
\begin{figure}[t]
\centering
\scriptsize
\begin{tabular}{l}
 \hspace{0.58cm} ToF \hspace{0.88cm} MVS\cite{schonberger2016pixelwise} \hspace{0.5cm} TSDF Fusion\cite{curless1996volumetric}\hspace{0.02cm}RoutedFusion\cite{Weder2020RoutedFusionLR} \hspace{0.51cm} Ours\hspace{0.82cm} Ground Truth\\
\includegraphics[align=c, width=1.0\linewidth]{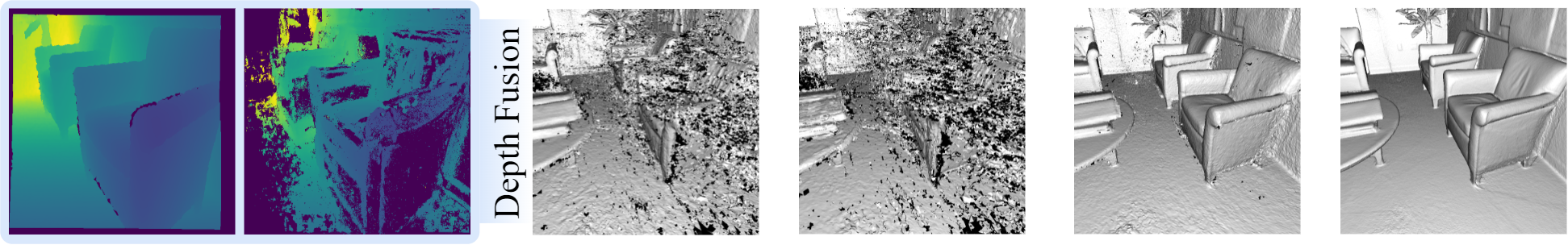} \\
\end{tabular}
\caption{\textbf{Online multi-sensor depth map fusion.}
We fuse depth streams from different sensors with a 3D late fusion approach, here, a time-of-flight (ToF) camera and multi-view stereo (MVS) depth. Compared to competitive depth fusion methods like TSDF Fusion or RoutedFusion, our learning-based approach handles multiple depth sensors and significantly reduces the amount of outliers without loss of completeness.}
\label{fig:teaser}
\end{figure}
\else
\begin{figure}[t]
\centering
\footnotesize
\begin{tabular}{l}
 \hspace{0.99cm} ToF \hspace{1.5cm} TSDF Fusion~\cite{curless1996volumetric} \hspace{0.2cm} RoutedFusion~\cite{Weder2020RoutedFusionLR} \\
\includegraphics[align=c, width=0.95\linewidth]{figures/scene3d/lounge/teaser_updated.png} \\
\hspace{0.55cm} MVS~\cite{schonberger2016pixelwise}
\hspace{1.21cm} SenFuNet (Ours) \hspace{0.64cm} Ground Truth\\
\end{tabular}
\caption{\textbf{Online multi-sensor depth map fusion.}
We fuse depth streams from a time-of-flight (ToF) camera and multi-view stereo (MVS) depth. Compared to competitive depth fusion methods such as TSDF Fusion and RoutedFusion, our learning-based approach can handle multiple depth sensors and significantly reduces the amount of outliers without loss of completeness.}
\label{fig:teaser}
\end{figure}
\fi

Real-time online 3D reconstruction has become increasingly important with the rise of applications like mixed reality, autonomous driving, robotics, or live 3D content creation via scanning. 
The majority of 3D reconstruction hardware platforms like phones, tablets, or mixed reality headsets contain a 
multitude of sensors, but few algorithms leverage them jointly to increase their accuracy, robustness, and reliability.
For instance, the HoloLens2 has four tracking cameras and a depth camera for mapping. Neither is its depth camera used for tracking, nor are the tracking cameras used for mapping.
Fusing the data from multiple sensors is challenging since different sensors typically operate in different domains, have diverse value ranges as well as noise and outlier statistics. 
This diversity is, however, what motivates sensor fusion. 
For example, RGB stereo cameras typically have a larger field of view and higher resolution than time-of-flight (ToF) cameras, but typically struggle on homogeneously textured surfaces. ToF cameras perform well regardless of texture, but show performance drops around edges.
Fig.~\ref{fig:teaser} shows the online fusion result of a ToF camera and a multi-view stereo (MVS) depth sensor.
Both traditional and recent learning-based techniques such as TSDF Fusion~\cite{curless1996volumetric} and RoutedFusion~\cite{Weder2020RoutedFusionLR} respectively, reveal a high degree of noise and outliers when fusing multi-sensor depth.
Although other recent works tackle depth map fusion~\cite{weder2021neuralfusion,yan2021continual,sucar2021iMap,huang2021di} with learning techniques, there is yet no work that considers multiple sensors for online dense reconstruction. 

In this paper, we present an
approach for sensor fusion (\ours{}) which jointly learns (1) the iterative online fusion of depth maps from a single sensor and (2) the effective fusion of depth data from multiple different sensors.
During training, our method learns relevant sensor properties which impact the reconstruction accuracy to locally emphasize the better sensor for particular input and geometry configurations (see Fig.~\ref{fig:teaser}).
We demonstrate with multiple sensor combinations that the learned sensor weighting is generic and can also be used as an expert system, \eg ~for fusing the results of different stereo methods. In this case, our method predicts which algorithm performs better on which part of the scene.
Since our approach handles time asynchronous sensor inputs, it is also applicable to collaborative multi-agent reconstruction. Our contributions are:

\begin{itemize}[itemsep=0pt,topsep=2pt,leftmargin=11pt]
    \item Our approach learns location-dependent fusion weights for the individual sensor contributions according to learned sensor statistics.
    For various sensor combinations our method extracts multi-sensor results that are consistently better than those obtained from the individual sensors.
    \item Our pipeline is end-to-end trained in an online manner, is light-weight, real-time capable and generalizes well even for small amounts of training data.
    \item In contrast to early fusion approaches which directly fuse depth values and thus generally assume a time synchronized sensor setup, our approach is flexible and can fuse the recovered scene reconstruction 
    from asynchronous sensors. Our system is therefore more robust (compared to early fusion) to sensor differences such as sampling frequency, pose and resolution differences. 
\end{itemize}

\section{Related Work}
In this section, we discuss dense online 3D reconstruction, multi-sensor depth fusion and multi-sensor dense 3D reconstruction.

\boldparagraph{Dense Online 3D Scene Reconstruction.} 
The foundation for many volumetric online 3D reconstruction methods via truncated signed distance functions (TSDF) was laid by Curless and Levoy~\cite{curless1996volumetric}. 
Popular extensions of this seminal work are KinectFusion~\cite{izadi2011kinectfusion} and scalable generalizations with voxel hashing~\cite{voxel_hashing,Kahler2015infiniTAM,Oleynikova2017voxblox}, octrees~\cite{6751517}, or increased pose robustness via sparse image features~\cite{7900211}.
Further extensions include tracking for Simultaneous Localization and Mapping (SLAM)~\cite{newcombe2011dtam,schops2019bad,sucar2021iMap,zhu2022nice} which potentially also handle loop closures, \eg~BundleFusion~\cite{dai2017bundlefusion}.
To account for greater depth noise, RoutedFusion~\cite{Weder2020RoutedFusionLR} learns online updates of the volumetric grid.
NeuralFusion~\cite{weder2021neuralfusion} extends this idea by additionally learning the scene representation which significantly improves robustness to outliers. DI-Fusion~\cite{huang2021di}, similarly to~\cite{weder2021neuralfusion}, learns the scene representation, but additionally decodes a confidence of the signed distance per voxel.
Continual Neural Mapping~\cite{yan2021continual} learns a continuous scene representation through a neural network from sequential depth maps.
Several recent works do not require depth input and instead perform online reconstruction from RGB-cameras such as Atlas~\cite{murez2020atlas}, VolumeFusion~\cite{choe2021volumefusion}, TransformerFusion~\cite{bovzivc2021transformerfusion} and NeuralRecon~\cite{sun2021neuralrecon}.
None of these approaches consider multiple sensors and their extensions to sensor-aware data fusion is often by no means straightforward.
Nevertheless,
by treating all sensors equally, they can be used as baseline methods.

The majority of the aforementioned traditional methods do not properly account for varying noise and outlier levels for different depth values, which are better handled by probabilistic fusion methods \cite{duan2012probabilistic,duan2015unified,lefloch2015anisotropic,dong2018psdf}.
Cao~\etal~\cite{cao2018real} introduced a probabilistic framework via a Gaussian mixture model into a surfel-based reconstruction framework to account for uncertainties in the observed depth. 
For a recent survey on online RGB-D 3D scene reconstruction, readers are referred to 
\cite{zollhofer2018state}. 
Overall, none of the state-of-the-art methods for dense online 3D scene reconstruction consider multiple sensors.

\boldparagraph{Multi-Sensor Depth Fusion.}
The task of fusing depth maps from diverse sensors has been studied extensively. 
Many works study the fusion of a specific set of sensors. For example, RGB stereo and time-of-flight (ToF)~\cite{van2012sensor,Choi2012fusion,agresti2017deep,evangelidis2015fusion,dal2015probabilistic,marin2016reliable,agresti2019stereo,deng2021tof}, RGB stereo and Lidar~\cite{maddern2016real}, RGB and Lidar~\cite{qiu2019deepLidar,park2018high,patil2020don}, RGB stereo and monocular depth~\cite{martins2018fusion} and the fusion of multiple RGB stereo algorithms~\cite{poggi2016deep}. All these methods only study a specific set of sensors, while we do not enforce such a limitation.
Few works study the fusion of arbitrary depth sensors~\cite{pu2019sdf}. Contrary to our method, all methods performing depth map fusion assume time synchronized sensors, which is hard, if not impossible, to achieve with realistic multi-sensor equipment. 

\boldparagraph{Multi-Sensor Dense 3D Reconstruction.}
Some works consider the problem of offline multi-sensor dense 3D reconstruction. For example, depth map fusion for semantic 3D reconstruction~\cite{rozumnyi2019learned}, combining multi-view stereo with a ToF sensor in a probabilistic framework~\cite{kim2009multi}, the combination of a depth sensor with photometric stereo~\cite{bylow2019combining} and large scene reconstruction using unsynchronized RGBD cameras mounted on an indoor robot~\cite{yang2020noise}.
These offline methods do not address the online problem setting that we are concerned with.
Some works use sensor fusion to achieve robust pose estimation in an online setting~\cite{yang2019heterofusion,gu20203d}.
In contrast to our method, these works do not leverage sensor fusion for mapping.
Ali~\etal~\cite{ali2019multi} present an online framework which perhaps is most closely related to our work.
They take Lidar and stereo depth maps as input and fuse the TSDF signals of both sensors with a linear average before updating the global grid using TSDF Fusion~\cite{curless1996volumetric}.
To reduce noise further, they optimize a least squares problem which encourages surface smoothing.
Contrary to our method, no learning is used and their system is only designed to fuse stereo depth with Lidar.

\section{Method}
\begin{figure*}[t]
\centering
 \includegraphics[trim={1.8cm 2.2cm 0cm 0cm}, align=c, width=\linewidth]{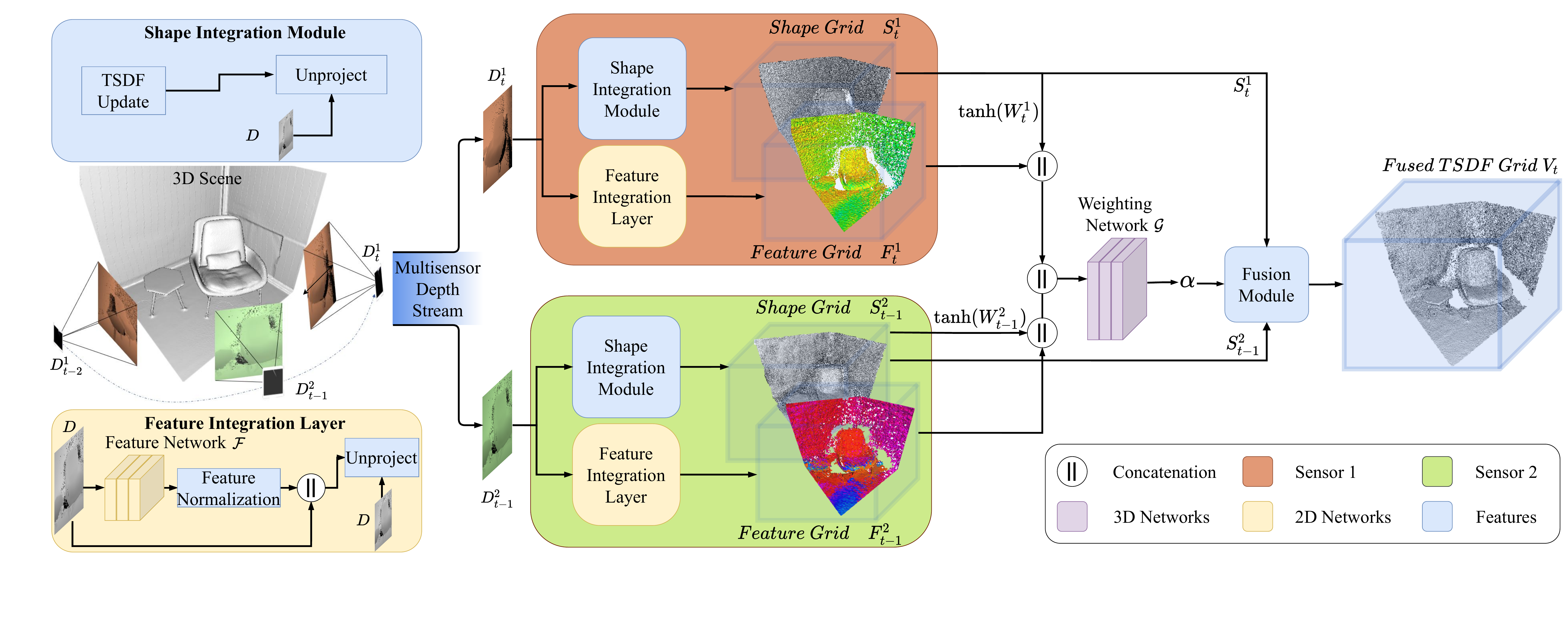}\\[-5pt]
\caption{\textbf{\ours{} Architecture.} Given a depth stream $D^i_t$, with known camera poses, our method fuses each frame at time $t$ from sensor $i$ into global sensor-specific shape $S^i_t$ and feature $F^i_t$ grids.
The \textbf{Shape Integration Module} fuses the frames into $S^i_t = \{V^i_t, W^i_t\}$ consisting of a TSDF grid $V^i_t$ and a weight counter grid $W^i_t$. 
In parallel, the \textbf{Feature Integration Layer} extracts features from the depth maps using a 2D Feature Network $\mathcal{F}^i$ and integrates them into the feature grid $F^i_t$. 
Next, $S^i_t$ and $F^i_t$ are combined and decoded through a 3D \textbf{Weighting Network} $\mathcal{G}$ into a sensor weighting $\alpha \in [0, 1]$. 
Together with $S^i_t$ and $\alpha$, the \textbf{Fusion Module} computes the fused grid $V_t$ at each voxel location.}
\label{fig:architecture}
\end{figure*}

\boldparagraph{Overview.} Given multiple noisy depth streams $D^i_t : \mathbb{R}^2 \rightarrow \mathbb{R}$ from different sensors with known camera calibration, \ie~extrinsics $P^i_t \in \mathbb{SE}(3)$ and intrinsics $K^i \in \mathbb{R}^{3 \times 3}$, our method integrates each depth frame at time $t \in \mathbb{N}$ from sensor $i \in \{1, \ 2\}$ into a globally consistent shape $S^i_t$ and feature $F^i_t$ grid.
Through a series of operations, we then decode $S^i_t$ and $F^i_t$ into a fused TSDF grid $V_t \in \mathbb{R}^{X \times Y \times Z}$, which can be converted into a mesh 
with marching cubes~\cite{lorensen1987marching}.
Our overall framework can be split into four components (see Fig.~\ref{fig:architecture}).
First, the \textbf{Shape Integration Module} integrates depth frames $D^i_t$ successively into the zero-initialized shape representation 
$S^i_t = \{V^i_t, W^i_t\}$. 
$S^i_t$ consists of a TSDF grid $V^i_t \in \mathbb{R}^{X \times Y \times Z}$ and a corresponding weight
grid $W^i_t \in \mathbb{N}^{X \times Y \times Z}$, which keeps track of the number of updates to each voxel.
In parallel, the \textbf{Feature Integration Layer} extracts features from the depth maps using a 2D feature network $\mathcal{F}^i : D^i_t \in \mathbb{R}^{W \times H \times 1} \rightarrow f^i_t \in \mathbb{R}^{W \times H \times n}$, with $n$ being the feature dimension.
We use separate feature networks per sensor to learn sensor-specific depth dependent statistics such as shape and edge information.
The extracted features $f^i_t$ are then integrated into the zero-initialized feature grid $F^i_t \in \mathbb{R}^{X \times Y \times Z \times n}$. 
Next, $S^i_t$ and $F^i_t$
are combined and decoded through a 3D \textbf{Weighting Network} $\mathcal{G}$ into a location-dependent
sensor weighting $\alpha \in [0, 1]$.
Together with $S^i_t$ and $\alpha$, the \textbf{Fusion Module} fuses the information into $V_t$ at each voxel location. Key to our approach is the separation of per sensor information into different representations along with the successive aggregation of shapes and features in the 3D domain.
This strategy enables $\mathcal{G}$ to learn a fusion strategy of the incoming multi-sensor depth stream. 
Our method is able to fuse the sensors in a spatially dependent manner from a smooth combination to a hard selection as illustrated in Fig.~\ref{fig:overview}. 
Our scheme hence avoids the need to perform post-outlier filtering by thresholding with the weight $W_t^i$, which is difficult to tune and is prone to reduce scene completion~\cite{Weder2020RoutedFusionLR}.
Another popular outlier filtering technique is free-space carving\footnote{Enforcing free space for voxels along the ray from the camera to the surface~\cite{newcombe2011kinectfusion}. Note that outliers behind surfaces are not removed with this technique.}, but this can be computationally expensive and is not required by our method. Instead, we use the learned $\alpha$ as part of an outlier filter at test time, requiring no manual tuning.
Next, we describe each component in detail.
\ifeccv
\begin{wrapfigure}[20]{R}{0.55\linewidth}
\vspace{-\intextsep}
\centering
\scriptsize
\setlength{\tabcolsep}{0pt}
\renewcommand{\arraystretch}{1}
\begin{tabular}{cccc}
 & \multicolumn{3}{l}{\rule[3pt]{0.5cm}{0.4pt} 25 \rule[3pt]{1.5cm}{0.4pt} 75 \rule[3pt]{1.525cm}{0.4pt} 120 $\xrightarrow{\hspace*{0.1cm}}$ Time} \\
\rotatebox[origin=c]{90}{\makecell{Online \\ Sensor Fusion}} & \includegraphics[align=c, width=.24\linewidth]{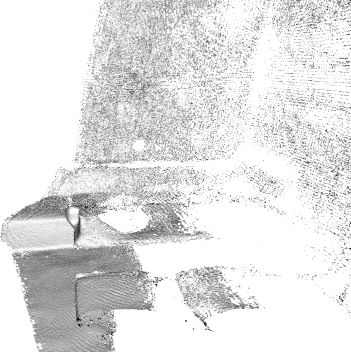} & \includegraphics[align=c, width=.32\linewidth]{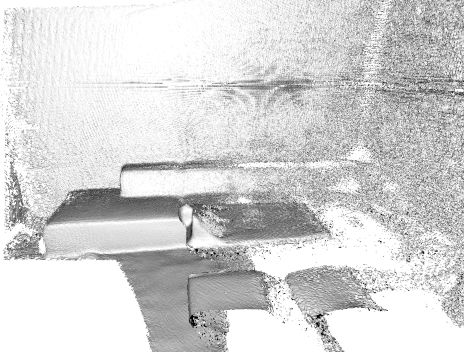} & \includegraphics[align=c, width=.345\linewidth]{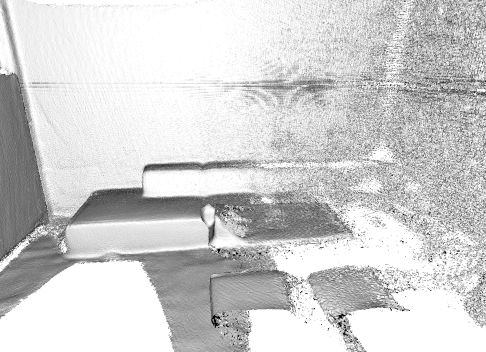} \\
\rotatebox[origin=c]{90}{\makecell{Sensor \\ Weighting}} & \includegraphics[align=c, width=.24\linewidth]{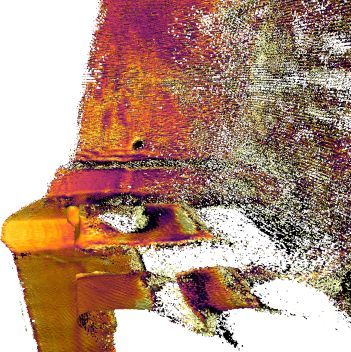} & \includegraphics[align=c, width=.32\linewidth]{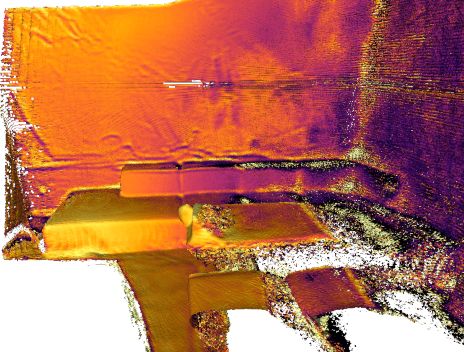} & \includegraphics[align=c, width=.345\linewidth]{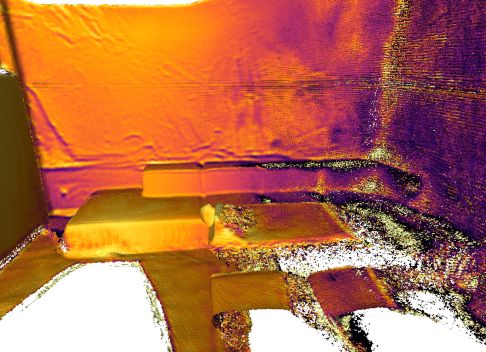} \\
& \multicolumn{3}{c}{\includegraphics[align=c, width=0.91\linewidth]{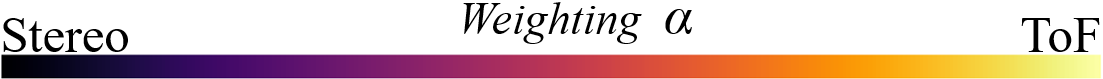}}
\end{tabular}
\begin{tabular}{cccccccc}
& ToF & Stereo & Stereo & ToF & Stereo & ToF & ToF \\
\rotatebox[origin=c]{90}{\makecell{Depth \\ Stream}} & \includegraphics[align=c, width=.13\linewidth]{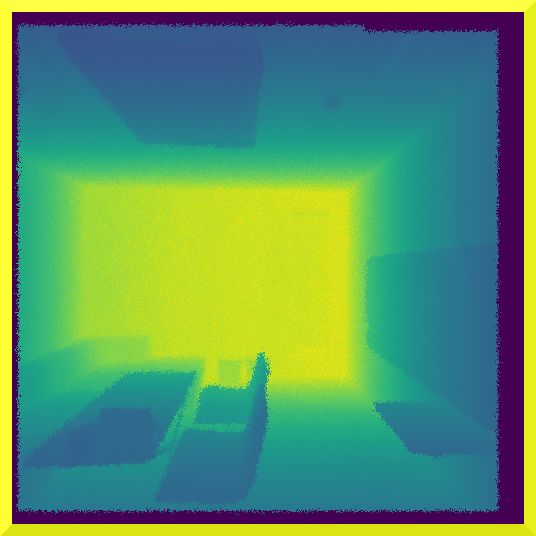} & \includegraphics[align=c, width=.13\linewidth]{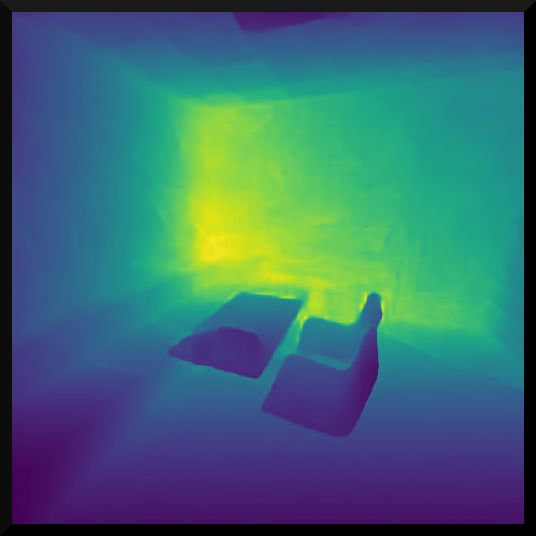} & \includegraphics[align=c, width=.13\linewidth]{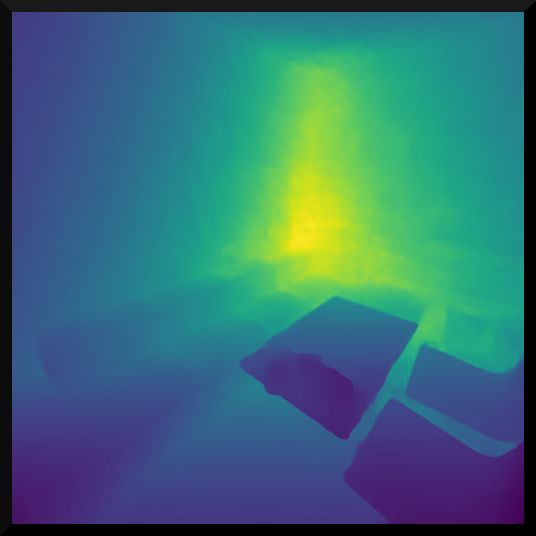}& \includegraphics[align=c, width=.13\linewidth]{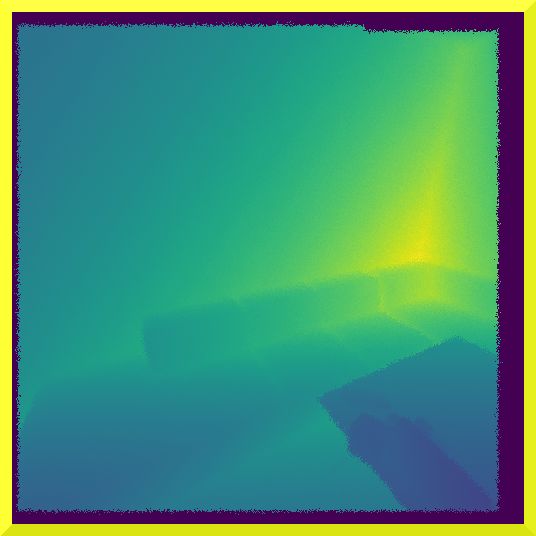} & \includegraphics[align=c, width=.13\linewidth]{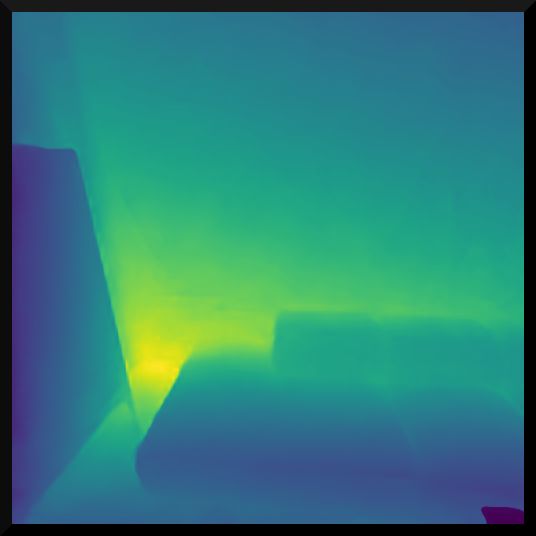} & \includegraphics[align=c, width=.13\linewidth]{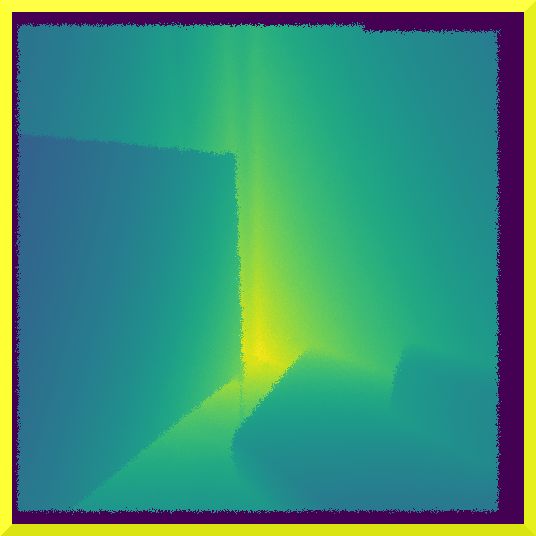} & \includegraphics[align=c, width=.13\linewidth]{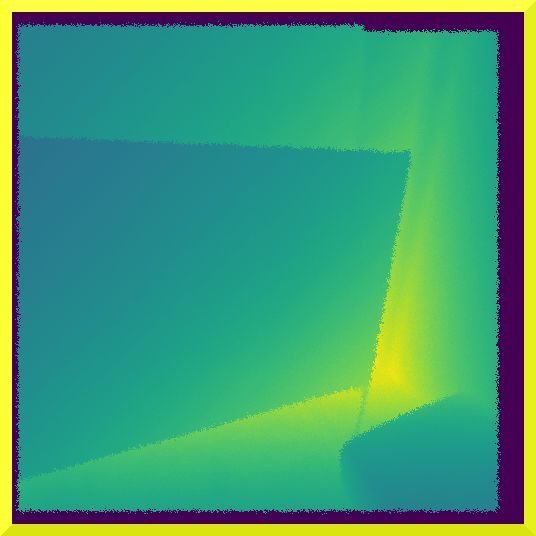}\\
$t=$ & $0$ & $26$ & $49$ & $65$ & $85$ & $103$ & $120$ \\ 
\end{tabular}
\caption{\textbf{Overview.} Left to right: Sequential fusion of a multi-sensor noisy depth stream. Our method integrates each depth frame at time $t$ and produces a sensor weighting which fuses the sensors in a spatially dependent manner. For example, areas in yellow denote high trust of the ToF sensor.}
\label{fig:overview}
\end{wrapfigure}
\fi

\boldparagraph{(a) Shape Integration Module.}
For each depth map $D^i_t$ and pixel, a full perspective unprojection of the depth into the world coordinate frame yields a 3D point $\textbf{x}_w \in \mathbb{R}^3$. 
Along each ray from the camera center, centered at $\textbf{x}_w$, we sample $T$ points uniformly over a predetermined distance $l$.
The coordinates are then converted
to the voxel space and a local shape grid $S^{i, *}_{t-1}$ is extracted from $S^i_{t-1}$ through nearest neighbor extraction. 
To incrementally update the local shape grid, we follow the moving average update scheme of TSDF Fusion~\cite{curless1996volumetric}. 
For numerical stability, the weights are clipped at a maximum weight $\omega_\textrm{max}$. 
For more details, see the suppl. material.

\boldparagraph{(b) Feature Integration Layer.}
Each depth map $D^i_t$ is passed through a 2D network $\mathcal{F}^i$ to extract context information $f^i_t$, which can be useful during the sensor fusion process. When fusing sensors based on the stereo matching principle, we provide the RGB frame as additional input channels to $\mathcal{F}^i$.
The network is fully convolutional and comprises 7 network blocks, each consisting of the following operations, 
1) a $3 \times 3$ convolution with zero padding $1$ and input channel dimension $4$ and output dimension $4$ (except the first block which takes $1$ channel as input when only depth is provided), 
2) a $\mathop{\mathrm{tanh}}$ activation, 
3) another $3 \times 3$ convolution with zero padding $1$ outputting $4$ channels and 
4) a $\mathop{\mathrm{tanh}}$ activation. 
The output of the first six blocks is added to the output of the next block via residual connections.
Finally, we normalize the feature vectors at each pixel location and concatenate the input depth.

Next, we repeat the features $T$ times along the direction of the viewing ray from the camera, $f^i_t \xrightarrow[\text{T times}]{\text{Repeat}}f^{i, T}_t \in \mathbb{R}^{W \times H \times T \times n}$.
The local feature grid $F^{i,*}_{t-1}$ is then updated using the precomputed update indices from the Shape Integration Module with a moving average update: $F^{i,*}_t = \frac{W^{i,*}_{t-1}F^{i,*}_{t-1} + f^{i,T}_t}{W^{i,*}_{t-1} + 1} .$
For all update locations the grid $F^i_t$ is replaced with $F^{i, *}_t$.

\boldparagraph{(c) Weighting Network.}
The task of the weighting network $\mathcal{G}$ is to predict the optimal fusion strategy of the surface hypotheses $V^i_t$.
The input to the network is prepared by first concatenating the features $F^i_t$ and the $\mathop{\mathrm{tanh}}$-transformed weight counters $W^i_t$ and second by concatenating the resulting vectors across the sensors. 
Due to memory constraints, the entire scene cannot be fit onto the GPU, and hence we use a sliding-window approach at test time to feed $\mathcal{G}$ chunks of data.  
First, the minimum bounding grid of the measured scene (\ie~where $W^i_t > 0$) is extracted from the global grids. 
Then, the extracted grid is parsed using chunks of size $d \times d \times d$ through $\mathcal{G} : \mathbb{R}^{d \times d \times d \times 2(n+1)} \rightarrow \alpha \in \mathbb{R}^{d \times d \times d \times 1}$ into $\alpha \in [0, 1]$.
To avoid edge effects, we use a stride of $d/2$ and update the central chunk of side length $d/2$. 
The architecture of $\mathcal{G}$ combines 2 layers of 3D-convolutions with kernel size 3 and replication padding 1 interleaved with $\mathop{\mathrm{ReLU}}$ activations. 
The first layer outputs $32$ and the second layer $16$ channels.
Finally, the $16$-dimensional features are decoded into the sensor weighting $\alpha$ by a $1\!\times\!1\!\times\!1$ convolution followed by a sigmoid activation. 

\boldparagraph{(d) Fusion Module.} The task of the fusion module is to combine $\alpha$ with the shapes $S^i_t$.
In the following, we define the set of voxels where only sensor 1 integrates as $C^1 = \{W^1_t > 0, \ W^2_{t-1} = 0\}$, the set where only sensor 2 integrates as $C^2 = \{W^1_t = 0, \ W^2_{t-1} > 0\}$ and the set where both sensors integrate as $C^{12} = \{W^1_t > 0, \ W^2_{t-1} > 0\}$. 
Let us also introduce $\alpha_1 = \alpha$ and $\alpha_2 = 1 - \alpha$.
The fusion module computes the fused grid $V_t$ as

\begin{equation}
  V_t =
    \begin{cases}
      \alpha_1 V_t^1 + \alpha_2V_{t-1}^2  & \mbox{if } C^{12} \\
      V_t^1                               & \mbox{if } C^1\\
      V_{t-1}^2                           & \mbox{if } C^2.
    \end{cases}       
    \label{eq:fusion}
\end{equation}
Depending on the voxel set, $V_t$ is computed either as a weighted average of the two surface hypotheses or by selecting one of them.
With only one sensor observation, a weighted average would corrupt the result.
Hence, the single observed surface is selected.
At test time, we additionally apply a learned \textbf{Outlier Filter} which utilizes 
$\alpha_i$ and 
$W^i_t$. 
The filter is formulated for sensors 1 and 2 as
%
%
%
\begin{equation}
    \hat{W}_t^1 = \1{C^1, \ \alpha_1 > 0.5}W_t^1 ,\quad \hat{W}_{t-1}^2 = \1{C^2, \ \alpha_2 > 0.5}W_{t-1}^2 ,
    \label{eq:outlier_filter}
\end{equation}
where $\1{.}$ denotes the indicator function\footnote{See supplementary material for a definition.}.
When only one sensor is observed at a certain voxel, we remove the observation if $\alpha_i$, which could be interpreted as a confidence, is below $0.5$. 
This is done by setting the weight counter to $0$.

\ifeccv
\else
\begin{figure}[t]
\centering
\footnotesize
\setlength{\tabcolsep}{0pt}
\renewcommand{\arraystretch}{1}
\begin{tabular}{cccc}
 & \multicolumn{3}{l}{\rule[3pt]{0.8cm}{0.4pt} 25 \rule[3pt]{1.85cm}{0.4pt} 75 \rule[3pt]{2.2cm}{0.4pt} 120 $\xrightarrow{\hspace*{0.125cm}}$ Time} \\
\rotatebox[origin=c]{90}{\makecell{Online \\ Fusion}} & \includegraphics[align=c, width=.24\linewidth]{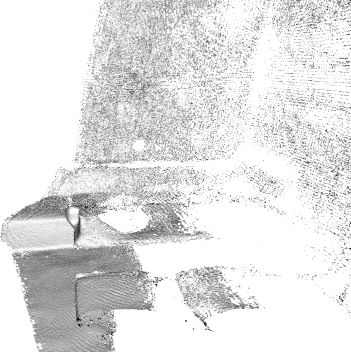} & \includegraphics[align=c, width=.32\linewidth]{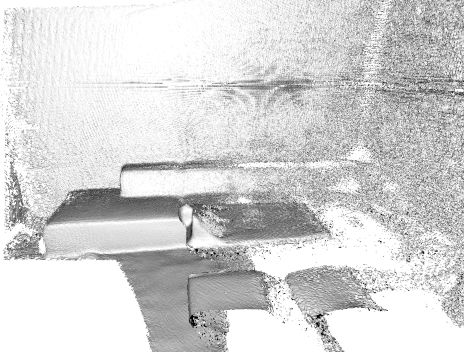} & \includegraphics[align=c, width=.345\linewidth]{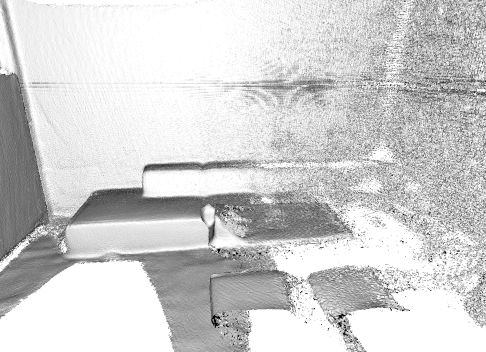} \\
\rotatebox[origin=c]{90}{\makecell{Sensor \\ Weighting}} & \includegraphics[align=c, width=.24\linewidth]{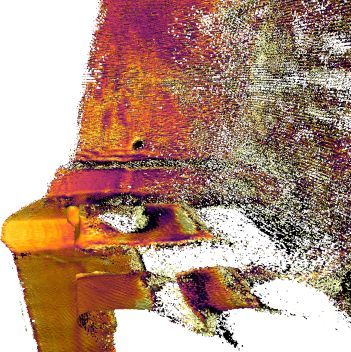} & \includegraphics[align=c, width=.32\linewidth]{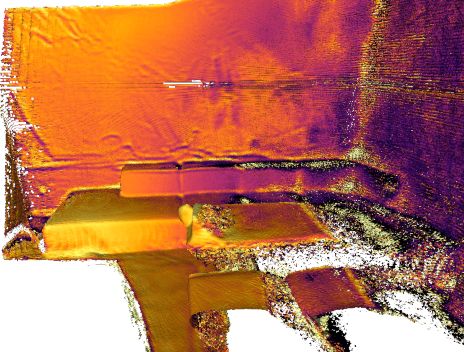} & \includegraphics[align=c, width=.345\linewidth]{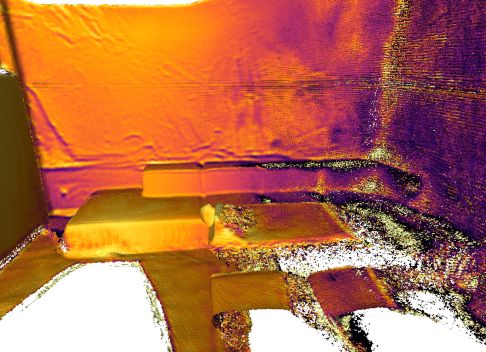} \\
& \multicolumn{3}{c}{\includegraphics[align=c, width=0.91\linewidth]{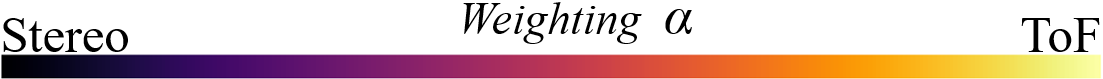}}
\end{tabular}
\begin{tabular}{cccccccc}
& ToF & Stereo & Stereo & ToF & Stereo & ToF & ToF \\
\rotatebox[origin=c]{90}{\makecell{Depth \\ Stream}} & \includegraphics[align=c, width=.13\linewidth]{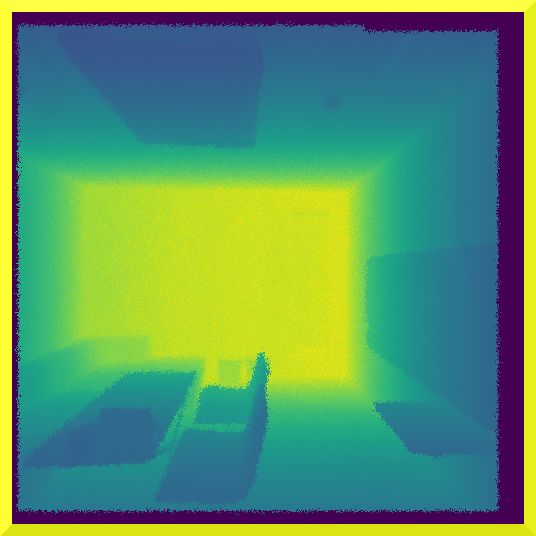} & \includegraphics[align=c, width=.13\linewidth]{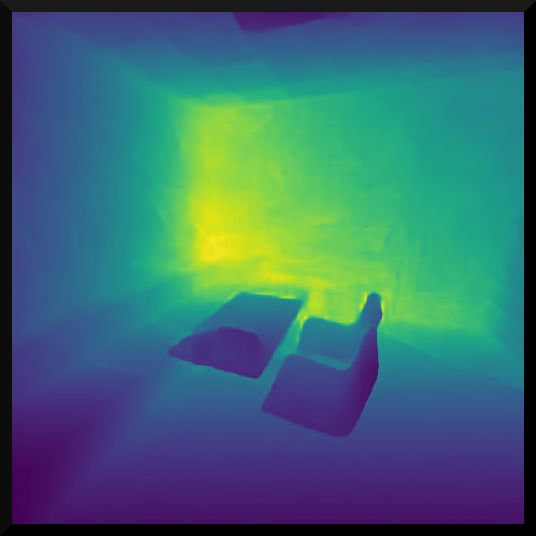} & \includegraphics[align=c, width=.13\linewidth]{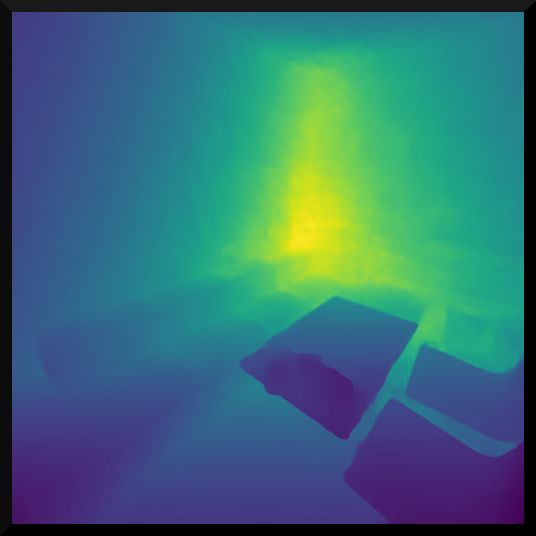}& \includegraphics[align=c, width=.13\linewidth]{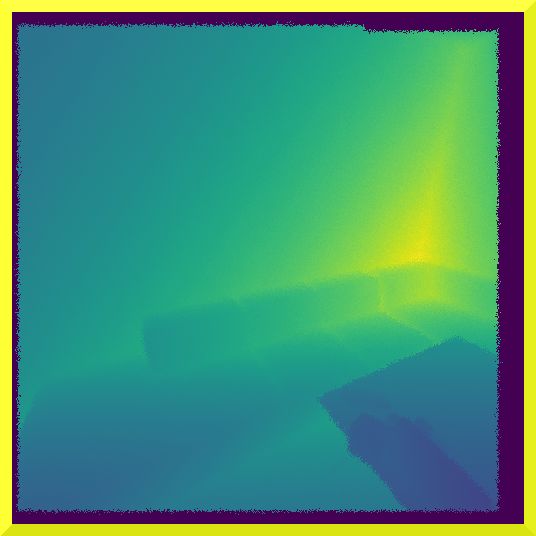} & \includegraphics[align=c, width=.13\linewidth]{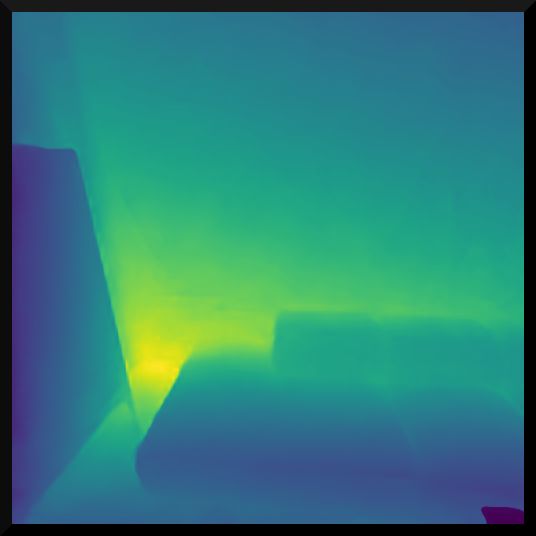} & \includegraphics[align=c, width=.13\linewidth]{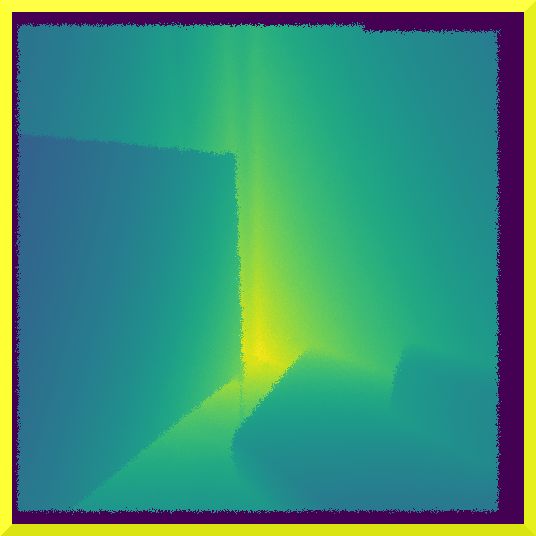} & \includegraphics[align=c, width=.13\linewidth]{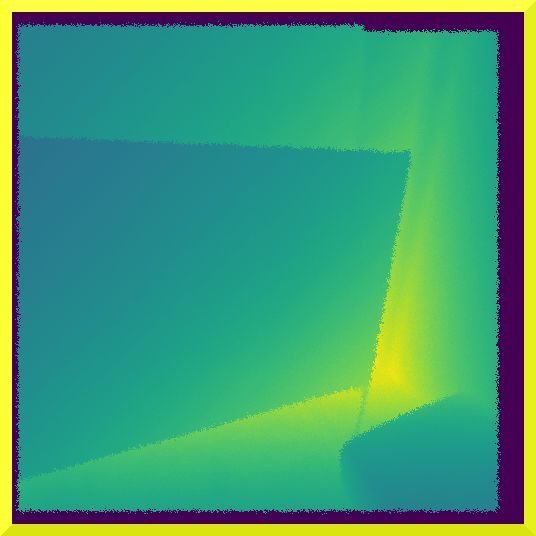}\\
$t=$ & $0$ & $26$ & $49$ & $65$ & $85$ & $103$ & $120$ \\ 
\end{tabular}
\caption{\textbf{Overview.} Left to right: Sequential fusion of a multi-sensor noisy depth stream. Our method fuses each depth frame at time $t$ to produce a sensor weighting which fuses the sensors in a spatially dependent manner.}
\label{fig:overview}
\end{figure}
\fi

\boldparagraph{Loss Function.}
The full pipeline is trained end-to-end using the overall loss
\begin{equation}
     \mathcal{L} = \mathcal{L}_{f} + \lambda_1 \sum_{i=1}^2\mathcal{L}_{C^i}^{in} + \lambda_2 \sum_{i=1}^2\mathcal{L}_{C^i}^{out}.
     \label{eq:loss}
\end{equation}
The term $\mathcal{L}_{f}$ computes the mean $L_1$ error to the ground truth TSDF masked by $C^{12}$ \eqref{eq:lf}.
To supervise the voxel sets $C^{1}$ and $C^{2}$, we introduce two additional terms, which penalize $L_1$ deviations from the optimal $\alpha$. 
The purpose of these terms is to provide a training signal for the outlier filter.
If the $L_1$ TSDF error is smaller than some threshold $\eta$, the observation is deemed to be an inlier, and the corresponding confidence $\alpha_i$ should be $1$, otherwise $0$. 
The loss is computed as the mean $L_1$ error to the optimal $\alpha$:

\ifeccv

\begin{align}
    \mathcal{L}_{f} = \frac{1}{N_{C^{12}}}\sum \1{C^{12}}|V_t & -  V^{GT}|_1, \
    \mathcal{L}_{C^i}^{in} = \frac{1}{N_{C^i}^{in}} \sum \1{C^i, \ |V_t-V^{GT}|_1 < \eta}| \alpha_i - 1|_1, \nonumber \\
    \mathcal{L}_{C^i}^{out} &= \frac{1}{N_{C^i}^{out}} \sum \1{C^i, \ |V_t-V^{GT}|_1 > \eta}| \alpha_i|_1,
    \label{eq:lf}
\end{align}
\else
\begin{align}
    \mathcal{L}_{f} &= \frac{1}{N_{C^{12}}}\sum \1{C^{12}}|V_t - V^{GT}|_1, \label{eq:lf} \\
    \mathcal{L}_{C^i}^{in} &= \frac{1}{N_{C^i}^{in}} \sum \1{C^i, \ |V_t-V^{GT}|_1 < \eta}| \alpha_i - 1|_1, \label{eq:lin} \\
    \mathcal{L}_{C^i}^{out} &= \frac{1}{N_{C^i}^{out}} \sum \1{C^i, \ |V_t-V^{GT}|_1 > \eta}| \alpha_i|_1,
    \label{eq:lout}
\end{align}
\fi

where the normalization factors are defined as
\begin{align}
    N_{C^{12}} = \sum &\1{C^{12}}, \ N_{C^i}^{in} = \sum \1{C^i, \ |V_t - V^{GT}|_1 < \eta}, \nonumber \\ N_{C^i}^{out} &= \sum \1{C^i, \ |V_t - V^{GT}|_1 > \eta} .
\end{align}

\boldparagraph{Training Forward Pass.}
After the integration of a new depth frame $D^i_t$ into the shape and feature grids, the update indices from the Shape Integration Module are used to compute the minimum bounding box of update voxels in $S^i_t$ and $F^i_t$.
The update box varies in size between frames and cannot always fit on the GPU.
Due to this and for the sake of training efficiency, we extract a $d \times d \times d$ chunk from within the box volume.
The chunk location is randomly selected using a uniform distribution along each axis of the bounding box. 
If the bounding box volume is smaller along any dimension than $d$, the chunk shrinks to the minimum size along the affected dimension.
To maximize the number of voxels that are used to train the networks $F^i$, we sample a chunk until we find one with at least 2000 update indices.
At most, we make 600 attempts. 
If not enough valid indices are found, the next frame is integrated. 
The update indices in the chunk are finally used to mask the loss.
We randomly reset the shape and feature grids with a probability of 0.01 at each frame integration to improve training robustness.
\section{Experiments}
We first describe our experimental setup and then evaluate our method against state-of-the-art online depth fusion methods on Replica, the real-world CoRBS and the Scene3D datasets.
All reported results are averages over the respective test scenes.
Further experiments and details are in the supplementary material.

\boldparagraph{Implementation Details.}
We use $\omega_\textrm{max} = 500$ and extract $T = 11$ points along the update band $l = 0.1$ m. We store $n = 5$ features at each voxel location and use a chunk side length of $d = 64$.
For the loss~\eqref{eq:loss} $\lambda_1 = 1/60$, $\lambda_2 = 1/600 $ and $\eta = 0.04$ m.
In total, the networks of our model comprise $27.7$K parameters, where $24.3$K are designated to $\mathcal{G}$ and the remaining parameters are split equally between $\mathcal{F}^1$ and $\mathcal{F}^2$. 
For our method and all baselines, the image size is $W=H = 256$, the voxel size is $0.01$ m and we mask the 10 pixel border of all depth maps to avoid edge artifacts, \ie~pixels belonging to the mask are not integrated into 3D. 
Since our TSDF updates cannot be larger than $0.05$ m, we truncate the ground truth TSDF grid at $l/2 = 0.05$ m.

\boldparagraph{Evaluation Metrics.} The TSDF grids are evaluated using the Mean Absolute Distance
(MAD), Mean Squared Error (MSE), Intersection over Union (IoU) and Accuracy (Acc.).
The meshes, produced by marching cubes~\cite{lorensen1987marching} from the TSDF grids, are evaluated using the F-score which is the harmonic mean of the Precision (P) and Recall (R).

\boldparagraph{Baseline Methods.} 
Since there is no other multi-sensor online 3D reconstruction method that addresses the same problem, we define our own baselines by generalizing single sensor fusion pipelines to multiple sensors.
TSDF Fusion~\cite{curless1996volumetric} is the gold standard for fast, dense mapping of posed depth maps.
It generalizes to the multi-sensor setting effortlessly by integrating all depth frames into the same TSDF grid at runtime.
RoutedFusion~\cite{Weder2020RoutedFusionLR} extends TSDF Fusion by learning the TSDF mapping. 
We generalize RoutedFusion to multiple sensors by feeding all depth frames into the same TSDF grid, but each sensor is assigned a separate fusion network to account for sensor-dependent noise\footnote{Additionally, we tweak the original implementation to get rid of outliers. See supplementary material.}. 
NeuralFusion~\cite{weder2021neuralfusion} extends RoutedFusion for better outlier handling, but despite efforts and help from the authors, the network did not converge during training due to heavy loss oscillations caused by integrating different sensors. DI-Fusion~\cite{huang2021di} learns the scene representation and predicts the signed distance value as well as the uncertainty $\sigma$ per voxel. We use the provided model from the authors and integrate all frames from both sensors into the same volumetric grid.
In the following, we refer to each multi-sensor baseline by using the corresponding single sensor name. 
For additional comparison, when time synchronized sensors with ground truth depth are available, we train a so-called ``Early Fusion'' baseline by fusing the 2D depth frames of both sensors.
The fusion is performed with a modified version of the 2D denoising network proposed by Weder~\etal~\cite{Weder2020RoutedFusionLR} followed by TSDF Fusion to attain the 3D model (see supplementary material). 
This baseline should be interpreted as a light-weight alternative to our proposed SenFuNet, but assumes synchronized sensors, which SenFuNet does not.
Finally, for the single-sensor results, we evaluate the TSDF grids $V^i_t$.
To make the comparisons fair, we do not use weight counter thresholding as post-outlier filter for any method. For DI-Fusion, we filter outliers by thresholding the learned voxel uncertainty. The default value provided in the implementation is used.

\subsection{Experiments on the Replica Dataset} 
%
The Replica dataset~\cite{straub2019replica} comprises high-quality 3D reconstructions of a variety of indoor scenes. We collect data from Replica to create a multi-sensor dataset suitable for depth map fusion.
To prepare ground truth signed distance grids, we first make the 3D meshes watertight using screened Poisson surface reconstruction~\cite{kazhdan2013screened}.
The meshes are then converted to signed distance grids using a modified version of mesh-to-sdf\footnote{\normalfont\url{https://github.com/marian42/mesh_to_sdf}} to accommodate non-cubic voxel grids. 
Ground truth depth and an RGB stereo pair are extracted using AI Habitat~\cite{habitat19iccv} along random trajectories.
In total, we collected $92698$ frames.
We use 7 training and 3 test scenes.
We simulate a depth sensor by adding noise to the GT depth of the left stereo view.
Correspondingly, from the RGB stereo pairs a left stereo view
depth map can be predicted using (optionally multi-view) stereo algorithms.
In the following, we construct two sensor combinations and evaluate our model.

\ifeccv
\begin{table}[tb]
\setlength{\belowcaptionskip}{0pt}
\begin{subtable}{.495\columnwidth}
\centering
\resizebox{\columnwidth}{!}
{
\setlength{\tabcolsep}{2pt}
\renewcommand{\arraystretch}{1.05}
\begin{tabular}{l|lllllll}
\cellcolor{gray}       & \cellcolor{gray}MSE$\downarrow$      & \cellcolor{gray}MAD$\downarrow$  & \cellcolor{gray}IoU$\uparrow$     & \cellcolor{gray}Acc.$\uparrow$ & \cellcolor{gray}F$\uparrow$ & \cellcolor{gray}P$\uparrow$ & \cellcolor{gray}R$\uparrow$ \\
\multirow{-2}{*}{\cellcolor{gray} \backslashbox[28.3mm]{Model}{Metric}} & \cellcolor{gray}*e-04 & \cellcolor{gray}*e-02 & \cellcolor{gray}[0,1] & \cellcolor{gray}$[\%]$ & \cellcolor{gray}$[\%]$ & \cellcolor{gray}$[\%]$ & \cellcolor{gray}$[\%]$ \\\hline
\multicolumn{8}{c}{\emph{Single Sensor}} \\ \hline
PSMNet~\cite{chang2018pyramid}  & 7.30 & 1.95 & 0.664 & 83.23 & 56.20 & 43.10 & 81.34 \\ 
ToF~\cite{handa2014benchmark}    & 7.48 & 1.99 & 0.664 & 83.65 & 58.52 & 45.84 & 84.85 \\ 
\hline
\multicolumn{8}{c}{\emph{Multi-Sensor Fusion}} \\ \hline
TSDF Fusion~\cite{curless1996volumetric} & 8.20 & 2.11 & 0.669 & 84.09 & 49.58 & 35.33 & \textbf{85.44}  \\  
RoutedFusion~\cite{Weder2020RoutedFusionLR} & 5.62 & 1.66 & 0.735 & 87.22 & 61.08 & 49.51 & 79.92 \\ 
DI-Fusion~\cite{huang2021di} $\sigma$=0.15 & - & - & - & - & 48.39 & 34.24 & 85.29 \\ 
\textbf{\ours{} (Ours)} & \textbf{4.65}  & \textbf{1.54}  & \textbf{0.753}  & \textbf{88.05}  & \textbf{69.29} & \textbf{62.05} & 79.81 \\ 
\end{tabular}
}
\subcaption{Without depth denoising.}
\label{tab:tof_psmnet_no_denoising}
\end{subtable}
\begin{subtable}{0.495\columnwidth}
\resizebox{\columnwidth}{!}
{
\setlength{\tabcolsep}{2pt}
\renewcommand{\arraystretch}{1.05}
\begin{tabular}{l|lllllll}
\cellcolor{gray}      & \cellcolor{gray}MSE$\downarrow$      & \cellcolor{gray}MAD$\downarrow$  & \cellcolor{gray}IoU$\uparrow$     & \cellcolor{gray}Acc.$\uparrow$ & \cellcolor{gray}F$\uparrow$ & \cellcolor{gray}P$\uparrow$ & \cellcolor{gray}R$\uparrow$   \\ 
\multirow{-2}{*}{\cellcolor{gray} \backslashbox[28.3mm]{Model}{Metric}}& \cellcolor{gray}*e-04 & \cellcolor{gray}*e-02 & \cellcolor{gray}[0,1] & \cellcolor{gray}$[\%]$ & \cellcolor{gray}$[\%]$ & \cellcolor{gray}$[\%]$ & \cellcolor{gray}$[\%]$ \\\hline
\multicolumn{8}{c}{\emph{Single Sensor}} \\ \hline
PSMNet~\cite{chang2018pyramid}   & 6.35 & 1.77 & 0.673 & 84.54 & 60.28 & 48.26 & 80.41 \\ 
ToF~\cite{handa2014benchmark} & 5.08 & 1.58 & 0.709 & 87.32 & 68.93 & 59.01  & 83.08 \\ 
\hline
\multicolumn{8}{c}{\emph{Multi-Sensor Fusion}} \\ \hline
TSDF Fusion~\cite{curless1996volumetric} & 6.40 & 1.80 & 0.681 & 85.31 & 52.93 & 38.95 & \bf 84.60  \\ 
RoutedFusion~\cite{Weder2020RoutedFusionLR} & 6.04 & 1.68 & 0.644 & 85.10 & 62.67 & 51.75 & 79.52 \\ 
Early Fusion & 6.40 & 1.40 & 0.760 & 89.02 & 74.60 & 67.46 & 83.47 \\ 
DI-Fusion~\cite{huang2021di} $\sigma$=0.15 & - & - & - & - & 55.66 & 41.49 & 85.33 \\ 
\textbf{\ours{} (Ours)} & \textbf{3.49}  & \textbf{1.31}  & \textbf{0.761}  & \textbf{89.61}  & \textbf{76.47} & \textbf{73.58} & 79.77 \\ 
\end{tabular}
}
\subcaption{With depth denoising}
\label{tab:tof_psmnet_denoising}
\end{subtable}
\caption{\textbf{Replica Dataset. ToF\bplus{}PSMNet Fusion.} (a) Our method outperforms the baselines as well as both of the sensor inputs and sets a new state-of-the-art for multi-sensor online depth fusion.
(b) The denoising network mitigates outliers along planar regions, compare to Tab.~\ref{tab:tof_psmnet_no_denoising}. Our method even outperforms the Early Fusion baseline, which assumes synchronized sensors.}
\label{tab:tof_psmnet}
\end{table}
\else
\begin{table}[tb]
\centering
\resizebox{\columnwidth}{!}
{
\begin{tabular}{l|lllllll}
\cellcolor{gray}       & \cellcolor{gray}MSE$\downarrow$      & \cellcolor{gray}MAD$\downarrow$  & \cellcolor{gray}IoU$\uparrow$     & \cellcolor{gray}Acc.$\uparrow$ & \cellcolor{gray}F$\uparrow$ & \cellcolor{gray}P$\uparrow$ & \cellcolor{gray}R$\uparrow$ \\
\multirow{-2}{*}{\cellcolor{gray} \backslashbox[28.3mm]{Model}{Metric}} & \cellcolor{gray}*e-04 & \cellcolor{gray}*e-02 & \cellcolor{gray}[0,1] & \cellcolor{gray}$[\%]$ & \cellcolor{gray}$[\%]$ & \cellcolor{gray}$[\%]$ & \cellcolor{gray}$[\%]$ \\\hline
\multicolumn{8}{c}{\emph{Single Sensor}} \\ \hline
PSMNet~\cite{chang2018pyramid}  & 7.30 & 1.95 & 0.664 & 83.23 & 56.20 & 43.10 & 81.34 \\ 
ToF~\cite{handa2014benchmark}    & 7.48 & 1.99 & 0.664 & 83.65 & 58.52 & 45.84 & 84.85 \\ 
\hline
\multicolumn{8}{c}{\emph{Multi-Sensor Fusion}} \\ \hline
TSDF Fusion~\cite{curless1996volumetric} & 8.20 & 2.11 & 0.669 & 84.09 & 49.58 & 35.33 & \textbf{85.44}  \\  
RoutedFusion~\cite{Weder2020RoutedFusionLR} & 5.62 & 1.66 & 0.735 & 87.22 & 61.08 & 49.51 & 79.92 \\ 
DI-Fusion~\cite{huang2021di} $\sigma$=0.15 & - & - & - & - & 48.39 & 34.24 & 85.29 \\ 
\textbf{\ours{} (Ours)} & \textbf{4.65}  & \textbf{1.54}  & \textbf{0.753}  & \textbf{88.05}  & \textbf{69.29} & \textbf{62.05} & 79.81 \\ 
\end{tabular}
}
\caption{\textbf{Replica Dataset. ToF\bplus{}PSMNet Fusion without denoising.} Our method outperforms the baselines as well as both of the sensor inputs and sets a new state-of-the-art for multi-sensor online depth fusion.}
\label{tab:tof_psmnet_no_denoising}
\end{table}
\fi

\ifeccv
\begin{figure*}[t]
\centering
{\tiny
\setlength{\tabcolsep}{1pt}
\renewcommand{\arraystretch}{1}
\newcommand{\sz}{0.153}
\begin{tabular}{ccccccccc}
\rotatebox[origin=c]{90}{Office 4} & 
\includegraphics[align=c, width=\sz\linewidth]{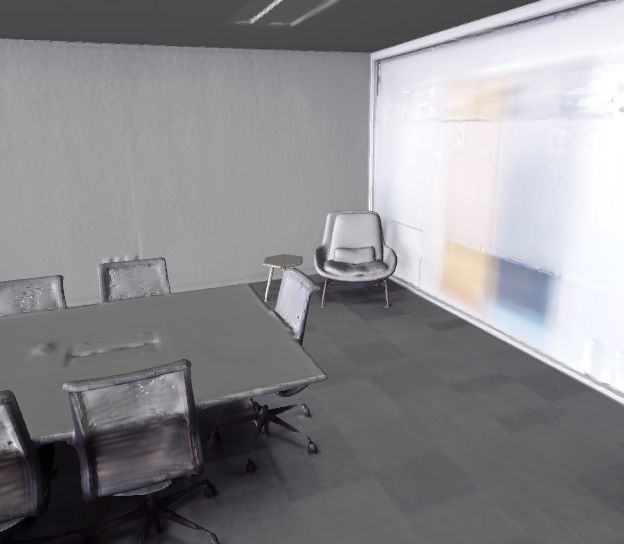} & 
\includegraphics[align=c, width=\sz\linewidth]{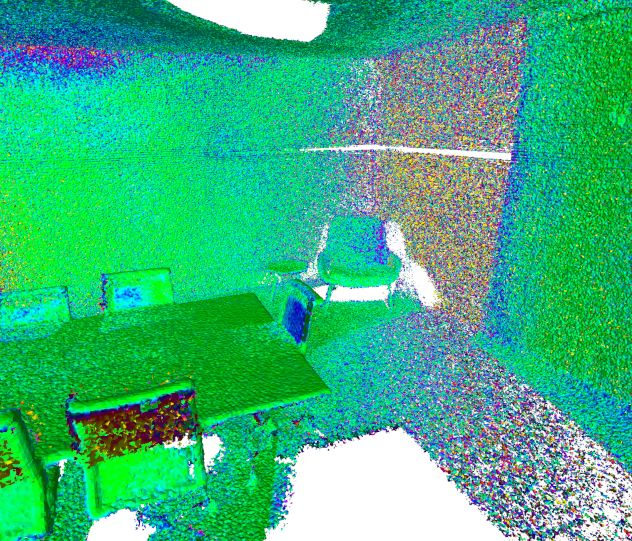} &
\includegraphics[align=c, width=\sz\linewidth]{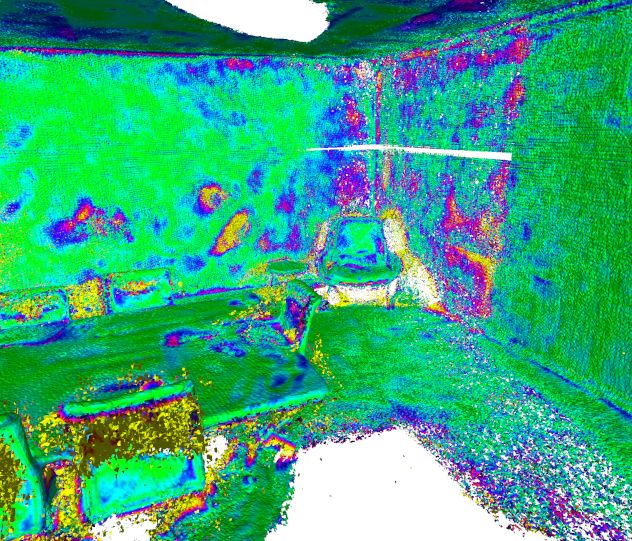} &
\includegraphics[align=c, width=\sz\linewidth]{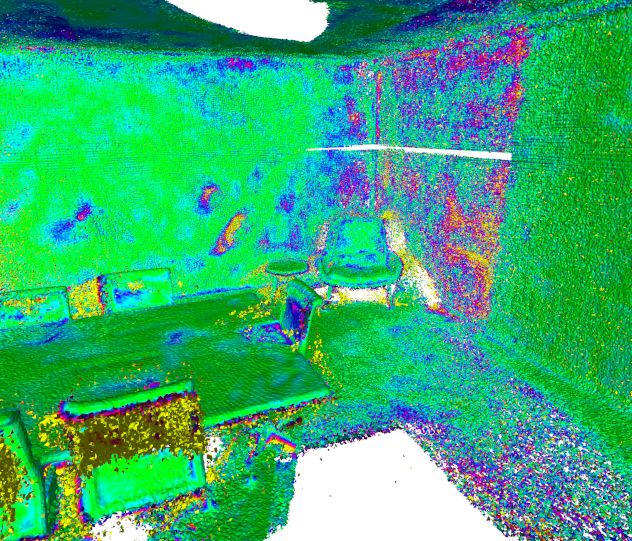} &
\includegraphics[align=c, width=\sz\linewidth]{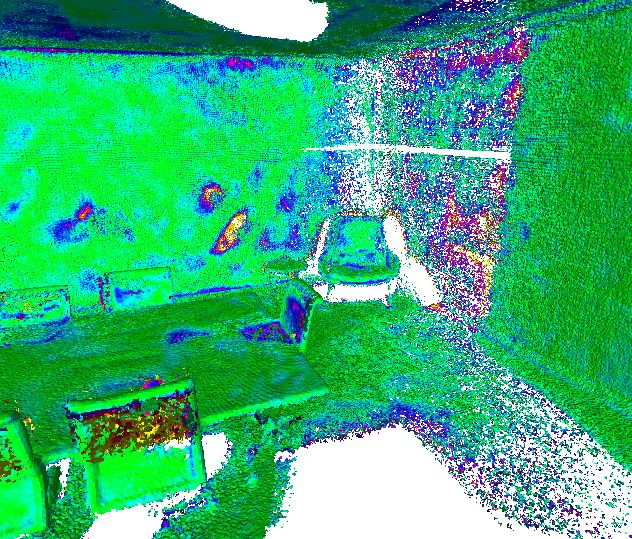} & 
\includegraphics[align=c, width=\sz\linewidth]{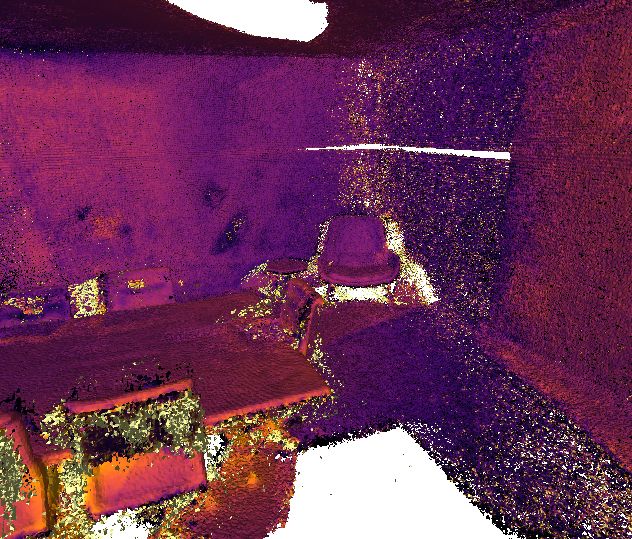} & \multirow{12}{*}[20.0pt]{\includegraphics[width=.0372\linewidth]{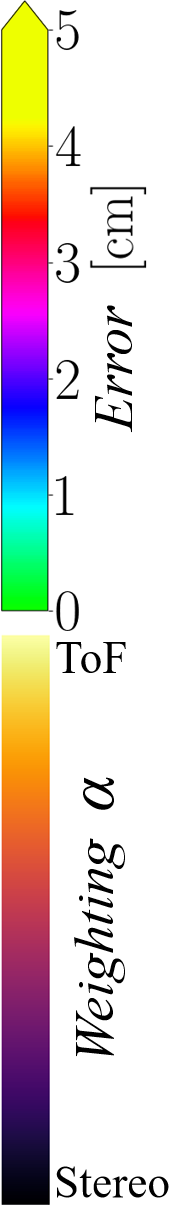}} \\
\rotatebox[origin=c]{90}{Office 4} & 
\includegraphics[align=c, width=\sz\linewidth]{figures/replica/tof_psmnet/office_4/model_small.jpg} & 
\includegraphics[align=c, width=\sz\linewidth]{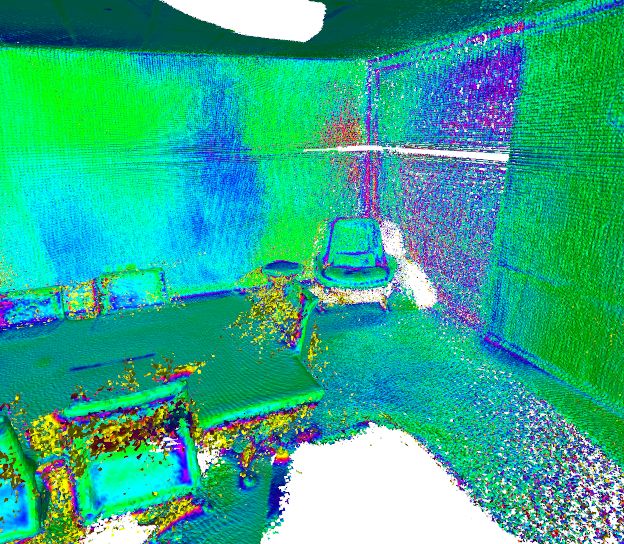} & 
\includegraphics[align=c, width=\sz\linewidth]{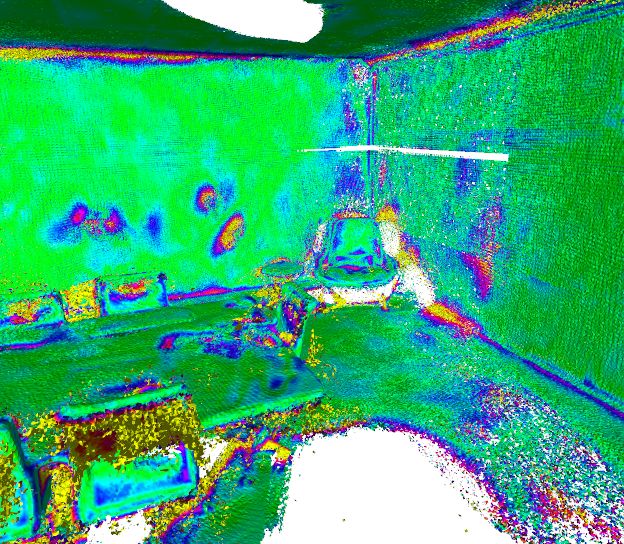} & 
\includegraphics[align=c, width=\sz\linewidth]{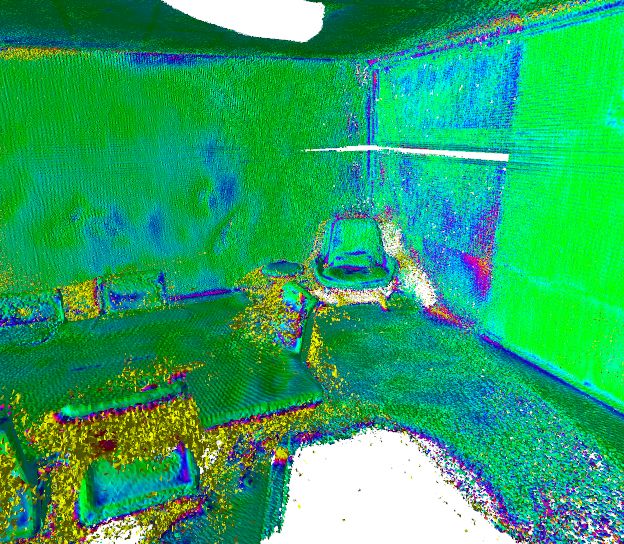} & 
\includegraphics[align=c, width=\sz\linewidth]{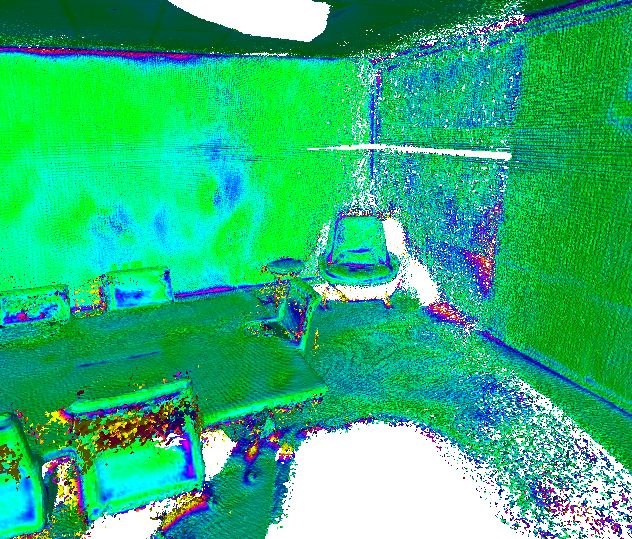} & 
\includegraphics[align=c, width=\sz\linewidth]{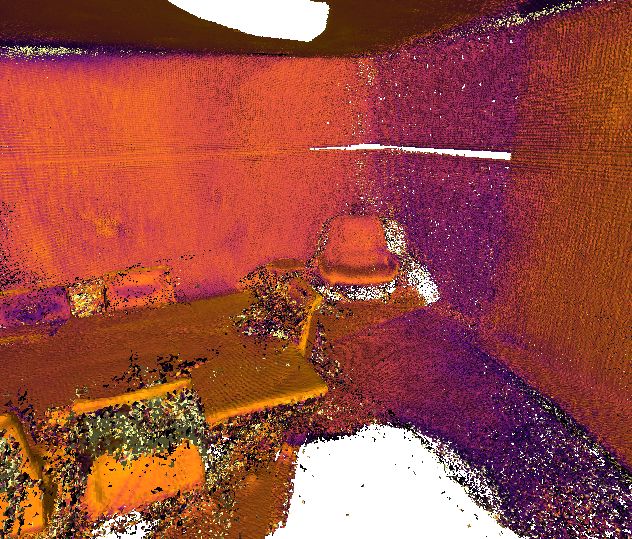} & \\
 & Model & ToF~\cite{handa2014benchmark} & PSMNet~\cite{chang2018pyramid} & TSDF Fusion~\cite{curless1996volumetric} & \ours{} (Ours) & Our sensor weight & \\
\end{tabular}
}
\caption{\textbf{Replica Dataset.} Our method fuses the sensors consistently better than the baselines.
Concretely, our method learns to detect and remove outliers much more effectively (best viewed on screen). \textbf{Top row:} ToF\bplus{}PSMNet Fusion without denoising. 
See also Tab.~\ref{tab:tof_psmnet_no_denoising}.
\textbf{Bottom row:} ToF\bplus{}PSMNet Fusion with denoising. See also Tab.~\ref{tab:tof_psmnet_denoising}.}
\label{fig:tof_psmnet}
\end{figure*}
\else
\begin{figure*}[t]
\centering
{\footnotesize
\setlength{\tabcolsep}{1pt}
\renewcommand{\arraystretch}{1}
\newcommand{\sz}{0.153}
\begin{tabular}{ccccccccc}
\rotatebox[origin=c]{90}{Office 4} & 
\includegraphics[align=c, width=\sz\linewidth]{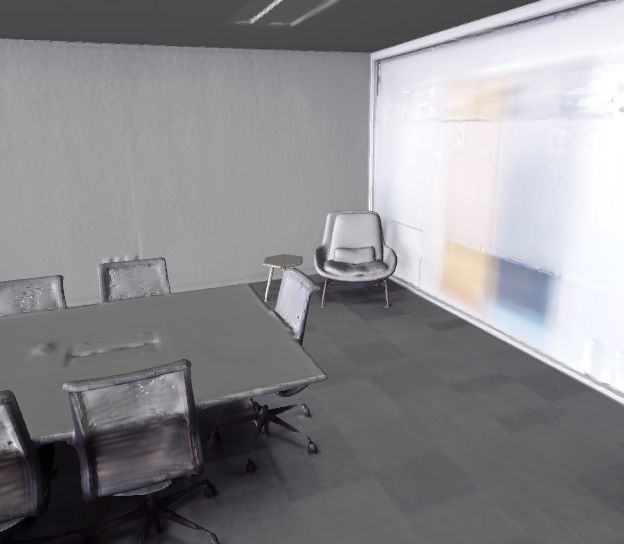} & 
\includegraphics[align=c, width=\sz\linewidth]{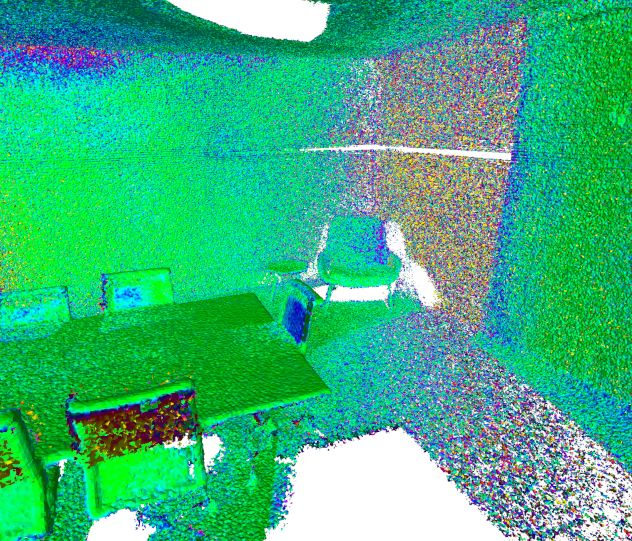} &
\includegraphics[align=c, width=\sz\linewidth]{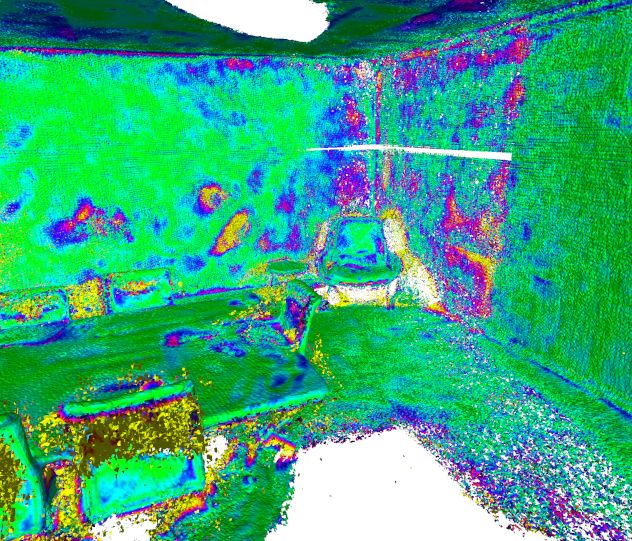} &
\includegraphics[align=c, width=\sz\linewidth]{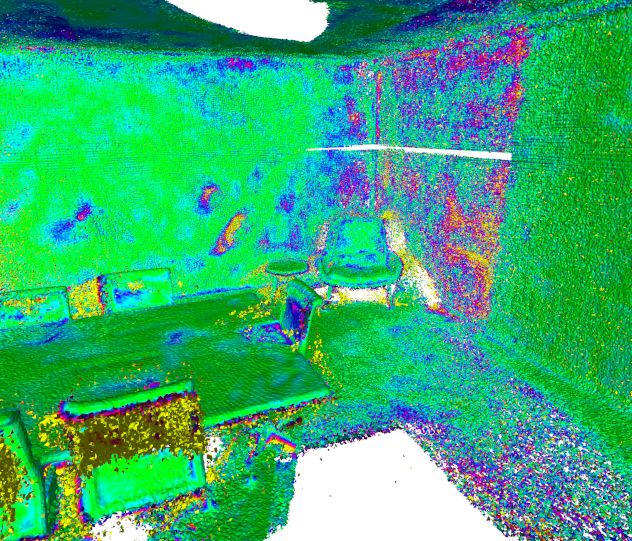} &
\includegraphics[align=c, width=\sz\linewidth]{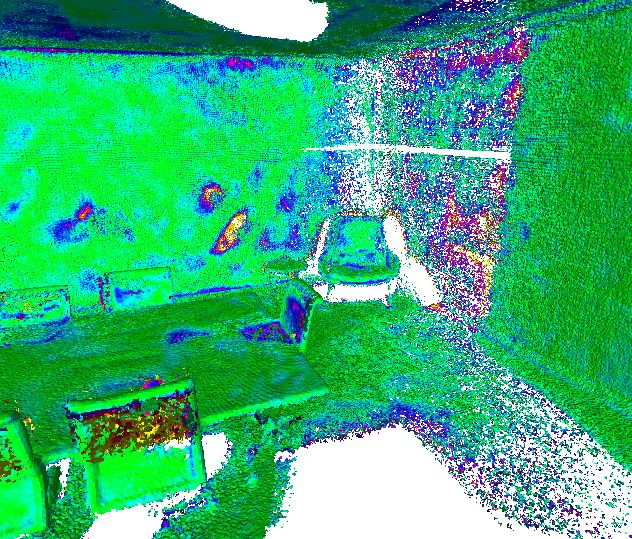} & 
\includegraphics[align=c, width=\sz\linewidth]{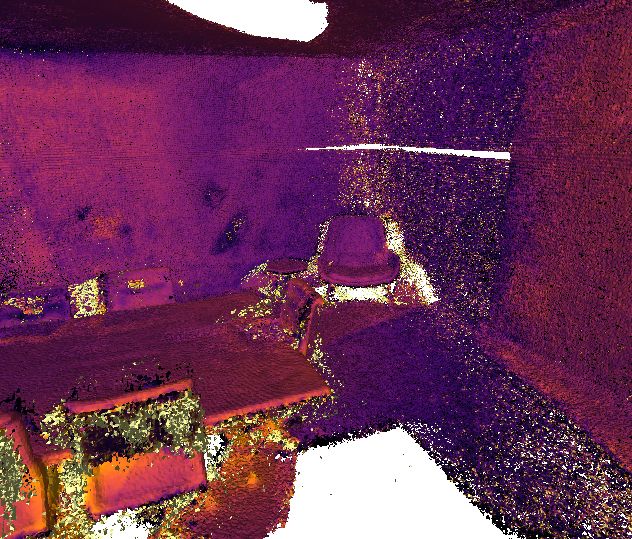} & \multirow{12}{*}[29.0pt]{\includegraphics[width=.0372\linewidth]{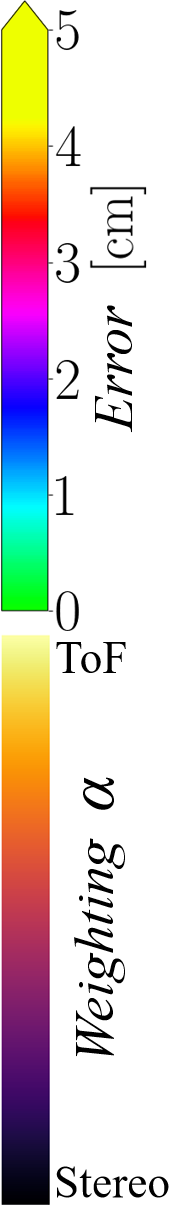}} \\
\rotatebox[origin=c]{90}{Office 4} & 
\includegraphics[align=c, width=\sz\linewidth]{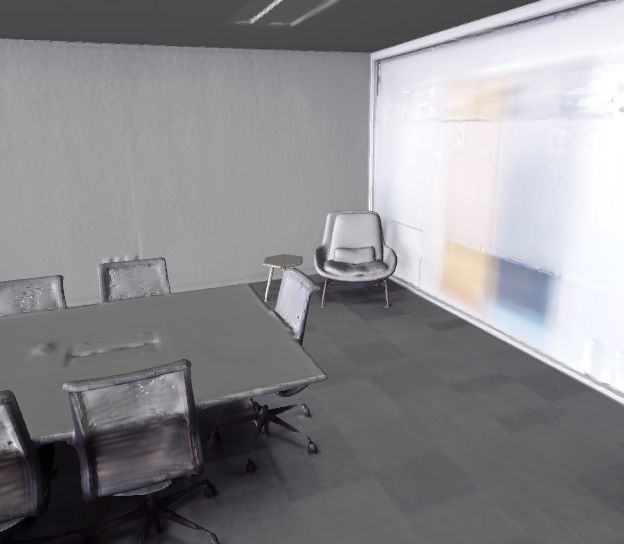} & 
\includegraphics[align=c, width=\sz\linewidth]{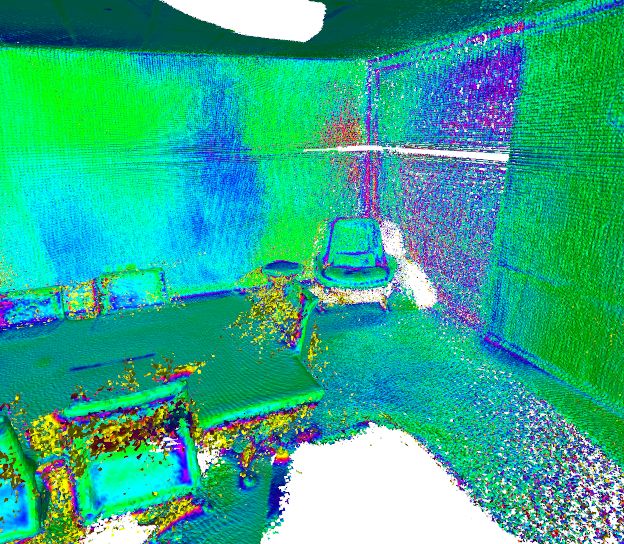} & 
\includegraphics[align=c, width=\sz\linewidth]{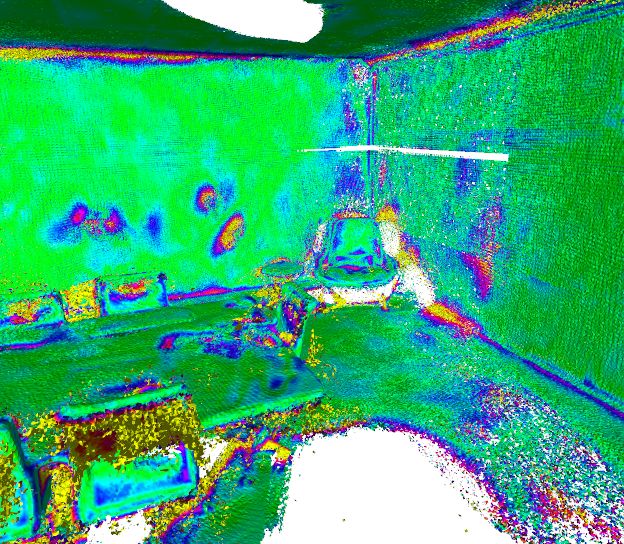} & 
\includegraphics[align=c, width=\sz\linewidth]{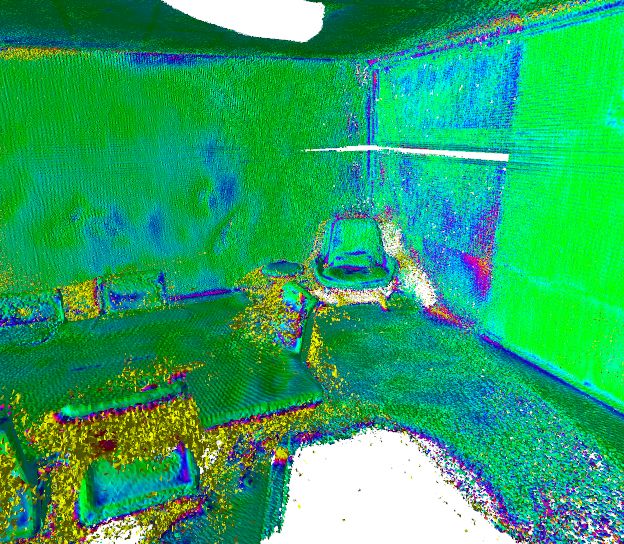} & 
\includegraphics[align=c, width=\sz\linewidth]{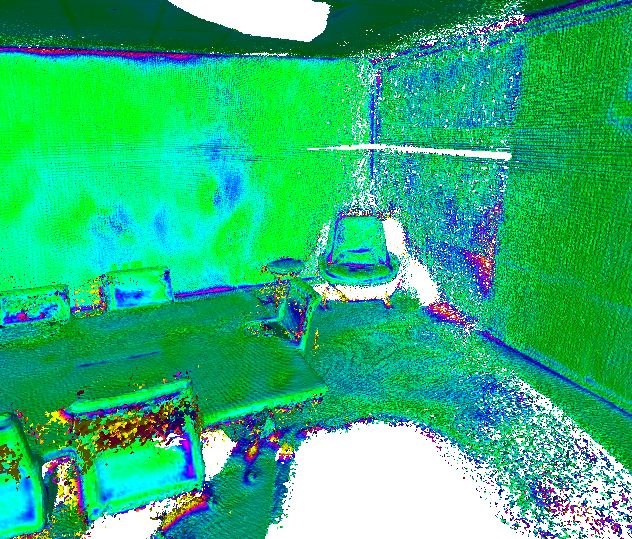} & 
\includegraphics[align=c, width=\sz\linewidth]{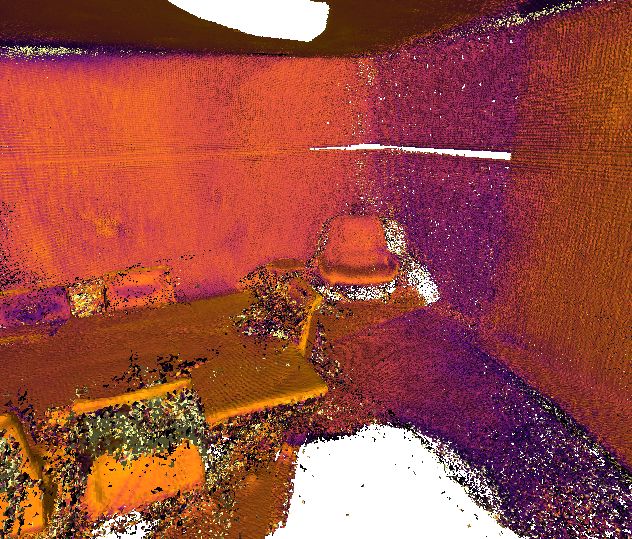} & \\
 & Model & ToF~\cite{handa2014benchmark} & PSMNet~\cite{chang2018pyramid} & TSDF Fusion~\cite{curless1996volumetric} & \ours{} (Ours) & Our sensor weighting & \\
\end{tabular}
}
\caption{\textbf{Replica Dataset. Top row: ToF\bplus{}PSMNet Fusion without denoising.} Our method fuses the sensors consistently better than the best baseline method.
In particular, our method learns to detect and remove outliers much more effectively.
See also Tab.~\ref{tab:tof_psmnet_no_denoising}.
\textbf{Bottom row: ToF\bplus{}PSMNet Fusion with denoising.} Our method fuses the sensors consistently better than TSDF Fusion.
In particular, our method learns to detect and remove outliers much more effectively (best viewed on screen). See also Tab.~\ref{tab:tof_psmnet_denoising}.}
\label{fig:tof_psmnet}
\end{figure*}
\fi

\boldparagraph{ToF\bplus{}PSMNet Fusion.} We simulate a ToF sensor by adding realistic noise to the ground truth depth maps\footnote{\label{footnote:tof}\normalfont\url{http://redwood-data.org/indoor/dataset.html}}~\cite{handa2014benchmark}.
To balance the two sensors, we increase the noise level by a factor $5$ compared to the original implementation.
We simulate another depth sensor from the RGB stereo pair using PSMNet~\cite{chang2018pyramid}.
We train the network on the Replica train set and keep it fixed while training our pipeline.
Tab.~\ref{tab:tof_psmnet_no_denoising} shows that our method outperforms both TSDF Fusion, RoutedFusion and DI-Fusion on all metrics except Recall with at least 13 $\%$ on the F-score.
Additionally, the F-score improves with a minimum of 18 $\%$ compared to the input sensors.
Specifically, note the absence of outliers (colored yellow) in Fig.~\ref{fig:tof_psmnet} \emph{Top row} when comparing our method to TSDF Fusion. 
Also note the sensor weighting \eg~we find lots of noise on the right wall of the ToF scene and thus, our method puts more weight on the stereo sensor in this region. 

Weder~\etal~\cite{Weder2020RoutedFusionLR} showed that a 2D denoising network (called routing network in the paper) that preprocesses the depth maps can improve performance when noise is present in planar regions. 
To this end, we train our own denoising network on the Replica train set and train a new model which applies a fixed denoising network.
According to Tab.~\ref{tab:tof_psmnet_denoising}, this yields a gain of 10 $\%$ on the F-score of the fused model compared to without using a denoising network, see also Fig.~\ref{fig:tof_psmnet} \emph{Bottom row}. 
Early Fusion is a strong alternative to our method when the sensors are synchronized. 
We want to highlight, however, that the resource overhead of our method is worthwhile since we outperform Early Fusion even in the synchronized setting.

\ifeccv
\else
\begin{table}[tb]
\centering
\resizebox{\columnwidth}{!}
{
\begin{tabular}{l|lllllll}
\cellcolor{gray}      & \cellcolor{gray}MSE$\downarrow$      & \cellcolor{gray}MAD$\downarrow$  & \cellcolor{gray}IoU$\uparrow$     & \cellcolor{gray}Acc.$\uparrow$ & \cellcolor{gray}F$\uparrow$ & \cellcolor{gray}P$\uparrow$ & \cellcolor{gray}R$\uparrow$   \\ 
\multirow{-2}{*}{\cellcolor{gray} \backslashbox[28.3mm]{Model}{Metric}}& \cellcolor{gray}*e-04 & \cellcolor{gray}*e-02 & \cellcolor{gray}[0,1] & \cellcolor{gray}$[\%]$ & \cellcolor{gray}$[\%]$ & \cellcolor{gray}$[\%]$ & \cellcolor{gray}$[\%]$ \\\hline
\multicolumn{8}{c}{\emph{Single Sensor}} \\ \hline
PSMNet~\cite{chang2018pyramid}   & 6.35 & 1.77 & 0.673 & 84.54 & 60.28 & 48.26 & 80.41 \\ 
ToF~\cite{handa2014benchmark} & 5.08 & 1.58 & 0.709 & 87.32 & 68.93 & 59.01  & 83.08 \\ 
\hline
\multicolumn{8}{c}{\emph{Multi-Sensor Fusion}} \\ \hline
TSDF Fusion~\cite{curless1996volumetric} & 6.40 & 1.80 & 0.681 & 85.31 & 52.93 & 38.95 & 84.60  \\ 
RoutedFusion~\cite{Weder2020RoutedFusionLR} & 6.04 & 1.68 & 0.644 & 85.10 & 62.67 & 51.75 & 79.52 \\ 
Early Fusion & 6.40 & 1.40 & 0.760 & 89.02 & 74.60 & 67.46 & 83.47 \\ 
DI-Fusion~\cite{huang2021di} $\sigma$=0.15 & - & - & - & - & 55.66 & 41.49 & \textbf{85.33} \\ 
\textbf{\ours{} (Ours)} & \textbf{3.49}  & \textbf{1.31}  & \textbf{0.761}  & \textbf{89.61}  & \textbf{76.47} & \textbf{73.58} & 79.77 \\ 
\end{tabular}
}
\caption{\textbf{Replica Dataset. ToF\bplus{}PSMNet Fusion with denoising.} The denoising network mitigates outliers along planar regions, compare to Tab.~\ref{tab:tof_psmnet_no_denoising}. Our method even outperforms the Early Fusion baseline, which assumes synchronized sensors.}
\label{tab:tof_psmnet_denoising}
\end{table}
\fi

\ifeccv
\begin{wraptable}[8]{R}{0.5\linewidth}
\vspace{-\intextsep}
\centering
\renewcommand{\arraystretch}{1.05}
\resizebox{\linewidth}{!}
{
\begin{tabular}{l|lllllll}
\cellcolor{gray}        & \cellcolor{gray}MSE$\downarrow$      & \cellcolor{gray}MAD$\downarrow$  & \cellcolor{gray}IoU$\uparrow$     & \cellcolor{gray}Acc.$\uparrow$ & \cellcolor{gray}F$\uparrow$ & \cellcolor{gray}P$\uparrow$ & \cellcolor{gray}R$\uparrow$   \\ 
\multirow{-2}{*}{\cellcolor{gray} \backslashbox[28mm]{Model}{Metric}} & \cellcolor{gray}*e-04 & \cellcolor{gray}*e-02 & \cellcolor{gray}[0,1] & \cellcolor{gray}$[\%]$ & \cellcolor{gray}$\%$ & \cellcolor{gray}$[\%]$ & \cellcolor{gray}$[\%]$ \\\hline
Early Fusion  & 7.66 & 1.99 & 0.642 & 84.65 & 61.34 & 48.47 & \textbf{83.63}  \\ 
SenFuNet (Ours) & 4.21 & 1.45 & 0.755 & 88.26 & 73.04 & 69.13 & 78.43 \\ 
SenFuNet (Ours)* & \textbf{3.15} & \textbf{1.23} & \textbf{0.760} & \textbf{89.52} & \textbf{79.26} & \textbf{79.91} & 78.79  \\ 
\end{tabular}
}
\caption{\textbf{Time Asynchronous Evaluation.} SenFuNet outperforms Early Fusion for sensors with different sampling frequencies. *With depth denoising.} 
\label{tab:synchronization}
\end{wraptable}

\boldparagraph{Time Asynchronous Evaluation.} 
RGB cameras often have higher frame rates than ToF sensors which makes Early Fusion more challenging as one sensor might lack new data.
We simulate this setting by giving the PSMNet sensor twice the sampling rate of the ToF sensor, \ie~we drop every second ToF frame.
To provide a corresponding ToF frame for Early Fusion, we reproject the latest observed ToF frame into the current view of the PSMNet sensor.
As demonstrated in Tab.~\ref{tab:synchronization} the gap between our SenFuNet late fusion approach and Early Fusion becomes even larger (cf. Tab.~\ref{tab:tof_psmnet} (b)). 
\fi

\ifeccv
\begin{table}[tb]
\setlength{\belowcaptionskip}{0pt}
\begin{subtable}{.495\columnwidth}
\centering
\resizebox{\columnwidth}{!}
{
\setlength{\tabcolsep}{2pt}
\renewcommand{\arraystretch}{1.05}
\begin{tabular}{l|lllllll}
\cellcolor{gray}       & \cellcolor{gray}MSE$\downarrow$      & \cellcolor{gray}MAD$\downarrow$  & \cellcolor{gray}IoU$\uparrow$     & \cellcolor{gray}Acc.$\uparrow$ & \cellcolor{gray}F$\uparrow$ & \cellcolor{gray}P$\uparrow$ & \cellcolor{gray}R$\uparrow$ \\
\multirow{-2}{*}{\cellcolor{gray} \backslashbox[28.3mm]{Model}{Metric}} & \cellcolor{gray}*e-04 & \cellcolor{gray}*e-02 & \cellcolor{gray}[0,1] & \cellcolor{gray}$[\%]$ & \cellcolor{gray}$[\%]$ & \cellcolor{gray}$[\%]$ & \cellcolor{gray}$[\%]$ \\\hline
\multicolumn{8}{c}{\emph{Single Sensor}} \\ \hline
PSMNet~\cite{chang2018pyramid} & 7.30 & 1.95 & 0.664 & 83.23 & 56.20 & 43.10 & 81.34 \\ 
SGM~\cite{hirschmuller2007stereo} & 8.90 & 2.17 & 0.610 & 81.48 & 57.71 & 44.40 & 84.08 \\ 
\hline
\multicolumn{8}{c}{\emph{Multi-Sensor Fusion}} \\ \hline
TSDF Fusion~\cite{curless1996volumetric} & 9.17 & 2.24 & 0.634 & 82.62 & 47.75 & 33.39 & 85.10 \\ 
RoutedFusion~\cite{Weder2020RoutedFusionLR} & 7.11 & 1.82 & 0.671 & 84.63 & 60.31 & 48.47 & 80.21 \\ 
DI-Fusion~\cite{huang2021di} $\sigma$=0.15 & - & - & - & - & 47.29 & 32.92 & \textbf{85.14} \\ 
\ours{} (Ours) &  \textbf{4.77} & \textbf{1.56}& \textbf{0.738} & \textbf{87.62} & \textbf{69.83} & \textbf{63.20} & 79.12 \\ 
\end{tabular}
}
\subcaption{Without depth denoising}
\label{tab:sgm_psmnet_no_denoising}
\end{subtable}
\begin{subtable}{0.495\columnwidth}
\resizebox{\columnwidth}{!}
{
\setlength{\tabcolsep}{2pt}
\renewcommand{\arraystretch}{1.05}
\begin{tabular}{l|lllllll}
\cellcolor{gray}      & \cellcolor{gray}MSE$\downarrow$      & \cellcolor{gray}MAD$\downarrow$  & \cellcolor{gray}IoU$\uparrow$     & \cellcolor{gray}Acc.$\uparrow$ & \cellcolor{gray}F$\uparrow$ & \cellcolor{gray}P$\uparrow$ & \cellcolor{gray}R$\uparrow$   \\ 
\multirow{-2}{*}{\cellcolor{gray} \backslashbox[28.3mm]{Model}{Metric}}& \cellcolor{gray}*e-04 & \cellcolor{gray}*e-02 & \cellcolor{gray}[0,1] & \cellcolor{gray}$[\%]$ & \cellcolor{gray}$[\%]$ & \cellcolor{gray}$[\%]$ & \cellcolor{gray}$[\%]$ \\\hline
\multicolumn{8}{c}{\emph{Single Sensor}} \\ \hline
PSMNet~\cite{chang2018pyramid} & 6.35 & 1.77 & 0.673 & 84.54 & 60.28 & 48.26 & 80.41 \\ 
SGM~\cite{hirschmuller2007stereo} & 6.60 & 1.80 & 0.659 & 84.29 & 60.79 & 49.78 & 78.13 \\ 
\hline
\multicolumn{8}{c}{\emph{Multi-Sensor Fusion}} \\ \hline
TSDF Fusion~\cite{curless1996volumetric} & 7.28 & 1.93 & 0.669 & 84.74 & 54.09 & 40.45 & 81.65 \\ 
RoutedFusion~\cite{Weder2020RoutedFusionLR}   & 8.09 & 2.05 & 0.580 & 80.18 & 59.88 & 47.13 & 82.12 \\ 
Early Fusion                                                        & 4.99 & 1.51 & 0.707 & 86.99 & 69.40 & 61.07 & 80.36 \\ 
DI-Fusion~\cite{huang2021di} $\sigma$=0.15 & - & - & - & - & 52.65 & 38.50 & \textbf{83.62} \\ 
\ours{} (Ours)  &  \textbf{4.04} & \textbf{1.41}  & \textbf{0.737} & \textbf{88.11} & \textbf{71.18} & \textbf{66.81} & 76.27 \\ 
\end{tabular}
}
\subcaption{With depth denoising}
\label{tab:sgm_psmnet_denoising}
\end{subtable}
\caption{\textbf{Replica Dataset. SGM\bplus{}PSMNet Fusion.} Our method does not assume a particular sensor pairing and works well for all tested sensors. The gain from the denoising network is marginal with a 2$\%$ F-score improvement since there are few outliers on planar regions of the stereo depth maps and the denoising network over-smooths the depth along discontinuities.
Our method outperforms Early Fusion (which generally assumes synchronized sensors) on most metrics even without depth denoising.
}
\label{tab:sgm_psmnet}
\end{table}
\fi

\boldparagraph{SGM\bplus{}PSMNet Fusion.} Our method does not assume a particular sensor pairing. 
We show this, by replacing the ToF sensor with a stereo sensor acquired using semi-global matching \cite{hirschmuller2007stereo}. 
In Tab.~\ref{tab:sgm_psmnet}, we show state-of-the-art sensor fusion performance both with and without a denoising network.
The denoising network tends to over-smooth depth discontinuities, which negatively affects performance when few outliers exist. Additionally, even without using a denoising network, we outperform Early Fusion on most metrics. TSDF Fusion, RoutedFusion and DI-Fusion aggregate outliers across the sensors leading to worse performance than the single sensor results.

\ifeccv
\else
\begin{table}[tb]
\centering
\resizebox{\columnwidth}{!}
{
\setlength{\tabcolsep}{3pt}
\renewcommand{\arraystretch}{1.05}
\begin{tabular}{l|lllllll}

\cellcolor{gray} & \cellcolor{gray}MSE$\downarrow$      & \cellcolor{gray}MAD$\downarrow$  & \cellcolor{gray}IoU$\uparrow$     & \cellcolor{gray}Acc.$\uparrow$ &\cellcolor{gray} F$\uparrow$ &\cellcolor{gray} P$\uparrow$ & \cellcolor{gray}R$\uparrow$   \\ 
\multirow{-2}{*}{\cellcolor{gray} \backslashbox[45mm]{Model}{Metric}} & \cellcolor{gray}*e-04 & \cellcolor{gray}*e-02 & \cellcolor{gray}[0,1] &\cellcolor{gray} $[\%]$ & \cellcolor{gray}$[\%]$ & \cellcolor{gray}$[\%]$ & \cellcolor{gray}$[\%]$ \\\hline
\multicolumn{8}{c}{\emph{Single Sensor}} \\ \hline
PSMNet~\cite{chang2018pyramid} w.\phantom{o} denoising                 & 6.35 & 1.77 & 0.673 & 84.54 & 60.28 & 48.26 & 80.41 \\ 
PSMNet~\cite{chang2018pyramid} w/o denoising                          & 7.30 & 1.95 & 0.664 & 83.23 & 56.20 & 43.10 & 81.34 \\ 
\hline
SGM~\cite{hirschmuller2007stereo} w.\phantom{o} denoising                      & 6.60 & 1.80 & 0.659 & 84.29 & 60.79 & 49.78 & 78.13 \\ 
SGM~\cite{hirschmuller2007stereo} w/o denoising                               & 8.90 & 2.17 & 0.610 & 81.48 & 57.71 & 44.40 & 84.08 \\ 
\hline
\multicolumn{8}{c}{\emph{Multi-Sensor Fusion}} \\ \hline
TSDF Fusion~\cite{curless1996volumetric} w.\phantom{o} denoising      & 7.28 & 1.93 & 0.669 & 84.74 & 54.09 & 40.45 & 81.65 \\ 
TSDF Fusion~\cite{curless1996volumetric} w/o denoising                & 9.17 & 2.24 & 0.634 & 82.62 & 47.75 & 33.39 & 85.10 \\ 
RoutedFusion~\cite{Weder2020RoutedFusionLR} w.\phantom{o} denoising   & 8.09 & 2.05 & 0.580 & 80.18 & 59.88 & 47.13 & 82.12 \\ 
RoutedFusion~\cite{Weder2020RoutedFusionLR} w/o denoising             & 7.11 & 1.82 & 0.671 & 84.63 & 60.31 & 48.47 & 80.21 \\ 
Early Fusion                                                        & 4.99 & 1.51 & 0.707 & 86.99 & 69.40 & 61.07 & 80.36 \\ 
DI-Fusion~\cite{huang2021di} w.\phantom{o} denoising $\sigma$=0.15 & - & - & - & - & 52.65 & 38.50 & 83.62 \\ 
DI-Fusion~\cite{huang2021di} w/o denoising $\sigma$=0.15 & - & - & - & - & 47.29 & 32.92 & \textbf{85.14} \\ 
\hline
\textbf{\ours{} (Ours) w.\phantom{o} denoising}  &  \textbf{4.04} & \textbf{1.41}  & \textbf{0.737} & \textbf{88.11} & \textbf{71.18} & \textbf{66.81} & 76.27 \\ 
%
\textcolor{Cerulean}{\ours{} (Ours) w/o denoising} &  \textcolor{Cerulean}{4.77} & \textcolor{Cerulean}{1.56} & \textcolor{Cerulean}{0.738} & \textcolor{Cerulean}{87.62} & \textcolor{Cerulean}{69.83} & \textcolor{Cerulean}{63.20} & \textcolor{Cerulean}{79.12} \\ 
\end{tabular}
}
\caption{\textbf{Replica Dataset. SGM\bplus{}PSMNet Fusion.} Our method does not assume a particular sensor pairing and works well for all tested sensors.
The gain from the denoising network is marginal with a 2$\%$ F-score improvement since there are few outliers on planar regions of the stereo depth maps and the denoising network over-smooths the depth along discontinuities.
Note that our method outperforms Early Fusion, which assumes synchronized sensors, on most metrics even without denoising.}
\label{tab:sgm_psmnet}
\end{table}
\fi

\ifeccv
\else
\begin{figure*}[t]
\centering
\scriptsize
\setlength{\tabcolsep}{1pt}
\renewcommand{\arraystretch}{1}
\begin{tabular}{ccccccccc}
& Frame & ToF & MVS\cite{schonberger2016pixelwise} & TSDF Fusion\cite{curless1996volumetric} & SenFuNet (Ours) & Our sensor weighting & &\\
\rotatebox[origin=c]{90}{Copy room} & 
\includegraphics[align=c, width=.153\linewidth]{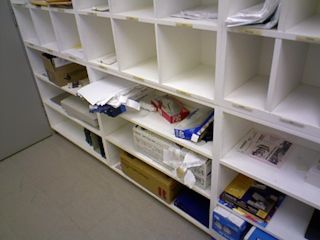} & 
\includegraphics[align=c, width=.153\linewidth]{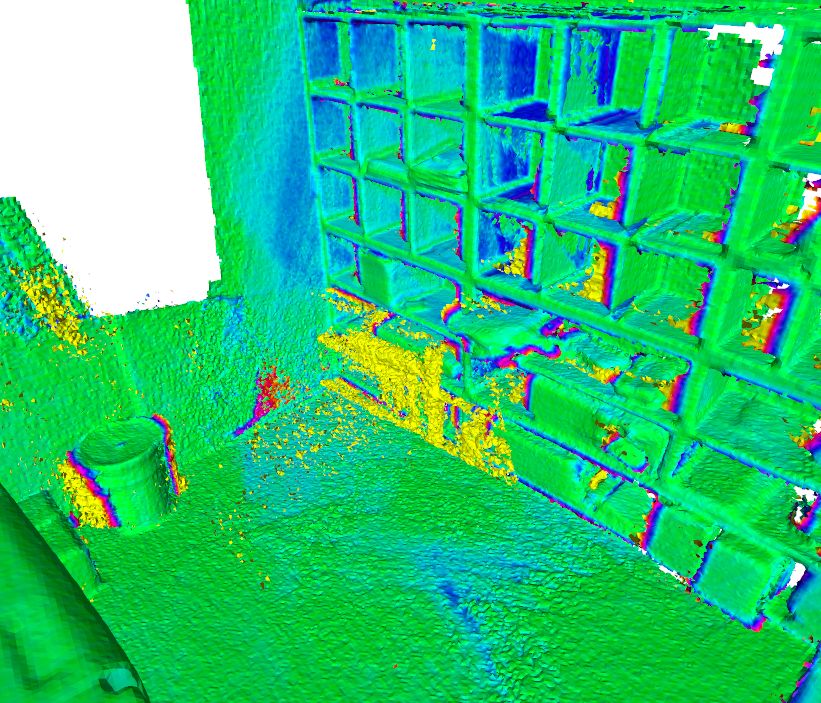} & 
\includegraphics[align=c, width=.153\linewidth]{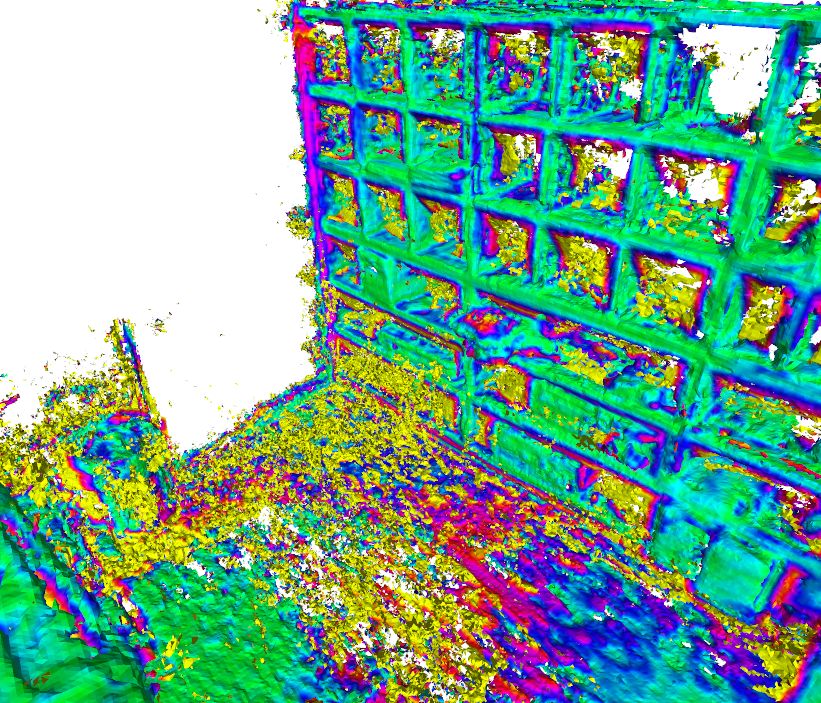} & 
\includegraphics[align=c, width=.153\linewidth]{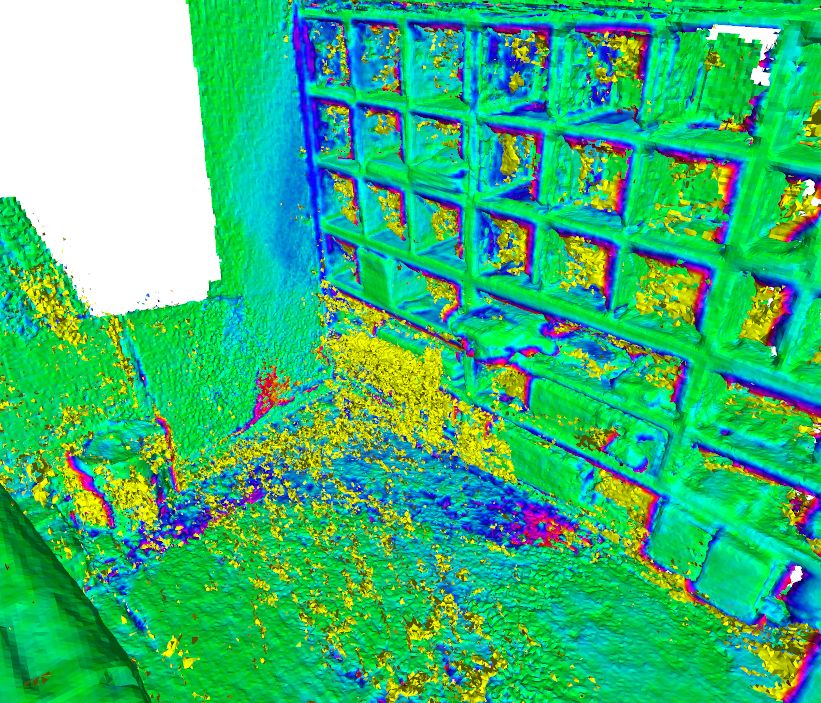} & 
\includegraphics[align=c, width=.153\linewidth]{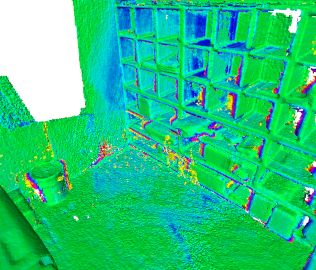} & 
\includegraphics[align=c, width=.153\linewidth]{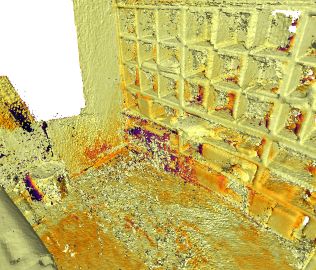} & \multirow{11}{*}[34pt]{\includegraphics[width=.038\linewidth]{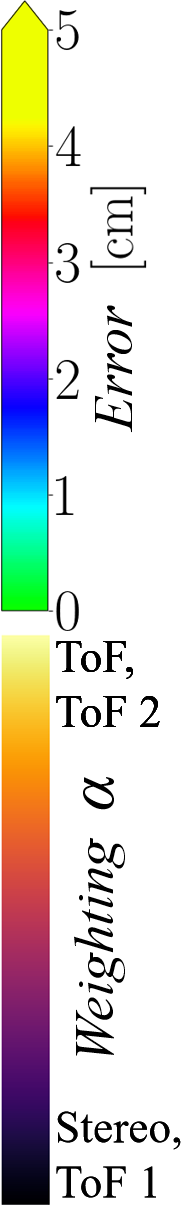}} \\
 \rotatebox[origin=c]{90}{Copy room} &
 \includegraphics[align=c, width=.153\linewidth]{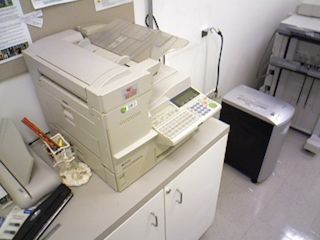} & 
 \includegraphics[trim={0cm 0cm 0cm 0cm}, clip=true, align=c, width=.153\linewidth]{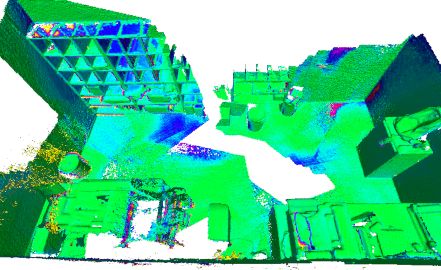} & 
 \includegraphics[trim={0cm 0cm 0cm 0cm}, clip=true,align=c, width=.153\linewidth]{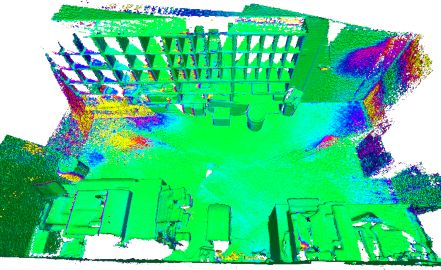} & 
 \includegraphics[trim={0cm 0cm 0cm 0cm}, clip=true,align=c, width=.153\linewidth]{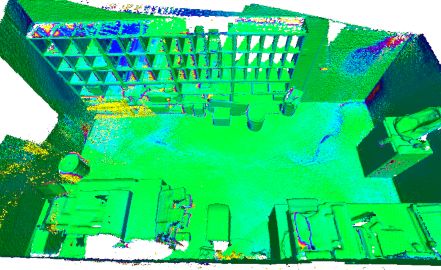}& 
 \includegraphics[trim={0cm 0cm 0cm 0cm}, clip=true,align=c, width=.153\linewidth]{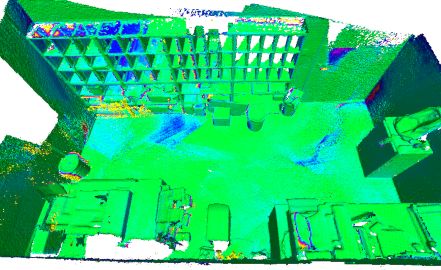} & 
 \includegraphics[trim={0cm 0cm 0cm 0cm}, clip=true,align=c, width=.153\linewidth]{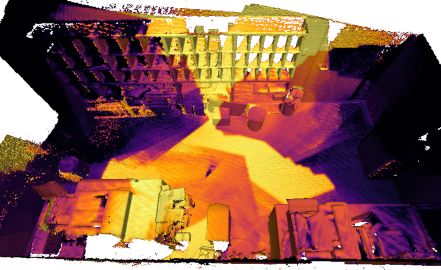}  &  \\
 & Frame & ToF 1 & ToF 2 & TSDF Fusion\cite{curless1996volumetric} & SenFuNet (Ours) & Our sensor weighting & &\\
\end{tabular}
\caption{\textbf{Scene3D Dataset. Top row: ToF\bplus{}MVS Fusion.} Our method effectively fuses the ToF and MVS sensors.
Note specifically the absence of yellow outliers in the corner of the bookshelf.
See also Tab.~\ref{tab:scene3d_tof_mvs}.
\textbf{Bottom row: Multi-Agent ToF Reconstruction.} Our method is flexible and can perform Multi-Agent reconstruction. 
Note that our model learns where to trust each agent at different spatial locations for maximum completeness while also being noise aware. See for instance the left bottom corner of the bookshelf where both agents integrate, but the noise-free agent is given a higher weighting (best viewed on screen).
See also Tab.~\ref{tab:scene3d_collab}.}
\label{fig:scene3d}
\end{figure*}
\fi

\ifeccv

\else
\begin{figure}[htb]
\centering
\resizebox{\columnwidth}{!}
{
\setlength{\tabcolsep}{6pt}
\renewcommand{\arraystretch}{1.1}
\begin{tabular}{l|lllllll}
\cellcolor{gray}        & \cellcolor{gray}MSE$\downarrow$      & \cellcolor{gray}MAD$\downarrow$  & \cellcolor{gray}IoU$\uparrow$     & \cellcolor{gray}Acc.$\uparrow$ & \cellcolor{gray}F$\uparrow$ & \cellcolor{gray}P$\uparrow$ & \cellcolor{gray}R$\uparrow$   \\
\multirow{-2}{*}{\cellcolor{gray} \backslashbox[28mm]{Model}{Metric}}& \cellcolor{gray}*e-04 & \cellcolor{gray}*e-02 & \cellcolor{gray}[0,1] & \cellcolor{gray}$[\%]$ & \cellcolor{gray}$[\%]$ & \cellcolor{gray}$[\%]$ & \cellcolor{gray}$[\%]$ \\\hline
\multicolumn{8}{c}{\emph{Single Sensor}} \\ \hline
MVS~\cite{schonberger2016pixelwise}   & 15.39 & 3.25 & 0.263 & 74.84 & 17.73 & 9.77  & 95.96 \\ 
ToF    & 7.11 & \textbf{1.52} & 0.486  & 83.13 & 72.98 & 57.85 & 98.82 \\ 
\hline
\multicolumn{8}{c}{\emph{Multi-Sensor Fusion}} \\ \hline
TSDF Fusion~\cite{curless1996volumetric}  & 15.09 & 3.15 & 0.290 & 75.96 & 18.13 & 9.98 & \textbf{99.19} \\ 
RoutedFusion~\cite{Weder2020RoutedFusionLR} & 23.09 & 4.40 & 0.222 & 49.42 & 2.36 & 1.27 & 16.43 \\ 
DI-Fusion~\cite{huang2021di} $\sigma$=0.15 & - & - & - & - & 28.19 & 16.71 & 90.15 \\ 
\textbf{\ours{} (Ours)}  & \textbf{6.53} & 1.55 & \textbf{0.510} & \textbf{84.88} & \textbf{74.56} & \textbf{59.74} & 99.16 \\ 
\end{tabular}
}
{\scriptsize
\setlength{\tabcolsep}{1pt}
\renewcommand{\arraystretch}{1}
\newcommand{\sz}{0.13}
\begin{tabular}{cccccccc}
\rotatebox[origin=c]{90}{Human} & 
\includegraphics[trim={0cm 0cm 0cm 0cm}, clip=true, align=c, width=0.115\linewidth]{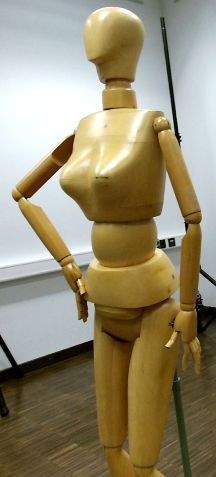} & 
\includegraphics[trim={2.5cm 0.5cm 2.5cm 0cm}, clip=true, align=c, width=\sz\linewidth]{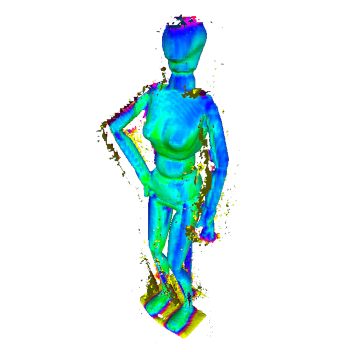} & 
\includegraphics[trim={2.5cm 0.5cm 2.5cm 0cm}, clip=true, align=c, width=\sz\linewidth]{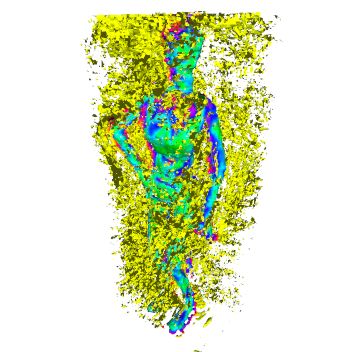} & 
\includegraphics[trim={2.5cm 0.5cm 2.5cm 0cm}, clip=true, align=c, width=\sz\linewidth]{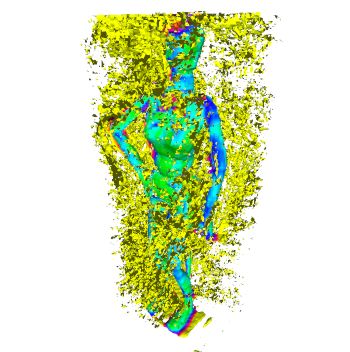} & 
\includegraphics[trim={2.5cm 0.5cm 2.5cm 0cm}, clip=true, align=c, width=\sz\linewidth]{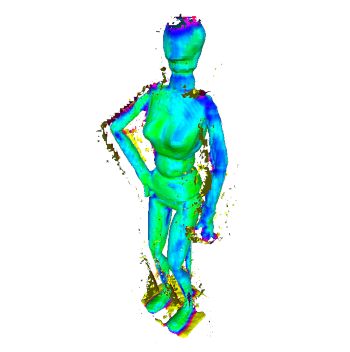} & 
\includegraphics[trim={2.5cm 0.5cm 2.5cm 0cm}, clip=true, align=c, width=\sz\linewidth]{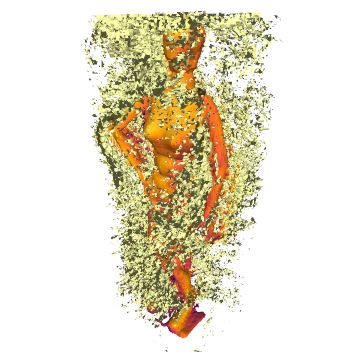} & 
\includegraphics[trim={0cm 0cm 0cm 0cm}, clip=true, align=c, width=\sz\linewidth]{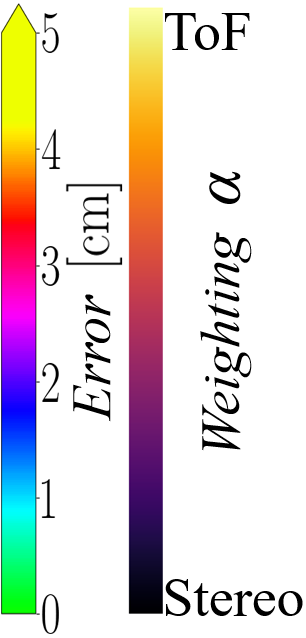}\\
 & Frame & ToF & MVS~\cite{schonberger2016pixelwise} & TSDF & \ours{} & Our sensor & \\
 & & & & Fusion~\cite{curless1996volumetric} & (Ours) & weighting & \\
\end{tabular}
}
\caption{\textbf{CoRBS Dataset. ToF\bplus{}MVS Fusion.} Our model can find synergies between very imbalanced sensors. The numerical results show that our fused model is better than the individual depth sensor inputs and significantly better than any of the baseline methods.
Contrary to our method, the baseline methods cannot handle the high degree of outliers present from the MVS sensor.}
\label{fig:corbs}
\end{figure}
\fi

\subsection{Experiments on the CoRBS Dataset}
%
The real-world CoRBS dataset~\cite{wasenmuller2016corbs} provides a selection of reconstructed objects with very accurate ground truth 3D and camera trajectories along with a consumer-grade RGBD camera.
We apply our method to the dataset by training a model on the desk scene and testing it on the human scene.
The procedure to create the ground truth signed distance grids is identical to the Replica dataset.
We create an additional depth sensor along the ToF depth using MVS with COLMAP~\cite{schonberger2016pixelwise}\footnote{Unfortunately, no suitable public real 3D dataset exists, which comprises binocular stereo pairs, and an active depth sensor, as well as ground truth geometry.}.
Fig.~\ref{fig:corbs} shows that our model can fuse very imbalanced sensors while the baseline methods fail severely. 
Even if one sensor (MVS) is significantly worse, our method still improves on most metrics. This confirms that our method learns to meaningfully fuse the sensors even if one sensor adds very little.

\ifeccv
\begin{figure}[b]
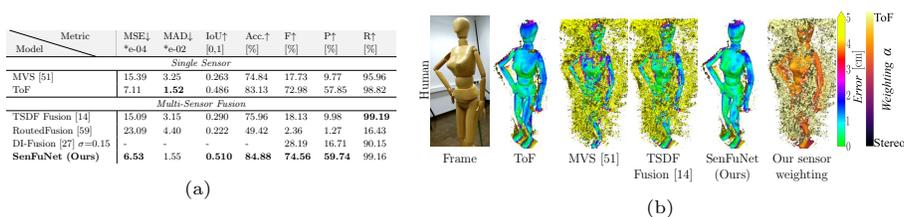

\setlength{\belowcaptionskip}{0pt}
\centering
\begin{subfigure}{0.44\columnwidth}
\resizebox{\columnwidth}{!}
{
\setlength{\tabcolsep}{6pt}
\renewcommand{\arraystretch}{1.1}
\begin{tabular}{l|lllllll}
\cellcolor{gray}        & \cellcolor{gray}MSE$\downarrow$      & \cellcolor{gray}MAD$\downarrow$  & \cellcolor{gray}IoU$\uparrow$     & \cellcolor{gray}Acc.$\uparrow$ & \cellcolor{gray}F$\uparrow$ & \cellcolor{gray}P$\uparrow$ & \cellcolor{gray}R$\uparrow$   \\
\multirow{-2}{*}{\cellcolor{gray} \backslashbox[28mm]{Model}{Metric}}& \cellcolor{gray}*e-04 & \cellcolor{gray}*e-02 & \cellcolor{gray}[0,1] & \cellcolor{gray}$[\%]$ & \cellcolor{gray}$[\%]$ & \cellcolor{gray}$[\%]$ & \cellcolor{gray}$[\%]$ \\\hline
\multicolumn{8}{c}{\emph{Single Sensor}} \\ \hline
MVS~\cite{schonberger2016pixelwise}   & 15.39 & 3.25 & 0.263 & 74.84 & 17.73 & 9.77  & 95.96 \\ 
ToF    & 7.11 & \textbf{1.52} & 0.486  & 83.13 & 72.98 & 57.85 & 98.82 \\ 
\hline
\multicolumn{8}{c}{\emph{Multi-Sensor Fusion}} \\ \hline
TSDF Fusion~\cite{curless1996volumetric}  & 15.09 & 3.15 & 0.290 & 75.96 & 18.13 & 9.98 & \textbf{99.19} \\ 
RoutedFusion~\cite{Weder2020RoutedFusionLR} & 23.09 & 4.40 & 0.222 & 49.42 & 2.36 & 1.27 & 16.43 \\ 
DI-Fusion~\cite{huang2021di} $\sigma$=0.15 & - & - & - & - & 28.19 & 16.71 & 90.15 \\ 
\textbf{\ours{} (Ours)}  & \textbf{6.53} & 1.55 & \textbf{0.510} & \textbf{84.88} & \textbf{74.56} & \textbf{59.74} & 99.16 \\ 
\end{tabular}
}
\subcaption{}
\label{fig:corbs_metric}
\end{subfigure}
\begin{subfigure}{0.55\columnwidth}
\resizebox{\columnwidth}{!}
{
\setlength{\tabcolsep}{1pt}
\renewcommand{\arraystretch}{1}
\newcommand{\sz}{0.4}
{\Large
\begin{tabular}{cccccccc}
\rotatebox[origin=c]{90}{Human} & 
\includegraphics[trim={0cm 0cm 0cm 0cm}, clip=true, align=c, width=0.35\linewidth]{figures/corbs/model_cropped.jpg} & 
\includegraphics[trim={2.5cm 0.5cm 2.5cm 0cm}, clip=true, align=c, width=\sz\linewidth]{figures/corbs/tof_small.jpg} & 
\includegraphics[trim={2.5cm 0.5cm 2.5cm 0cm}, clip=true, align=c, width=\sz\linewidth]{figures/corbs/mvs_small.jpg} & 
\includegraphics[trim={2.5cm 0.5cm 2.5cm 0cm}, clip=true, align=c, width=\sz\linewidth]{figures/corbs/tsdf_small.jpg} & 
\includegraphics[trim={2.5cm 0.5cm 2.5cm 0cm}, clip=true, align=c, width=\sz\linewidth]{figures/corbs/fused_small.jpg} & 
\includegraphics[trim={2.5cm 0.5cm 2.5cm 0cm}, clip=true, align=c, width=\sz\linewidth]{figures/corbs/weighting_small.jpg} & 
\includegraphics[trim={0cm 0cm 0cm 0cm}, clip=true, align=c, width=\sz\linewidth]{figures/colorbars/corbs_colorbar.png}\\
 & Frame & ToF & MVS~\cite{schonberger2016pixelwise} & TSDF & \ours{} & Our sensor & \\
 & & & & Fusion~\cite{curless1996volumetric} & (Ours) & weighting & \\
\end{tabular}
}
}
\subcaption{}
\label{fig:corbs_visual}
\end{subfigure}
\caption{\textbf{CoRBS Dataset. ToF\bplus{}MVS Fusion.} Our model can find synergies between very imbalanced sensors. (a) The numerical results show that our fused model is better than the individual depth sensor inputs and significantly better than any of the baseline methods.
(b) Contrary to our method, the baseline methods cannot handle the high degree of outliers
from the MVS sensor.}
\label{fig:corbs}
\end{figure}
\fi

\ifeccv
\begin{figure*}[ht]
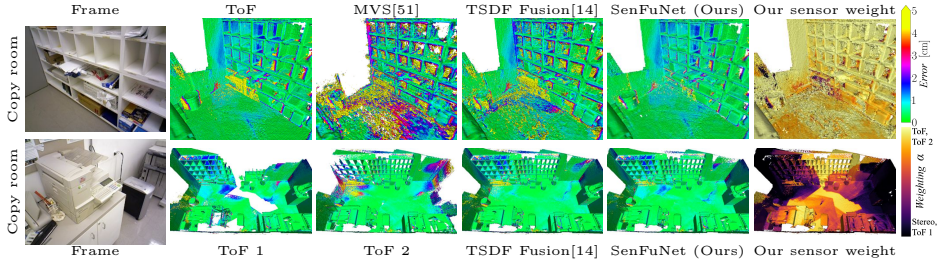

\centering
\tiny
\setlength{\tabcolsep}{1pt}
\renewcommand{\arraystretch}{1}
\begin{tabular}{ccccccccc}
& Frame & ToF & MVS\cite{schonberger2016pixelwise} & TSDF Fusion\cite{curless1996volumetric} & SenFuNet (Ours) & Our sensor weight & &\\
\rotatebox[origin=c]{90}{Copy room} & 
\includegraphics[align=c, width=.153\linewidth]{figures/scene3d/copyroom_mvs_tof/model_small.jpg} & 
\includegraphics[align=c, width=.153\linewidth]{figures/scene3d/copyroom_mvs_tof/tof_small.jpg} & 
\includegraphics[align=c, width=.153\linewidth]{figures/scene3d/copyroom_mvs_tof/mvs_small.jpg} & 
\includegraphics[align=c, width=.153\linewidth]{figures/scene3d/copyroom_mvs_tof/tsdf_small.jpg} & 
\includegraphics[align=c, width=.153\linewidth]{figures/scene3d/copyroom_mvs_tof/fused_small.jpg} & 
\includegraphics[align=c, width=.153\linewidth]{figures/scene3d/copyroom_mvs_tof/weighting_small.jpg} & \multirow{11}{*}[25pt]{\includegraphics[width=.038\linewidth]{figures/colorbars/multi_agent_colorbar.png}} \\
 \rotatebox[origin=c]{90}{Copy room} &
 \includegraphics[align=c, width=.153\linewidth]{figures/scene3d/copyroom_collab/model_small.jpg} & 
 \includegraphics[trim={0cm 0cm 0cm 0cm}, clip=true, align=c, width=.153\linewidth]{figures/scene3d/copyroom_collab/tof_1_small.jpg} & 
 \includegraphics[trim={0cm 0cm 0cm 0cm}, clip=true,align=c, width=.153\linewidth]{figures/scene3d/copyroom_collab/tof_2_small.jpg} & 
 \includegraphics[trim={0cm 0cm 0cm 0cm}, clip=true,align=c, width=.153\linewidth]{figures/scene3d/copyroom_collab/tsdf_small.jpg}& 
 \includegraphics[trim={0cm 0cm 0cm 0cm}, clip=true,align=c, width=.153\linewidth]{figures/scene3d/copyroom_collab/fused_small.jpg} & 
 \includegraphics[trim={0cm 0cm 0cm 0cm}, clip=true,align=c, width=.153\linewidth]{figures/scene3d/copyroom_collab/weighting_small.jpg}  &  \\
 & Frame & ToF 1 & ToF 2 & TSDF Fusion\cite{curless1996volumetric} & SenFuNet (Ours) & Our sensor weight & &\\
\end{tabular}
\caption{ \textbf{Top row:}  Our method effectively fuses the ToF and MVS sensors.
Note specifically the absence of yellow outliers in the corner of the bookshelf.
See also Tab.~\ref{tab:scene3d_tof_mvs}.
\textbf{Bottom row:} Multi-Agent ToF Reconstruction. Our method is flexible and can perform Multi-Agent reconstruction. 
Note that our model learns where to trust each agent at different spatial locations for maximum completeness, while also being noise aware. See for instance the left bottom corner of the bookshelf where both agents integrate, but the noise-free agent is given a higher weighting. The above scene is taken from the Scene3D Dataset \cite{zhou2013dense} (best viewed on screen).
See also Tab.~\ref{tab:scene3d_collab}.}
\label{fig:scene3d}
\end{figure*}
\fi

\ifeccv
\begin{wraptable}[12]{R}{0.60\linewidth}
\vspace{-\intextsep}
\setlength{\belowcaptionskip}{0pt}
\begin{subtable}{.49\linewidth}
\centering
\resizebox{\linewidth}{!}
{
\setlength{\tabcolsep}{2pt}
\renewcommand{\arraystretch}{1.05}
\begin{tabular}{l|lll}
\cellcolor{gray}Model   &\cellcolor{gray} F$\uparrow$ $[\%]$ & \cellcolor{gray}P$\uparrow$ $[\%]$ & \cellcolor{gray}R$\uparrow$ $[\%]$   \\ \hline
\multicolumn{4}{c}{\emph{Single Sensor}} \\ \hline
MVS~\cite{schonberger2016pixelwise}    & 44.26 & 32.08 & 71.33 \\ 
ToF      & 90.73 & 85.53 & 96.61 \\ 
\hline
\multicolumn{4}{c}{\emph{Multi-Sensor Fusion}} \\ \hline
TSDF Fusion~\cite{curless1996volumetric}   & 54.77 & 38.11 & 97.29 \\ 
RoutedFusion~\cite{Weder2020RoutedFusionLR} & 53.98 & 37.73 & 94.86 \\ 
DI-Fusion~\cite{huang2021di} $\sigma$=0.15 & 48.84 & 32.50 & \textbf{98.24} \\ 
\textbf{\ours{} (Ours)}  & \textbf{93.39} & \textbf{91.28} & 95.60 \\[-10pt] 
\end{tabular}
}
\subcaption{}
\label{tab:scene3d_tof_mvs}
\end{subtable}
\begin{subtable}{0.49\linewidth}
\resizebox{\linewidth}{!}
{
\setlength{\tabcolsep}{2pt}
\renewcommand{\arraystretch}{1.05}
\begin{tabular}{l|lll}
\cellcolor{gray}Model   & \cellcolor{gray}F$\uparrow$ $[\%]$ & \cellcolor{gray}P$\uparrow$ $[\%]$ & \cellcolor{gray}R$\uparrow$ $[\%]$   \\ \hline
\multicolumn{4}{c}{\emph{Single Sensor}} \\ \hline
ToF 1  & 84.68 & 89.92 & 80.01\\ 
ToF 2    & 77.77  & 79.65 & 75.97  \\ 
\hline
\multicolumn{4}{c}{\emph{Multi-Sensor Fusion}} \\ \hline
TSDF Fusion~\cite{curless1996volumetric}   & 90.73  & 85.53 & 96.61 \\ 
RoutedFusion\cite{Weder2020RoutedFusionLR}  & 84.11 & 74.16 & 97.14 \\ 
DI-Fusion~\cite{huang2021di} $\sigma$=0.15 & 86.31 & 77.27 & \textbf{97.74} \\ 
\textbf{\ours{} (Ours)}  & \textbf{93.73} & \textbf{91.56} & 96.00 \\[-10pt] 
\end{tabular}
}
\subcaption{}
\label{tab:scene3d_collab}
\end{subtable}
\caption{\textbf{Scene3D Dataset.} (a) \textbf{ToF\bplus{}MVS Fusion.} Our method outperforms the baselines on real-world data on a room-sized scene.
(b) \textbf{Multi-Agent ToF Reconstruction.} Our method is flexible and can perform collaborative sensor fusion from multiple sensors with different camera trajectories.}
\label{tab:scene3d}
\end{wraptable}
\else
\begin{table}
\setlength{\belowcaptionskip}{0pt}
\begin{subtable}{.495\columnwidth}
\centering
\resizebox{\columnwidth}{!}
{
\setlength{\tabcolsep}{2pt}
\renewcommand{\arraystretch}{1.05}
\begin{tabular}{l|lll}
\cellcolor{gray}Model   &\cellcolor{gray} F$\uparrow$ $[\%]$ & \cellcolor{gray}P$\uparrow$ $[\%]$ & \cellcolor{gray}R$\uparrow$ $[\%]$   \\ \hline
\multicolumn{4}{c}{\emph{Single Sensor}} \\ \hline
MVS~\cite{schonberger2016pixelwise}    & 44.26 & 32.08 & 71.33 \\ 
ToF      & 90.73 & 85.53 & 96.61 \\ 
\hline
\multicolumn{4}{c}{\emph{Multi-Sensor Fusion}} \\ \hline
TSDF Fusion~\cite{curless1996volumetric}   & 54.77 & 38.11 & 97.29 \\ 
RoutedFusion~\cite{Weder2020RoutedFusionLR} & 53.98 & 37.73 & 94.86 \\ 
DI-Fusion~\cite{huang2021di} $\sigma$=0.15 & 48.84 & 32.50 & \textbf{98.24} \\ 
\textbf{\ours{} (Ours)}  & \textbf{93.39} & \textbf{91.28} & 95.60 \\[-10pt] 
\end{tabular}
}
\subcaption{}
\label{tab:scene3d_tof_mvs}
\end{subtable}
\begin{subtable}{0.495\columnwidth}
\resizebox{\columnwidth}{!}
{
\setlength{\tabcolsep}{2pt}
\renewcommand{\arraystretch}{1.05}
\begin{tabular}{l|lll}
\cellcolor{gray}Model   & \cellcolor{gray}F$\uparrow$ $[\%]$ & \cellcolor{gray}P$\uparrow$ $[\%]$ & \cellcolor{gray}R$\uparrow$ $[\%]$   \\ \hline
\multicolumn{4}{c}{\emph{Single Sensor}} \\ \hline
ToF 1  & 84.68 & 89.92 & 80.01\\ 
ToF 2    & 77.77  & 79.65 & 75.97  \\ 
\hline
\multicolumn{4}{c}{\emph{Multi-Sensor Fusion}} \\ \hline
TSDF Fusion~\cite{curless1996volumetric}   & 90.73  & 85.53 & 96.61 \\ 
RoutedFusion\cite{Weder2020RoutedFusionLR}  & 84.11 & 74.16 & 97.14 \\ 
DI-Fusion~\cite{huang2021di} $\sigma$=0.15 & 86.31 & 77.27 & \textbf{97.74} \\ 
\textbf{\ours{} (Ours)}  & \textbf{93.73} & \textbf{91.56} & 96.00 \\[-10pt] 
\end{tabular}
}
\subcaption{}
\label{tab:scene3d_collab}
\end{subtable}
\caption{\textbf{Scene3D Dataset.} \textbf{(a) ToF\bplus{}MVS Fusion.} Our method outperforms the baselines on real-world data on a room-sized scene.
\textbf{(b) Multi-Agent ToF Reconstruction.} Our method is flexible and can perform collaborative sensor fusion when multiple sensors with different camera trajectories are available.}
\label{tab:scene3d}
\end{table}
\fi

\subsection{Experiments on the Scene3D Dataset} 
%
We demonstrate that our framework can fuse imbalanced sensors on room-sized real-world scenes using the RGBD Scene3D dataset~\cite{zhou2013dense}. The Scene3D dataset comprises a collection of 3D models of indoor and outdoor scenes. 
We train our model on the stonewall scene and test it on the copy room scene. To create the ground truth training grid, we follow the steps outlined previously except that it was not necessary to make the mesh watertight.
We fuse every 10th frame during train and test time. Equivalent to the study on CoRBS, we create an MVS depth sensor using COLMAP and perform ToF and MVS fusion. 
We only integrate MVS depth in the interval $[0.5, 3.0]$ m.
Tab.~\ref{tab:scene3d_tof_mvs} along with Fig.~\ref{fig:scene3d} \emph{Top row} shows that our method yields a fused result better than the individual input sensors and the baseline methods. Further, Fig.~\ref{fig:teaser} shows our method in comparison with TSDF Fusion~\cite{curless1996volumetric} and RoutedFusion~\cite{Weder2020RoutedFusionLR} on the lounge scene.

\boldparagraph{Multi-Agent ToF Fusion.} Our method is not exclusively applicable to sensor fusion, but more flexible.
We demonstrate this by formulating a Multi-Agent reconstruction problem, which assumes that two identical ToF sensors with different camera trajectories are provided. 
The task is to fuse the reconstructions from the two agents.
This requires an understanding of when to perform sensor selection for increased completeness and smooth fusion when both sensors have registered observations.
Note that this formulation is different from typical works on collaborative 3D reconstruction \eg~\cite{golodetz2018collaborative} where the goal is to align 3D reconstruction fragments to produce a complete model.
In our Multi-Agent setting, the alignment is given and the task is instead to perform data fusion on the 3D fragments. No modification of our method is required to perform this task.
We set $\lambda_1 = 1/1200$ and $\lambda_2 = 1/12000$ and split the original trajectory into 100 frame chunks that are divided between the agents.
Tab.~\ref{tab:scene3d_collab} and Fig.~\ref{fig:scene3d} \emph{Bottom row} show that our method effectively fuses the incoming data streams and yields a 4 $\%$ F-score gain on the TSDF Fusion baseline. 

\subsection{More Statistical Analysis}
%

\boldparagraph{Performance over Camera Trajectory.} To show that our fused output is not only better at the end of the fusion process, we visualize the quantitative performance across the accumulated trajectory. In Fig.~\ref{fig:plot_trajectory_fusion}, we evaluate the office 0 scene on the sensors $\{$ToF, PSMNet$\}$ with depth denoising. Our fused model consistently improves on the inputs.

\ifeccv
\begin{figure}[bt]
\setlength{\belowcaptionskip}{0pt}
\begin{subfigure}{.50\linewidth}
\centering
\setlength{\tabcolsep}{1pt}
\renewcommand{\arraystretch}{1}
\begin{tabular}{cc}
 \includegraphics[align=c, width=.49\linewidth]{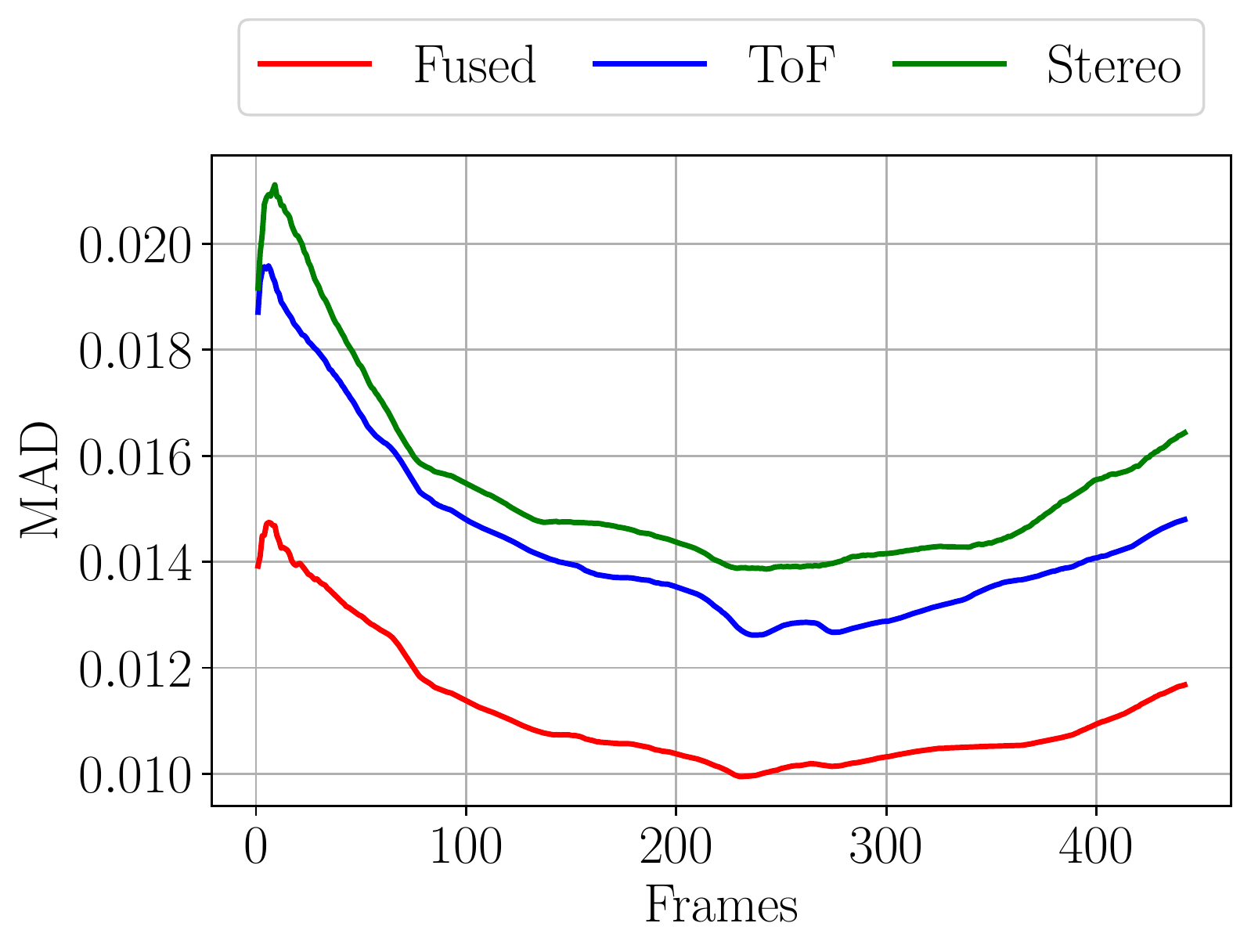} & \includegraphics[align=c, width=.49\linewidth]{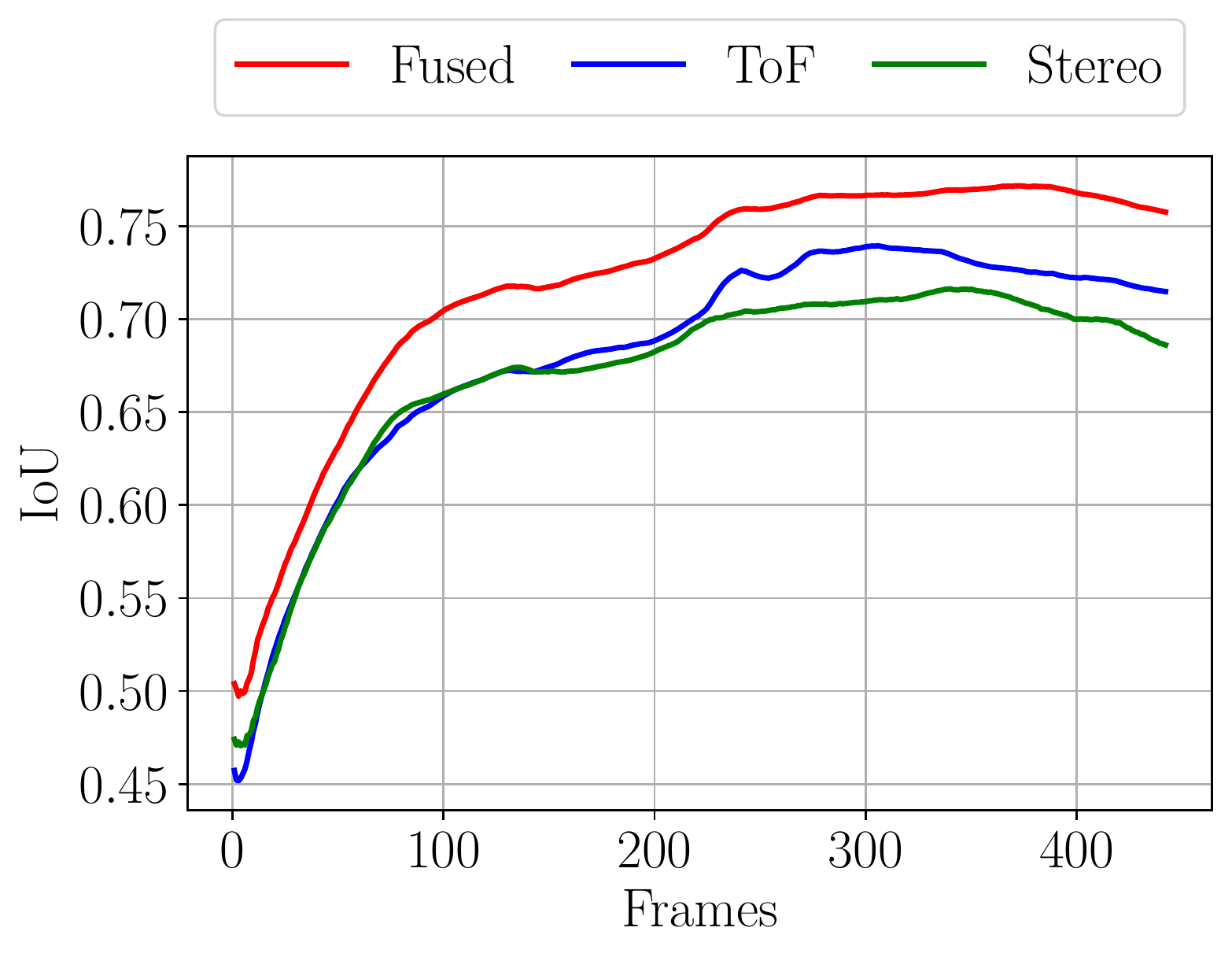} \\
\end{tabular}
\subcaption{}
\label{fig:plot_trajectory_fusion}
\end{subfigure}
\begin{subfigure}{0.50\linewidth}
\tiny
\setlength{\tabcolsep}{1pt}
\renewcommand{\arraystretch}{1}
\begin{tabular}{cccc}
 & \hspace{0.2cm} Without Outlier Filter & \hspace{0.2cm} With Outlier Filter &\\
\raisebox{0.2cm}[0pt][0pt]{\rotatebox[origin=c]{90}{\makecell{Feature Space}}} & \includegraphics[align=c, width=.42\linewidth]{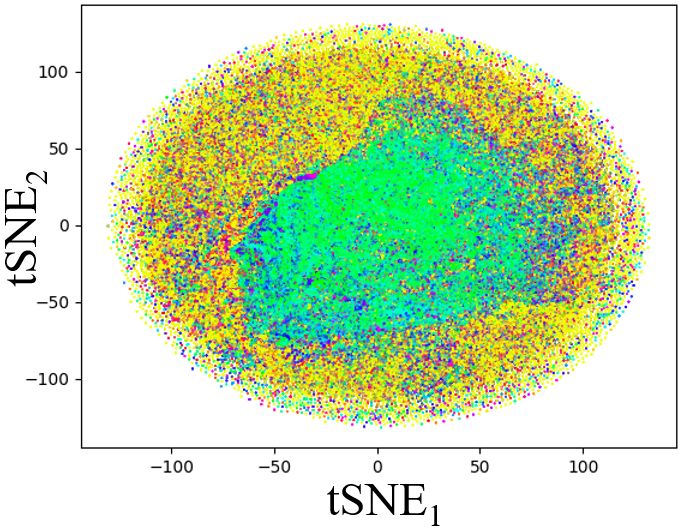} & \includegraphics[align=c, width=.42\linewidth]{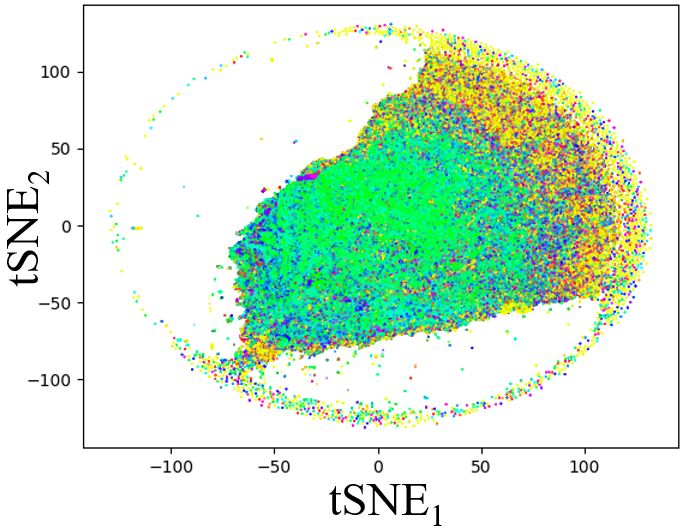} & \raisebox{0.2cm}[0pt][0pt]{\includegraphics[align=c,width=.075\linewidth]{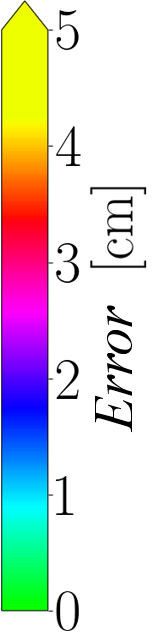}} \\
\end{tabular}
\subcaption{}
\label{fig:outlier_filter}
\end{subfigure}
\caption{(a) \textbf{Performance over Camera Trajectory.} Our fused output
outperforms the single sensor reconstructions ($V_{t}^{i}$) for all frames along the camera trajectory. The above results are evaluated on the Replica \emph{office 0} scene using $\{$ToF, PSMNet$\}$ with depth denoising. Note that the results get slightly worse after 300 frames. This is due to additional noise from the depth sensors when viewing the scene from further away.
(b) \boldparagraph{Effect of Learned Outlier Filter.} The learned filter is crucial for robust outlier handling. Erroneous outlier predictions shown in yellow are effectively removed by our approach while keeping the correct green-colored predictions.
}
\end{figure}
\else
\begin{figure}[bt]
\centering
\setlength{\tabcolsep}{1pt}
\renewcommand{\arraystretch}{1}
\begin{tabular}{cc}
 \includegraphics[align=c, width=.50\linewidth]{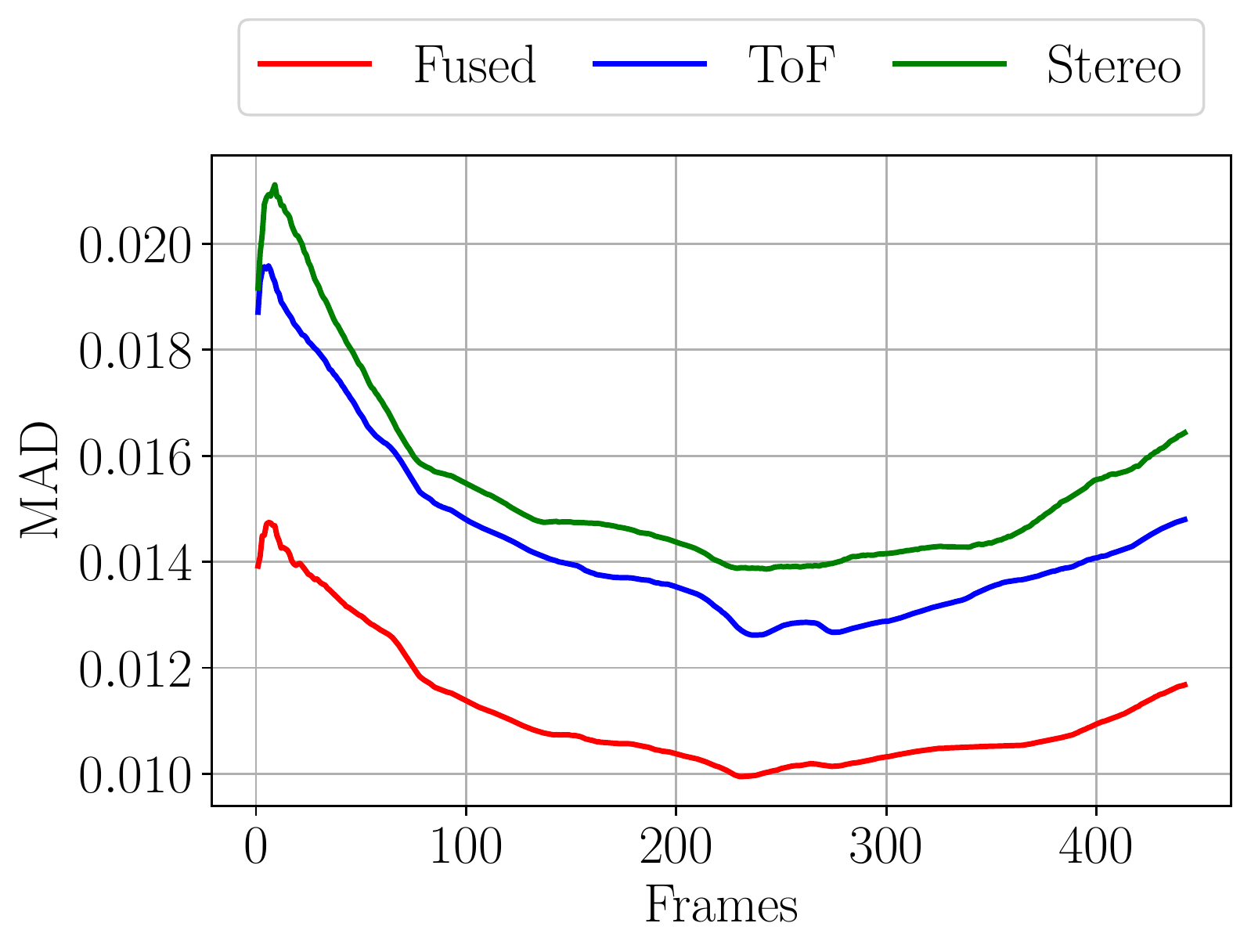} & \includegraphics[align=c, width=.50\linewidth]{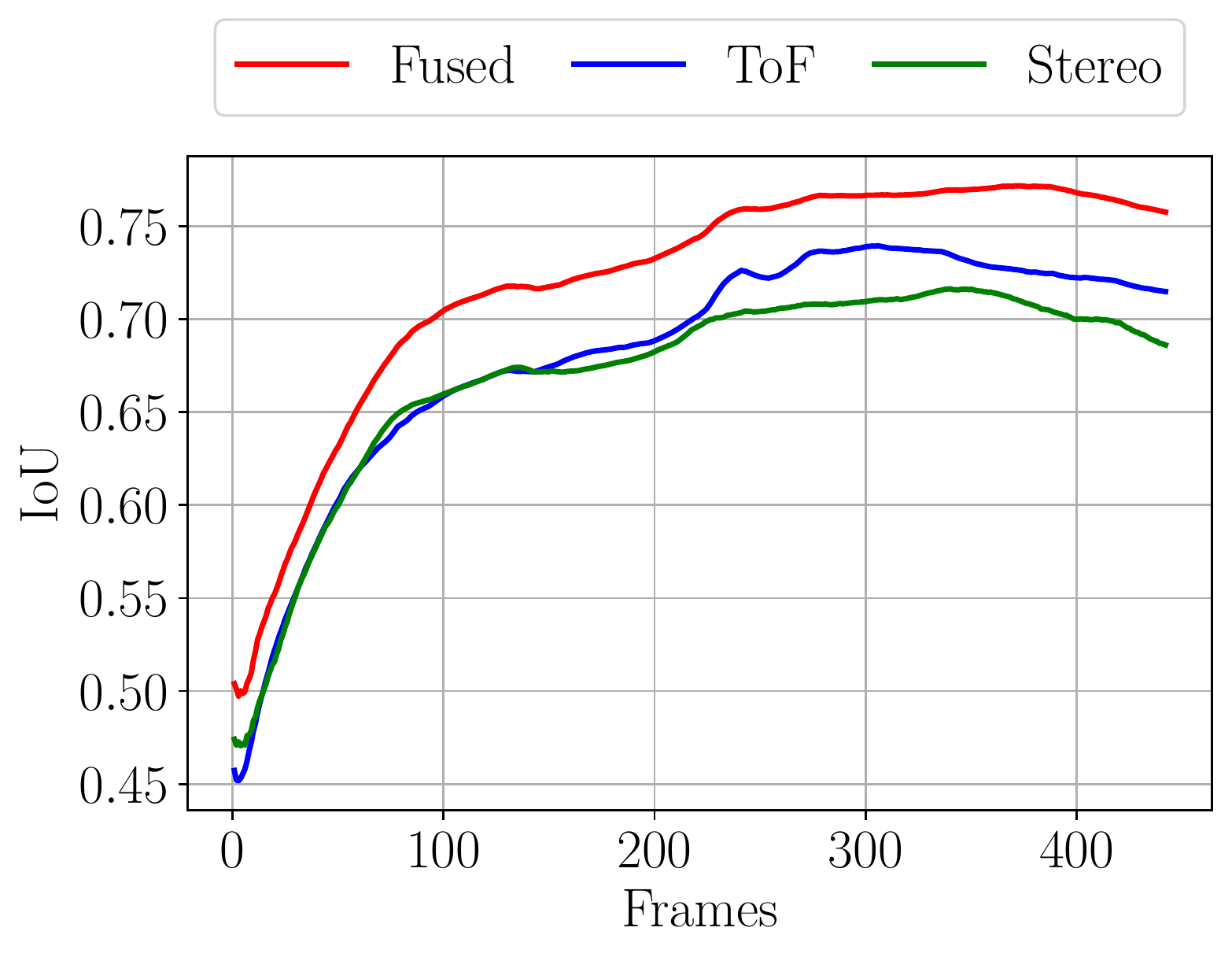} \\
\end{tabular}
\caption{\textbf{Performance over Camera Trajectory.} The fused output of our method outperforms the single input reconstructions for all frames along the trajectory.
Experiment done on the Replica \emph{office 0} scene on the sensors $\{$ToF, PSMNet$\}$ with denoising. Note: The metrics worsen slightly after $\sim$300 frames. This is trajectory dependent - the scene is viewed from further away.}
\label{fig:plot_trajectory_fusion}
\end{figure}
\fi

\ifeccv
\begin{wraptable}[6]{R}{0.55\linewidth}
\vspace{-\intextsep}
\centering
\resizebox{\linewidth}{!}
{
\footnotesize
\setlength{\tabcolsep}{9.6pt}
\renewcommand{\arraystretch}{1.1}
\begin{tabular}{l|lllll}
\cellcolor{gray} \# Layers &\cellcolor{gray} 0 &\cellcolor{gray} 1 &\cellcolor{gray} 2 &\cellcolor{gray} 3 &\cellcolor{gray} 4 \\ 
\hline
F1 & 48.57 & 68.47 & \textbf{69.86} & 69.45 & 68.21\\ 
\end{tabular}
}
\caption{\boldparagraph{Architecture Ablation.} We vary the number of 3D convolutional layers with kernel size 3. 
Performance is optimal at 2 layers, equivalent to a receptive field of $9^3$.}
\label{tab:ablation_layers_weighting_net}
\end{wraptable}
\fi

\boldparagraph{Architecture Ablation.} We perform a network ablation on the Replica dataset with the sensors $\{$SGM, PSMNet$\}$ without depth denoising. 
In Tab.~\ref{tab:ablation_layers_weighting_net}, we investigate the number of layers with kernel size 3 in the Weighting Network $\mathcal{G}$.
Two layers yield optimal performance which amounts to a receptive field of $9 \!\times\! 9 \!\times\! 9$.
This is realistic given that the support for a specific sensor is local by nature.

\ifeccv
\else
\begin{table}
\centering
\footnotesize
\setlength{\tabcolsep}{9.6pt}
\renewcommand{\arraystretch}{1.1}
\begin{tabular}{l|lllll}
\cellcolor{gray} \# Layers &\cellcolor{gray} 0 &\cellcolor{gray} 1 &\cellcolor{gray} 2 &\cellcolor{gray} 3 &\cellcolor{gray} 4 \\ 
\hline
F1 & 48.57 & 68.47 & \textbf{69.86} & 69.45 & 68.21\\ 
\end{tabular}
\caption{\boldparagraph{Architecture Ablation.} We vary the number of 3D convolutional layers with kernel size 3. 
Performance is optimal at 2 layers, equivalent to a receptive field of $9 \!\times\! 9 \!\times\! 9$.}
\label{tab:ablation_layers_weighting_net}
\end{table}
\fi


\ifeccv
\begin{wraptable}[7]{R}{0.5\linewidth}
\vspace{-\intextsep}
\centering
\renewcommand{\arraystretch}{1.05}
\resizebox{\linewidth}{!}
{
\begin{tabular}{l|lllllll}
\cellcolor{gray}Train Set        & \cellcolor{gray}MSE$\downarrow$      & \cellcolor{gray}MAD$\downarrow$  & \cellcolor{gray}IoU$\uparrow$     & \cellcolor{gray}Acc.$\uparrow$ & \cellcolor{gray}F$\uparrow$ & \cellcolor{gray}P$\uparrow$ & \cellcolor{gray}R$\uparrow$   \\ 
\cellcolor{gray}& \cellcolor{gray}*e-04 & \cellcolor{gray}*e-02 & \cellcolor{gray}[0,1] & \cellcolor{gray}$[\%]$ & \cellcolor{gray}$\%$ & \cellcolor{gray}$[\%]$ & \cellcolor{gray}$[\%]$ \\\hline
Full Train Set  & \textbf{4.51} & 1.54 & 0.748 & 87.78 & 68.70 & \textbf{59.49} & 81.28  \\ 
Office 0 & 4.54 & \textbf{1.53} & \textbf{0.752}  & \textbf{87.97} & \textbf{69.23} & 59.45 & \textbf{82.86}  \\ 
\end{tabular}
}
\caption{\textbf{Generalization Capability.} Our model generalizes well when evaluated on the \emph{office 0} scene. We conclude this by training a model only on \emph{office 0}.} 
\label{tab:overfitting}
\end{wraptable}
\else
\begin{table}
\centering
\renewcommand{\arraystretch}{1.05}
\resizebox{\columnwidth}{!}
{
\begin{tabular}{l|lllllll}
\cellcolor{gray}Train Set        & \cellcolor{gray}MSE$\downarrow$      & \cellcolor{gray}MAD$\downarrow$  & \cellcolor{gray}IoU$\uparrow$     & \cellcolor{gray}Acc.$\uparrow$ & \cellcolor{gray}F$\uparrow$ & \cellcolor{gray}P$\uparrow$ & \cellcolor{gray}R$\uparrow$   \\ 
\cellcolor{gray}& \cellcolor{gray}*e-04 & \cellcolor{gray}*e-02 & \cellcolor{gray}[0,1] & \cellcolor{gray}$[\%]$ & \cellcolor{gray}$\%$ & \cellcolor{gray}$[\%]$ & \cellcolor{gray}$[\%]$ \\\hline
Full Train Set  & \textbf{4.51} & 1.54 & 0.748 & 87.78 & 68.70 & \textbf{59.49} & 81.28  \\ 
Office 0 & 4.54 & \textbf{1.53} & \textbf{0.752}  & \textbf{87.97} & \textbf{69.23} & 59.45 & \textbf{82.86}  \\ 
\end{tabular}
}
\caption{\textbf{Generalization Capability.} Our model generalizes well when evaluated on the \emph{office 0} scene. We conclude this by training an additional model only on \emph{office 0}.} 
\label{tab:overfitting}
\end{table}
\fi

\boldparagraph{Generalization Capability.} 
Tab.~\ref{tab:overfitting} shows our model's generalization ability for $\{$SGM, PSMNet$\}$ fusion when evaluated against a model trained and tested on the \emph{office 0} scene. 
Our model generalizes well and performs almost on par with one which is only trained on the \emph{office 0} scene. 
The generalization capability is not surprising since $\mathcal{G}$ has a limited receptive field of $9 \!\times\! 9 \!\times\! 9$.

\boldparagraph{Effect of Learned Outlier Filter.} To show the effectiveness of the filter, we study the feature space on the input side of $\mathcal{G}$. 
Specifically, we consider the \emph{hotel 0} scene and sensors $\{$ToF, PSMNet$\}$ with depth denoising.
First, we concatenate both feature grids and flatten the resulting grid.
Then, we reduce the observations of the 12-dim feature space to a 2-dim representation using tSNE~\cite{van2008visualizing}.
We then colorize each point with the corresponding signed distance error at the original voxel position.
We repeat the visualization with and without the learned outlier filter. 
Fig.~\ref{fig:outlier_filter} shows that the filter effectively removes outliers while keeping good predictions.

\ifeccv
\else
\begin{figure}[t]
\centering
\scriptsize
\setlength{\tabcolsep}{1pt}
\renewcommand{\arraystretch}{1}
\begin{tabular}{cccc}
 & \hspace{0.2cm} Without Outlier Filter & \hspace{0.2cm} With Outlier Filter &\\
\raisebox{0.2cm}[0pt][0pt]{\rotatebox[origin=c]{90}{\makecell{Feature Space}}} & \includegraphics[align=c, width=.42\linewidth]{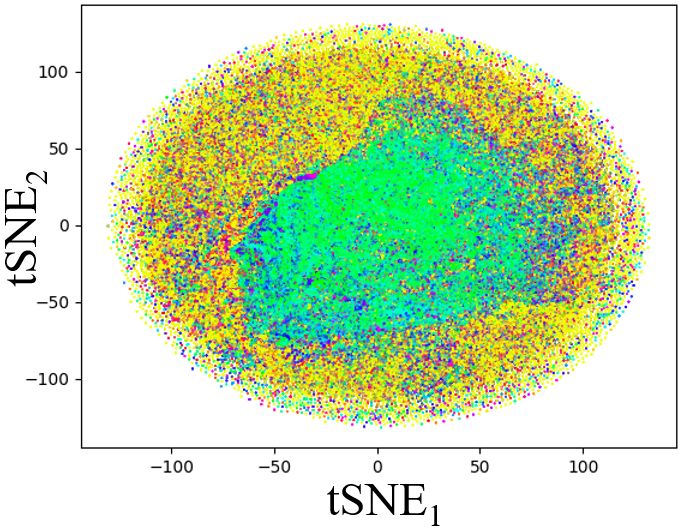} & \includegraphics[align=c, width=.42\linewidth]{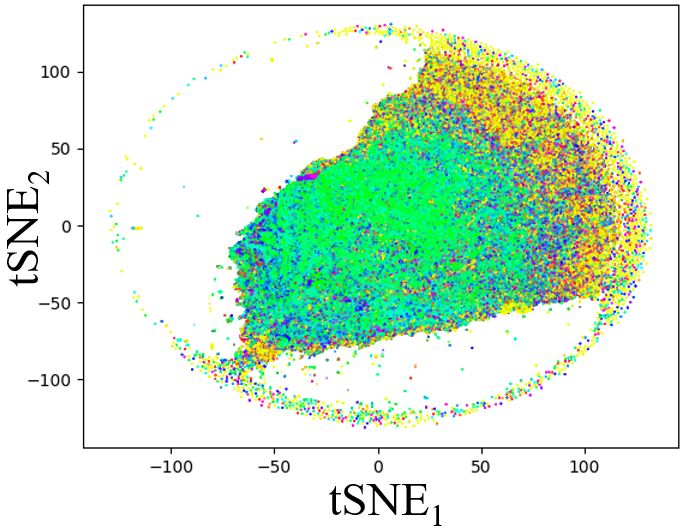} & \raisebox{0.2cm}[0pt][0pt]{\includegraphics[align=c,width=.075\linewidth]{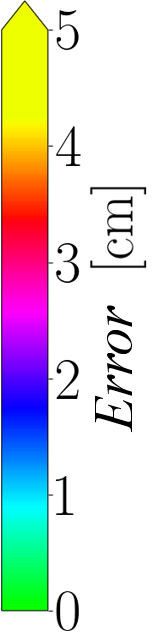}} \\
\end{tabular}
\caption{\boldparagraph{Effect of Learned Outlier Filter.} The filter is crucial for achieving robust outlier handling.
A lot of the erroneous yellow-colored predictions are removed by the filter, while keeping the correct green-colored predictions.}
\label{fig:outlier_filter}
\end{figure}
\fi

\ifeccv

\begin{wraptable}[9]{R}{0.5\linewidth}
\centering
\vspace{-\intextsep}
\resizebox{1.0\linewidth}{!}
{
\begin{tabular}{l|lllllll}
\cellcolor{gray}Loss $\mathcal{L}$       \cellcolor{gray}& \cellcolor{gray}MSE$\downarrow$      & \cellcolor{gray}MAD$\downarrow$  & \cellcolor{gray}IoU$\uparrow$     & \cellcolor{gray}Acc.$\uparrow$ & \cellcolor{gray}F$\uparrow$ & \cellcolor{gray}P$\uparrow$ & \cellcolor{gray}R$\uparrow$   \\ 
\cellcolor{gray} & \cellcolor{gray}*e-04 & \cellcolor{gray}*e-02 & \cellcolor{gray}[0,1] & \cellcolor{gray}$[\%]$ & \cellcolor{gray}$[\%]$ & \cellcolor{gray}$[\%]$ & \cellcolor{gray}$[\%]$ \\\hline
Only $\mathcal{L}_f$ & 6.04 & 1.78 & 0.710 & 86.20 & 62.65 & 50.88 & \textbf{82.17} \\ 
Full Loss  & \textbf{4.77} & \textbf{1.56}  & \textbf{0.738}  & \textbf{87.62}  & \textbf{69.83} & \textbf{63.20} & 79.12  \\ 
\end{tabular}
}
\caption{\textbf{Loss Ablation.} When only the term $\mathcal{L}_f$ is used, we observe a significant performance drop compared to the full loss. Note, however, that only with the term $\mathcal{L}_f$, our model still improves on the single sensor input metrics compared to Tab.~\ref{tab:sgm_psmnet_no_denoising}.}
\label{tab:single_sensor_supervision}
\end{wraptable}
\fi

\boldparagraph{Loss Ablation.} Tab.~\ref{tab:single_sensor_supervision} shows the performance difference when the model is trained only with the term $\mathcal{L}_f$ compared to the full loss \eqref{eq:loss}. We perform $\{$SGM, PSMNet$\}$ fusion on the Replica dataset. The extra terms of the full loss clearly help improve overall performance and specifically to filter outliers.

\boldparagraph{Limitations.}
Our framework currently supports two sensors. 
Its extension to a $k$-sensor setting is straightforward, but the memory footprint grows linearly with the number of sensors. 
However, few devices have more than two or three depth sensors.
While our method generates better results on average, some local regions may not improve and our method struggles with overlapping outliers from both sensors.
For qualitative examples, see the supplementary material.
Ideally, outliers can be filtered and the data fused with a learned scene representation as in~\cite{weder2021neuralfusion}, but our efforts to make~\cite{weder2021neuralfusion} work with multiple sensors suggests that this is a harder learning problem which deserves attention in future work.

\section{Conclusion}
We propose a machine learning approach for online multi-sensor 3D reconstruction using depth maps. 
We show that a fusion decision on 3D features rather than directly on 2D depth maps generally improves surface accuracy and outlier robustness. 
This also holds when 2D fusion is straightforward, \ie~for time synchronized sensors, equal sensor resolution and calibration.
The experiments demonstrate that our model handles various sensors, can scale to room-sized real-world scenes and produce a fused result that is quantitatively and 
qualitatively better than the single sensor inputs and the compared baselines methods. \\

\boldparagraph{Acknowledgements}: This work was supported by the Google Focused Research Award 2019-HE-318, 2019-HE-323, 2020-FS-351, 2020-HS-411, as well as by research grants from FIFA and Toshiba. 
We further thank Hugo Sellerberg for helping with video editing.

\clearpage

\title{Learning Online Multi-Sensor Depth Fusion \\ Supplementary Material} 

\author{Erik Sandström$^{1}$
\and
Martin R. Oswald$^{1,2}$
\and
Suryansh Kumar$^{1}$
\and
Silvan Weder$^{1}$
\and
Fisher Yu$^{1}$
\and
Cristian Sminchisescu$^{3,5}$
\and
Luc Van Gool$^{1,4}$
}

\authorrunning{E. Sandström \etal}
\institute{$^{1}$ETH Zürich, $^{2}$University of Amsterdam, $^{3}$Lund University, $^{4}$KU Leuven, $^{5}$Google Research}

\maketitle
\thispagestyle{empty}

\begin{abstract}
    This supplementary material accompanies the main paper by providing further information for better reproducibility as well as additional evaluations and qualitative results.
\end{abstract}


\appendix
\section*{A. Videos}
\label{sec:videos}
\ifeccv
\else
\addcontentsline{toc}{section}{\nameref{sec:videos}}
\fi
We provide an introductory video that presents an overview of our method (available here: \url{https://youtu.be/woA8FU05AM0}) as well as a summary of the most important results.
Additionally, a selection of short videos is linked in the description of the introductory video showing the online reconstruction process for various sensors and scenes.

\section*{B. Method}
\label{sec:method}
\ifeccv
\else
\addcontentsline{toc}{section}{\nameref{sec:method}}
\fi
In the following, we provide more details about our proposed method, specifically the Shape Integration Module which updates the shape grid $S^i_t$.

\boldparagraph{Shape Integration Module.}
The shape integration module takes as input a depth map $D^i_t$ from sensor $i \in \{0, 1\}$ at time $t \in \mathbb{N}$, with known camera calibration $P^{c \rightarrow w}_t \in \mathbb{SE}(3)$ from camera to world space and intrinsics $K_t \in \mathbb{R}^{3\times3}$ and performs a full perspective unprojection to attain a point cloud $\textbf{X}_w$ in the world coordinate space. Each 3D point $\textbf{x}_w$ $\in$ $\textbf{X}_w$ is computed by transforming each pixel $(u,\ v)$ of the depth map into the camera space $\textbf{x}_c$ according to \eqref{eq:im_cam},

\begin{equation}
    \textbf{x}_{c} = D_t^i(u, v)K_t^{-1}\begin{bmatrix}
           u \\
           v \\
           1
         \end{bmatrix}
    \label{eq:im_cam}
\end{equation}
and then into the world camera space according to \eqref{eq:cam_world}.

\begin{equation}
    \textbf{x}_w = P^{c \rightarrow w}\begin{bmatrix}
           \textbf{x}_c \\
            1
         \end{bmatrix}
    \label{eq:cam_world}
\end{equation}
Along each ray from the camera center, centered at $\textbf{x}_w$, we sample T points evenly over a predetermined distance $l$. The distance $l$ and the number of points $T$ are selected based on the noise level of the depth sensor and could be interpreted similarly as the truncation distance in standard TSDF Fusion~\cite{curless1996volumetric}. The distance between sampled points should ideally be the same as the voxel side length.
After computing $T$ points along each ray, we convert the coordinates to the voxel space and extract a local shape grid $S^{i, *}_{t-1}$ from $S^i_{t-1}$ through nearest neighbor extraction. 
To incrementally update $S^{i, *}_{t-1}$, we follow the moving average update scheme of TSDF Fusion,
\begin{equation}
    V^{i,*}_t = \frac{W^{i,*}_{t-1}V^{i,*}_{t-1} + v^i_t}{W^{i,*}_{t-1} + 1},
    \label{eq:v_update}
\end{equation}
where $v^i_t$ is the TSDF update. The local weight grid $W^{i, *}_t$ is updated as
\begin{equation}
    W^{i,*}_t = \max(\omega_{\textrm{max}}, W^{i,*}_{t-1} + 1).
    \label{eq:w_update}
\end{equation}
The weights are clipped at $\omega_{\textrm{max}}$ to prevent numerical instabilities.

\boldparagraph{Denoising Network.}
We use the identical denoising network as described by Weder~\etal~ \cite{Weder2020RoutedFusionLR} (called routing network) except that we change the loss hyperparameter to $\lambda=0.06$.

\boldparagraph{Indicator Function.}
We define the indicator function $\1{A}(x)$ for a voxel index $x = (x_1, \ x_2, \ x_3)$, where $x_i \in \mathbb{N}$ as
\begin{equation}
\1{A}(x) = \begin{cases}
      0 & \text{if}\ x \notin A\\
      1 & \text{if}\ x \in A,
    \end{cases}
    \label{eq:indicator_function}
\end{equation}
and for two sets $A$ and $B$
\begin{equation}
\1{A, \ B}(x) = \begin{cases}
      0 & \text{if}\ x \notin A\cap B\\
      1 & \text{if}\ x \in A\cap B.
    \end{cases}
    \label{eq:indicator_function_and}
\end{equation}
In the main paper, we omit $x$ for brevity. 

\section*{C. Implementation Details}
\label{sec:imp_details}
\ifeccv
\else
\addcontentsline{toc}{section}{\nameref{sec:imp_details}}
\fi
We use PyTorch 1.7.1 and Python 3.8.5 to implement the pipeline. 
Training is done with the Adam optimizer using an Nvidia TITAN RTX with 24 GB of GPU memory.
We use a learning rate of $1e$-$04$ and otherwise the default Adam hyperparameters \emph{betas} $= (0.9, 0.999)$, \emph{eps} = $1e$-$08$ and \emph{weight$\_$decay} = 0.
We use a batch size of 1 due to the online nature of the pipeline, but accumulate the gradients over 20 frames before updating all network weights.
We shuffle the frames during train time and for training efficiency, we integrate every 10th frame during validation. We record a runtime of $\sim$15 fps on our unoptimized implementation on the Human CoRBS scene. For our largest scenes (\eg~\emph{Hotel 0}), our integration frame rate is between 1-2 fps. The bottleneck is cpu-gpu communication where our implementation loads the full voxel grid to the gpu for fast updates and then loads the grid back to the cpu to allow for scene reconstruction of multiple scenes in parallel.
We train our network until convergence which takes between 12-24 hours on the Replica dataset and a few hours on the CoRBS and Scene3D datasets.

\section*{D. Evaluation Metrics}
\label{sec:metrics}
\ifeccv
\else
\addcontentsline{toc}{section}{\nameref{sec:metrics}}
\fi
We use the following seven metrics to quantify the reconstruction performance.

\boldparagraph{Voxel Grid Metrics.} We use four metrics on the TSDF voxel grid. We mask the evaluation so that only voxels with a non-zero weight $W_k$ are considered. Mean Absolute Distance (MAD): Computed as the mean of the $L_1$ error to the ground truth signed distance grid MAD$ = \frac{1}{N}\sum_{k=0}^N |V_k - V_k^{GT}|_1$, where N is the total number of valid voxels. Mean Squared Error (MSE): Computed as the mean squared error to the ground truth signed distance grid MSE$ = \frac{1}{N}\sum_{k=0}^N (V_k - V_k^{GT})^2$, where N is the total number of valid voxels. Intersection over Union (IoU): Computed on the occupancy grid of the voxel grid as IoU$ = \frac{tn}{tn + fp + fn}$ and Accuracy as Acc$ = \frac{tn + tp}{tp + tn + fp + fn}$,
where 
\begin{align}
    tn &= \sum \{\sign(V) < 0 \ \text{and} \ \sign(V^{GT}) < 0\}  \\
    tp &= \sum \{\sign(V) >= 0 \ \text{and} \ \sign(V^{GT}) >= 0\} \\ 
    fp &= \sum \{\sign(V) >= 0 \ \text{and} \ \sign(V^{GT}) < 0\}  \\ 
    fn &= \sum \{\sign(V) < 0 \ \text{and} \ \sign(V^{GT}) >= 0\} 
\end{align}

\boldparagraph{Mesh Metrics.} We run marching cubes~\cite{lorensen1987marching} on the predicted voxel grid $V$ and the ground truth voxel grid $V^{GT}$ and compare the two meshes. 
The F-score is defined as the harmonic mean between Recall (R) and Precision (P), $F = 2\frac{PR}{P+R}$. 
Precision is defined as the percentage of vertices on the predicted mesh $V_m$ which lie within some distance $\tau$ from a vertex on the ground truth mesh $V_m^{GT}$.
Vice versa, Recall is defined as the percentage of points on the ground truth mesh $V_m^{GT}$ which lie within the same distance $\tau$ from a vertex on the predicted mesh $V_m$. In all our experiments, we use a distance threshold $\tau = 0.02$ m. 
We use the provided evaluation script of the Tanks and Temples dataset~\cite{knapitsch2017tanks}, but modify it to our needs.
For a more accurate evaluation, we do not downsample or crop the meshes and we do not utilize the automatic alignment procedure since our meshes are already aligned.

\section*{E. Replica Dataset Collection}
\label{sec:replica}
\ifeccv
\else
\addcontentsline{toc}{section}{\nameref{sec:replica}}
\fi
Due to the lack of available data for the study of multi-sensor depth fusion, we construct our own 2D dataset from the 3D Replica dataset~\cite{straub2019replica}, which comprises 18 high-quality scenes. 
To compute the ground truth signed distance value at each voxel grid point, we require a well-defined normal direction. 
This is not provided by the non-watertight Replica meshes. 
Thus, we apply screened Poisson surface reconstruction with CloudCompare~\cite{girardeau2016cloudcompare}, with an octree depth of 12. 
Otherwise, the default settings are used. 
Additionally, we found that the Poisson surface reconstructions are not clean enough to produce high-quality signed distance grids. 
Thus, each watertight mesh is cleaned with the Meshlab~\cite{cignoni2008meshlab} filter function \texttt{remove isolated pieces with respect to face number}. We set the face number threshold to 100.
The signed distance grids are then computed from the meshes with a modified version of the mesh-to-sdf library\footnote{{\normalfont\url{https://github.com/marian42/mesh_to_sdf}}} to accommodate non-cubic voxel grids. 

\begin{figure}[t]
\centering
\scriptsize
\setlength{\tabcolsep}{1pt}
\renewcommand{\arraystretch}{1}
\begin{tabular}{cccc}
\rotatebox[origin=c]{90}{\makecell{Room 0}} &
\includegraphics[align=c, width=.315\linewidth]{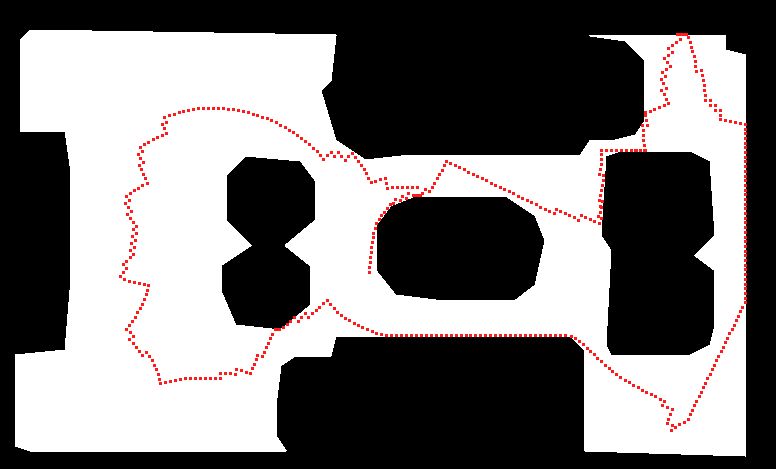} & 
\includegraphics[align=c, width=.315\linewidth]{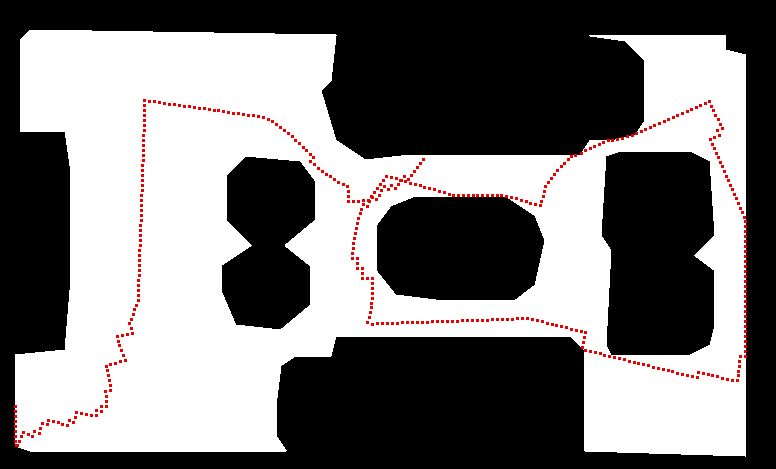} &
\includegraphics[align=c,width=.315\linewidth]{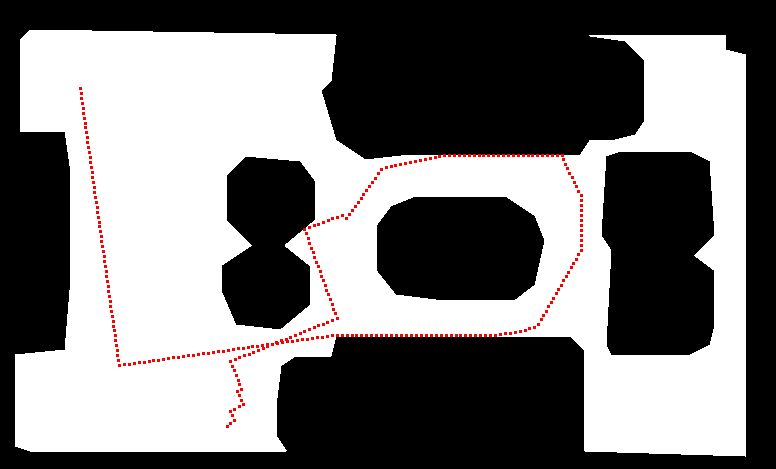} \\
 & Trajectory 1 & Trajectory 2 & Trajectory 3 \\
\end{tabular}
\caption{\boldparagraph{Trajectory Visualization.} Top-down visualization of the three manually traversed trajectories for the \emph{room 0} scene. Navigable space is colored white.}
\label{fig:trajectories}
\end{figure}

Using the Habitat AI platform~\cite{habitat19iccv}, we define an agent that moves around in the watertight 3D scenes by traversing the scenes manually using the keyboard. 
We sample three trajectories per scene with diverse starting points and capture the scene content to simulate a realistic capturing scenario \ie~for instance a human moving a mobile device.
For each step that the agent takes, it moves $0.05$ m along the x- or y-axis. 
When the agent rotates, it rotates $2.5$ degrees.
The step sizes were chosen so that the dataset also can be used to evaluate multi-sensor tracking methods in the future.
The agent is equipped with two pairs of RGBD cameras. Both cameras are located at a fixed height of $1.5$ m above the floor and the baseline between the cameras is $0.1$ m. We use an identical resolution of 512$\times$512 for both cameras and a field of view of 90 degrees. 
The full dataset comprises $92698$ frames, of which we use a subset to produce a training, validation and testing set. For example, a dense voxel grid of the $apartment \ 0$ scene did not fit on the GPU, and the variations of the \emph{frl apartment} scenes did not provide any further diversity in the data. The train set consists of the scenes $\{$\emph{apartment 1, frl apartment 0, office 1, room 2, office 3, room 0}$\}$ and contains $22358$ frames, the validation set consists of the scene $\{$\emph{frl apartment 1}$\}$ and contains $1958$ frames and the test set consists of the scenes $\{$\emph{office 0, office 4, hotel 0}$\}$ and contains $1891$ frames. When training our model, we use all three trajectories from each train scene. During validation, only trajectory $1$ is used. 
During testing, we use trajectory 3 for \emph{hotel 0} and \emph{office 4} and trajectory 1 for \emph{office 0}. As an example, in Fig.~\ref{fig:trajectories}, we visualize a top-down view of the three manually traversed trajectories for the \emph{room 0} scene. The trajectory is visualized in red and the navigable floor is white. 

\section*{F. Scene3D Dataset}
\label{sec:scene3d}
\ifeccv
\else
\addcontentsline{toc}{section}{\nameref{sec:scene3d}}
\fi
The Scene3D dataset comprises multiple scenes, but we only evaluate the Copy room scene. We found that the ground truth meshes of all other scenes were not complete enough, which made the evaluation inaccurate.

\section*{G. Baselines}
\label{sec:baselines}
\ifeccv
\else
\addcontentsline{toc}{section}{\nameref{sec:baselines}}
\fi

\begin{figure}[t]
\centering
\scriptsize
\setlength{\tabcolsep}{1pt}
\renewcommand{\arraystretch}{1}
\begin{tabular}{cccc}
\rotatebox[origin=c]{90}{\makecell{Lounge}} & 
\includegraphics[align=c, width=.34\linewidth]{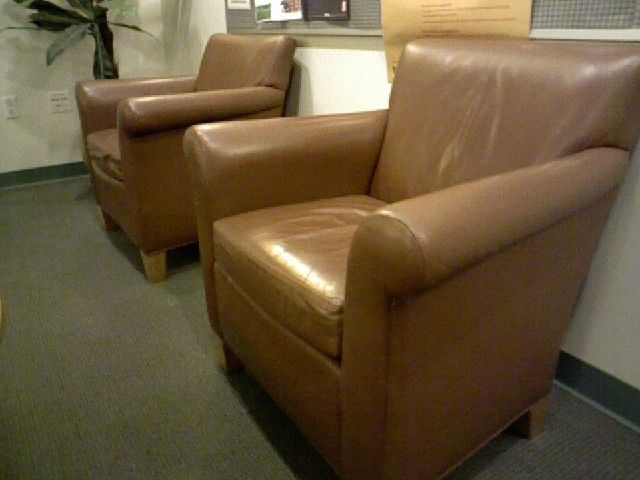} & 
\includegraphics[align=c, width=.31\linewidth]{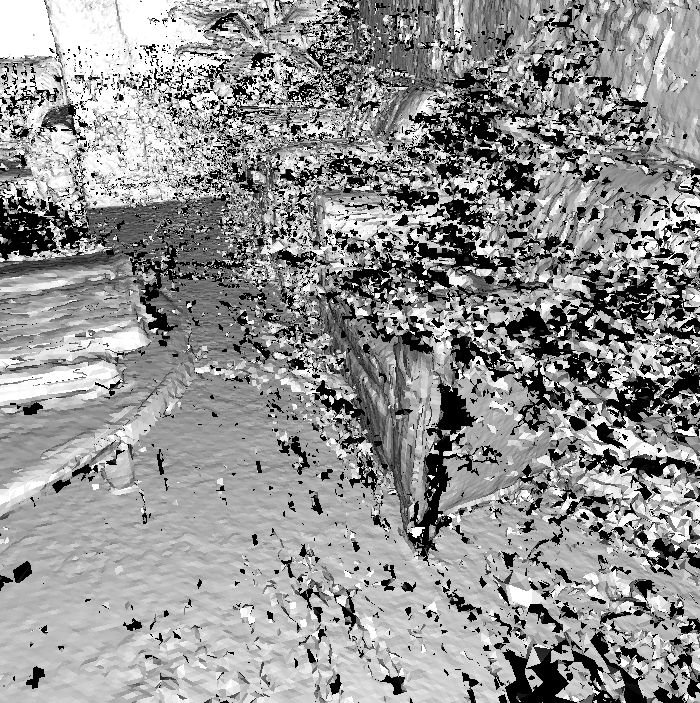} &
\includegraphics[align=c,width=.31\linewidth]{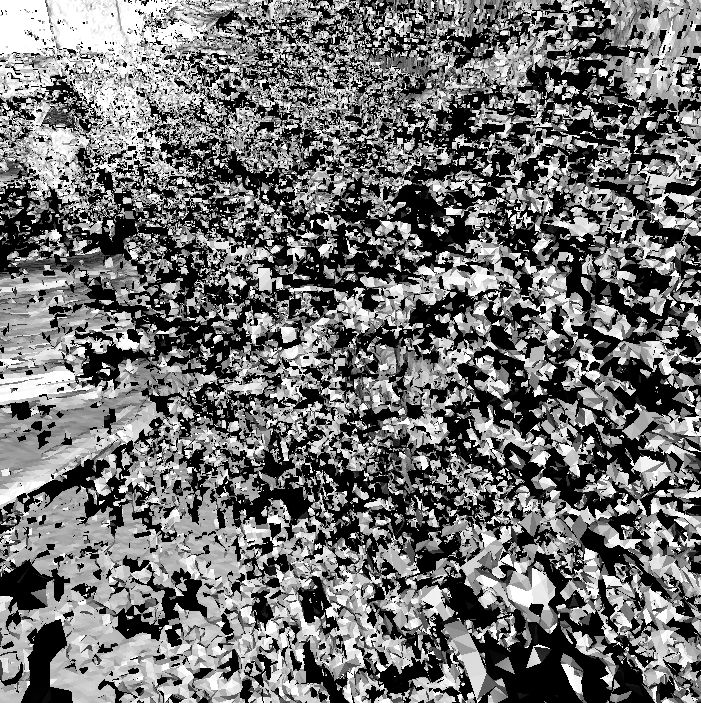} \\
 & Frame & Nearest & Trilinear  \\
 & & Neighbor Mask & Interpolation Mask \\
\end{tabular}
\caption{\boldparagraph{RoutedFusion Masking.} The trilinear interpolation mask of RoutedFusion~\cite{Weder2020RoutedFusionLR} results in significantly more outliers than using the nearest neighbor mask. Trilinear interpolation updates eight grid points per sampled point along the ray instead of two. This exposes more outliers as marching cubes~\cite{lorensen1987marching} requires that all eight grid points have a non-zero weight for a surface to be drawn.}
\label{fig:routedfusion_masking}
\end{figure}

\boldparagraph{RoutedFusion.}
We found that the original implementation of RoutedFusion generates significantly more outliers than TSDF Fusion~\cite{curless1996volumetric} when no weight thresholding is applied as post-outlier filter. This is due to the trilinear interpolation extraction step of RoutedFusion compared to the nearest neighbor extraction of TSDF Fusion. Trilinear interpolation updates eight grid points per sampled point along the ray instead of one. This exposes more outliers as marching cubes~\cite{lorensen1987marching} requires that all eight grid points have a non-zero weight for a surface to be drawn. Fig.~\ref{fig:routedfusion_masking} compares the meshes of two masks (the sets of non-zero weights) on the same RoutedFusion model. The trilinear interpolation mask is the standard output of RoutedFusion while the nearest neighbor mask is taken from running TSDF Fusion on the same scene. Given the significantly better result with the nearest neighbor mask, we report all results in the paper using the nearest neighbor mask.

\boldparagraph{Early Fusion.}
We use the denoising network described by Weder~\etal \cite{Weder2020RoutedFusionLR} (called routing network) as basis for our Early Fusion baseline. We increase the number of input channels by one so that the sensor depth maps can be fused and we use the loss hyperparameter $\lambda=0.06$.

\boldparagraph{DI-Fusion.} For a fair comparison between all models, we make the following modifications to the original implementation of DI-Fusion~\cite{huang2021di}. 1) We turn off the camera tracker and provide the ground truth camera poses. 2) We turn off the heuristic pre-outlier filter used on the incoming depth maps. 3) We turn off the heuristic weight counter thresholding post-outlier filter. 4) We use a voxel size of 2 cm, but sample the grid at a simulated resolution of 1 cm. This is achieved by setting the \texttt{resolution} variable in the config file to 2. We note that the implementation does not support \texttt{resolution < 2}. Furthermore, the implementation does not allow for convenient access the dense voxel grid and thus, we only report the mesh metrics.

\section*{H. Depth Sensor Details}
\label{sec:sensors}
\ifeccv
\else
\addcontentsline{toc}{section}{\nameref{sec:sensors}}
\fi
\boldparagraph{Synthesized ToF Depth.}
The noise model\footnote{\label{footnote:tof_supp}\normalfont\url{http://redwood-data.org/indoor/dataset.html}}~\cite{handa2014benchmark} incorporates disparity-based quantization, high-frequency noise, and a model of low-frequency distortion estimated on a real depth camera. We increase the noise level of the depth by increasing the standard deviation of the high-frequency noise by a factor of $5$. We also multiply the standard deviation of the pixel shuffling with the same factor. Other works have previously used this model for the evaluation of dense surface reconstruction methods~\cite{choi2015robust,zhou2014simultaneous,zhou2013elastic}.

\boldparagraph{PSMNet Stereo Depth.}
We first pretrain PSMNet~\cite{chang2018pyramid} on the SceneFlow dataset according to the documentation provided by Chang \etal.
The model is then fine-tuned on our Replica dataset. 
During training we use the default parameters.

\boldparagraph{SGM Stereo Depth.}
We generate depth maps with Semi-Global Matching~\cite{hirschmuller2007stereo} implemented in OpenCV~\cite{opencv_library}.
We set the number of disparities $numDisparities=64$ and use the full variant of the algorithm which considers 8 directions instead of 5 by setting $mode=MODE\_HH$.
Otherwise, we use the default settings.

\boldparagraph{COLMAP MVS Depth.}
Dense depth maps are computed with known camera poses and intrinsics using the sequential matcher with a 10 image overlap. 

\section*{I. More Experiments}
\label{sec:exp}
\ifeccv
\else
\addcontentsline{toc}{section}{\nameref{sec:exp}}
\fi

\ifeccv
\else
\boldparagraph{Loss Ablation.}  In Tab.~\ref{tab:single_sensor_supervision}, we show the performance difference when the model is trained only with the term $\mathcal{L}_f$ compared to the full loss (\textcolor{red}{4}) in the main paper. The terms (\textcolor{red}{6}) and (\textcolor{red}{7}) in the main paper clearly help improve overall performance and specifically to filter outliers.
\fi

\ifeccv
\else
\begin{table}
\centering
\resizebox{0.75\columnwidth}{!}
{
\begin{tabular}{l|lllllll}
\cellcolor{gray}Loss $\mathcal{L}$       \cellcolor{gray}& \cellcolor{gray}MSE$\downarrow$      & \cellcolor{gray}MAD$\downarrow$  & \cellcolor{gray}IoU$\uparrow$     & \cellcolor{gray}Acc.$\uparrow$ & \cellcolor{gray}F$\uparrow$ & \cellcolor{gray}P$\uparrow$ & \cellcolor{gray}R$\uparrow$   \\ 
\cellcolor{gray} & \cellcolor{gray}*e-04 & \cellcolor{gray}*e-02 & \cellcolor{gray}[0,1] & \cellcolor{gray}$[\%]$ & \cellcolor{gray}$[\%]$ & \cellcolor{gray}$[\%]$ & \cellcolor{gray}$[\%]$ \\\hline
Only $\mathcal{L}_f$ & 6.04 & 1.78 & 0.710 & 86.20 & 62.65 & 50.88 & \textbf{82.17} \\ 
Full Loss  & \textbf{4.77} & \textbf{1.56}  & \textbf{0.738}  & \textbf{87.62}  & \textbf{69.83} & \textbf{63.20} & 79.12  \\ 
\end{tabular}
}
\caption{\textbf{Loss Ablation.} When only the term $\mathcal{L}_f$ is used, we observe a significant performance drop compared to the full loss. Note, however, that only with the term $\mathcal{L}_f$, our model still improves on the single sensor input metrics compared to Tab.~\textcolor{red}{3} in the main paper. This shows all terms are helpful during training.}
\label{tab:single_sensor_supervision}
\end{table}
\fi

\ifeccv
\boldparagraph{Time Asynchronous Evaluation.} RGB cameras often have higher frame rates than ToF sensors which makes Early Fusion more challenging as one sensor might lack new data. Tab.~2 in the main paper provides performance results when the ToF sensor has half the sampling rate compared to the PSMNet sensor. In Tab.\ref{tab:asynchronous}, we show that the performance gap to our method grows even more when the sampling rate is decreased to a third (of the PSMNet sensor) for the ToF sensor. The drop in performance compared to our method can be attributed to the reprojection of the ToF frames. The reprojection step introduces occlusions and pixel quantization errors. The performance of our method actually slightly improves on some metrics. This can be explained by the fact that the completeness of the scene is saturated and dropping ToF frames removes noise and outliers that would otherwise have been integrated. See also Fig.\ref{fig:asynch} for a visualization. We do not retrain any model for this experiment.

\begin{wraptable}[8]{R}{0.63\linewidth}
\centering
\renewcommand{\arraystretch}{1.05}
\resizebox{\linewidth}{!}
{
\begin{tabular}{c|l|lllllll}
\cellcolor{gray} Sampling        & \cellcolor{gray}        & \cellcolor{gray}MSE$\downarrow$      & \cellcolor{gray}MAD$\downarrow$  & \cellcolor{gray}IoU$\uparrow$     & \cellcolor{gray}Acc.$\uparrow$ & \cellcolor{gray}F$\uparrow$ & \cellcolor{gray}P$\uparrow$ & \cellcolor{gray}R$\uparrow$   \\ 
\cellcolor{gray} Rate ToF & \multirow{-2}{*}{\cellcolor{gray} \backslashbox[28mm]{Model}{Metric}} & \cellcolor{gray}*e-04 & \cellcolor{gray}*e-02 & \cellcolor{gray}[0,1] & \cellcolor{gray}$[\%]$ & \cellcolor{gray}$\%$ & \cellcolor{gray}$[\%]$ & \cellcolor{gray}$[\%]$ \\\hline
1/2 & Early Fusion & 7.66 & 1.99 & 0.642 & 84.65 & 61.34 & 48.47 & \textbf{83.63}  \\ 
1/3 & Early Fusion & 8.52 & 2.15 & 0.610 & 83.60 & 55.52 & 41.63 & 83.40 \\ 
1/2 & SenFuNet (Ours) & 4.21 & 1.45 & \textbf{0.755} & \textbf{88.26} & 73.04 & 69.13 & 78.43 \\ 
1/3 & SenFuNet (Ours) & \textbf{4.00} & \textbf{1.41} & 0.751 & 88.14 & \textbf{74.93} & \textbf{73.57} & 77.05 \\ 
\end{tabular}
}
\caption{\textbf{Time Asynchronous Evaluation.} The gap between SenFuNet and Early Fusion increases further when the ToF camera has a sampling rate of one third compared to the PSMNet sensor.} 
\label{tab:asynchronous}
\end{wraptable}
\fi

\begin{figure}[t]
\centering
\setlength{\tabcolsep}{1pt}
\renewcommand{\arraystretch}{1}
\begin{tabular}{cc}
 \includegraphics[align=c, width=.5\linewidth]{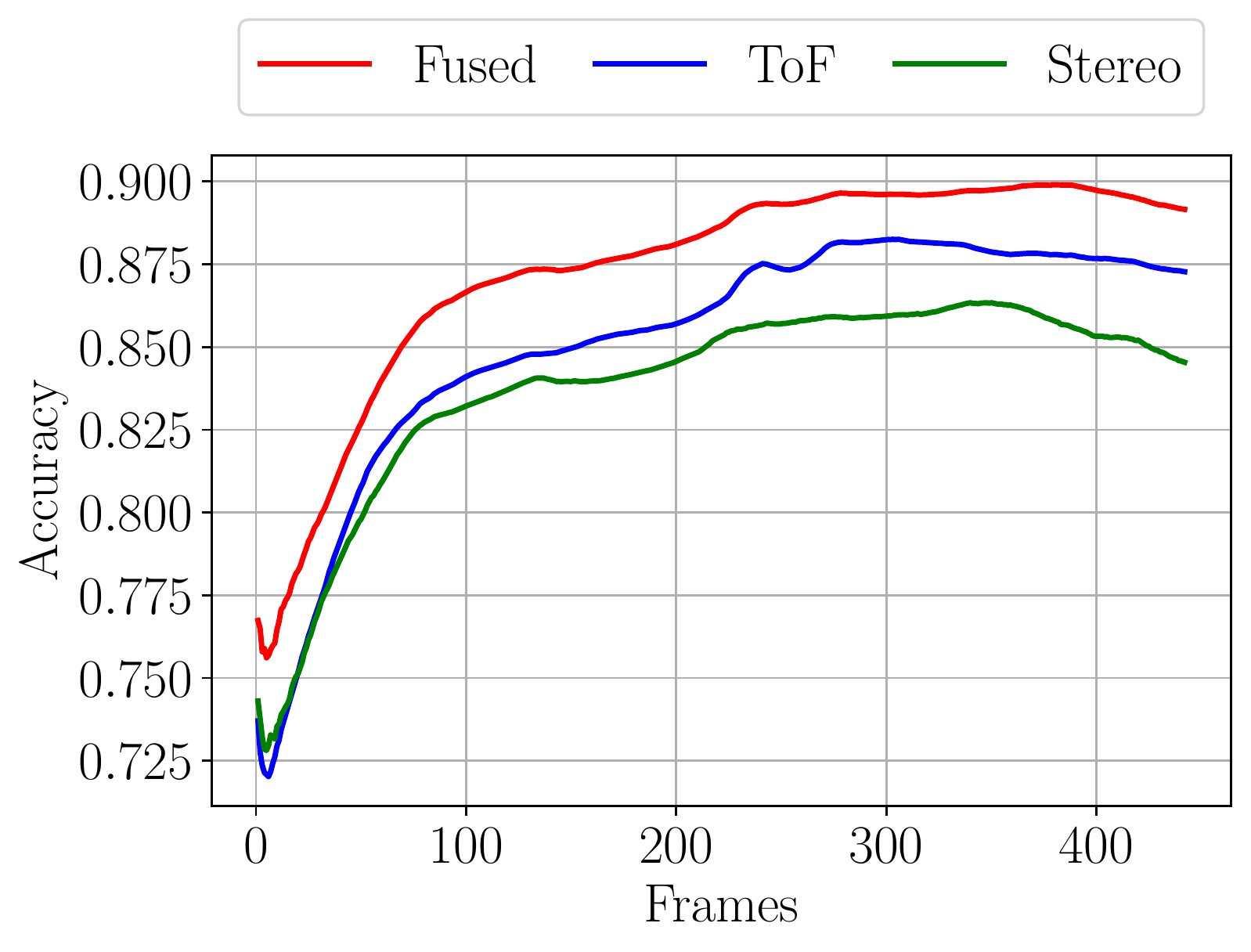} & \includegraphics[align=c, width=.5\linewidth]{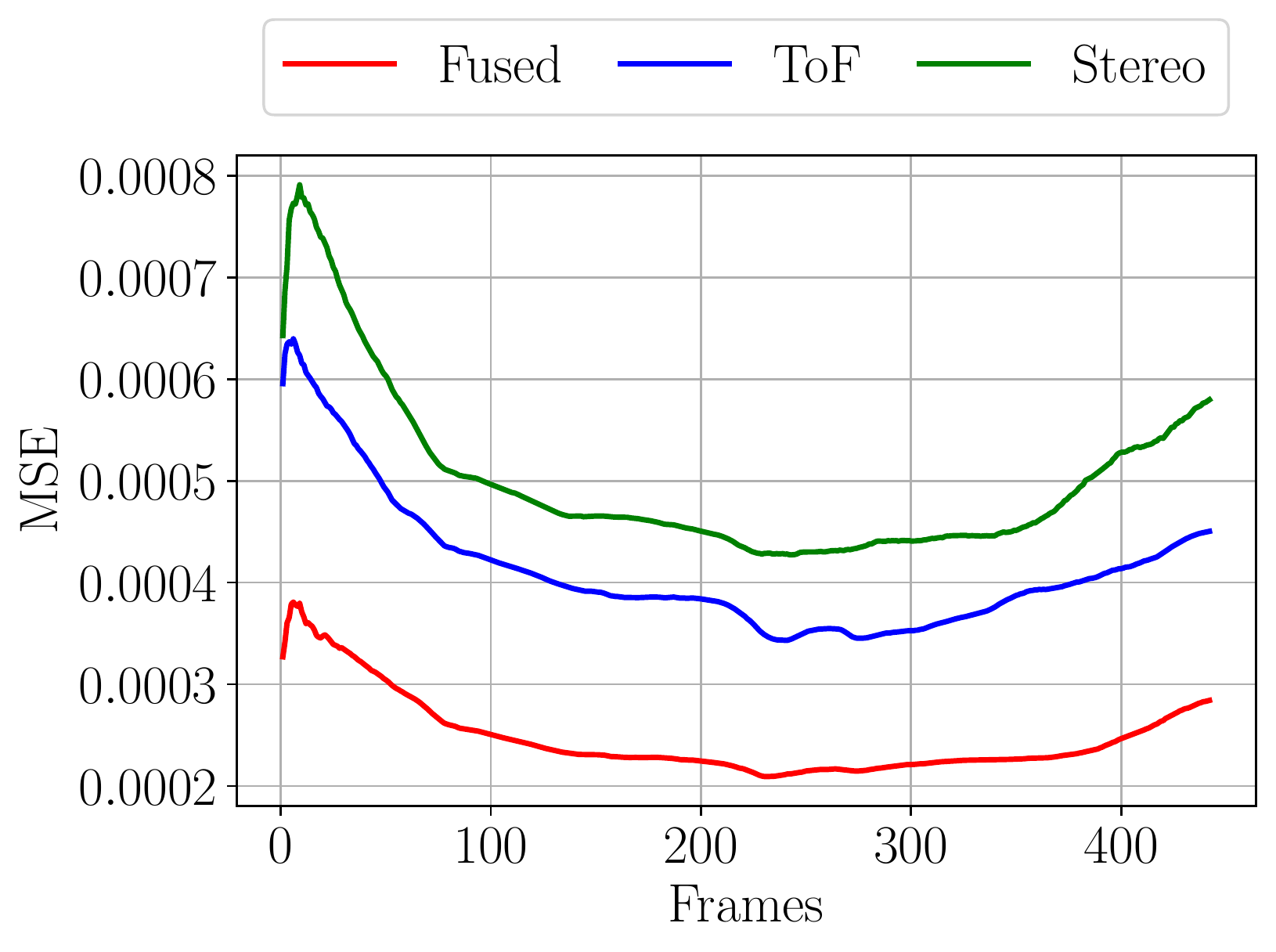} \\
\end{tabular}
\caption{\textbf{Performance over Camera Trajectory.} The fused output of our method outperforms the single-input reconstructions for all frames along the trajectory.
The experiment was done on the \emph{office 0} scene for the sensors $\{$ToF, PSMNet$\}$ with denoising. Note that the results get slightly worse after 300 frames. This is due to additional noise from the depth sensors when viewing the scene from further away.}
\label{fig:plot_trajectory_fusion_supp}
\end{figure}

\boldparagraph{Performance over Camera Trajectory.} To show that our fused output is not only better at the end of the fusion process, we visualize the quantitative performance across the accumulated trajectory for the \emph{office 0} scene for the model $\{$ToF, PSMNet$\}$ with denoising in Fig.~\ref{fig:plot_trajectory_fusion_supp}. Our fused model consistently improves on the inputs. Note that we only show the metrics IoU and MAD in the main paper.

\boldparagraph{Weight Ablation.}
In Tab.~\ref{tab:ablation_tanh_transform} we evaluate our full model against two models which only use the weight ($W_t^i$) as input to the Weighting Network $\mathcal{G}$. We perform $\{$SGM, PSMNet$\}$ fusion without denoising on the Replica dataset. We observe that our full model provides a gain on the model which only uses the $\mathop{\mathrm{tanh}}$-transformed weights. This justifies our Feature Network. We also show that a normalization of the weight with a $\mathop{\mathrm{tanh}}$-transformation improves performance over no normalization. 

\ifeccv
\else
\begin{table}
\centering
\footnotesize
\setlength{\tabcolsep}{15pt}
\begin{tabular}{l|lllll}
\cellcolor{gray}Model & \cellcolor{gray}F$\uparrow$ $[\%]$\\ 
\hline
Only Weights & 66.69 \\ 
Only \emph{tanh}(Weights) & 68.92 \\ 
Full Model & \textbf{69.83} \\ 
\end{tabular}
\caption{\boldparagraph{Weight Counter Ablation.} Our full model outperforms models which only use the weight ($W_t^i$) as input to the Weighting Network $\mathcal{G}$. Normalization of the weight with a \emph{tanh}-transformation improves performance over no normalization.}
\label{tab:ablation_tanh_transform}
\end{table}
\fi

\boldparagraph{Architecture Ablation.}
For the architecture ablations, we perform $\{$SGM, PSMNet$\}$ fusion without denoising on the Replica dataset. For all experiments, unless otherwise specified, we use $4$ layers of 3D-convolutions with kernel size 3 in $\mathcal{G}$, $6$ network blocks in the Feature Networks $\mathcal{F}^i$, and store feature vectors of dimension $n = 5$ in the feature grids $F^i_t$.

In Tab.~\ref{tab:ablation_blocks_feature_net}, we investigate the effect of using different number of network blocks in the feature network.
Performance is maximized when using $5$ blocks.

In Tab.~\ref{tab:ablation_feature_dims} we vary the number of feature dimensions that is stored in the grids. $4$ dimensions yield optimal performance.

Finally, in Tab.~\ref{tab:ablation_bypass_feature_net}, we study the importance of the Feature Network by testing it against a model which bypasses the network and hence unprojects the 2D features without any 2D processing. 
For this experiment, both models use $4$ feature dimensions to make the comparison fair. Note that no weights were used as input to the Weighting Network.
We gain performance by using the feature network, which justifies our design choice. 

\ifeccv
\else
\begin{table}
\centering
\resizebox{\columnwidth}{!}
{
\begin{tabular}{l|llllll}
\cellcolor{gray}Nbr Blocks & \cellcolor{gray}1 & \cellcolor{gray}2 & \cellcolor{gray}3 & \cellcolor{gray}4 & \cellcolor{gray}5 & \cellcolor{gray}6 \\ 
\hline
F$\uparrow$ $[\%]$ & 67.45 & 67.68 & 66.79 & 67.59 & \textbf{68.39} & 68.21 \\ 
\end{tabular}
}
\caption{\boldparagraph{Ablation Study.} We change the number of blocks for the feature network.
Optimal performance is achieved with $5$ blocks.}
\label{tab:ablation_blocks_feature_net}
\end{table}
\fi

\ifeccv
\begin{table}[ht]
\begin{minipage}[t]{0.495\linewidth}
\centering
\footnotesize
\setlength{\tabcolsep}{15pt}
\begin{tabular}{l|lllll}
\cellcolor{gray}Model & \cellcolor{gray}F$\uparrow$ $[\%]$\\ 
\hline
Only Weights & 66.69 \\ 
Only \emph{tanh}(Weights) & 68.92 \\ 
Full Model & \textbf{69.83} \\ 
\end{tabular}
\caption{\boldparagraph{Weight Counter Ablation.} Our full model outperforms models which only use the weight ($W_t^i$) as input to the Weighting Network $\mathcal{G}$. Normalization of the weight with a \emph{tanh}-transformation improves performance over no normalization.}
\label{tab:ablation_tanh_transform}
\end{minipage}
\hspace{0.1cm}
\begin{minipage}[t]{0.495\linewidth}
\centering
\resizebox{\columnwidth}{!}
{
\begin{tabular}{l|llllll}
\cellcolor{gray}Nbr Blocks & \cellcolor{gray}1 & \cellcolor{gray}2 & \cellcolor{gray}3 & \cellcolor{gray}4 & \cellcolor{gray}5 & \cellcolor{gray}6 \\ 
\hline
F$\uparrow$ $[\%]$ & 67.45 & 67.68 & 66.79 & 67.59 & \textbf{68.39} & 68.21 \\ 
\end{tabular}
}
\caption{\boldparagraph{Ablation Study.} We change the number of blocks for the feature network.
Optimal performance is achieved with $5$ blocks.}
\label{tab:ablation_blocks_feature_net}
\end{minipage}
\begin{minipage}[t]{0.495\linewidth}
\centering
\footnotesize
\setlength{\tabcolsep}{10pt}
\resizebox{\columnwidth}{!}
{
\begin{tabular}{l|llll}
\cellcolor{gray}n & \cellcolor{gray}2 & \cellcolor{gray}3 & \cellcolor{gray}4 & \cellcolor{gray}5 \\ 
\hline
F$\uparrow$ $[\%]$ & 68.86 & 68.20 & \textbf{68.87} & 68.21 \\ 
\end{tabular}
}
\caption{\boldparagraph{Ablation Study.} We alter the dimension of the feature vector.
The performance is maximized at $n = 4$.}
\label{tab:ablation_feature_dims}
\end{minipage}
\hspace{0.1cm}
\begin{minipage}[t]{0.495\linewidth}
\centering
\footnotesize
\setlength{\tabcolsep}{15pt}
\resizebox{1.0\columnwidth}{!}
{
\begin{tabular}{l|lllll}
\cellcolor{gray}Model & \cellcolor{gray}F$\uparrow$ $[\%]$ \\ 
\hline
Without Feature Net & 67.44 \\ 
With Feature Net & \textbf{68.87} \\ 
\end{tabular}
}
\caption{\boldparagraph{Architecture Ablation.} We demonstrate the difference in performance with and without the feature network.}
\label{tab:ablation_bypass_feature_net}
\end{minipage}
\end{table}
\fi

\ifeccv
\else
\begin{table}
\centering
\footnotesize
\setlength{\tabcolsep}{10pt}
\begin{tabular}{l|llll}
\cellcolor{gray}n & \cellcolor{gray}2 & \cellcolor{gray}3 & \cellcolor{gray}4 & \cellcolor{gray}5 \\ 
\hline
F$\uparrow$ $[\%]$ & 68.86 & 68.20 & \textbf{68.87} & 68.21 \\ 
\end{tabular}

\caption{\boldparagraph{Ablation Study.} We alter the dimension of the feature vector.
The performance is maximized at $n = 4$.}
\label{tab:ablation_feature_dims}
\end{table}
\fi

\ifeccv
\else
\begin{table}
\centering
\footnotesize
\setlength{\tabcolsep}{15pt}
\begin{tabular}{l|lllll}
\cellcolor{gray}Model & \cellcolor{gray}F$\uparrow$ $[\%]$ \\ 
\hline
Without Feature Net & 67.44 \\ 
With Feature Net & \textbf{68.87} \\ 
\end{tabular}
\caption{\boldparagraph{Architecture Ablation.} We demonstrate the difference in performance with and without the feature network.}
\label{tab:ablation_bypass_feature_net}
\end{table}
\fi

The ablations suggest that the best performance is achieved with $n = 4$, the number of blocks in the feature networks is $5$ and when we use $2$ 3D-convolutional layers with kernel size 3.
Empirically, we found that $n = 5$, $6$ feature network blocks and $2$ kernel size 3 3D convolutions gave marginally better results and this is what we report throughout the paper.

\boldparagraph{Effect of Weight Thresholding.} 
RoutedFusion~\cite{Weder2020RoutedFusionLR} applies weight thresholding to filter outliers \ie~a TSDF surface observation needs to have a weight larger than some threshold to not be removed during post-processing. Weight thresholding was not applied to any model in the main paper to avoid the need of tuning an additional parameter and to make the comparison fair between the models. For example, on the Replica scenes, the optimal weight threshold is typically within the range 1-10, while for the CoRBS human scene it is around 500. Tab.~\ref{tab:weight_thresholding} shows the effect when weight thresholding is applied on the Replica test set on the sensors $\{$ToF, PSMNet$\}$ with denoising. Regardless of the weight threshold, our method outperforms RoutedFusion.

\boldparagraph{DI-Fusion Ablation.} 
Tab.~\ref{tab:di_fusion} shows the performance of DI-Fusion~\cite{huang2021di} for $\sigma=\{0.15, 0.06\}$ compared to our method. 
The results when $\sigma=0.15$ is reported in the main paper and this number is also specified in the implementation provided by the authors. 
$\sigma=0.06$ is suggested for one experiment in the DI-Fusion paper.
SenFuNet outperforms DI-Fusion for both choices of the $\sigma$ threshold. 
In general, a high $\sigma$ yields high recall, but poor precision and vice versa. 
This is, however, not true for the Human scene where SenFuNet outperforms DI-Fusion on all metrics.
On the copyroom scene, SenFuNet outperforms DI-Fusion both in terms of the F-score and precision. 
On the Replica dataset, SenFuNet attains around 10 percent points higher F-score compared to the best DI-Fusion model.

\ifeccv
\else
\begin{table}
\centering
\resizebox{\columnwidth}{!}
{
\setlength{\tabcolsep}{2pt}
\renewcommand{\arraystretch}{1.05}
\begin{tabular}{l|lll|lll}
\cellcolor{gray}Model   & \cellcolor{gray}F$\uparrow$ $[\%]$ & \cellcolor{gray}P$\uparrow$ $[\%]$ & \cellcolor{gray}R$\uparrow$ $[\%]$ & \cellcolor{gray}F$\uparrow$ $[\%]$ & \cellcolor{gray}P$\uparrow$ $[\%]$ & \cellcolor{gray}R$\uparrow$ $[\%]$  \\ \hline
\multicolumn{1}{c}{\emph{ToF\bplus{}MVS}} & \multicolumn{3}{c}{\emph{Copyroom}} & \multicolumn{3}{c}{\emph{Human}}\\ \hline
DI-Fusion~\cite{huang2021di} $\sigma$=0.15 & 86.31 & 77.27 & \textbf{97.74} & 28.19 & 16.71 & 90.15 \\ 
DI-Fusion~\cite{huang2021di} $\sigma$=0.06 & 79.21 & 91.03 & 70.11 & 13.52 & 18.75 & 10.56 \\
\textbf{\ours{} (Ours)}  & \textbf{93.73} & \textbf{91.56} & 96.00 & \textbf{74.56} & \textbf{59.74} & \textbf{99.16} \\ 
\hline
\multicolumn{1}{c}{\emph{Replica: ToF\bplus{}PSMNet}} & \multicolumn{3}{c}{\emph{w/o denoising}} & \multicolumn{3}{c}{\emph{w. denoising}}\\ \hline
DI-Fusion~\cite{huang2021di} $\sigma$=0.15 & 48.39 & 34.24 & \textbf{85.29} & 55.66 & 41.49 & \textbf{85.33} \\ 
DI-Fusion~\cite{huang2021di} $\sigma$=0.06  & 60.30 & \textbf{72.49} & 51.88 & 63.02 & \textbf{75.14} & 54.46 \\ 
\textbf{\ours{} (Ours)} & \textbf{69.29} & 62.05 & 79.81 & \textbf{76.47} & 73.58 & 79.77 \\ 
\hline
\multicolumn{1}{c}{\emph{Replica: SGM\bplus{}PSMNet}} & \multicolumn{3}{c}{\emph{w/o denoising}} & \multicolumn{3}{c}{\emph{w. denoising}}\\ \hline
DI-Fusion~\cite{huang2021di} $\sigma$=0.15 & 47.29 & 32.92 & \textbf{85.14} & 52.65 & 38.50 & \textbf{83.62}\\ 
DI-Fusion~\cite{huang2021di} $\sigma$=0.06 & 59.24 & \textbf{71.73} & 50.61 & 63.39 & \textbf{75.59} & 54.73\\ 
\textbf{\ours{} (Ours)} & \textbf{69.83} & 63.20 & 79.12 & \textbf{71.18} & 66.81 & 76.27\\ 
\end{tabular}
}
\caption{\textbf{DI-Fusion Ablation.} SenFuNet outperforms DI-Fusion for various choices of the $\sigma$ threshold. In general, a high $\sigma$ yields high recall, but poor precision and vice versa. This is, however, not true for the Human scene where SenFuNet outperforms DI-Fusion on all metrics. On the Replica dataset, SenFuNet attains around a 10 higher F-score compared to the best DI-Fusion model.}
\label{tab:di_fusion}
\end{table}
\fi

\ifeccv
\begin{table}[ht]
\begin{minipage}[c]{0.5\linewidth}
\centering
\resizebox{\columnwidth}{!}
{
\setlength{\tabcolsep}{1pt}
\renewcommand{\arraystretch}{1.05}
\begin{tabular}{l|lll|lll}
\cellcolor{gray}Model   & \cellcolor{gray}F$\uparrow$ $[\%]$ & \cellcolor{gray}P$\uparrow$ $[\%]$ & \cellcolor{gray}R$\uparrow$ $[\%]$ & \cellcolor{gray}F$\uparrow$ $[\%]$ & \cellcolor{gray}P$\uparrow$ $[\%]$ & \cellcolor{gray}R$\uparrow$ $[\%]$  \\ \hline
\multicolumn{1}{c}{\emph{ToF\bplus{}MVS}} & \multicolumn{3}{c}{\emph{Copyroom}} & \multicolumn{3}{c}{\emph{Human}}\\ \hline
DI-Fusion~\cite{huang2021di} $\sigma$=0.15 & 86.31 & 77.27 & \textbf{97.74} & 28.19 & 16.71 & 90.15 \\ 
DI-Fusion~\cite{huang2021di} $\sigma$=0.06 & 79.21 & 91.03 & 70.11 & 13.52 & 18.75 & 10.56 \\
\textbf{\ours{} (Ours)}  & \textbf{93.73} & \textbf{91.56} & 96.00 & \textbf{74.56} & \textbf{59.74} & \textbf{99.16} \\ 
\hline
\multicolumn{1}{c}{\emph{Replica: ToF\bplus{}PSMNet}} & \multicolumn{3}{c}{\emph{w/o denoising}} & \multicolumn{3}{c}{\emph{w. denoising}}\\ \hline
DI-Fusion~\cite{huang2021di} $\sigma$=0.15 & 48.39 & 34.24 & \textbf{85.29} & 55.66 & 41.49 & \textbf{85.33} \\ 
DI-Fusion~\cite{huang2021di} $\sigma$=0.06  & 60.30 & \textbf{72.49} & 51.88 & 63.02 & \textbf{75.14} & 54.46 \\ 
\textbf{\ours{} (Ours)} & \textbf{69.29} & 62.05 & 79.81 & \textbf{76.47} & 73.58 & 79.77 \\ 
\hline
\multicolumn{1}{c}{\emph{Replica: SGM\bplus{}PSMNet}} & \multicolumn{3}{c}{\emph{w/o denoising}} & \multicolumn{3}{c}{\emph{w. denoising}}\\ \hline
DI-Fusion~\cite{huang2021di} $\sigma$=0.15 & 47.29 & 32.92 & \textbf{85.14} & 52.65 & 38.50 & \textbf{83.62}\\ 
DI-Fusion~\cite{huang2021di} $\sigma$=0.06 & 59.24 & \textbf{71.73} & 50.61 & 63.39 & \textbf{75.59} & 54.73\\ 
\textbf{\ours{} (Ours)} & \textbf{69.83} & 63.20 & 79.12 & \textbf{71.18} & 66.81 & 76.27\\ 
\end{tabular}
}
\caption{\textbf{DI-Fusion Ablation.} SenFuNet outperforms DI-Fusion for various choices of the $\sigma$ threshold. In general, a high $\sigma$ yields high recall, but poor precision and vice versa. This is, however, not true for the Human scene where SenFuNet outperforms DI-Fusion on all metrics. On the Replica dataset, SenFuNet attains around a 10 higher F-score compared to the best DI-Fusion model.}
\label{tab:di_fusion}
\end{minipage}
\hspace{0.1cm}
\begin{minipage}[b]{0.48\linewidth}
\centering
\footnotesize
\setlength{\tabcolsep}{1pt}
\resizebox{\columnwidth}{!}
{
\begin{tabular}{l|lllllll}
\cellcolor{gray}Weight Threshold & \cellcolor{gray}1 & \cellcolor{gray}3 & \cellcolor{gray}5 & \cellcolor{gray}7 & \cellcolor{gray}9 & \cellcolor{gray}11 \\ 
\hline
RoutedFusion~\cite{Weder2020RoutedFusionLR} F$\uparrow$ $[\%]$& 62.12 & 71.40 & 74.12 & 75.07 & 75.11 & 74.99 \\ 
Ours F$\uparrow$ $[\%]$ & \textbf{77.24} & \textbf{79.35} & \textbf{79.67} & \textbf{79.39} & \textbf{78.47} & \textbf{77.40} \\ 
\end{tabular}
}
\caption{\boldparagraph{Weight Thresholding.} Our model outperforms RoutedFusion on the Replica test set also when weight thresholding is applied.}
\label{tab:weight_thresholding}
\begin{minipage}[t]{\linewidth}
\centering
\footnotesize
\setlength{\tabcolsep}{1pt}
\resizebox{\columnwidth}{!}
{
\begin{tabular}{l|lllllll}
\cellcolor{gray}       & \cellcolor{gray}MSE$\downarrow$      & \cellcolor{gray}MAD$\downarrow$  & \cellcolor{gray}IoU$\uparrow$     & \cellcolor{gray}Acc.$\uparrow$ & \cellcolor{gray}F$\uparrow$ & \cellcolor{gray}P$\uparrow$ & \cellcolor{gray}R$\uparrow$ \\
\multirow{-2}{*}{\cellcolor{gray} \backslashbox[23mm]{Model}{Metric}} & \cellcolor{gray}*e-04 & \cellcolor{gray}*e-02 & \cellcolor{gray}[0,1] & \cellcolor{gray}$[\%]$ & \cellcolor{gray}$[\%]$ & \cellcolor{gray}$[\%]$ & \cellcolor{gray}$[\%]$ \\\hline
\multicolumn{8}{c}{\emph{Single Sensor}} \\ \hline
ToF~\cite{handa2014benchmark} & 7.48 & 1.99 & 0.664 & 83.65 & 58.52 & 45.84 & \textbf{84.85} \\ 
ToF denoising~\cite{handa2014benchmark}   & 5.08 & 1.58 & 0.709 & 87.32 & 68.93 & 59.01  & 83.08 \\ 
\hline
\multicolumn{8}{c}{\emph{Multi-Sensor Fusion}} \\ \hline
Ours & \textbf{3.19}  & \textbf{1.26}  & \textbf{0.791}  & \textbf{90.99}  & \textbf{78.87} & \textbf{76.56} & 81.68 \\ 
\end{tabular}
}
\caption{\textbf{ToF\bplus{}ToF denoising Fusion.} Our method learns to combine the preprocessed depth and the raw depth in a fashion that improves the overall reconstruction.}
\label{tab:tof_tof_denoising}
\end{minipage}
\end{minipage}
\end{table}
\fi

\ifeccv
\else
\begin{table}
\centering
\setlength{\tabcolsep}{3pt}
\resizebox{\columnwidth}{!}
{
\begin{tabular}{l|lllllll}
\cellcolor{gray}Weight Threshold & \cellcolor{gray}1 & \cellcolor{gray}3 & \cellcolor{gray}5 & \cellcolor{gray}7 & \cellcolor{gray}9 & \cellcolor{gray}11 \\ 
\hline
RoutedFusion~\cite{Weder2020RoutedFusionLR} F$\uparrow$ $[\%]$& 62.12 & 71.40 & 74.12 & 75.07 & 75.11 & 74.99 \\ 
Ours F$\uparrow$ $[\%]$ & \textbf{77.24} & \textbf{79.35} & \textbf{79.67} & \textbf{79.39} & \textbf{78.47} & \textbf{77.40} \\ 
\end{tabular}
}
\caption{\boldparagraph{Weight Thresholding.} Our model outperforms RoutedFusion on the Replica test set also when weight thresholding is applied.}
\label{tab:weight_thresholding}
\end{table}
\fi

\ifeccv
\begin{figure*}[t]
\centering
\tiny
\setlength{\tabcolsep}{1pt}
\renewcommand{\arraystretch}{1}
\begin{tabular}{ccccccc}
\rotatebox[origin=c]{90}{Office 0} & \includegraphics[align=c, width=.18\linewidth]{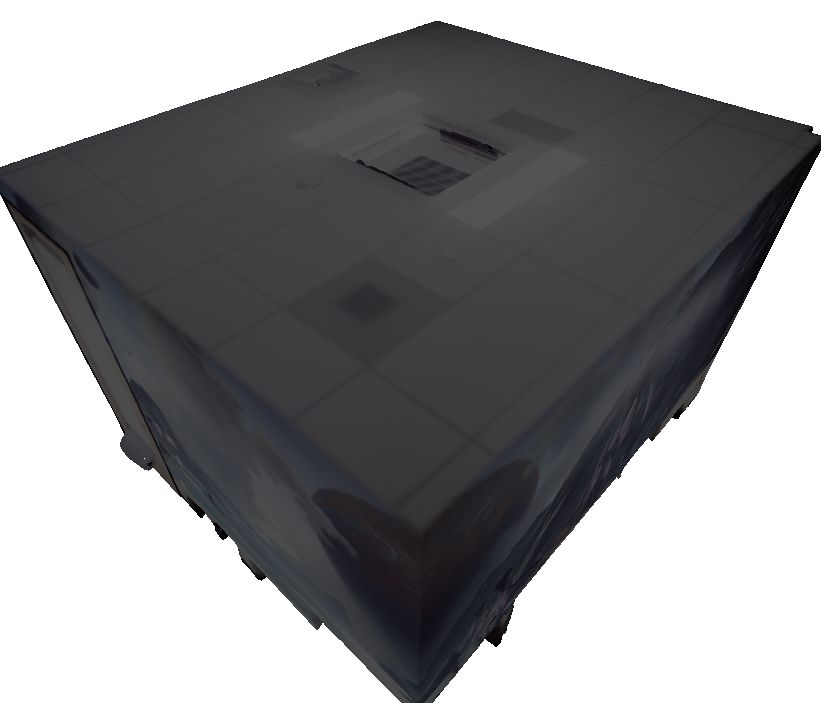} & \includegraphics[align=c, width=.18\linewidth]{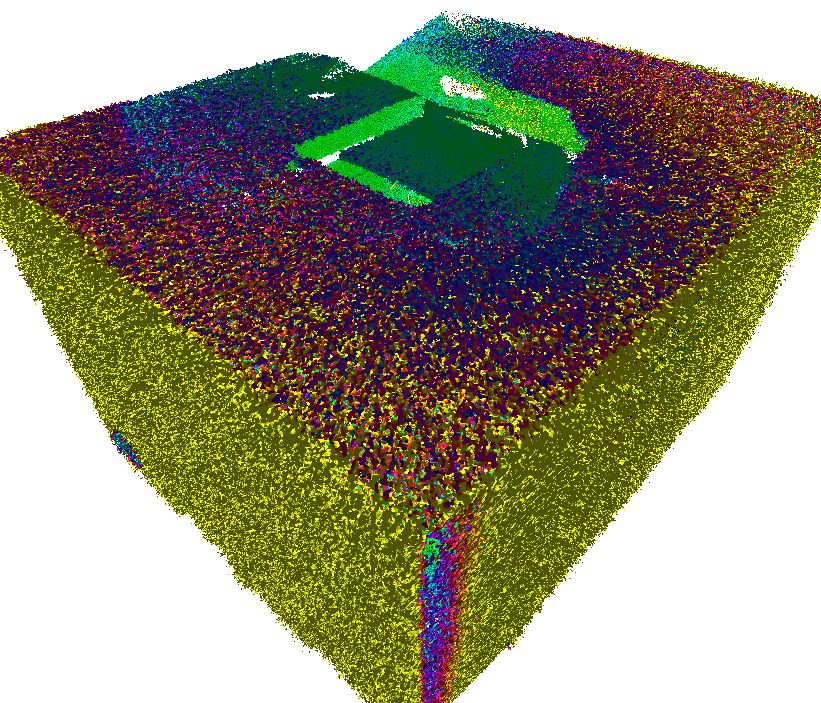} & \includegraphics[align=c, width=.18\linewidth]{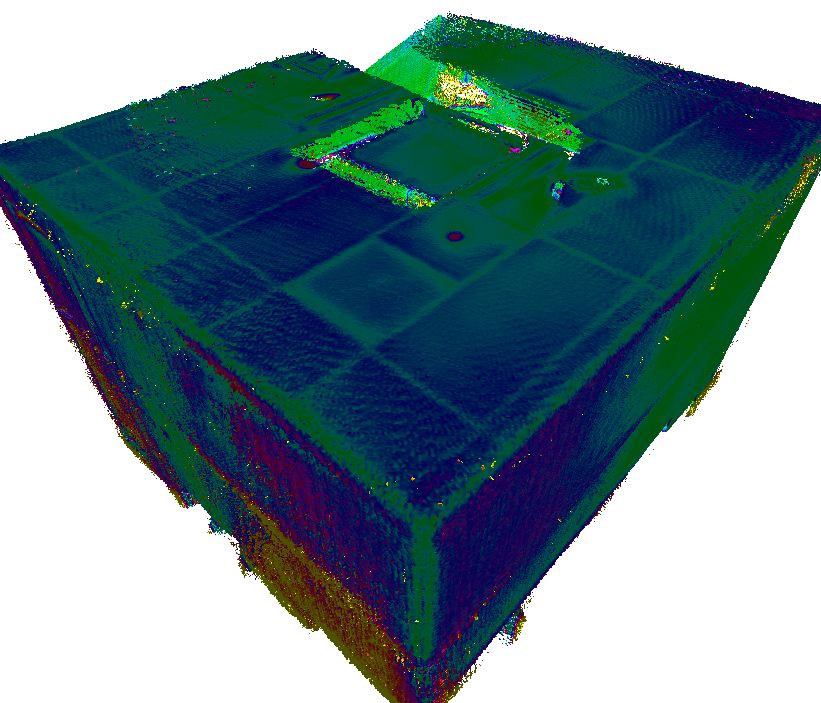} & \includegraphics[align=c, width=.18\linewidth]{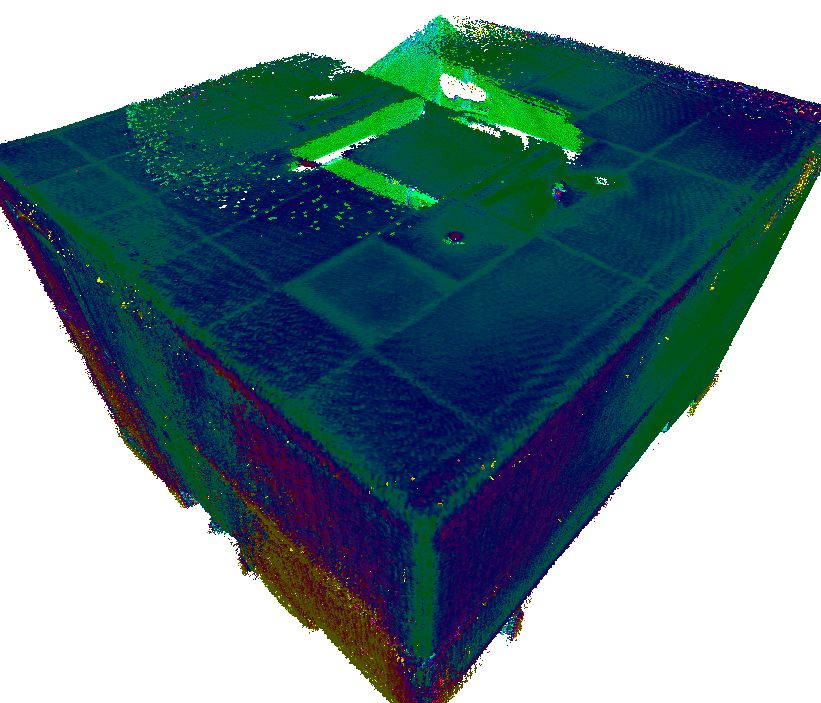} & \includegraphics[align=c, width=.18\linewidth]{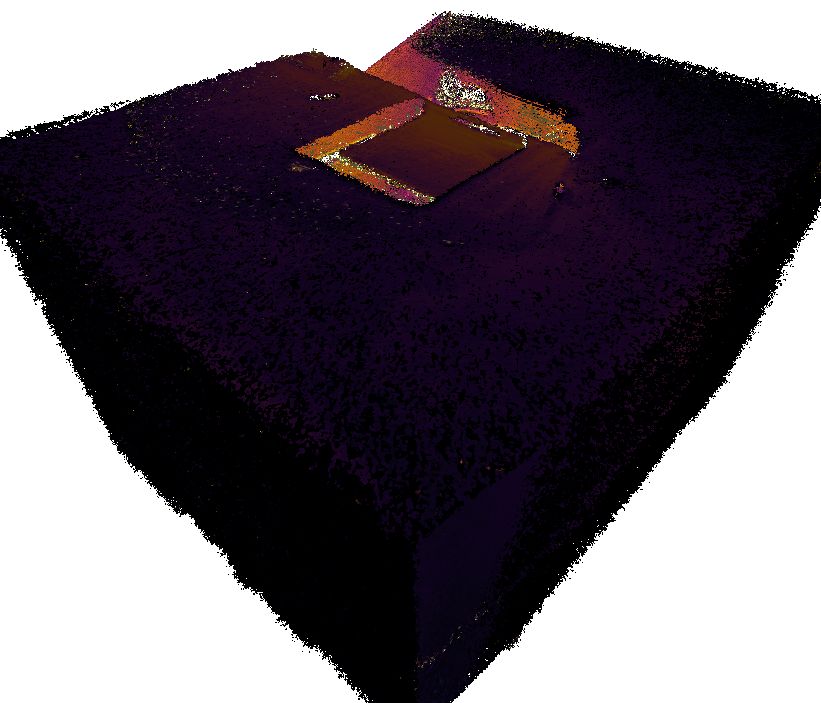} & \multirow[t]{12}{*}[25pt]{\includegraphics[align=t, width=.051\linewidth]{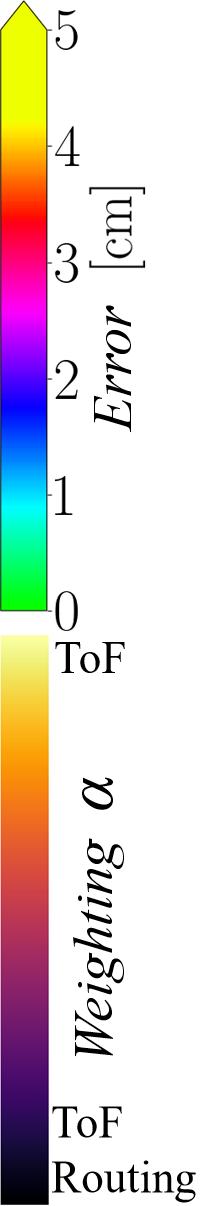}} \\
\rotatebox[origin=c]{90}{Office 0} & \includegraphics[align=c, width=.18\linewidth]{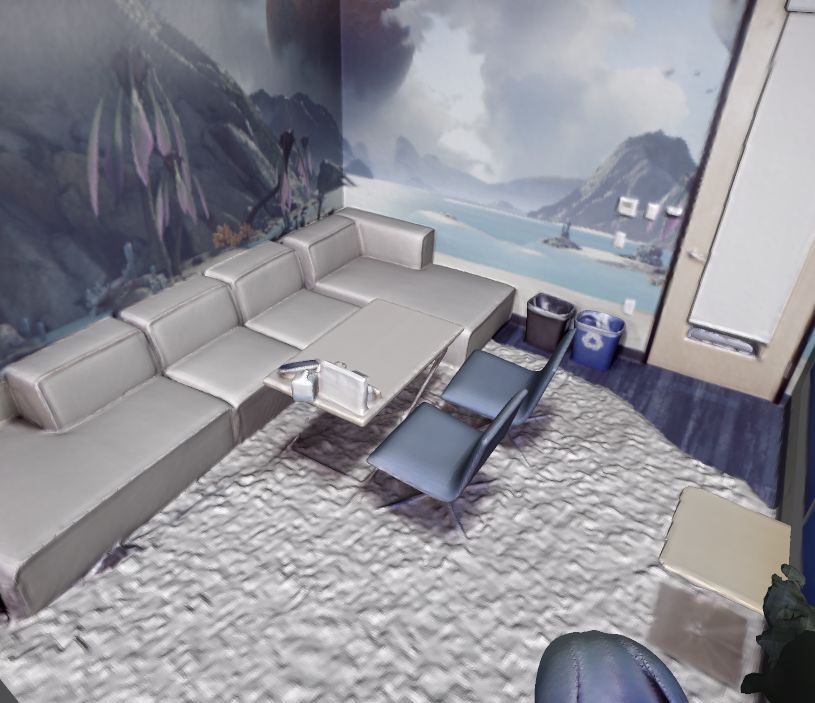} &  \includegraphics[align=c, width=.18\linewidth]{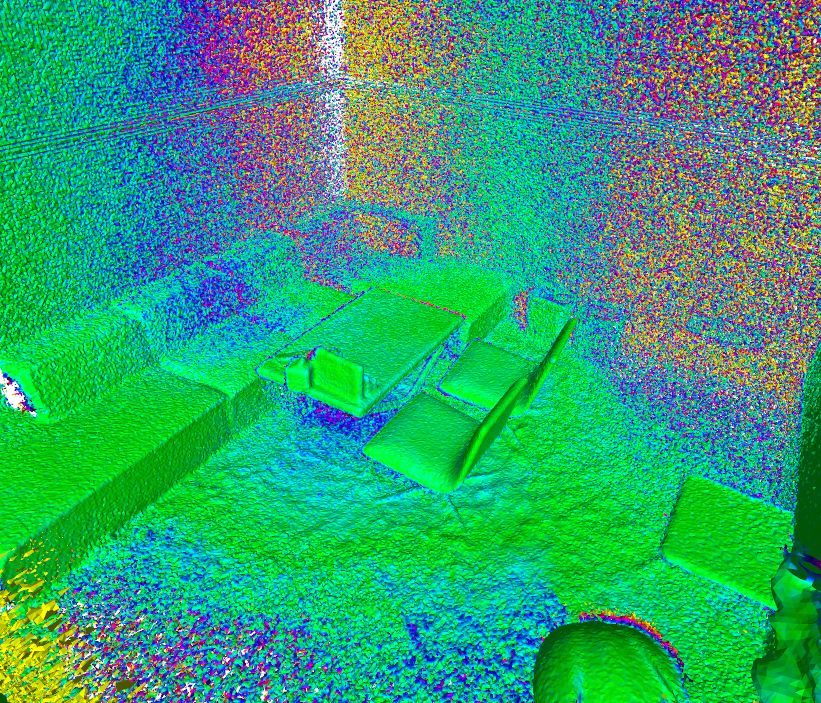} & \includegraphics[align=c, width=.18\linewidth]{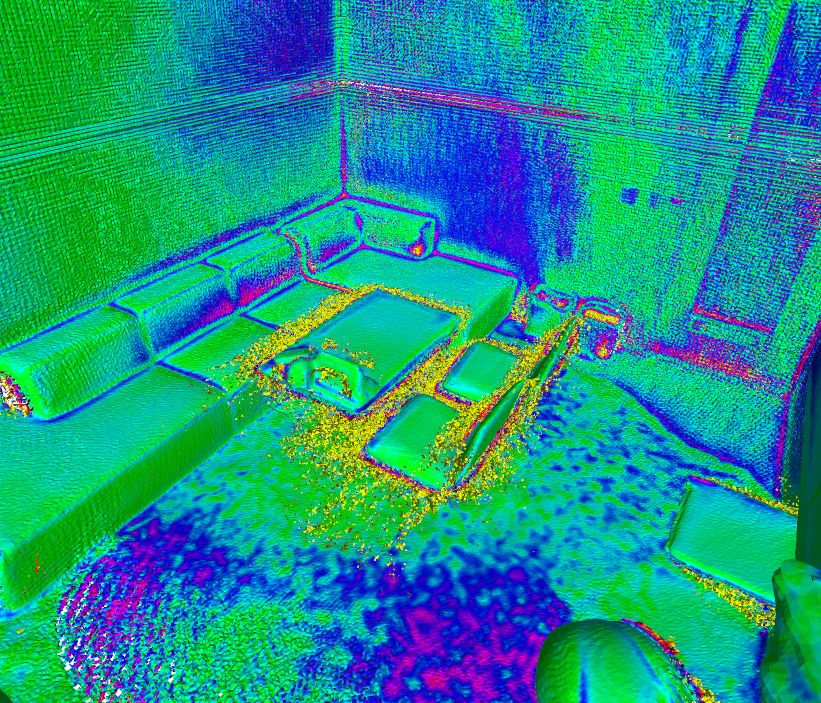} & \includegraphics[align=c, width=.18\linewidth]{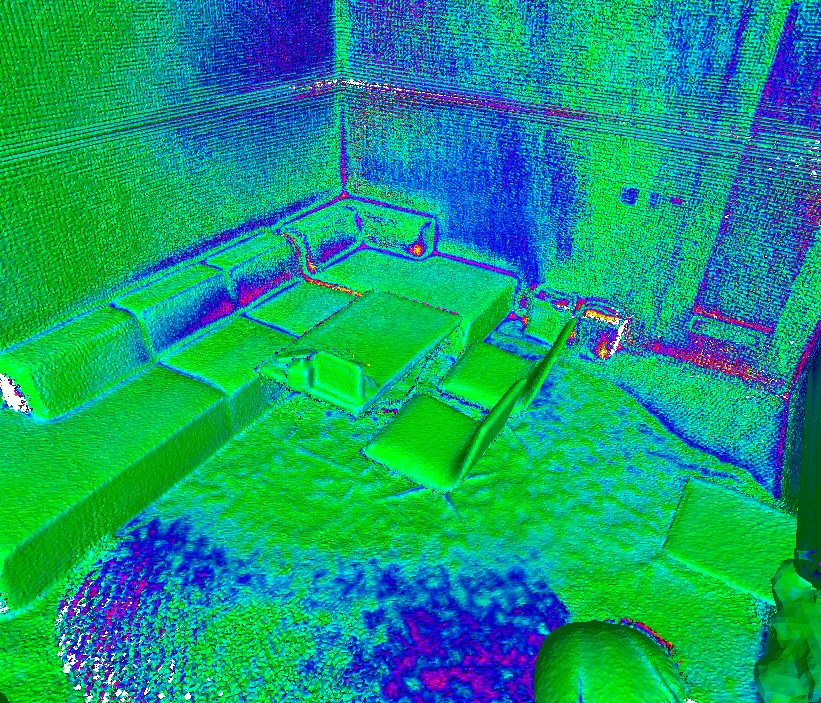} & \includegraphics[align=c, width=.18\linewidth]{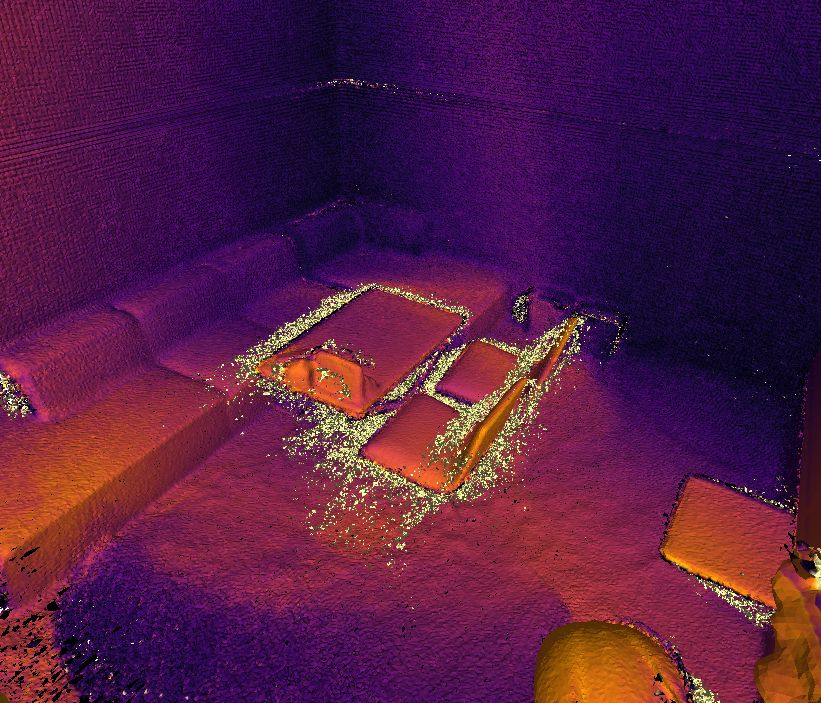} & \\
 & Model & ToF~\cite{handa2014benchmark}  & ToF Denoising~\cite{handa2014benchmark} & Our Fused & Our Sensor Weight & \\
\end{tabular}

\caption{\boldparagraph{ToF\bplus{}ToF Denoising Fusion.} Our method fuses the raw ToF sensor with the depth denoising preprocessed ToF sensor such that the fused result is improved. Note for example that the outliers from outside the walls from the raw sensor are removed and so are the outliers around the table from the denoising ToF sensor. Fig.~\ref{fig:trajectory_office_0} provides the camera trajectory that was used for the evaluation. Note that the raw sensor is favored on surfaces that were viewed close to the camera while the denoising ToF sensor is favored when the measured depth is large.
See also Tab.~\ref{tab:tof_tof_denoising}.}
\label{fig:tof_tof_denoising}
\end{figure*}
\else
\begin{figure*}[t]
\centering
\scriptsize
\setlength{\tabcolsep}{1pt}
\renewcommand{\arraystretch}{1}
\begin{tabular}{ccccccc}
\rotatebox[origin=c]{90}{Office 0} & \includegraphics[align=c, width=.18\linewidth]{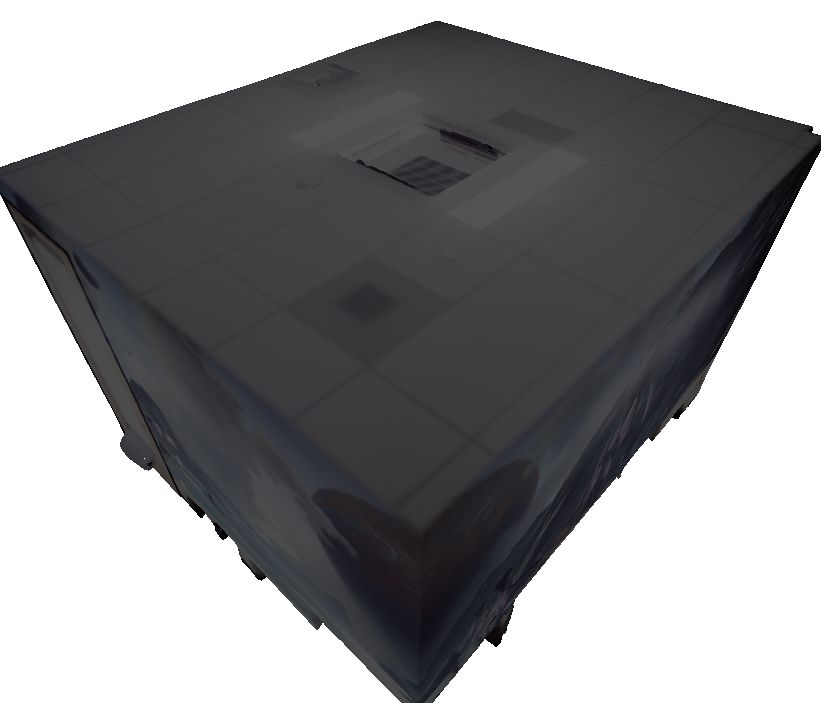} & \includegraphics[align=c, width=.18\linewidth]{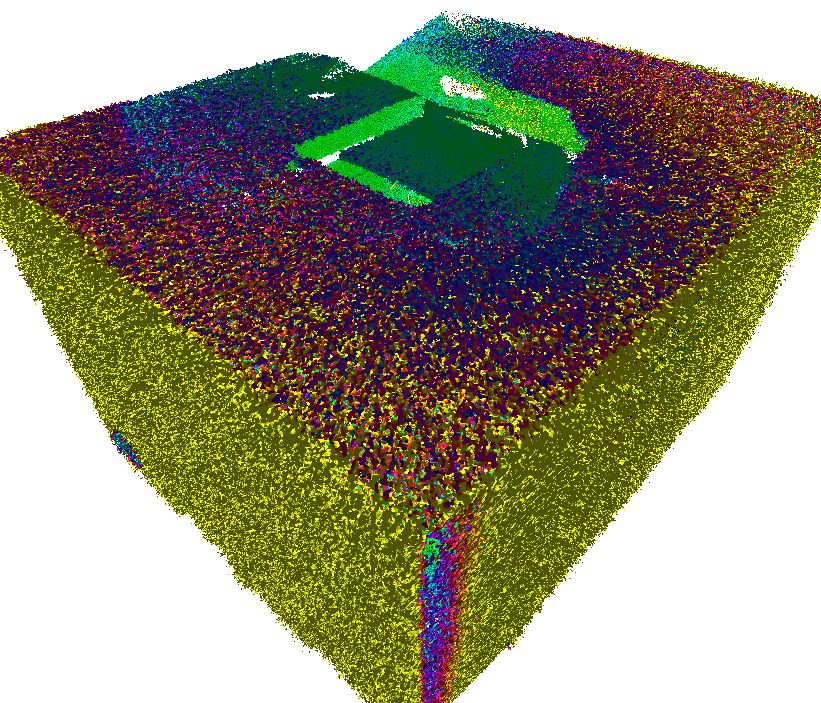} & \includegraphics[align=c, width=.18\linewidth]{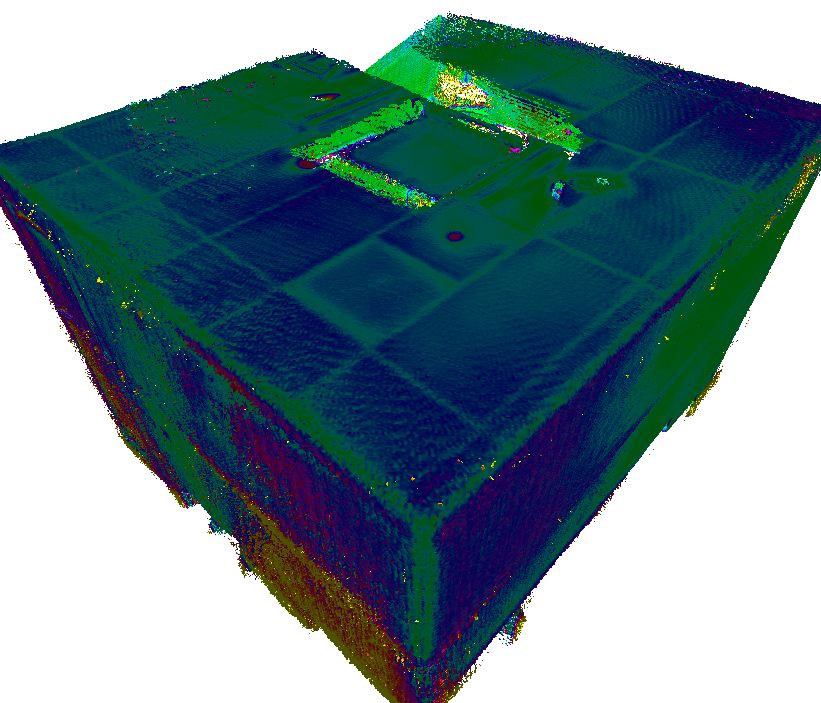} & \includegraphics[align=c, width=.18\linewidth]{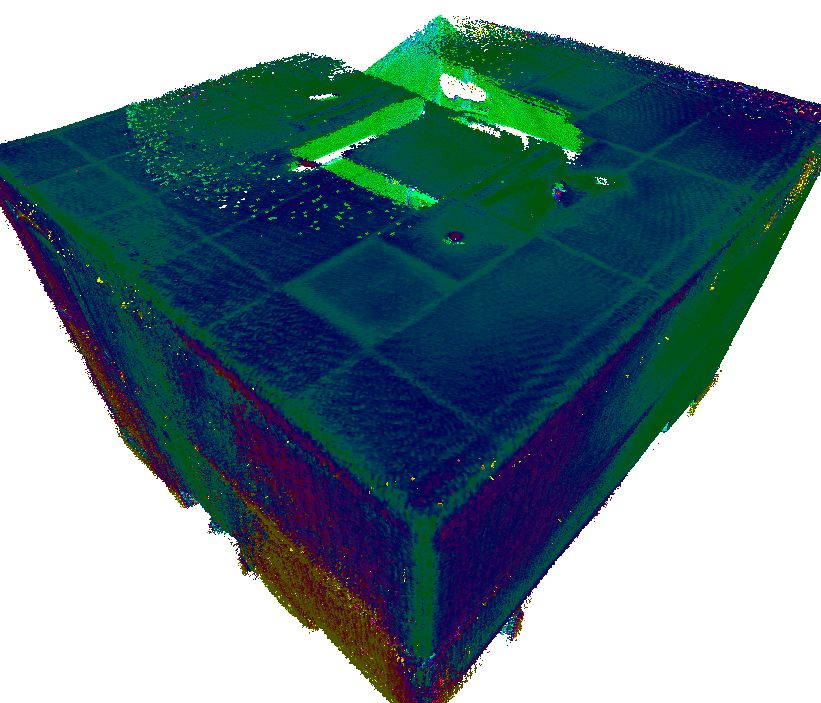} & \includegraphics[align=c, width=.18\linewidth]{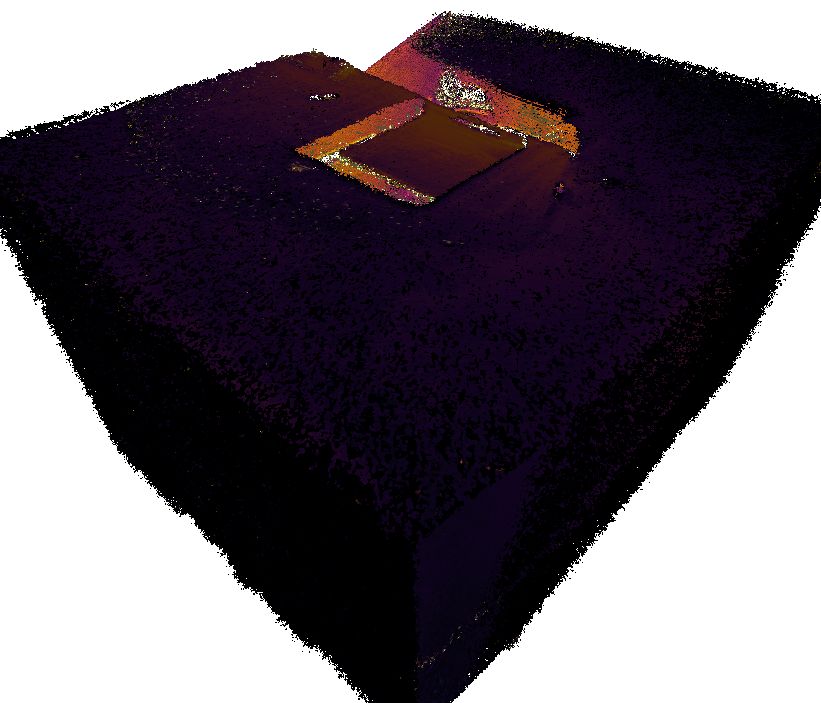} & \multirow[t]{12}{*}[35pt]{\includegraphics[align=t, width=.051\linewidth]{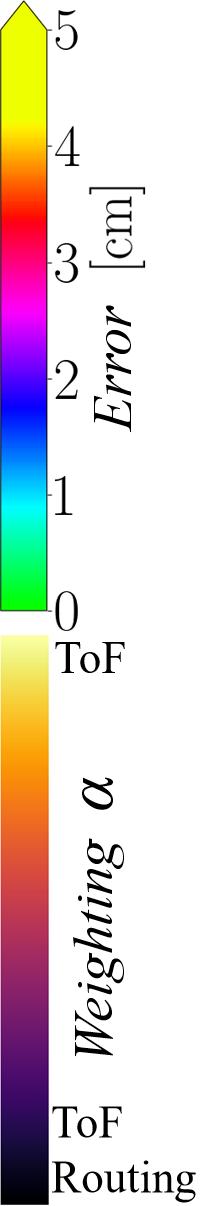}} \\
\rotatebox[origin=c]{90}{Office 0} & \includegraphics[align=c, width=.18\linewidth]{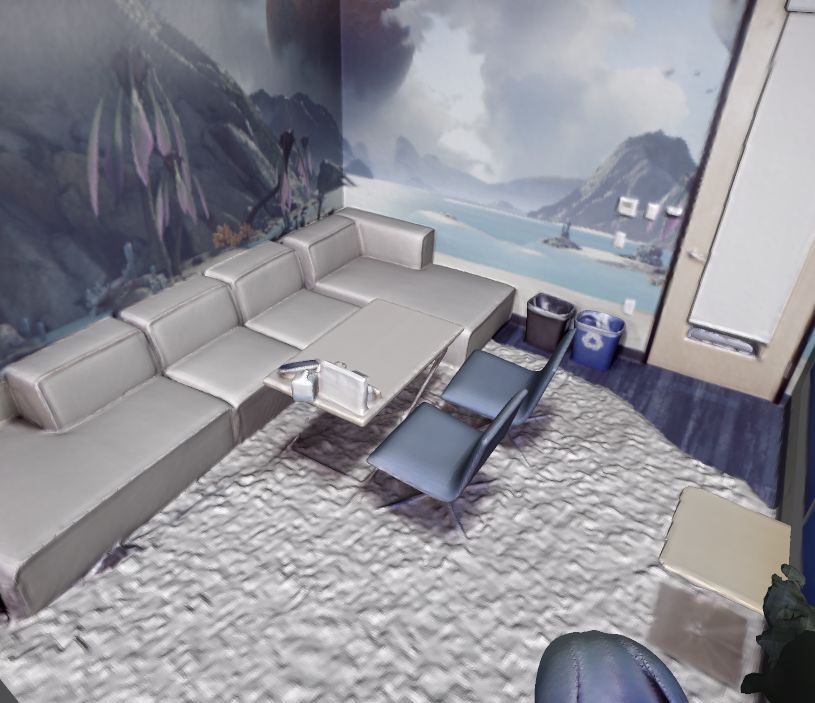} &  \includegraphics[align=c, width=.18\linewidth]{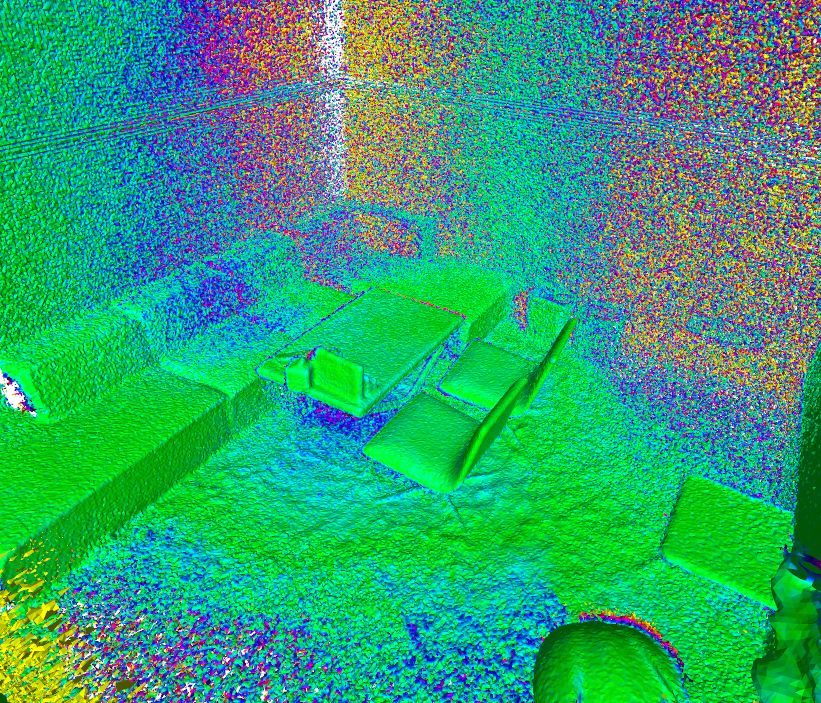} & \includegraphics[align=c, width=.18\linewidth]{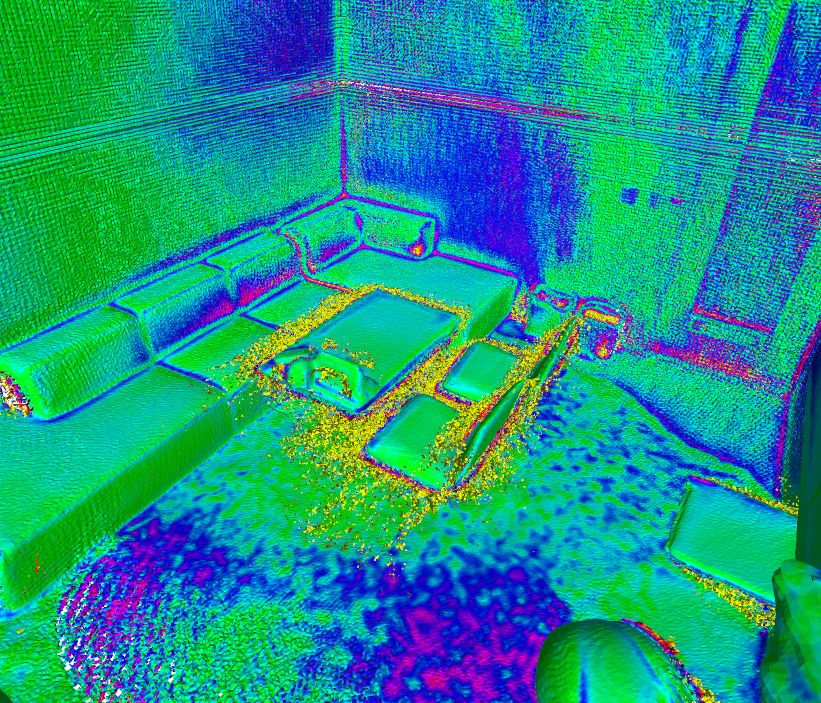} & \includegraphics[align=c, width=.18\linewidth]{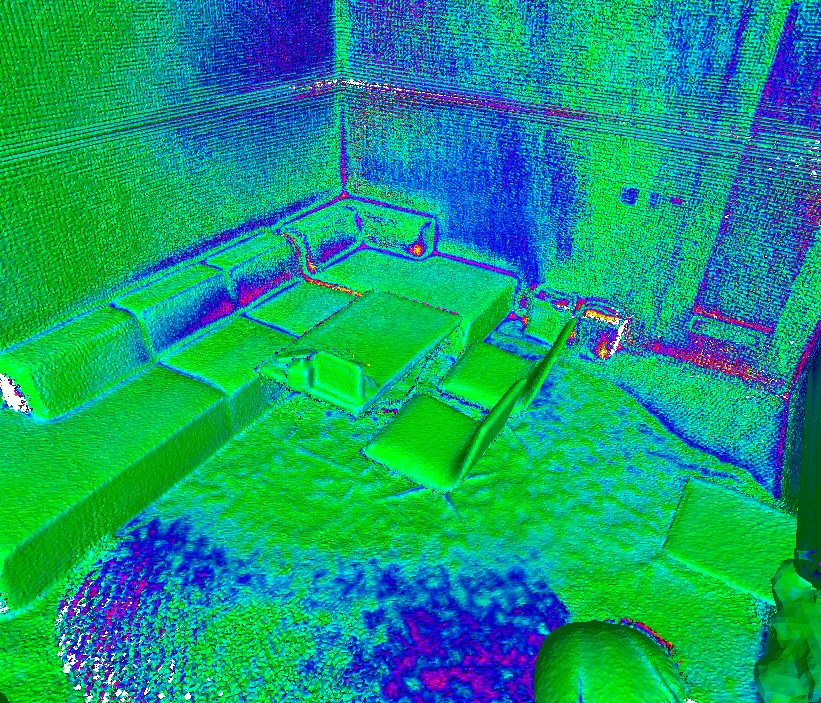} & \includegraphics[align=c, width=.18\linewidth]{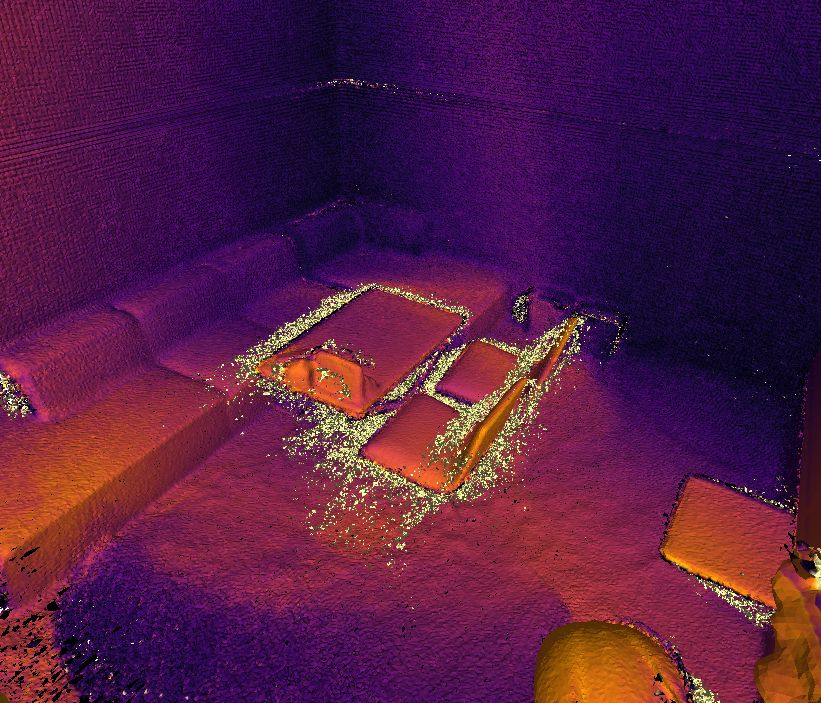} & \\
 & Model & ToF~\cite{handa2014benchmark}  & ToF Denoising~\cite{handa2014benchmark} & Our Fused & Our Sensor Weighting & \\
\end{tabular}

\caption{\boldparagraph{ToF\bplus{}ToF denoising Fusion.} Our method fuses the raw ToF sensor with the denoising preprocessed ToF sensor such that the fused result is improved. Note for example that the outliers from outside the walls from the raw sensor are removed and so are the outliers around the table from the denoising ToF sensor. Fig.~\ref{fig:trajectory_office_0} provides the camera trajectory that was used for the evaluation. Note that the raw sensor is favored on surfaces that were viewed close to the camera while the denoising ToF sensor is favored when the measured depth is large.
See also Tab.~\ref{tab:tof_tof_denoising}.}
\label{fig:tof_tof_denoising}
\end{figure*}
\fi

\ifeccv
\else
\begin{table}
\centering
\resizebox{\columnwidth}{!}
{
\begin{tabular}{l|lllllll}
\cellcolor{gray}       & \cellcolor{gray}MSE$\downarrow$      & \cellcolor{gray}MAD$\downarrow$  & \cellcolor{gray}IoU$\uparrow$     & \cellcolor{gray}Acc.$\uparrow$ & \cellcolor{gray}F$\uparrow$ & \cellcolor{gray}P$\uparrow$ & \cellcolor{gray}R$\uparrow$ \\
\multirow{-2}{*}{\cellcolor{gray} \backslashbox[23mm]{Model}{Metric}} & \cellcolor{gray}*e-04 & \cellcolor{gray}*e-02 & \cellcolor{gray}[0,1] & \cellcolor{gray}$[\%]$ & \cellcolor{gray}$[\%]$ & \cellcolor{gray}$[\%]$ & \cellcolor{gray}$[\%]$ \\\hline
\multicolumn{8}{c}{\emph{Single Sensor}} \\ \hline
ToF~\cite{handa2014benchmark} & 7.48 & 1.99 & 0.664 & 83.65 & 58.52 & 45.84 & \textbf{84.85} \\ 
ToF denoising~\cite{handa2014benchmark}   & 5.08 & 1.58 & 0.709 & 87.32 & 68.93 & 59.01  & 83.08 \\ 
\hline
\multicolumn{8}{c}{\emph{Multi-Sensor Fusion}} \\ \hline
Ours & \textbf{3.19}  & \textbf{1.26}  & \textbf{0.791}  & \textbf{90.99}  & \textbf{78.87} & \textbf{76.56} & 81.68 \\ 
\end{tabular}
}
\caption{\textbf{ToF\bplus{}ToF denoising Fusion.} Our method learns to combine the preprocessed depth and the raw depth in a fashion that improves the overall reconstruction.}
\label{tab:tof_tof_denoising}
\end{table}
\fi

\ifeccv
\begin{wrapfigure}[14]{R}{0.55\linewidth}
\vspace{-\intextsep}
\centering
\includegraphics[align=c, width=0.6\linewidth]{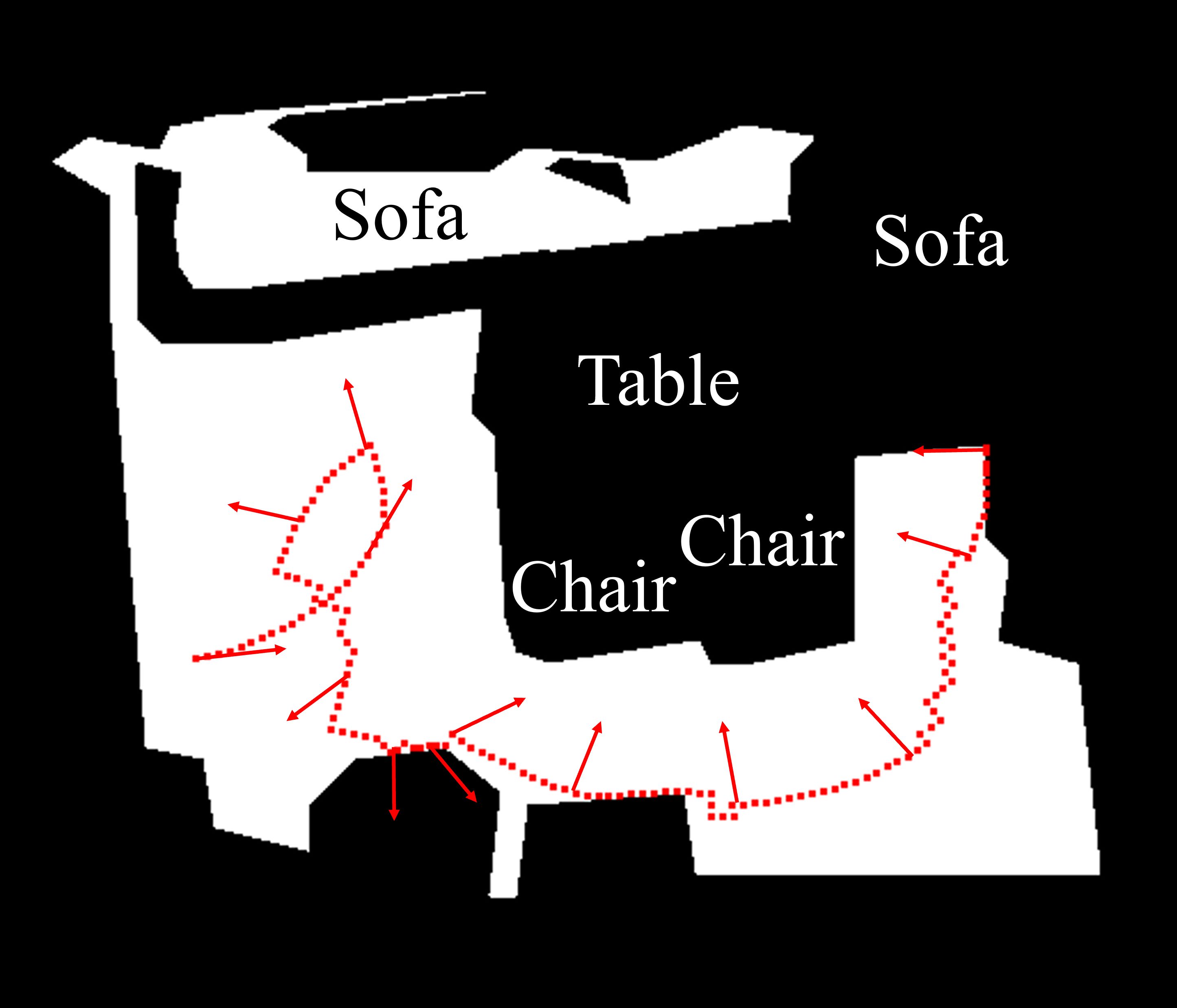} 
\caption{\boldparagraph{Camera Trajectory.} Top-down visualization of the camera trajectory used for the office 0 test scene in Fig.~\ref{fig:tof_tof_denoising}. Navigable space is colored white and the arrows show the camera direction.}
\label{fig:trajectory_office_0}
\end{wrapfigure}
\else
\begin{figure}[t]
\centering
\includegraphics[align=c, width=0.6\linewidth]{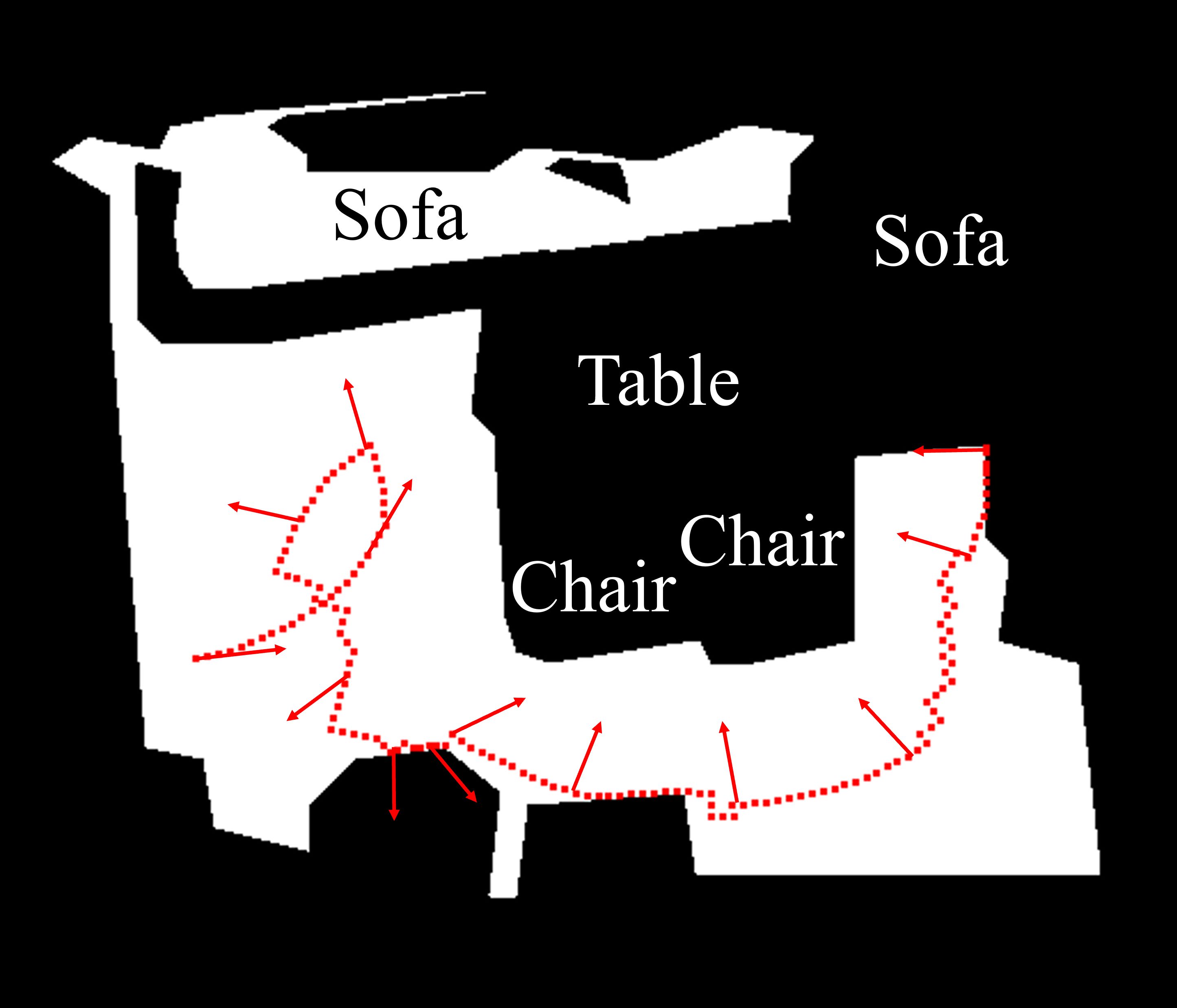} 
\caption{\boldparagraph{Camera Trajectory.} Top-down visualization of the camera trajectory used for the office 0 test scene in Fig.~\ref{fig:tof_tof_denoising}. Navigable space is colored white and the arrows show the camera direction.}
\label{fig:trajectory_office_0}
\end{figure}
\fi

\boldparagraph{ToF\bplus{}ToF denoising Fusion.} From the experiments with and without depth denoising in the main paper, we note that the depth denoising network brings advantages where planar noise is present, but disadvantages due to over-smoothing around depth discontinuities. A natural question arises: Can our framework combine the best of both worlds by fusing a raw ToF frame with a ToF frame preprocessed by a denoising network? Tab.~\ref{tab:tof_tof_denoising} along with Fig.~\ref{fig:tof_tof_denoising} show that our fused output significantly improves on the inputs. For example, the raw ToF contains many outliers behind the walls, while the ToF denoising sensor does not. Vice versa, the raw ToF does not contain outliers around the table and chairs, while the ToF denoising sensor does. As a result, our method selects the appropriate noise-free sensor where needed. Fig.~\ref{fig:trajectory_office_0} provides the camera trajectory that was used for the evaluation. Note that the raw sensor is favored on surfaces that are viewed close to the camera while the ToF denoising sensor is favored when the measured depth is large.

\ifeccv
\begin{figure}[t]
\centering
{\scriptsize
\setlength{\tabcolsep}{1pt}
\renewcommand{\arraystretch}{1}
\begin{tabular}{cccccc}
SR* & PSMNet 1, ToF 1 & PSMNet 1, ToF 1/2 & PSMNet 1, ToF 1/3 & PSMNet 1, ToF 1/3 & \\
\rotatebox[origin=c]{90}{Hotel 0} & \includegraphics[align=c, width=.22\linewidth]{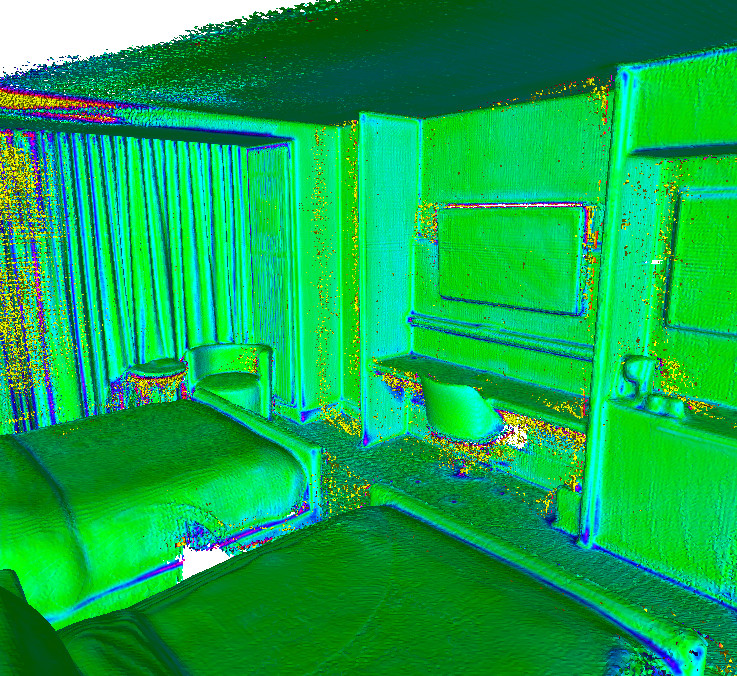} & \includegraphics[align=c, width=.22\linewidth]{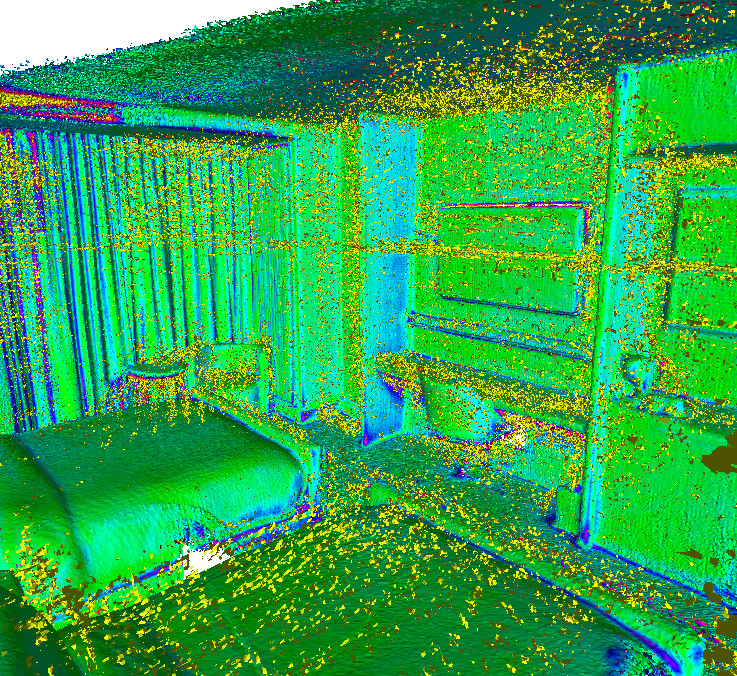} & \includegraphics[align=c, width=.22\linewidth]{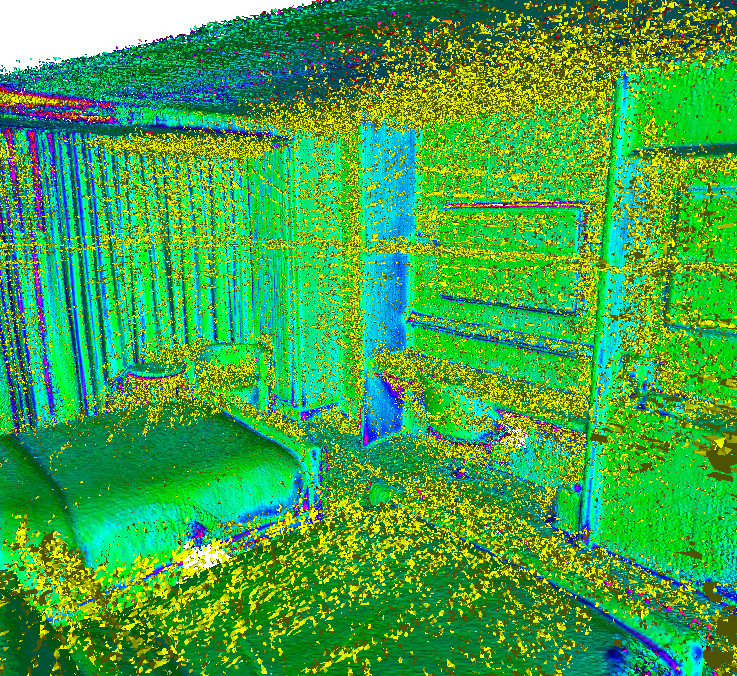} & \includegraphics[align=c, width=.22\linewidth]{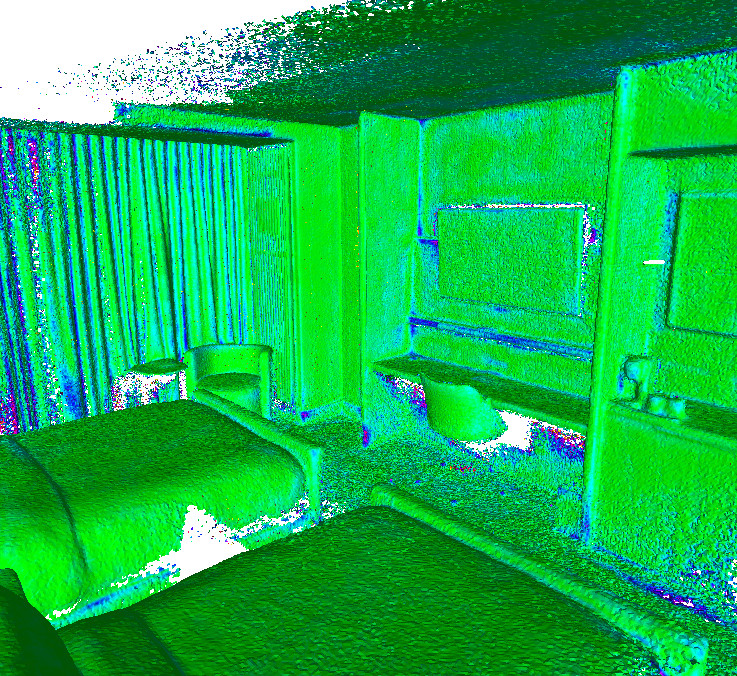} & \multirow[t]{12}{*}[37pt]{\includegraphics[align=t, width=.056\linewidth]{figures/colorbars/colorbar_outlier_filter.png}} \\
 & Early Fusion & Early Fusion & Early Fusion & SenFuNet (Ours) & \\
\end{tabular}
}
\caption{\boldparagraph{ToF\bplus{}PSMNet Fusion.} The performance gap to our method grows when asynchronous sensors are considered. The performance decreases further for the Early Fusion when the sampling rate is reduced to 1/3 compared to the PSMNet sensor while out method remains robust (best viewed on screen). SR* = Sampling Rate.}
\label{fig:asynch}
\end{figure}
\fi

\ifeccv
\begin{figure}[t]
\centering
{\scriptsize
\setlength{\tabcolsep}{1pt}
\renewcommand{\arraystretch}{1}
\begin{tabular}{cccccc}
\rotatebox[origin=c]{90}{Hotel 0} & \includegraphics[align=c, width=.23\linewidth]{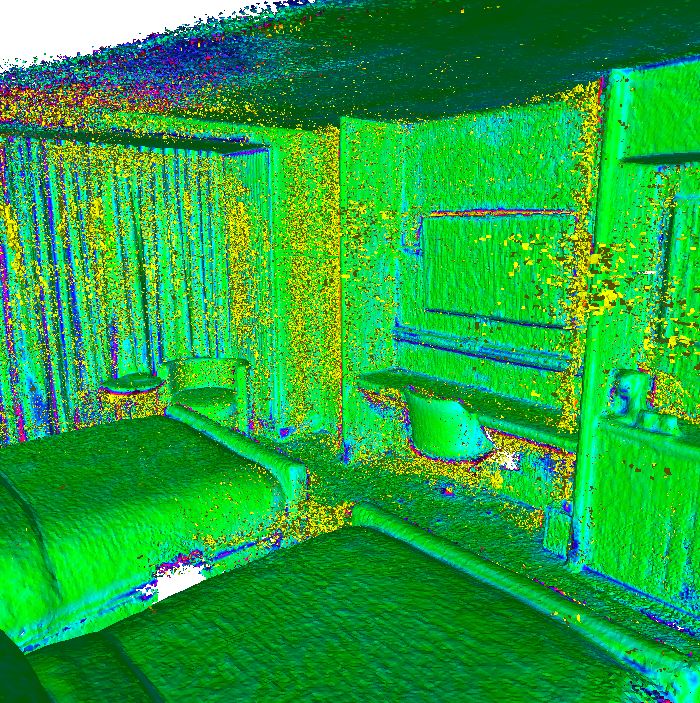} & \includegraphics[align=c, width=.23\linewidth]{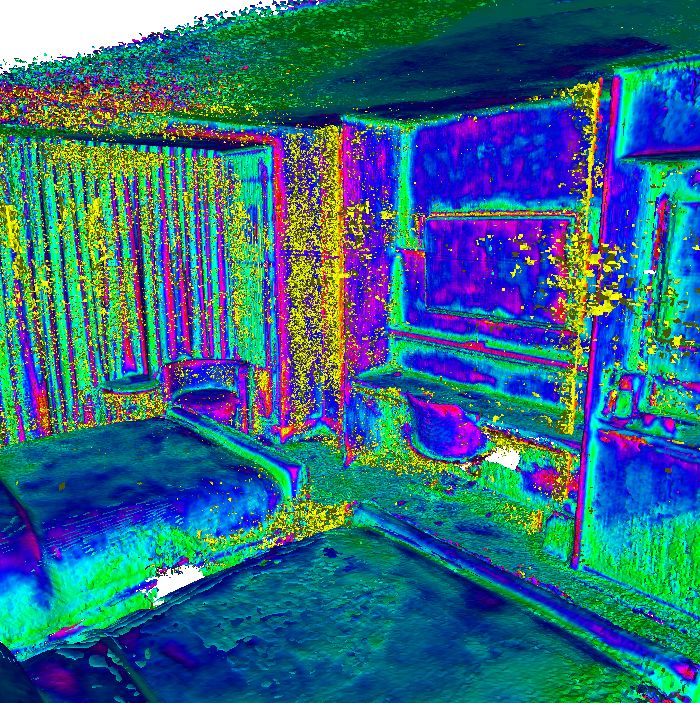} & \includegraphics[align=c, width=.23\linewidth]{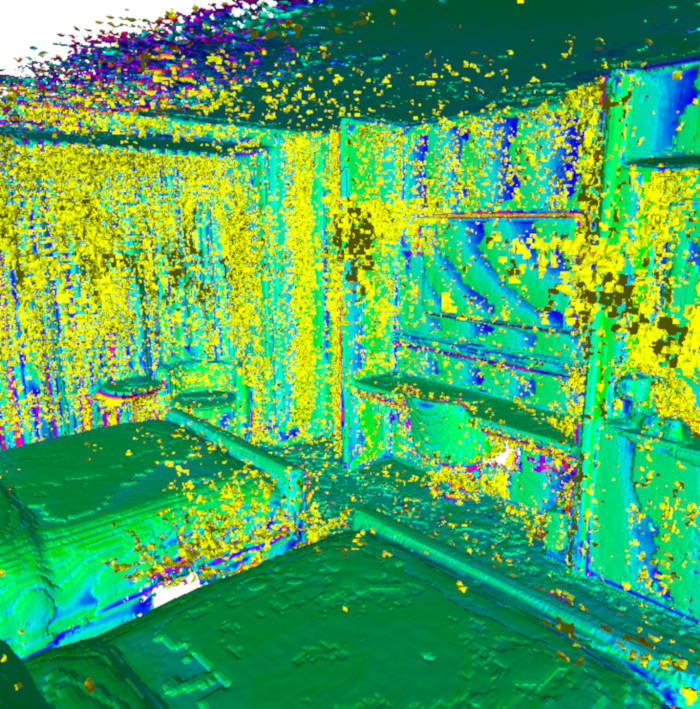} & \includegraphics[align=c, width=.23\linewidth]{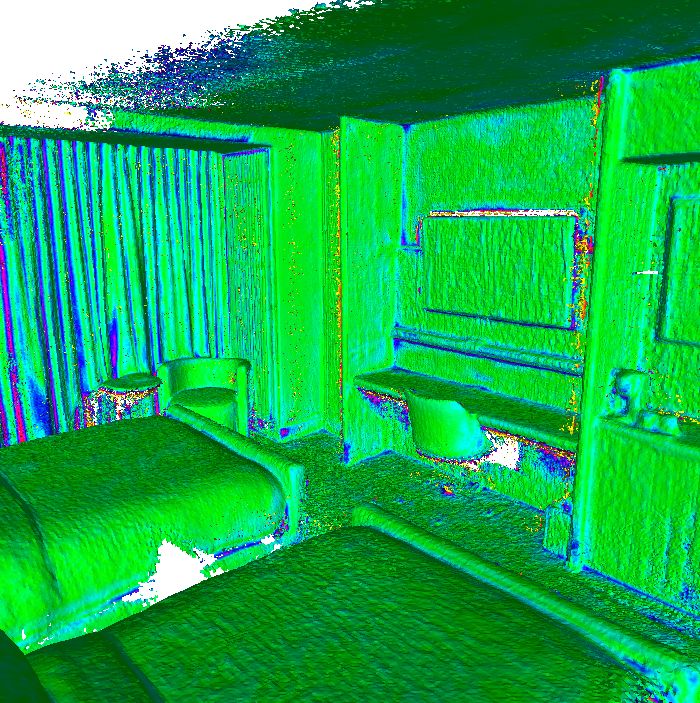} & \multirow[t]{12}{*}[43pt]{\includegraphics[align=t, width=.064\linewidth]{figures/colorbars/colorbar_outlier_filter.png}} \\
\rotatebox[origin=c]{90}{Office 0} & \includegraphics[align=c, width=.23\linewidth]{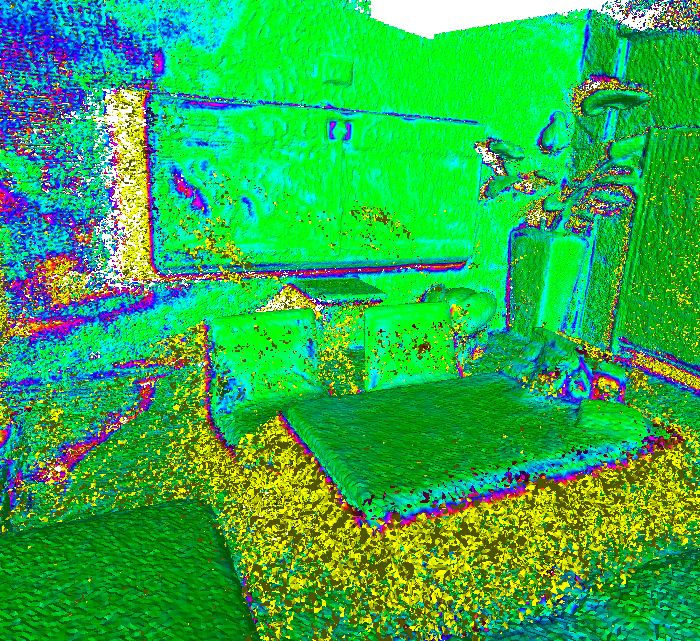} & \includegraphics[align=c, width=.23\linewidth]{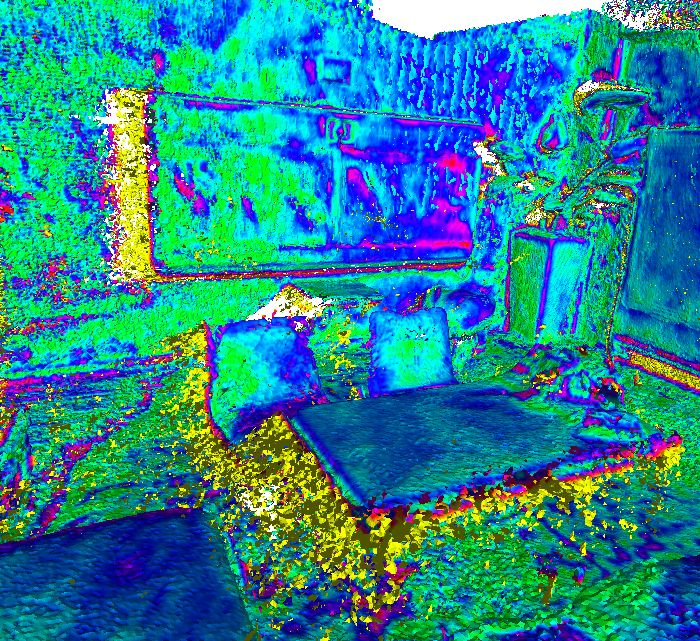} & \includegraphics[align=c, width=.23\linewidth]{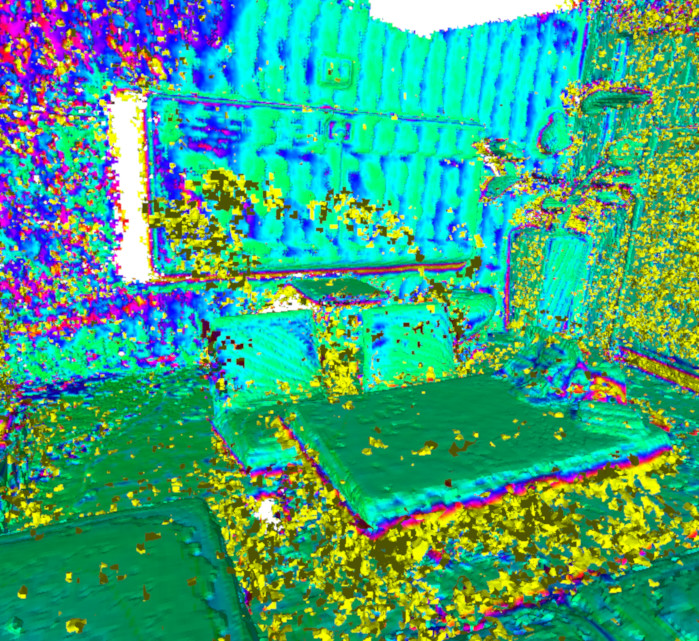} & \includegraphics[align=c, width=.23\linewidth]{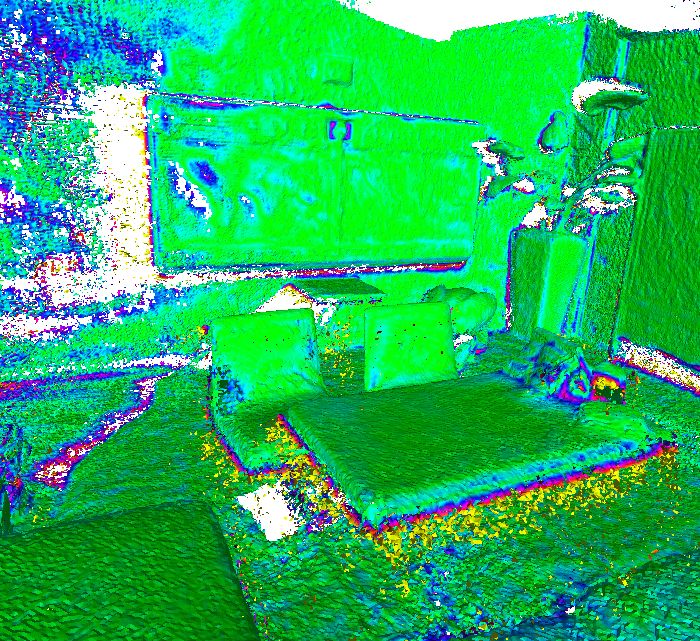} & \\
 & TSDF Fusion~\cite{curless1996volumetric} & RoutedFusion~\cite{Weder2020RoutedFusionLR} & DI-Fusion~\cite{huang2021di} $\sigma$=0.15 & SenFuNet (Ours) & \\
\end{tabular}
}
\caption{\boldparagraph{SGM\bplus{}PSMNet Fusion without denoising.} Our method fuses the sensors consistently better than the baseline methods.
In particular, our method learns to detect and remove outliers much more effectively (best viewed on screen).}
\label{fig:all_methods}
\end{figure}
\else
\begin{figure}[t]
\centering
{\scriptsize
\setlength{\tabcolsep}{1pt}
\renewcommand{\arraystretch}{1}
\begin{tabular}{ccccc}
\rotatebox[origin=c]{90}{Hotel 0} & \includegraphics[align=c, width=.30\linewidth]{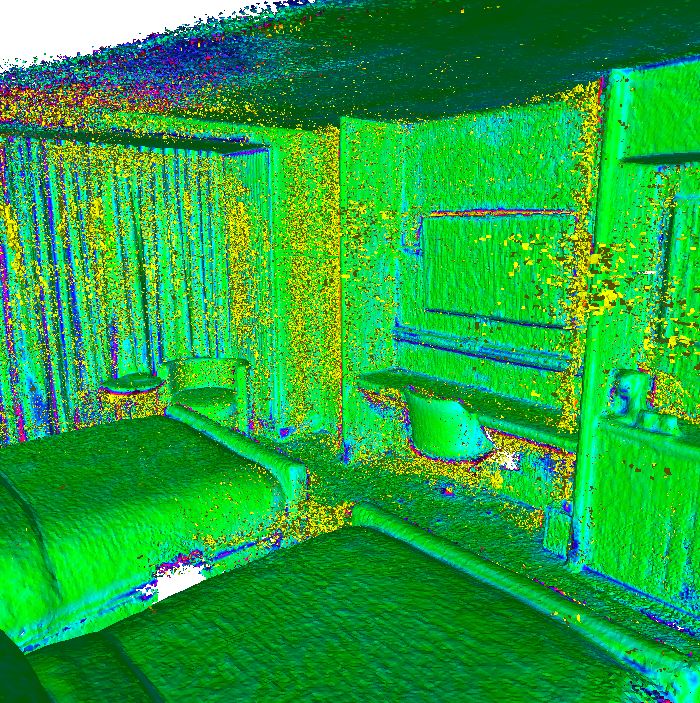} & \includegraphics[align=c, width=.30\linewidth]{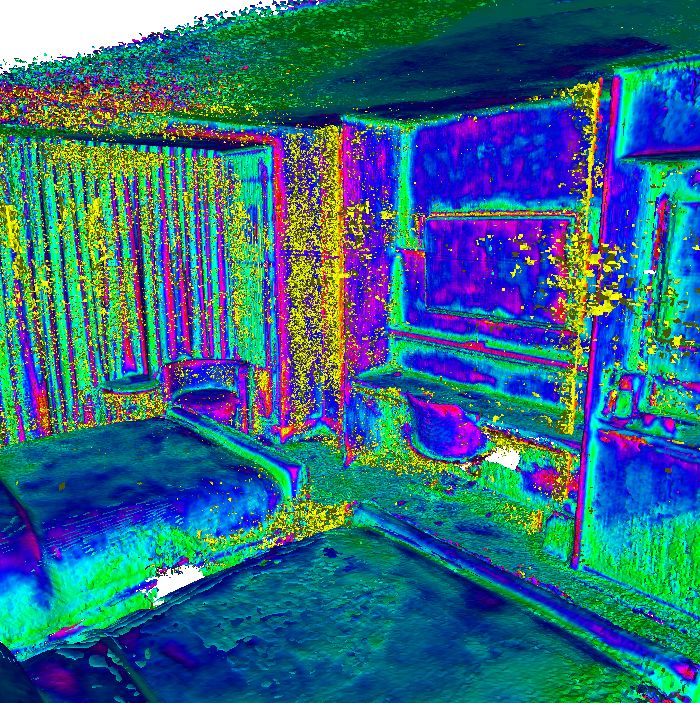} & \includegraphics[align=c, width=.30\linewidth]{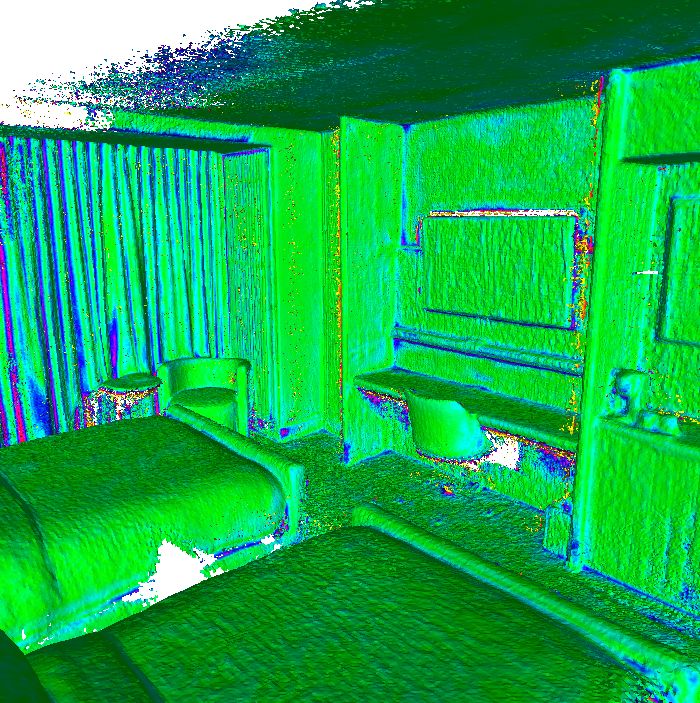} & \multirow[t]{12}{*}[37pt]{\includegraphics[align=t, width=.0835\linewidth]{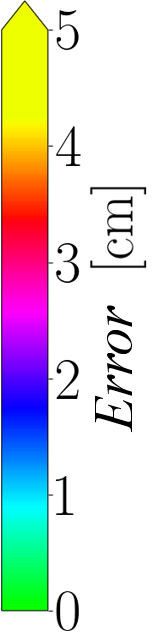}} \\
\rotatebox[origin=c]{90}{Office 0} & \includegraphics[align=c, width=.30\linewidth]{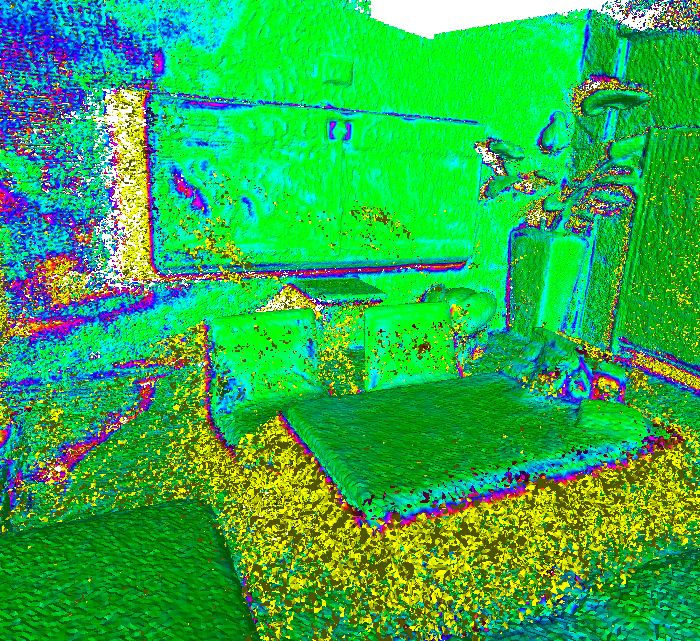} & \includegraphics[align=c, width=.30\linewidth]{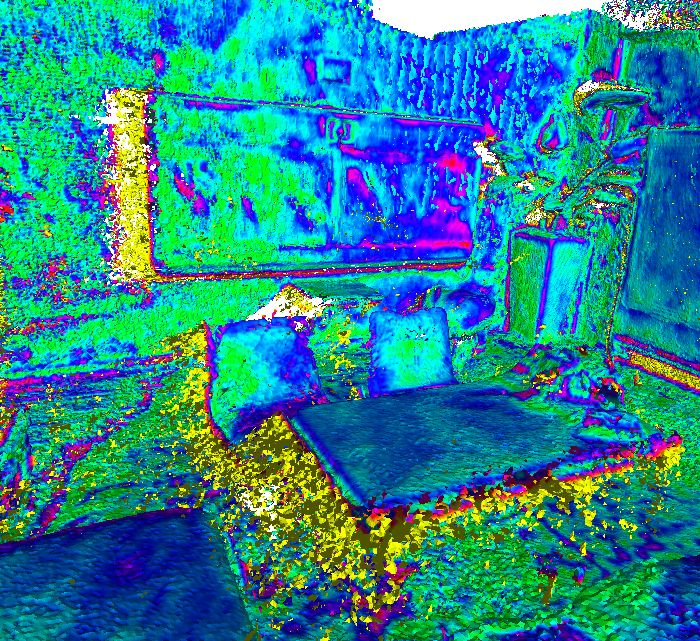} & \includegraphics[align=c, width=.30\linewidth]{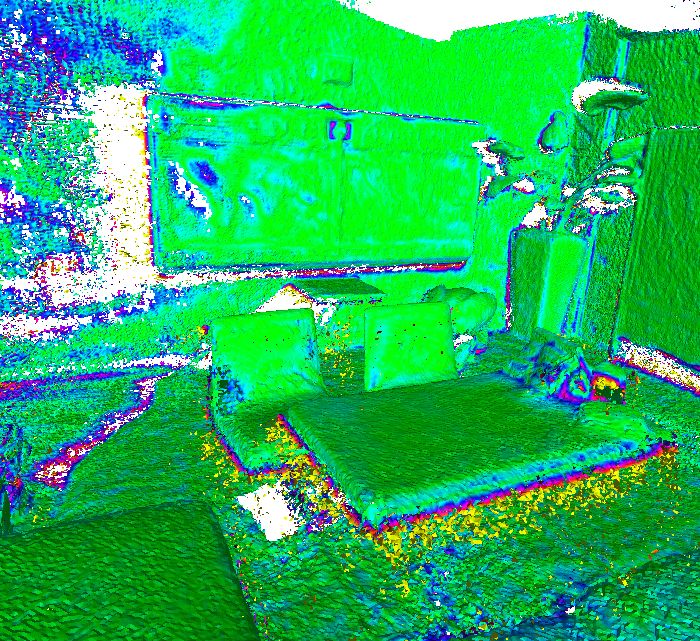} & \\
 & TSDF Fusion~\cite{curless1996volumetric} & RoutedFusion~\cite{Weder2020RoutedFusionLR} & SenFuNet (Ours) & \\
\end{tabular}
}
\caption{\boldparagraph{SGM\bplus{}PSMNet Fusion without denoising.} Our method fuses the sensors consistently better than the baseline methods.
In particular, our method learns to detect and remove outliers much more effectively (best viewed on screen).}
\label{fig:all_methods}
\end{figure}
\fi

\ifeccv
\boldparagraph{Visualizations.}
In Fig.~\ref{fig:tof_psmnet_supp} we show qualitative results on the sensors $\{$ToF, PSMNet$\}$. As concluded from the experiment on the sensors $\{$ToF, ToF denoising$\}$, the ToF sensor (without denoising) is not favored when the depth is large. We observe the same prediction on the office 0 scene in Fig.~\ref{fig:tof_psmnet_supp}.
Fig.~\ref{fig:all_methods} shows a comparison between our method on the sensors $\{$SGM, PSMNet$\}$ without denoising and TSDF Fusion~\cite{curless1996volumetric}, RoutedFusion~\cite{Weder2020RoutedFusionLR} and DI-Fusion~\cite{huang2021di}. We achieve better surface reconstruction performance than RoutedFusion and DI-Fusion and better outlier handling than all baseline methods.
\begin{figure*}[h!]
\centering
{\tiny
\setlength{\tabcolsep}{1pt}
\renewcommand{\arraystretch}{1}
\begin{tabular}{ccccccccc}
\multirow[c]{2}{*}[13pt]{\rotatebox[origin=c]{90}{\small Without Denoising}} & \rotatebox[origin=c]{90}{Hotel 0} & \includegraphics[align=c, width=.15\linewidth]{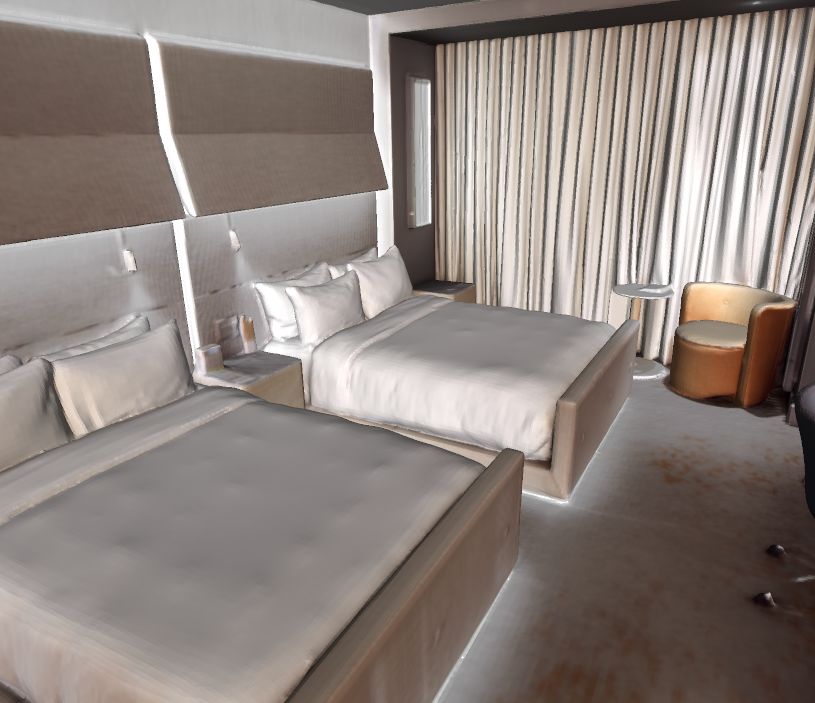} & \includegraphics[align=c, width=.15\linewidth]{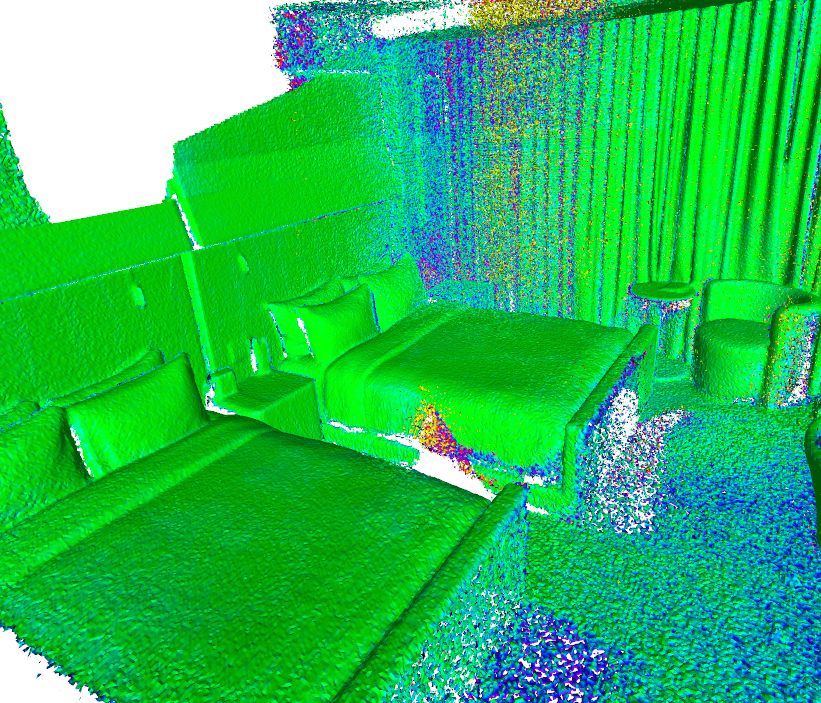} & \includegraphics[align=c, width=.15\linewidth]{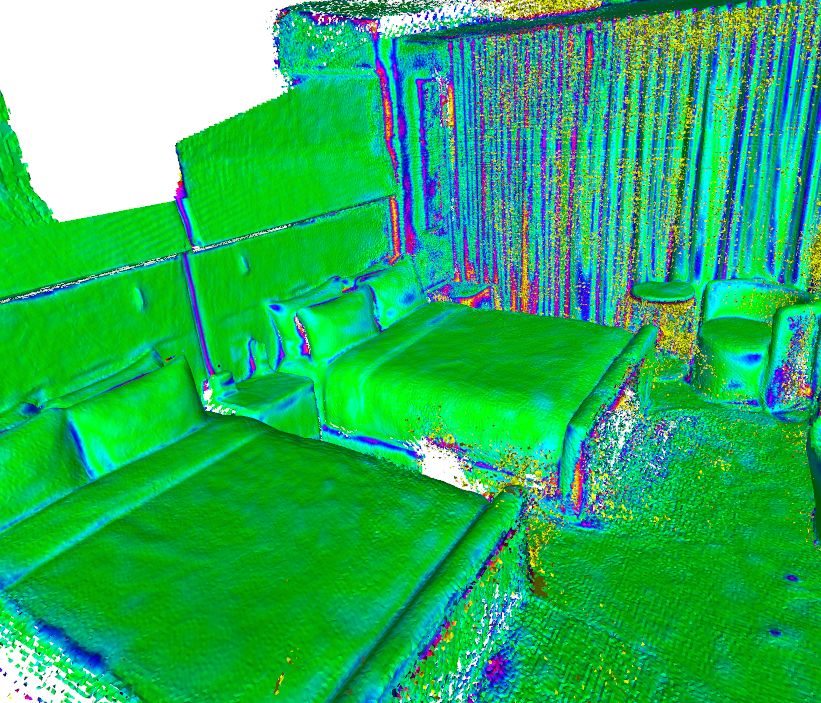} & \includegraphics[align=c, width=.15\linewidth]{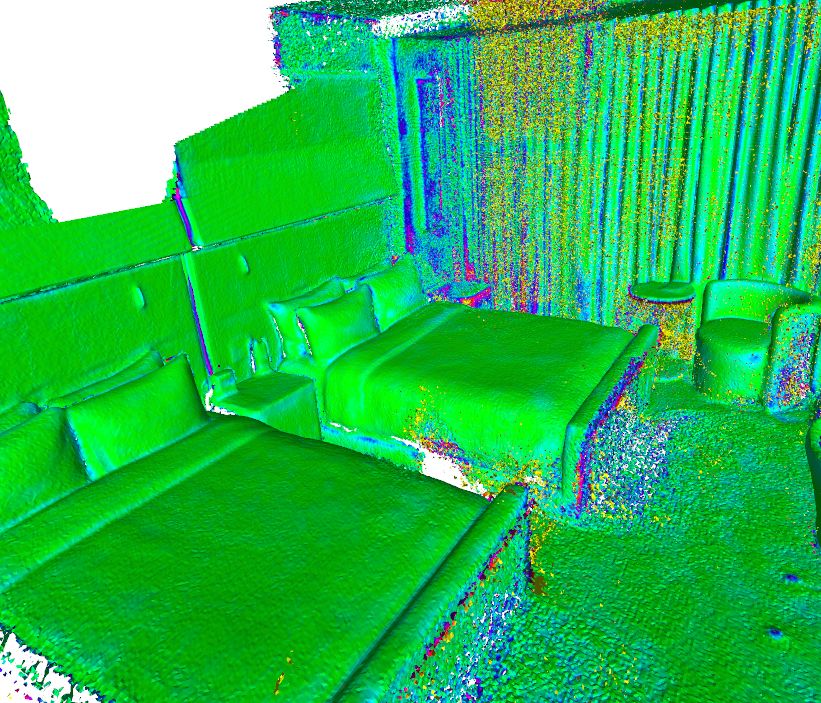} & \includegraphics[align=c, width=.15\linewidth]{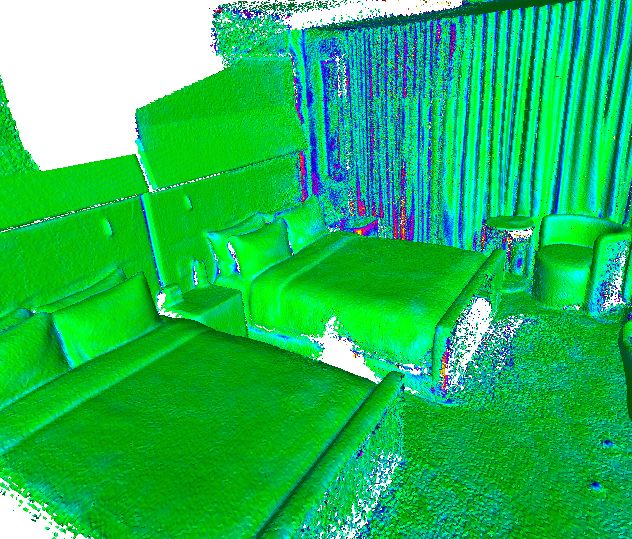} & \includegraphics[align=c, width=.15\linewidth]{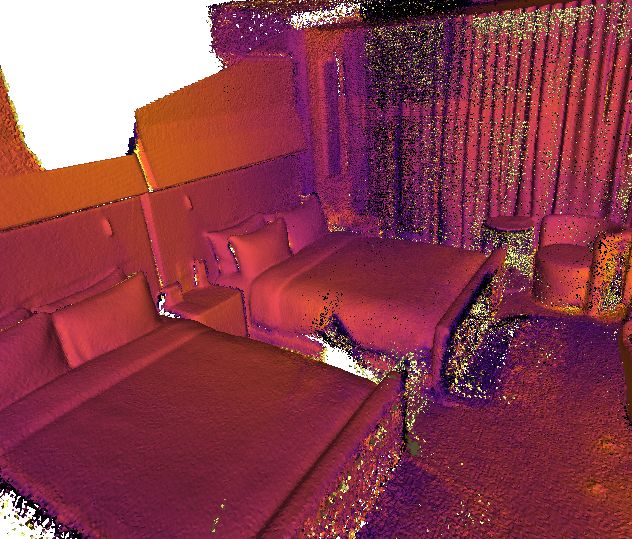} &  \multirow[t]{12}{*}[21pt]{\includegraphics[align=t, width=.036\linewidth]{figures/colorbars/office_4_colorbar.png}}\\  & \rotatebox[origin=c]{90}{Office 0} & \includegraphics[align=c, width=.15\linewidth]{figures/replica/tof_psmnet/office_0/model_small.jpg} & \includegraphics[align=c, width=.15\linewidth]{figures/replica/tof_psmnet_wo_routing/office_0/tof_small.jpg} & \includegraphics[align=c, width=.15\linewidth]{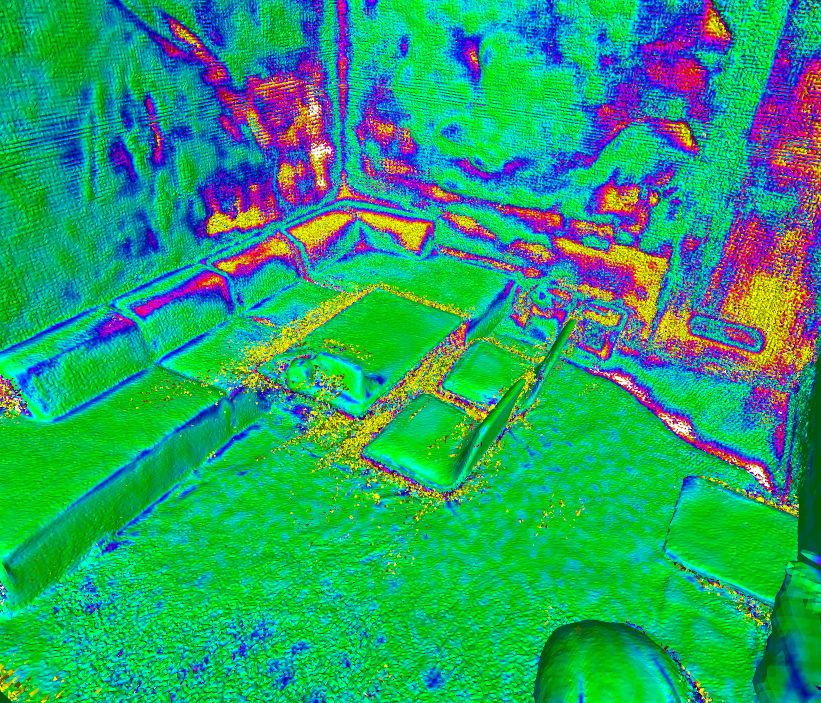} & \includegraphics[align=c, width=.15\linewidth]{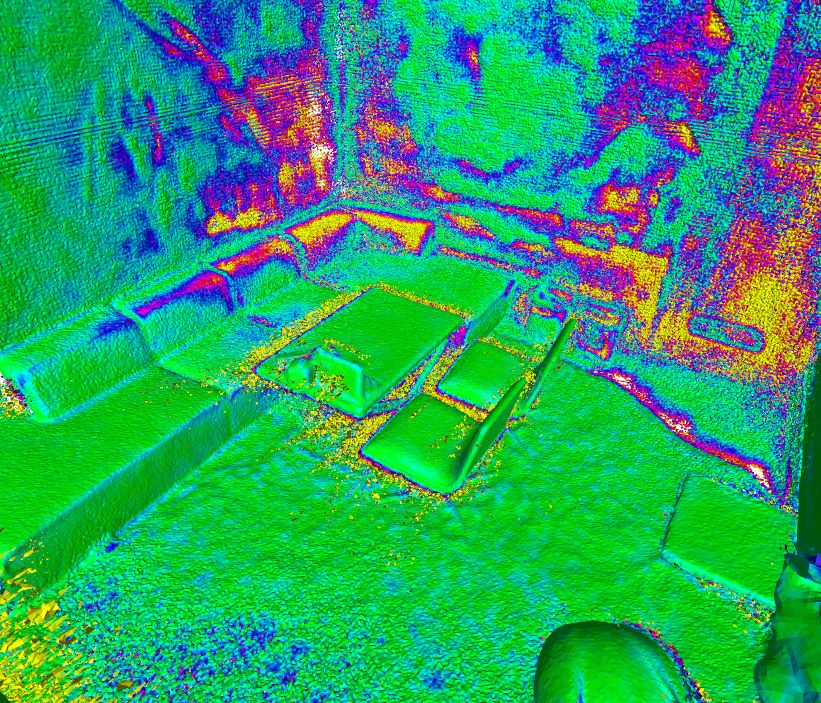} & \includegraphics[align=c, width=.15\linewidth]{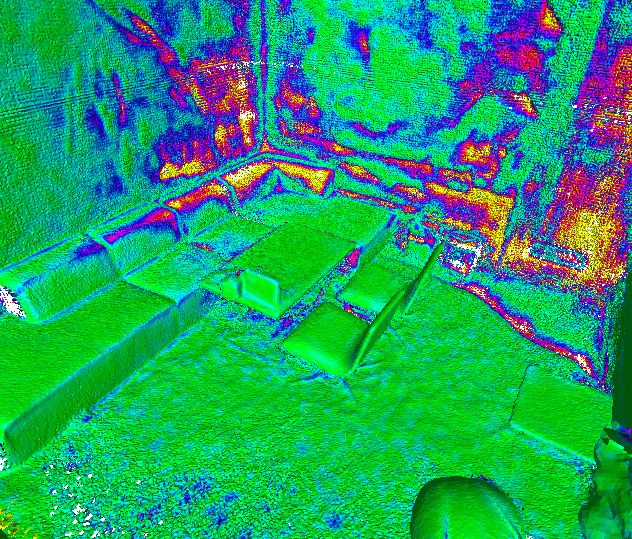} & \includegraphics[align=c, width=.15\linewidth]{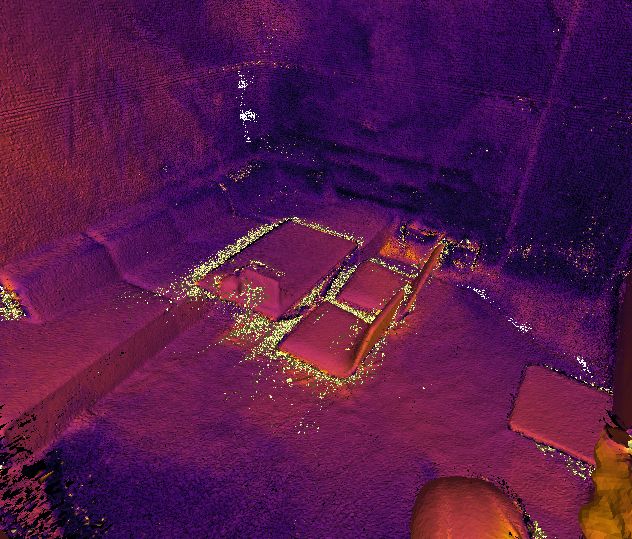} & \\
\multirow[c]{2}{*}[7pt]{\rotatebox[origin=c]{90}{\small With Denoising}} & \rotatebox[origin=c]{90}{Hotel 0} & \includegraphics[align=c, width=.15\linewidth]{figures/replica/tof_psmnet/hotel_0/model.jpg} & \includegraphics[align=c, width=.15\linewidth]{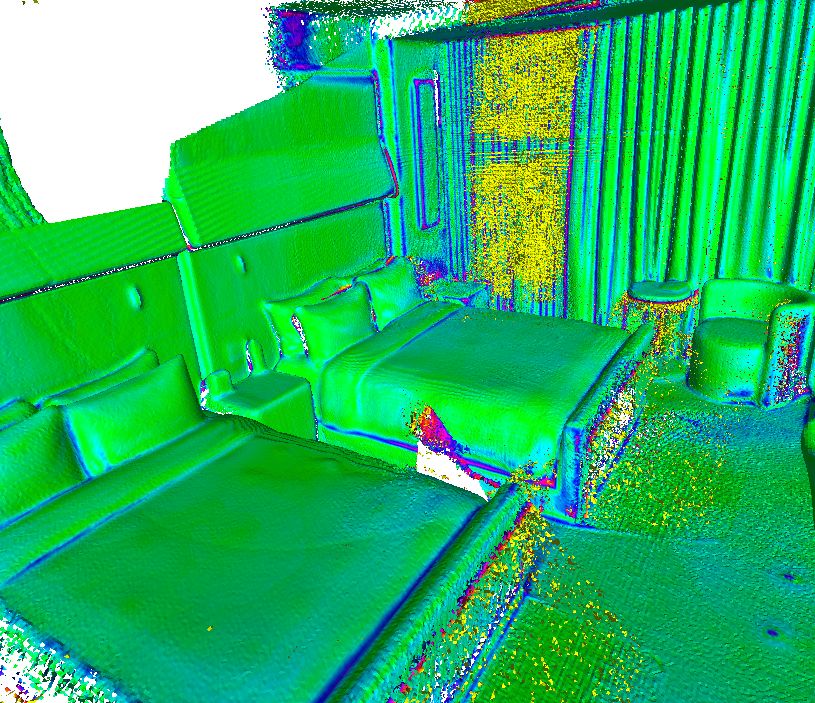} & \includegraphics[align=c, width=.15\linewidth]{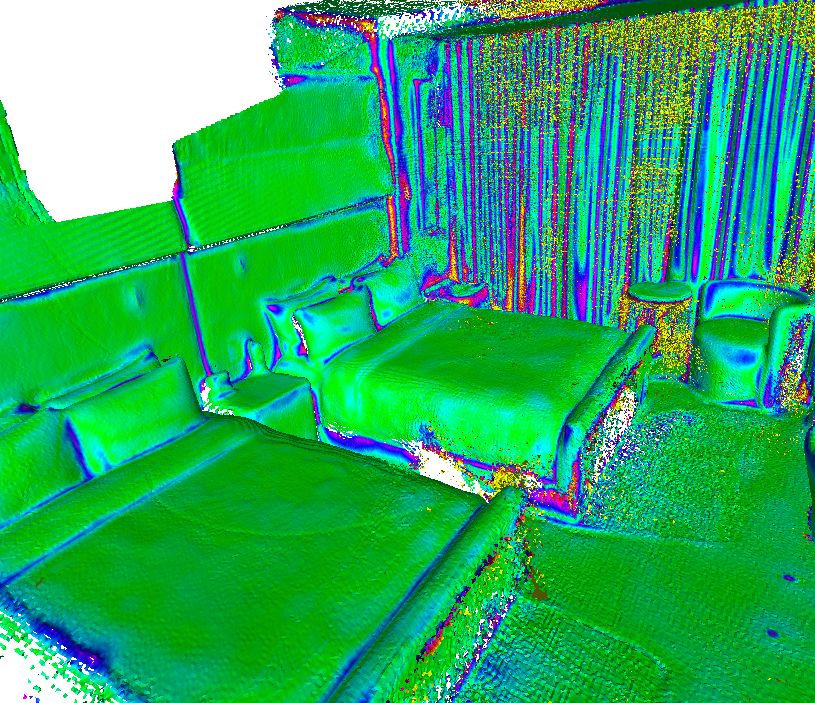} & \includegraphics[align=c, width=.15\linewidth]{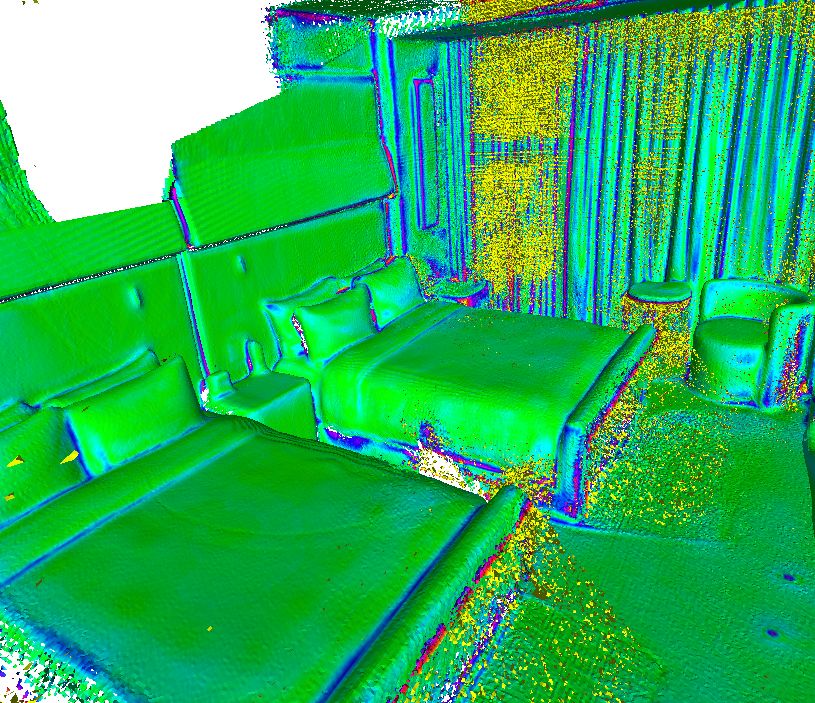} & \includegraphics[align=c, width=.15\linewidth]{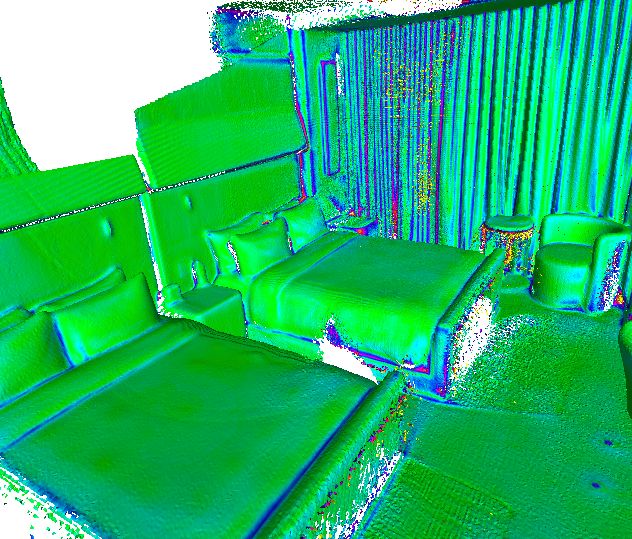} & \includegraphics[align=c, width=.15\linewidth]{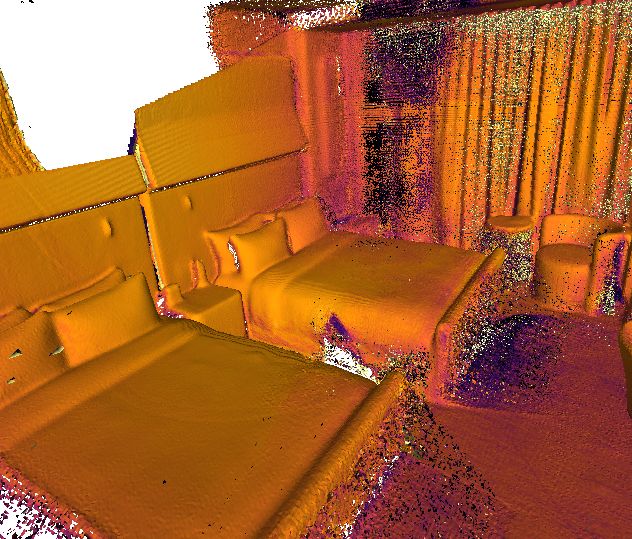} & \\ 
& \rotatebox[origin=c]{90}{Office 0} & \includegraphics[align=c, width=.15\linewidth]{figures/replica/tof_psmnet/office_0/model_small.jpg} & \includegraphics[align=c, width=.15\linewidth]{figures/replica/tof_psmnet/office_0/tof_small.jpg} & \includegraphics[align=c, width=.15\linewidth]{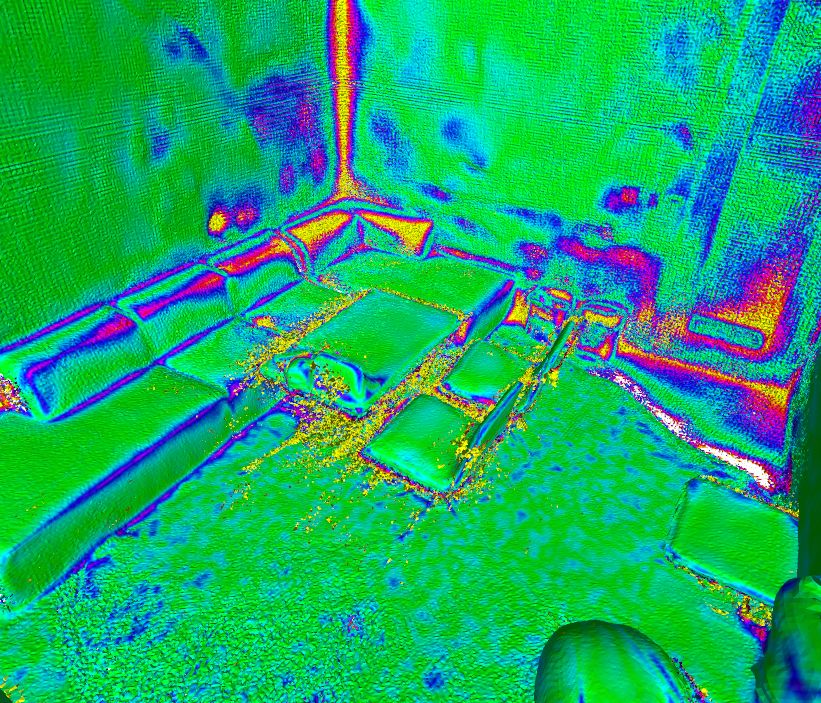} & \includegraphics[align=c, width=.15\linewidth]{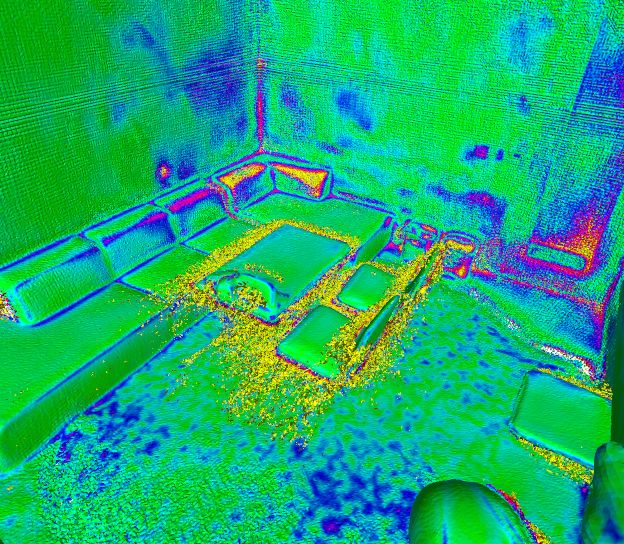} & \includegraphics[align=c, width=.15\linewidth]{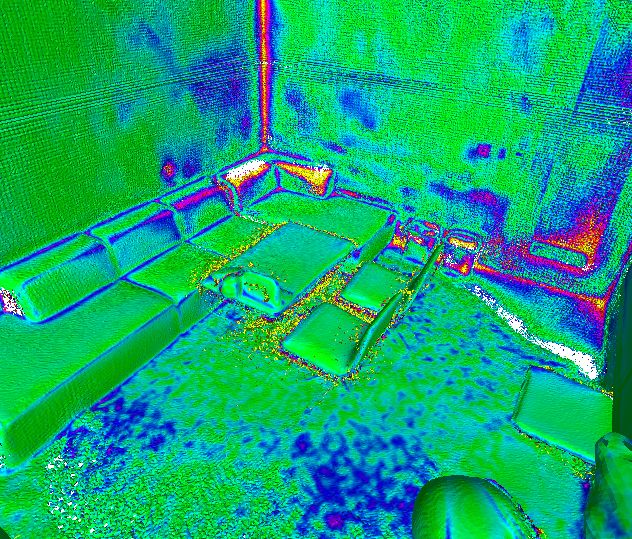} & \includegraphics[align=c, width=.15\linewidth]{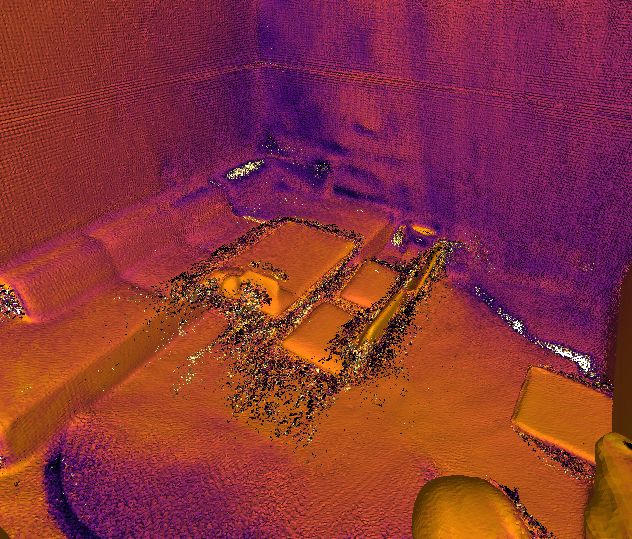} & \\
& & Model & ToF~\cite{handa2014benchmark}  & PSMNet~\cite{chang2018pyramid} & TSDF Fusion~\cite{curless1996volumetric} & SenFuNet (Ours) & Our Sensor Weight & \\
\end{tabular}
}
\caption{\boldparagraph{ToF\bplus{}PSMNet Fusion.} Our method fuses the sensors consistently better than TSDF Fusion~\cite{curless1996volumetric}.
In particular, our method learns to detect and remove outliers much more effectively (best viewed on screen).}
\label{fig:tof_psmnet_supp}
\end{figure*}
In Fig.~\ref{fig:all_methods_denoising}, we add depth denoising to the model and evaluate the baseline methods against our method again. Our model achieves better outlier handling overall and more precise surface reconstruction in most regions compared to the Early Fusion method.

For more visual results, we refer to the supplementary videos.
\else
\boldparagraph{Visualizations.}
In Fig.~\ref{fig:tof_psmnet_supp} we show qualitative results on the sensors $\{$ToF, PSMNet$\}$. As concluded from the experiment on the sensors $\{$ToF, ToF denoising$\}$, the ToF sensor (without denoising) is not favored when the depth is large. We observe the same prediction on the office 0 scene in Fig.~\ref{fig:tof_psmnet_supp}.
Fig.~\ref{fig:all_methods} shows a comparison between our method on the sensors $\{$SGM, PSMNet$\}$ without denoising and TSDF Fusion~\cite{curless1996volumetric} and RoutedFusion~\cite{Weder2020RoutedFusionLR}. We achieve better surface
\FloatBarrier
\begin{figure*}[h!]
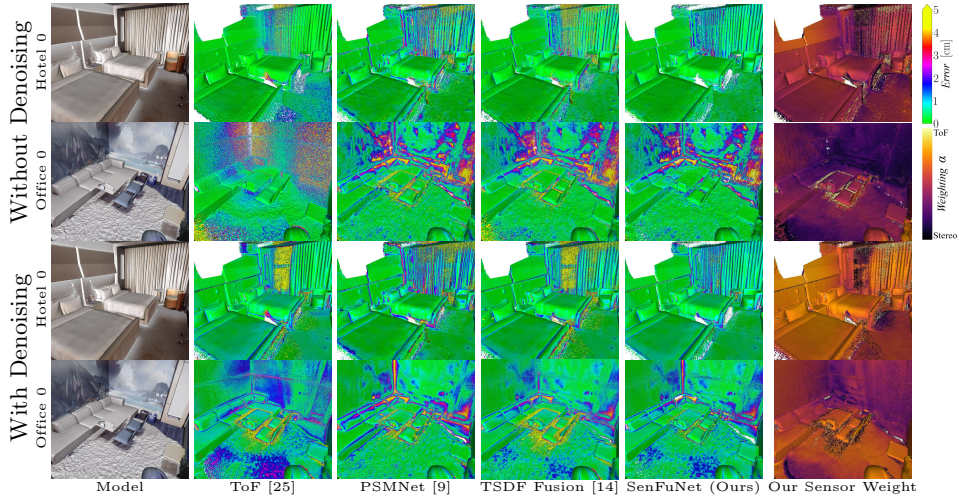

\centering
{\scriptsize
\setlength{\tabcolsep}{1pt}
\renewcommand{\arraystretch}{1}
\begin{tabular}{ccccccccc}
\multirow[c]{2}{*}[0pt]{\rotatebox[origin=c]{90}{\small Without Denoising}} & \rotatebox[origin=c]{90}{Hotel 0} & \includegraphics[align=c, width=.15\linewidth]{figures/replica/tof_psmnet/hotel_0/model.jpg} & \includegraphics[align=c, width=.15\linewidth]{figures/replica/tof_psmnet_wo_routing/hotel_0/tof_small.jpg} & \includegraphics[align=c, width=.15\linewidth]{figures/replica/tof_psmnet_wo_routing/hotel_0/psmnet_stereo_small.jpg} & \includegraphics[align=c, width=.15\linewidth]{figures/replica/tof_psmnet_wo_routing/hotel_0/tsdf_middle_small.jpg} & \includegraphics[align=c, width=.15\linewidth]{figures/replica/tof_psmnet_wo_routing/hotel_0/fused.jpg} & \includegraphics[align=c, width=.15\linewidth]{figures/replica/tof_psmnet_wo_routing/hotel_0/weighting.jpg} &  \multirow[t]{12}{*}[29pt]{\includegraphics[align=t, width=.036\linewidth]{figures/colorbars/office_4_colorbar.png}}\\  & \rotatebox[origin=c]{90}{Office 0} & \includegraphics[align=c, width=.15\linewidth]{figures/replica/tof_psmnet/office_0/model_small.jpg} & \includegraphics[align=c, width=.15\linewidth]{figures/replica/tof_psmnet_wo_routing/office_0/tof_small.jpg} & \includegraphics[align=c, width=.15\linewidth]{figures/replica/tof_psmnet_wo_routing/office_0/psmnet_stereo_small.jpg} & \includegraphics[align=c, width=.15\linewidth]{figures/replica/tof_psmnet_wo_routing/office_0/tsdf_middle_small.jpg} & \includegraphics[align=c, width=.15\linewidth]{figures/replica/tof_psmnet_wo_routing/office_0/fused.jpg} & \includegraphics[align=c, width=.15\linewidth]{figures/replica/tof_psmnet_wo_routing/office_0/weighting.jpg} & \\
\multirow[c]{2}{*}[0pt]{\rotatebox[origin=c]{90}{\small With Denoising}} & \rotatebox[origin=c]{90}{Hotel 0} & \includegraphics[align=c, width=.15\linewidth]{figures/replica/tof_psmnet/hotel_0/model.jpg} & \includegraphics[align=c, width=.15\linewidth]{figures/replica/tof_psmnet/hotel_0/tof.jpg} & \includegraphics[align=c, width=.15\linewidth]{figures/replica/tof_psmnet/hotel_0/psmnet_stereo.jpg} & \includegraphics[align=c, width=.15\linewidth]{figures/replica/tof_psmnet/hotel_0/tsdf_middle.jpg} & \includegraphics[align=c, width=.15\linewidth]{figures/replica/tof_psmnet/hotel_0/fused.jpg} & \includegraphics[align=c, width=.15\linewidth]{figures/replica/tof_psmnet/hotel_0/weighting.jpg} & \\ 
& \rotatebox[origin=c]{90}{Office 0} & \includegraphics[align=c, width=.15\linewidth]{figures/replica/tof_psmnet/office_0/model_small.jpg} & \includegraphics[align=c, width=.15\linewidth]{figures/replica/tof_psmnet/office_0/tof_small.jpg} & \includegraphics[align=c, width=.15\linewidth]{figures/replica/tof_psmnet/office_0/psmnet_stereo_small.jpg} & \includegraphics[align=c, width=.15\linewidth]{figures/replica/tof_psmnet/office_0/tsdf_middle_small.jpg} & \includegraphics[align=c, width=.15\linewidth]{figures/replica/tof_psmnet/office_0/fused.jpg} & \includegraphics[align=c, width=.15\linewidth]{figures/replica/tof_psmnet/office_0/weighting.jpg} & \\
& & Model & ToF~\cite{handa2014benchmark}  & PSMNet~\cite{chang2018pyramid} & TSDF Fusion~\cite{curless1996volumetric} & SenFuNet (Ours) & Our Sensor Weighting & \\
\end{tabular}
}
\caption{\boldparagraph{ToF\bplus{}PSMNet Fusion.} Our method fuses the sensors consistently better than TSDF Fusion~\cite{curless1996volumetric}.
In particular, our method learns to detect and remove outliers much more effectively (best viewed on screen).}
\label{fig:tof_psmnet_supp}
\end{figure*}
\FloatBarrier
\noindent reconstruction performance than RoutedFusion and better outlier handling than both RoutedFusion and TSDF Fusion.
In Fig.~\ref{fig:all_methods_denoising}, we add denoising to the model and evaluate the baseline methods against our method again. Our model achieves better outlier handling overall and more precise surface reconstruction in most regions compared to the Early Fusion method.

For more visual results, we refer to the supplementary videos.
\fi

\section*{J. Limitations}
\label{sec:limits}
\ifeccv
\else
\addcontentsline{toc}{section}{\nameref{sec:limits}}
\fi
While our method generates better reconstructions on average, specific local regions may still not improve if the wrong sensor weighting is estimated. Fig.~\ref{fig:fail_cases} shows four failure cases of our method. The top left visualization of $\{$SGM, PSMNet$\}$ without depth denoising shows that the PSMNet surface is selected to a large degree. Our method typically selects the more smooth surface (PSMNet), when compared to a noisy surface (SGM), even though the noisier surface (SGM) may be better on average. The red rectangles on the bottom row and in the top right example show less severe failure cases when our method performs smoothing when selection would have resulted in a more accurate surface prediction. This typically happens around edges, but may in rare cases happen on planar regions containing repetitive textures, for example a tiled bathroom shower (see bottom left example).
Lastly, our method has difficulties handling overlapping outliers from both sensors \ie~where both sensors have registered an outlier at the same voxel. See the orange rectangle in the top right example. This is due to the fact that the Outlier Filter can only be applied on voxels with a single sensor observation.
\ifeccv
\begin{figure*}[t]
\centering
{\scriptsize
\setlength{\tabcolsep}{1pt}
\renewcommand{\arraystretch}{1}
\begin{tabular}{ccccccc}
\rotatebox[origin=c]{90}{Hotel 0} & \includegraphics[align=c, width=.18\linewidth]{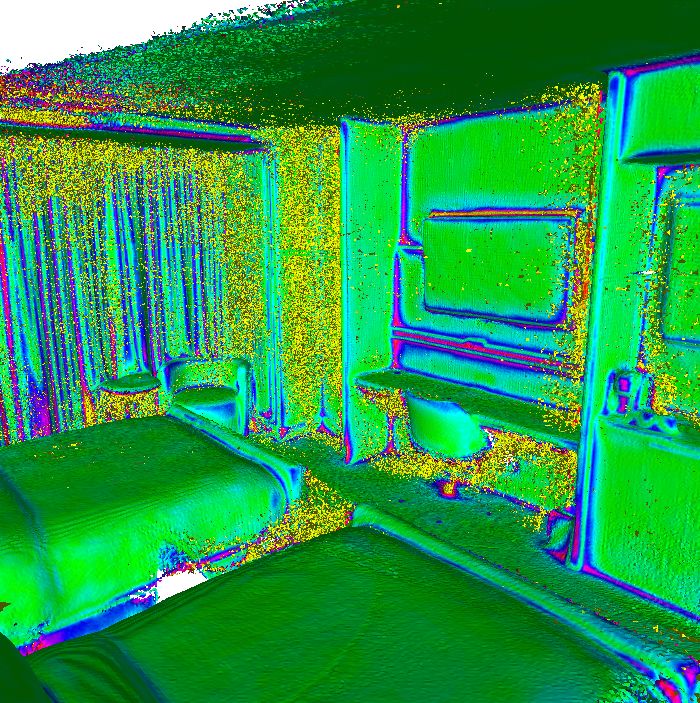} & \includegraphics[align=c, width=.18\linewidth]{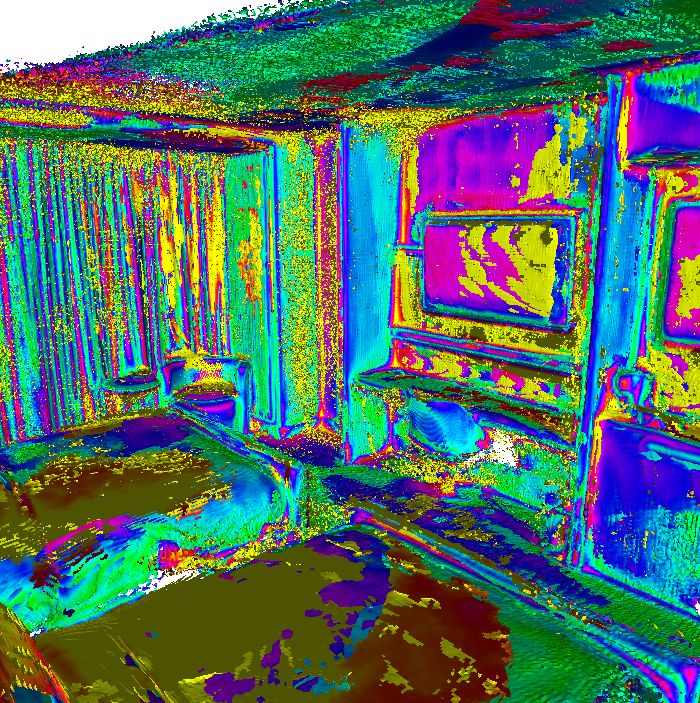} & \includegraphics[align=c, width=.18\linewidth]{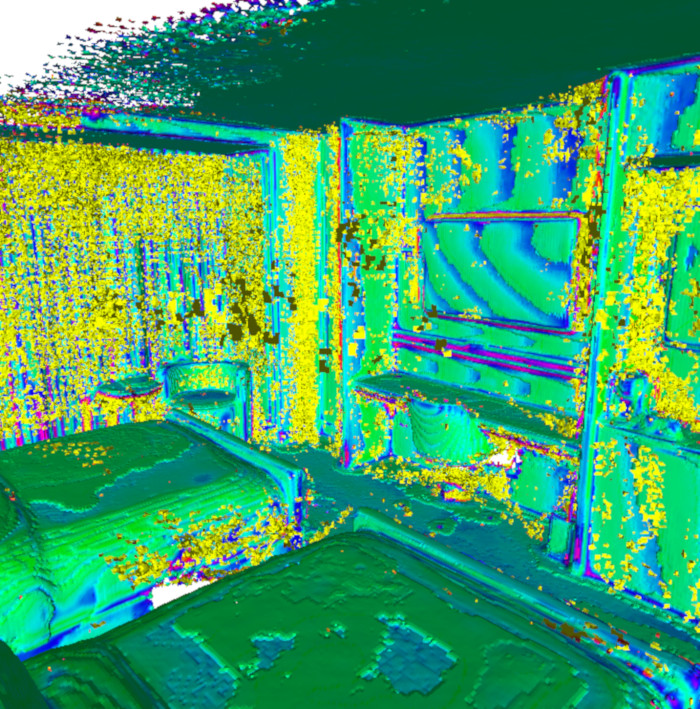} & \includegraphics[align=c, width=.18\linewidth]{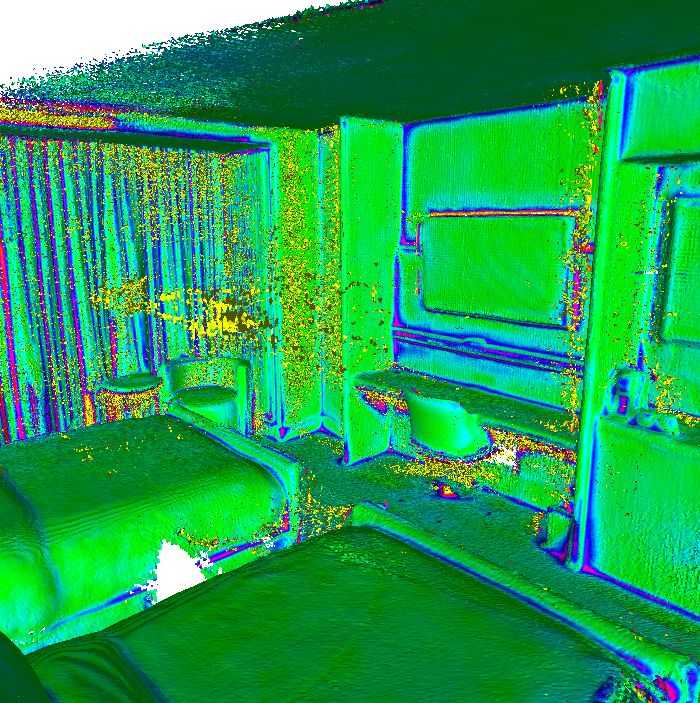} & \includegraphics[align=c, width=.18\linewidth]{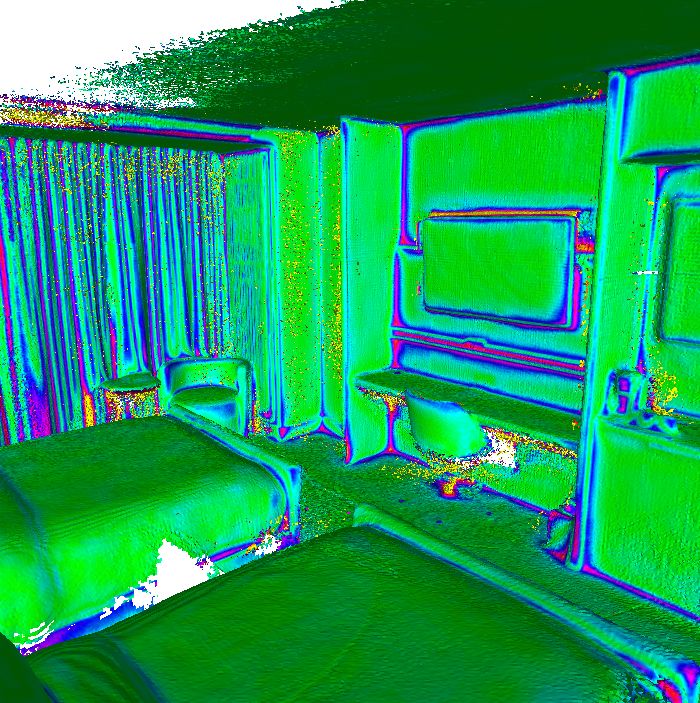} & \multirow[t]{12}{*}[31pt]{\includegraphics[align=t, width=.048\linewidth]{figures/colorbars/colorbar_outlier_filter.png}} \\
\rotatebox[origin=c]{90}{Office 0} & \includegraphics[align=c, width=.18\linewidth]{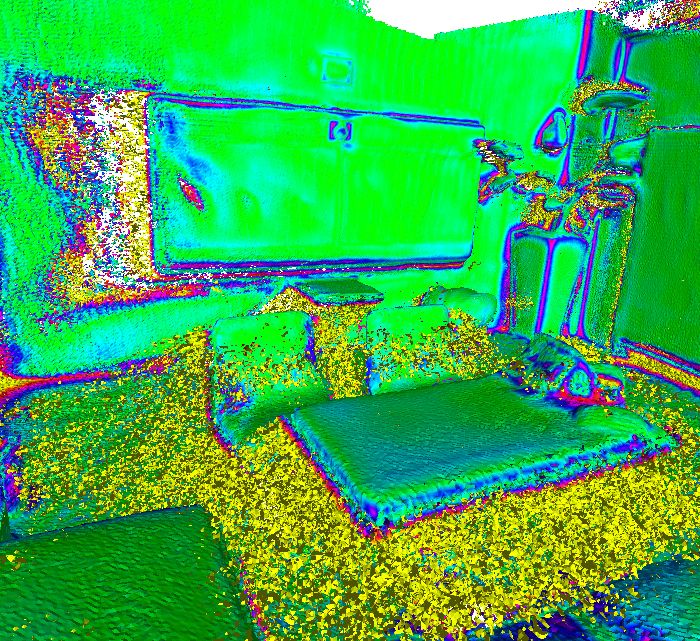} & \includegraphics[align=c, width=.18\linewidth]{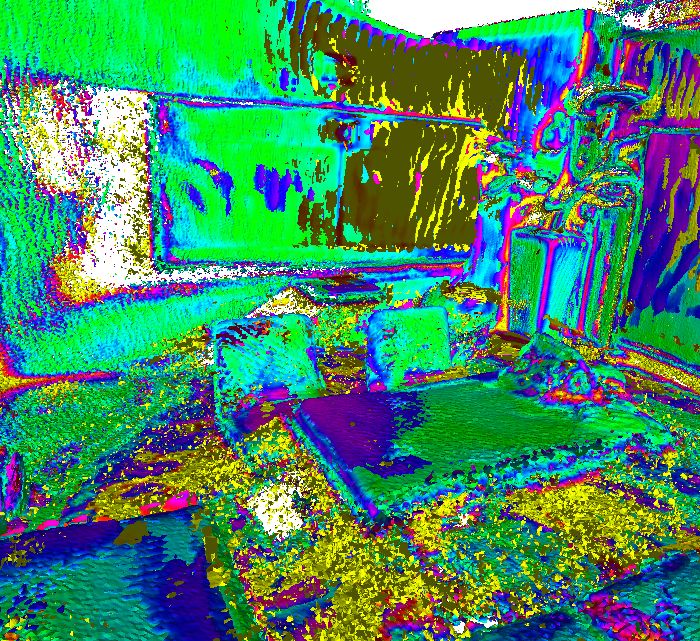} & \includegraphics[align=c, width=.18\linewidth]{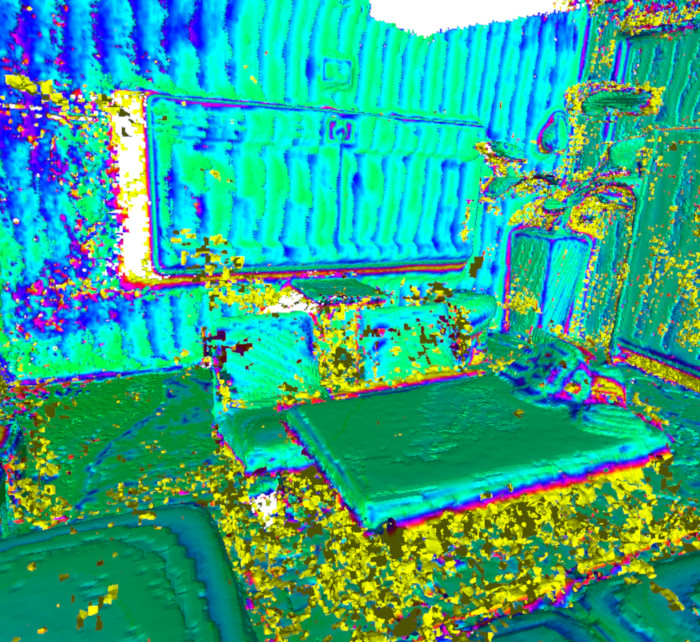} & \includegraphics[align=c, width=.18\linewidth]{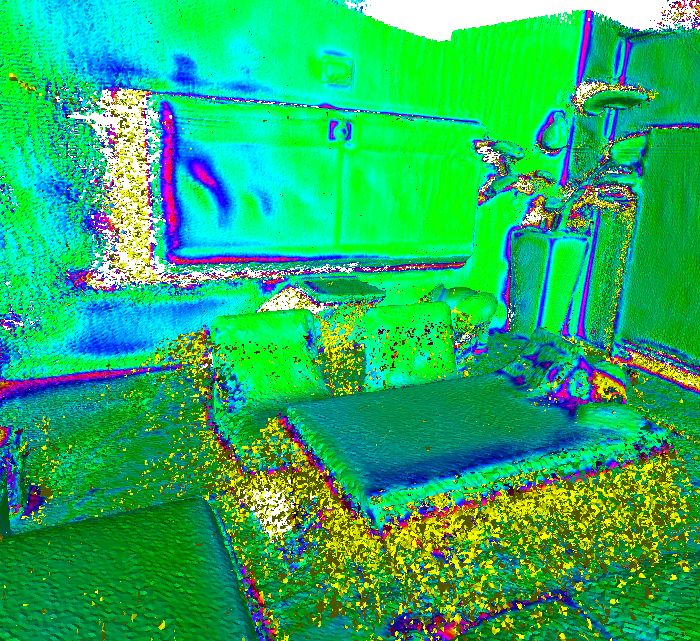} & \includegraphics[align=c, width=.18\linewidth]{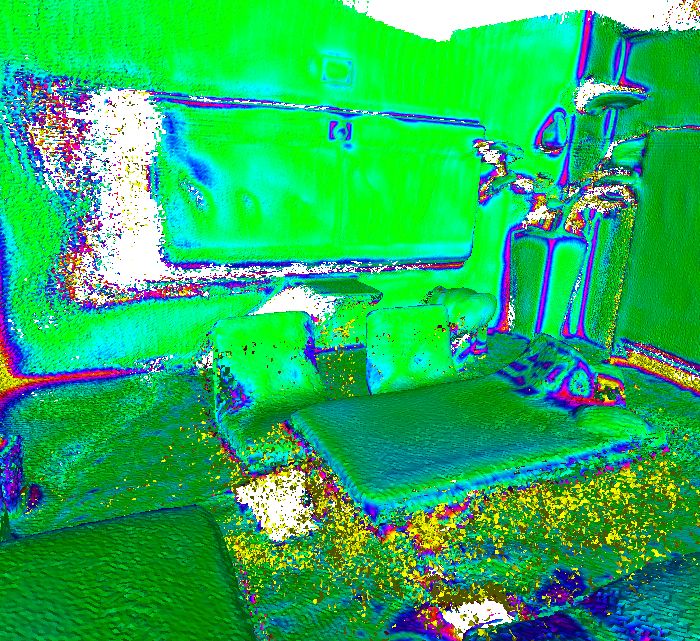} & \\
 & TSDF Fusion~\cite{curless1996volumetric} & RoutedFusion~\cite{Weder2020RoutedFusionLR} & DI-Fusion~\cite{huang2021di} & Eary Fusion & SenFuNet (Ours) & \\
 & & &  $\sigma$=0.15 & & &
\end{tabular}
}
\caption{\boldparagraph{SGM\bplus{}PSMNet Fusion with denoising.} Our method fuses the sensors consistently better than the baseline methods. Compared to the Early Fusion baseline, our method removes more outliers and reconstructs most surfaces better (best viewed on screen).}
\label{fig:all_methods_denoising}
\end{figure*}
\else
\begin{figure*}[t]
\centering
{\scriptsize
\setlength{\tabcolsep}{1pt}
\renewcommand{\arraystretch}{1}
\begin{tabular}{cccccc}
\rotatebox[origin=c]{90}{Hotel 0} & \includegraphics[align=c, width=.23\linewidth]{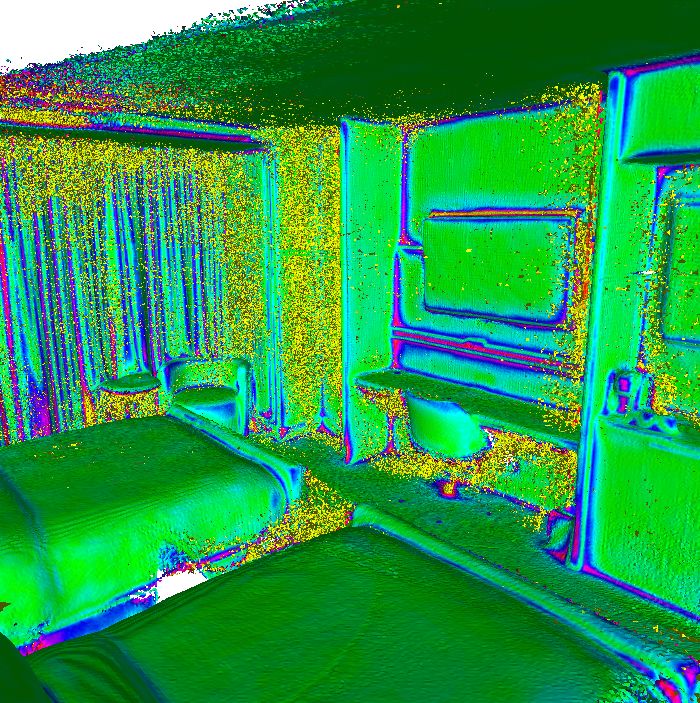} & \includegraphics[align=c, width=.23\linewidth]{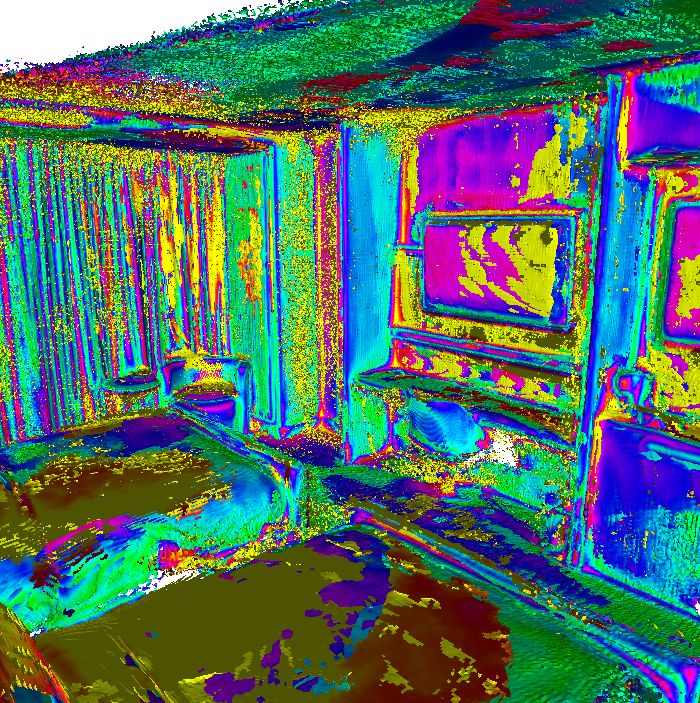} & \includegraphics[align=c, width=.23\linewidth]{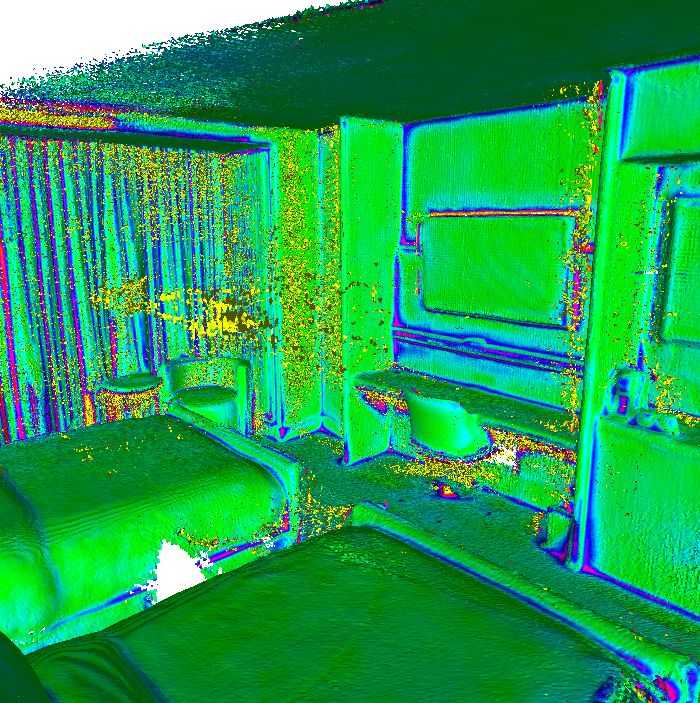} & \includegraphics[align=c, width=.23\linewidth]{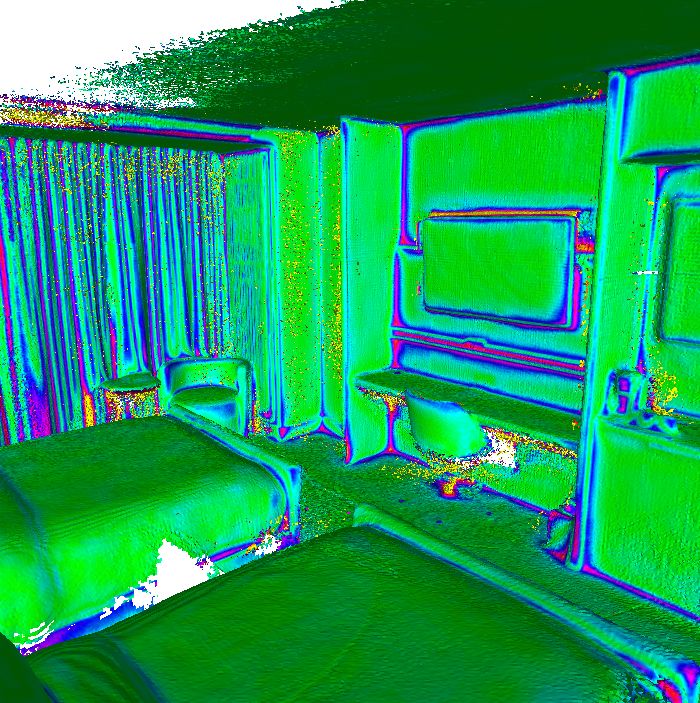} & \multirow[t]{12}{*}[62.5pt]{\includegraphics[align=t, width=.065\linewidth]{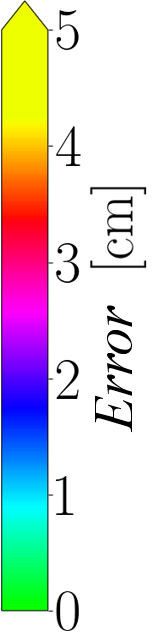}} \\
\rotatebox[origin=c]{90}{Office 0} & \includegraphics[align=c, width=.23\linewidth]{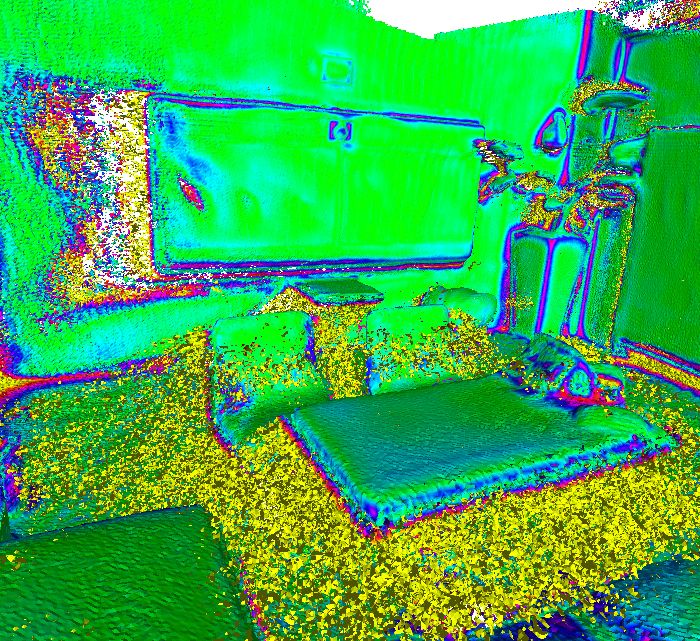} & \includegraphics[align=c, width=.23\linewidth]{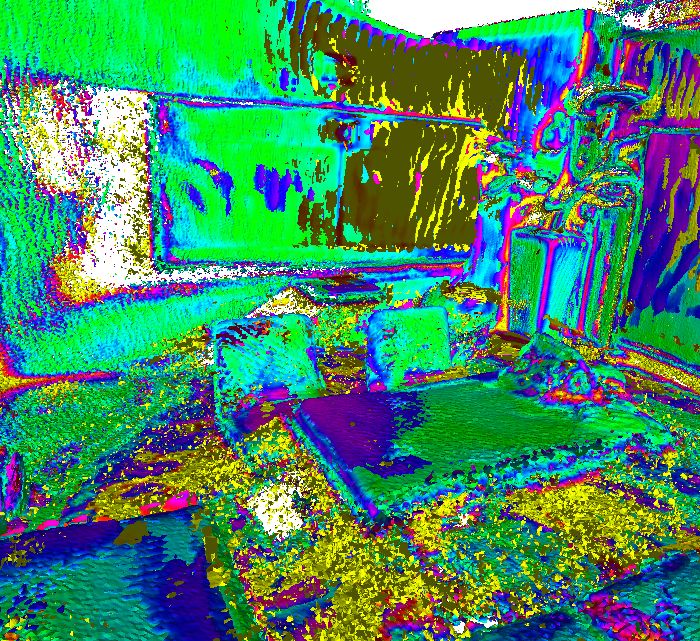} & \includegraphics[align=c, width=.23\linewidth]{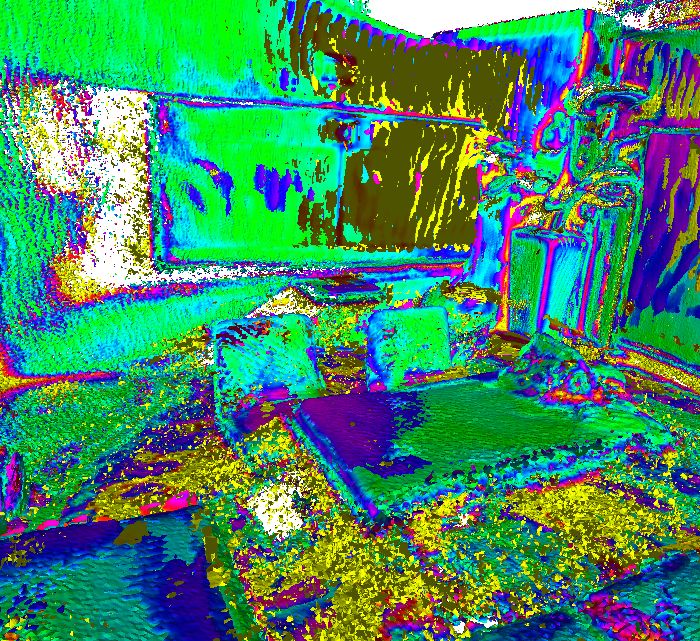} & \includegraphics[align=c, width=.23\linewidth]{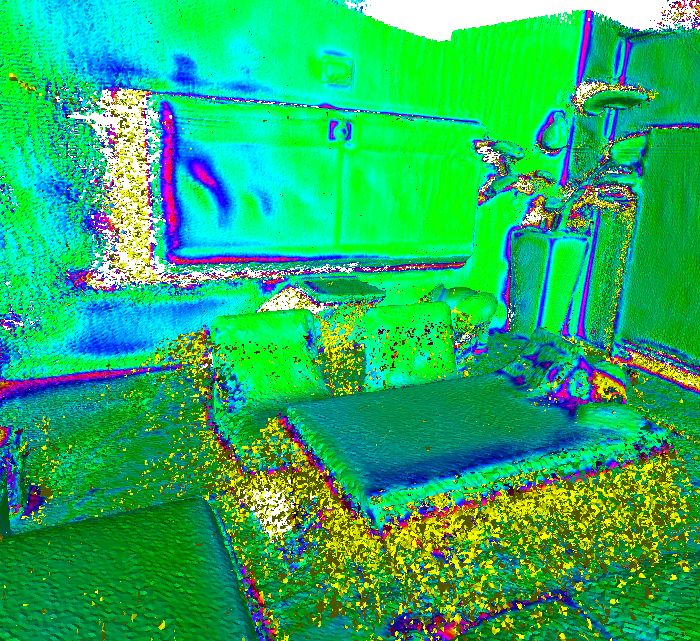} & \includegraphics[align=c, width=.23\linewidth]{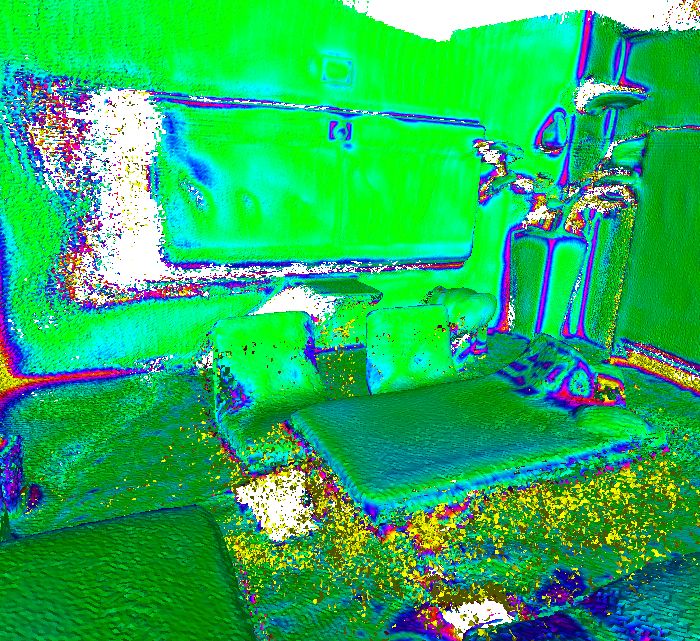} & \\
 & TSDF Fusion~\cite{curless1996volumetric} & RoutedFusion~\cite{Weder2020RoutedFusionLR} & Eary Fusion & SenFuNet (Ours) & \\
\end{tabular}
}
\caption{\boldparagraph{SGM\bplus{}PSMNet Fusion with denoising.} Our method fuses the sensors consistently better than the baseline methods. Compared to the Early Fusion baseline, our method removes more outliers and reconstructs most surfaces better (best viewed on screen).}
\label{fig:all_methods_denoising}
\end{figure*}
\fi

\ifeccv
\begin{figure*}[t]
\centering
\tiny
\setlength{\tabcolsep}{1pt}
\renewcommand{\arraystretch}{2.0}
\hspace*{-0.3cm}
\begin{tabular}{c|m{0.464\linewidth}<{\centering}|m{0.464\linewidth}<{\centering}|c}
 \cline{2-3}
& \normalsize Without Denoising &
\normalsize With Denoising & \\ \cline{2-3}
\end{tabular}
\renewcommand{\arraystretch}{1}
\begin{tabular}{cccccccc}
\rotatebox[origin=c]{90}{Office 0} &
\includegraphics[align=c, width=.153\linewidth]{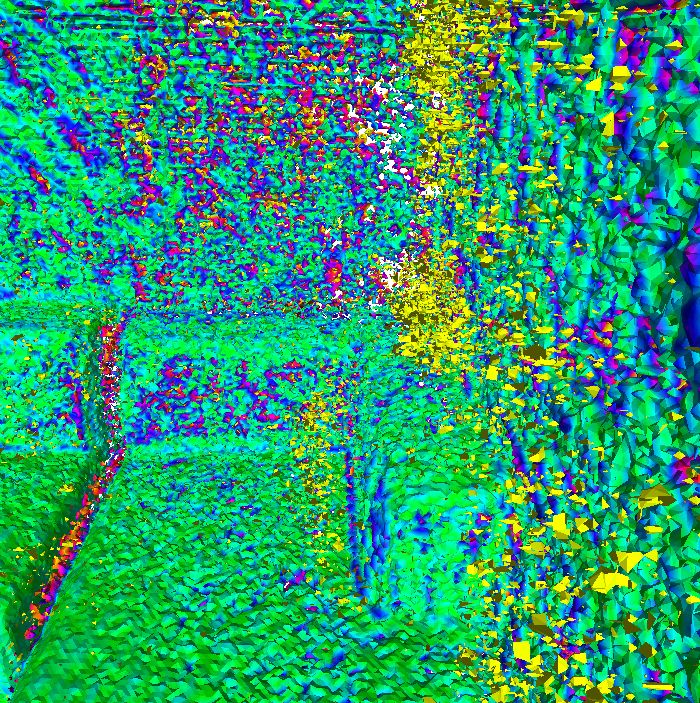} & 
\includegraphics[align=c, width=.153\linewidth]{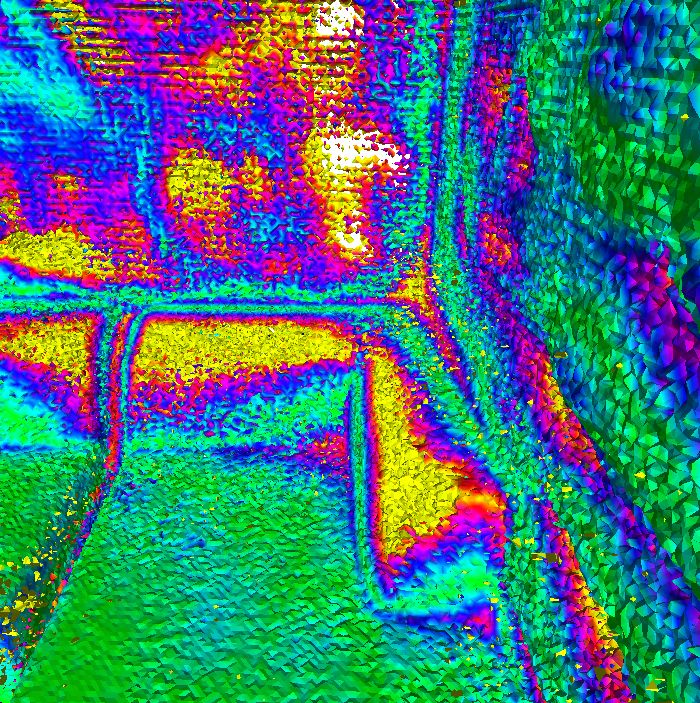} & 
\includegraphics[align=c, width=.153\linewidth]{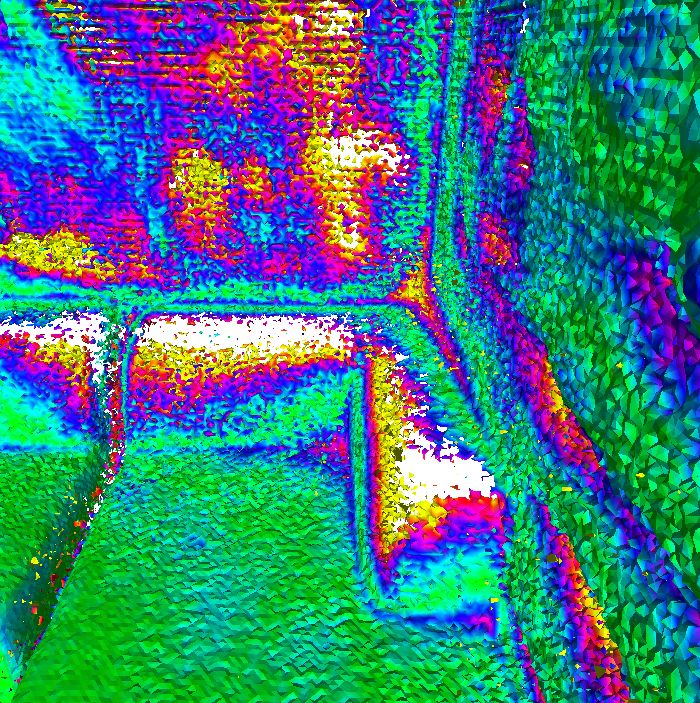} &
\includegraphics[align=c, trim={0 0 0.5cm 0}, clip, width=.153\linewidth]{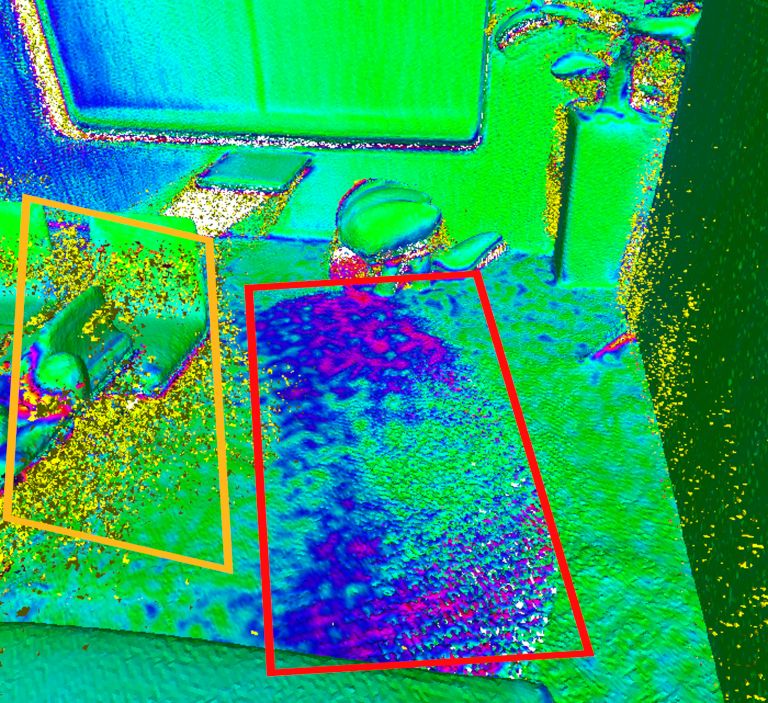} &
\includegraphics[align=c, trim={0 0 0.5cm 0}, clip, width=.153\linewidth]{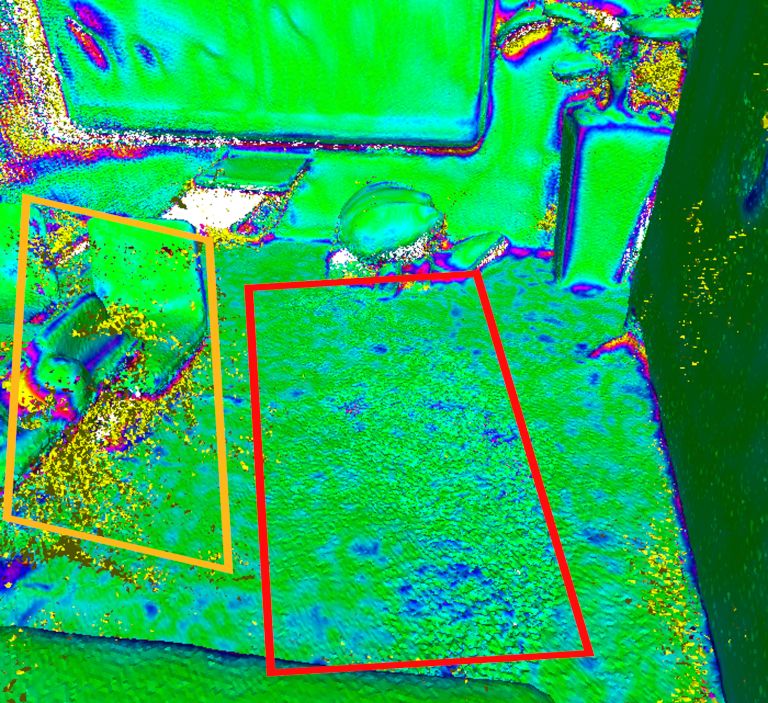} & 
\includegraphics[align=c, trim={0 0 0.5cm 0}, clip, width=.153\linewidth]{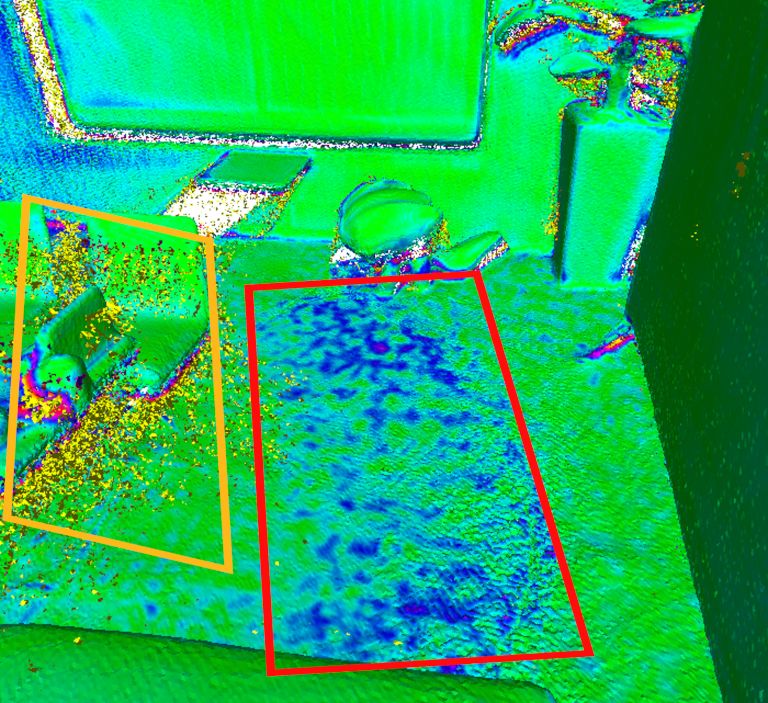} &
\multirow[t]{12}{*}[25pt]{\includegraphics[align=t, width=.04\linewidth]{figures/colorbars/colorbar_outlier_filter.png}} \\
 & SGM~\cite{hirschmuller2007stereo} & PSMNet~\cite{chang2018pyramid} & SenFuNet (Ours) & ToF~\cite{handa2014benchmark} & PSMNet~\cite{chang2018pyramid} & SenFuNet (Ours) \\
\rotatebox[origin=c]{90}{Hotel 0} & \includegraphics[align=c, width=.153\linewidth]{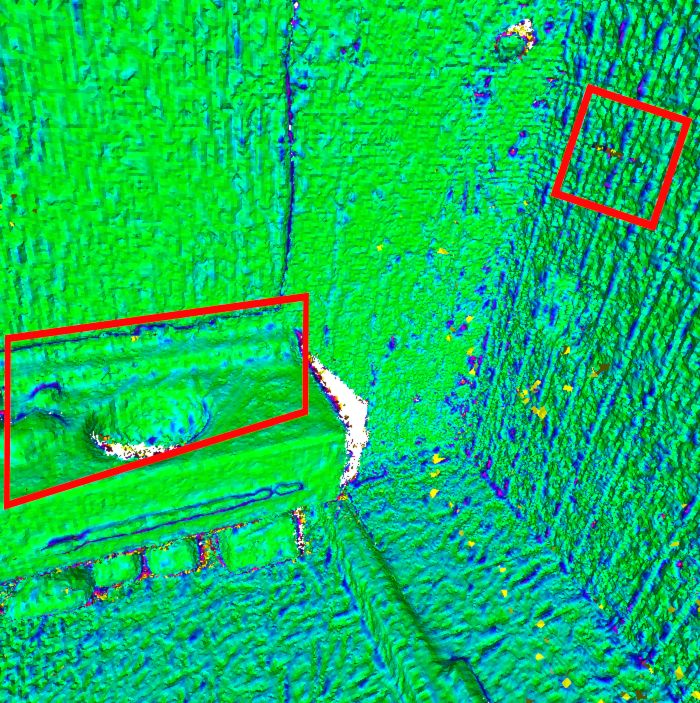} &  \includegraphics[align=c, width=.153\linewidth]{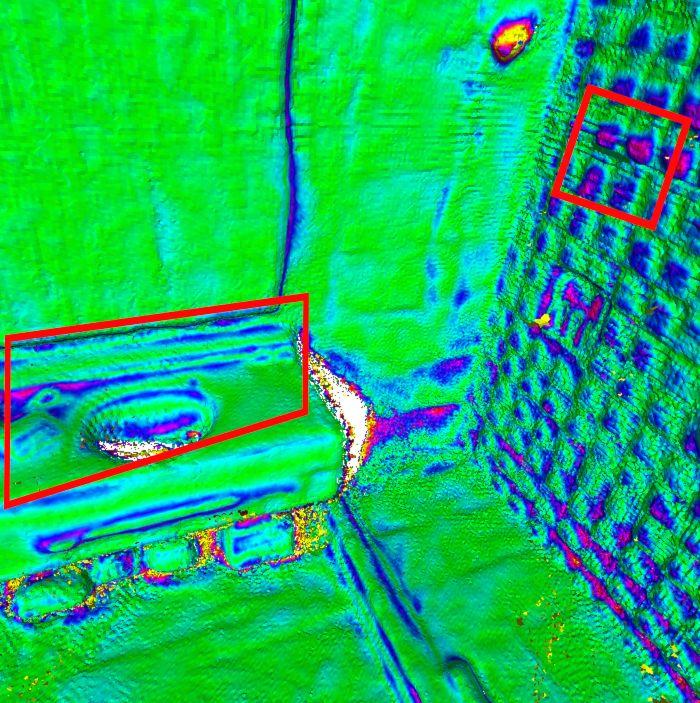} & \includegraphics[align=c, width=.153\linewidth]{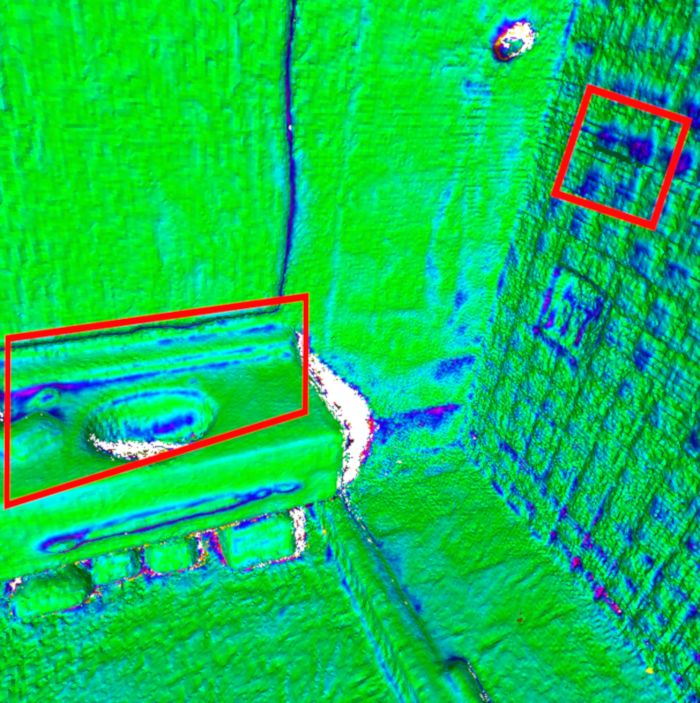} & \includegraphics[align=c, width=.153\linewidth]{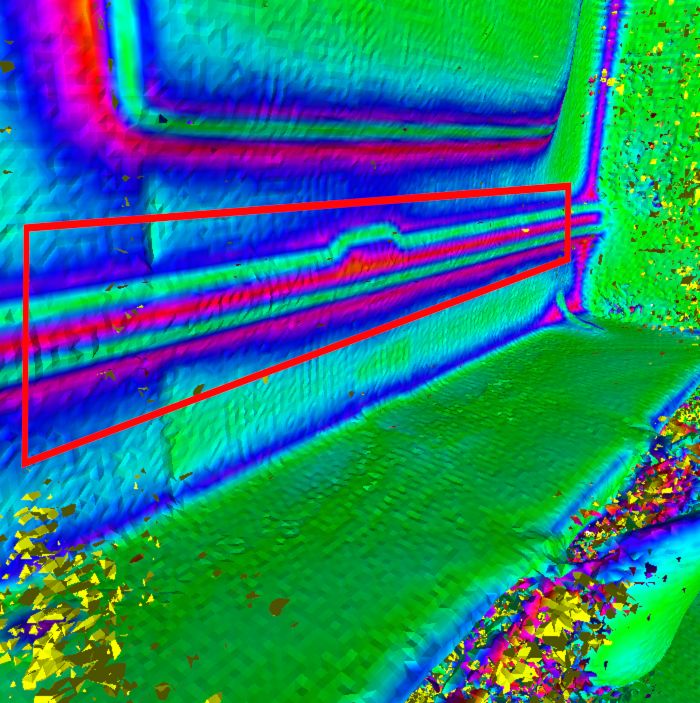} & \includegraphics[align=c, width=.153\linewidth]{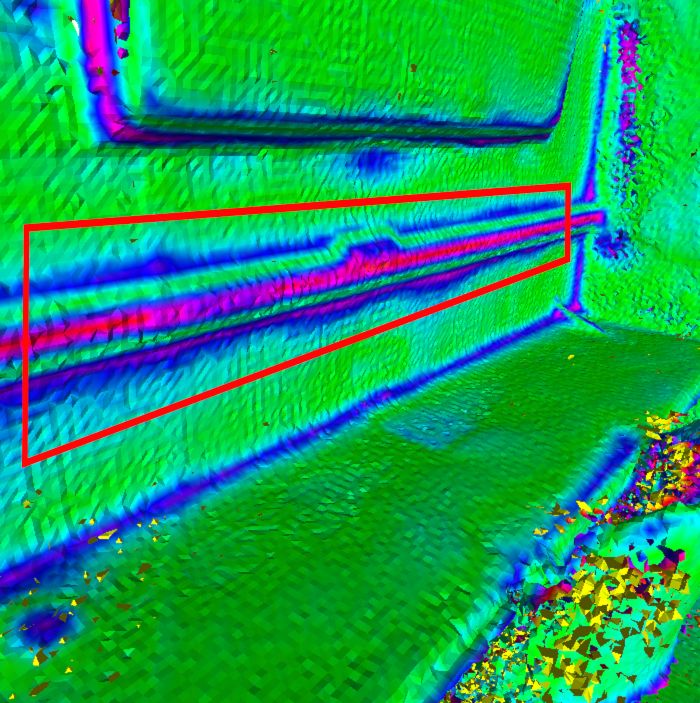} & \includegraphics[align=c, width=.153\linewidth]{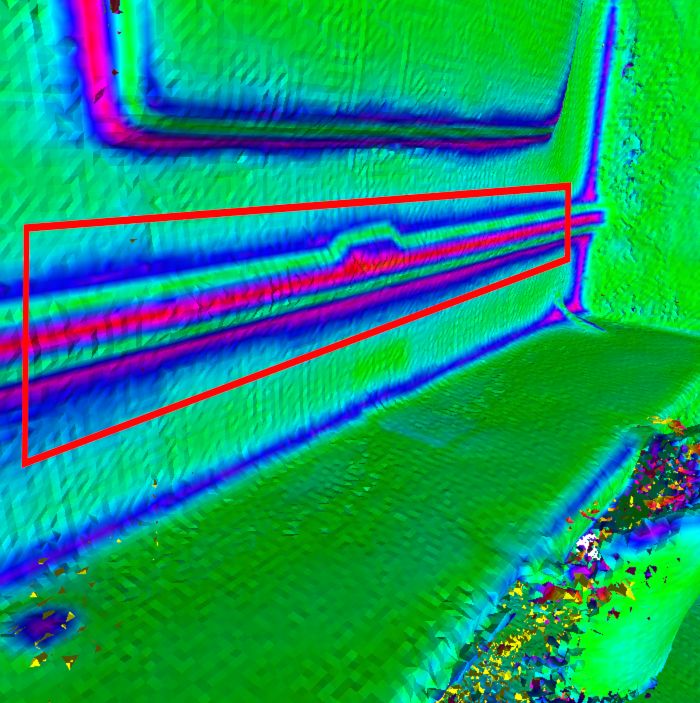} & \\
 & SGM~\cite{hirschmuller2007stereo} & PSMNet~\cite{chang2018pyramid} & SenFuNet (Ours) & SGM~\cite{hirschmuller2007stereo} & PSMNet~\cite{chang2018pyramid} & SenFuNet (Ours) & \\
\end{tabular}

\caption{\boldparagraph{Failure Cases.} While our method generates better reconstructions on average, specific local regions may not improve. Overlapping outliers, some edges and cases where one sensor looks noisy but is quantitatively good, are especially difficult to handle (best viewed on screen).}
\label{fig:fail_cases}
\end{figure*}
\else
\begin{figure*}[t]
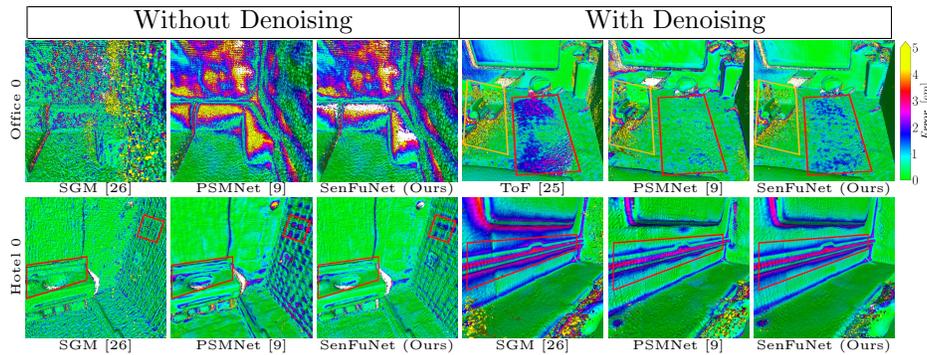

\centering
\scriptsize
\setlength{\tabcolsep}{1pt}
\renewcommand{\arraystretch}{2.0}
\hspace*{-0.53cm}
\begin{tabular}{c|m{0.464\linewidth}<{\centering}|m{0.464\linewidth}<{\centering}|c}
 \cline{2-3}
& \normalsize Without Denoising  &
\normalsize With Denoising & \\ \cline{2-3}
\end{tabular}
\renewcommand{\arraystretch}{1}
\begin{tabular}{cccccccc}
\rotatebox[origin=c]{90}{Office 0} &
\includegraphics[align=c, width=.153\linewidth]{figures/replica/fail_cases/office_0/sgm_psmnet_no_routing/sgm.jpg} & 
\includegraphics[align=c, width=.153\linewidth]{figures/replica/fail_cases/office_0/sgm_psmnet_no_routing/psmnet.jpg} & 
\includegraphics[align=c, width=.153\linewidth]{figures/replica/fail_cases/office_0/sgm_psmnet_no_routing/fused.jpg} &
\includegraphics[align=c, trim={0 0 0.5cm 0}, clip, width=.153\linewidth]{figures/replica/fail_cases/office_0/tof_psmnet_routing/tof.jpg} &
\includegraphics[align=c, trim={0 0 0.5cm 0}, clip, width=.153\linewidth]{figures/replica/fail_cases/office_0/tof_psmnet_routing/psmnet.jpg} & 
\includegraphics[align=c, trim={0 0 0.5cm 0}, clip, width=.153\linewidth]{figures/replica/fail_cases/office_0/tof_psmnet_routing/fused.jpg} &
\multirow[t]{12}{*}[35pt]{\includegraphics[align=t, width=.04\linewidth]{figures/colorbars/colorbar_outlier_filter.png}} \\
 & SGM~\cite{hirschmuller2007stereo} & PSMNet~\cite{chang2018pyramid} & SenFuNet (Ours) & ToF~\cite{handa2014benchmark} & PSMNet~\cite{chang2018pyramid} & SenFuNet (Ours) \\
\rotatebox[origin=c]{90}{Hotel 0} & \includegraphics[align=c, width=.153\linewidth]{figures/replica/fail_cases/hotel_0/sgm_psmnet_no_routing/sgm.jpg} &  \includegraphics[align=c, width=.153\linewidth]{figures/replica/fail_cases/hotel_0/sgm_psmnet_no_routing/psmnet.jpg} & \includegraphics[align=c, width=.153\linewidth]{figures/replica/fail_cases/hotel_0/sgm_psmnet_no_routing/fused.jpg} & \includegraphics[align=c, width=.153\linewidth]{figures/replica/fail_cases/hotel_0/sgm_psmnet_routing/sgm.jpg} & \includegraphics[align=c, width=.153\linewidth]{figures/replica/fail_cases/hotel_0/sgm_psmnet_routing/psmnet.jpg} & \includegraphics[align=c, width=.153\linewidth]{figures/replica/fail_cases/hotel_0/sgm_psmnet_routing/fused.jpg} & \\
 & SGM~\cite{hirschmuller2007stereo} & PSMNet~\cite{chang2018pyramid} & SenFuNet (Ours) & SGM~\cite{hirschmuller2007stereo} & PSMNet~\cite{chang2018pyramid} & SenFuNet (Ours) & \\
\end{tabular}

\caption{\boldparagraph{Failure Cases.} While our method generates better reconstructions on average, specific local regions may not improve. Overlapping outliers, some edges and cases where one sensor looks noisy but is quantitatively good, are especially difficult to handle (best viewed on screen).}
\label{fig:fail_cases}
\end{figure*}
\fi

\clearpage
%
%
\bibliographystyle{splncs04_eccv}
\bibliography{egbib}

\begin{thebibliography}{10}
\providecommand{\url}[1]{\texttt{#1}}
\providecommand{\urlprefix}{URL }
\providecommand{\doi}[1]{https://doi.org/#1}

\bibitem{agresti2017deep}
Agresti, G., Minto, L., Marin, G., Zanuttigh, P.: Deep learning for confidence
  information in stereo and tof data fusion. In: Proceedings of the IEEE
  International Conference on Computer Vision. pp. 697--705 (2017)

\bibitem{agresti2019stereo}
Agresti, G., Minto, L., Marin, G., Zanuttigh, P.: Stereo and tof data fusion by
  learning from synthetic data. Information Fusion  \textbf{49},  161--173
  (2019)

\bibitem{ali2019multi}
Ali, M.K., Rajput, A., Shahzad, M., Khan, F., Akhtar, F., B{\"o}rner, A.:
  Multi-sensor depth fusion framework for real-time 3d reconstruction. IEEE
  Access  \textbf{7},  136471--136480 (2019)

\bibitem{bovzivc2021transformerfusion}
Bo{\v{z}}i{\v{c}}, A., Palafox, P., Thies, J., Dai, A., Nie{\ss}ner, M.:
  Transformerfusion: Monocular rgb scene reconstruction using transformers.
  arXiv preprint arXiv:2107.02191  (2021)

\bibitem{opencv_library}
Bradski, G.: {The OpenCV Library}. Dr. Dobb's Journal of Software Tools  (2000)

\bibitem{7900211}
{Bylow}, E., {Olsson}, C., {Kahl}, F.: Robust online 3d reconstruction
  combining a depth sensor and sparse feature points. In: 2016 23rd
  International Conference on Pattern Recognition (ICPR). pp. 3709--3714 (2016)

\bibitem{bylow2019combining}
Bylow, E., Maier, R., Kahl, F., Olsson, C.: Combining depth fusion and
  photometric stereo for fine-detailed 3d models. In: Scandinavian Conference
  on Image Analysis. pp. 261--274. Springer (2019)

\bibitem{cao2018real}
Cao, Y.P., Kobbelt, L., Hu, S.M.: Real-time high-accuracy three-dimensional
  reconstruction with consumer rgb-d cameras. ACM Transactions on Graphics
  (TOG)  \textbf{37}(5),  1--16 (2018)

\bibitem{chang2018pyramid}
Chang, J.R., Chen, Y.S.: Pyramid stereo matching network. In: Proceedings of
  the IEEE Conference on Computer Vision and Pattern Recognition. pp.
  5410--5418 (2018)

\bibitem{choe2021volumefusion}
Choe, J., Im, S., Rameau, F., Kang, M., Kweon, I.S.: Volumefusion: Deep depth
  fusion for 3d scene reconstruction. In: Proceedings of the IEEE/CVF
  International Conference on Computer Vision (ICCV). pp. 16086--16095 (October
  2021)

\bibitem{Choi2012fusion}
Choi, O., Lee, S.: Fusion of time-of-flight and stereo for disambiguation of
  depth measurements. In: Lee, K.M., Matsushita, Y., Rehg, J.M., Hu, Z. (eds.)
  Computer Vision - {ACCV} 2012, 11th Asian Conference on Computer Vision,
  Daejeon, Korea, November 5-9, 2012, Revised Selected Papers, Part {IV}.
  Lecture Notes in Computer Science, vol.~7727, pp. 640--653. Springer (2012).
  \doi{10.1007/978-3-642-37447-0\_49},
  \url{https://doi.org/10.1007/978-3-642-37447-0\_49}

\bibitem{choi2015robust}
Choi, S., Zhou, Q.Y., Koltun, V.: Robust reconstruction of indoor scenes. In:
  Proceedings of the IEEE Conference on Computer Vision and Pattern
  Recognition. pp. 5556--5565 (2015)

\bibitem{cignoni2008meshlab}
Cignoni, P., Callieri, M., Corsini, M., Dellepiane, M., Ganovelli, F.,
  Ranzuglia, G., et~al.: Meshlab: an open-source mesh processing tool. In:
  Eurographics Italian chapter conference. vol.~2008, pp. 129--136. Salerno,
  Italy (2008)

\bibitem{curless1996volumetric}
Curless, B., Levoy, M.: A volumetric method for building complex models from
  range images. In: Proceedings of the 23rd annual conference on Computer
  graphics and interactive techniques. pp. 303--312 (1996)

\bibitem{dai2017bundlefusion}
Dai, A., Nie{\ss}ner, M., Zollh{\"o}fer, M., Izadi, S., Theobalt, C.:
  Bundlefusion: Real-time globally consistent 3d reconstruction using
  on-the-fly surface reintegration. ACM Transactions on Graphics (ToG)
  \textbf{36}(4), ~1 (2017)

\bibitem{dal2015probabilistic}
Dal~Mutto, C., Zanuttigh, P., Cortelazzo, G.M.: Probabilistic tof and stereo
  data fusion based on mixed pixels measurement models. IEEE transactions on
  pattern analysis and machine intelligence  \textbf{37}(11),  2260--2272
  (2015)

\bibitem{deng2021tof}
Deng, Y., Xiao, J., Zhou, S.Z.: Tof and stereo data fusion using dynamic search
  range stereo matching. IEEE Transactions on Multimedia  \textbf{24},
  2739--2751 (2021)

\bibitem{dong2018psdf}
Dong, W., Wang, Q., Wang, X., Zha, H.: Psdf fusion: Probabilistic signed
  distance function for on-the-fly 3d data fusion and scene reconstruction. In:
  Proceedings of the European Conference on Computer Vision (ECCV). pp.
  701--717 (2018)

\bibitem{duan2012probabilistic}
Duan, Y., Pei, M., Wang, Y.: Probabilistic depth map fusion of kinect and
  stereo in real-time. In: 2012 IEEE International Conference on Robotics and
  Biomimetics (ROBIO). pp. 2317--2322. IEEE (2012)

\bibitem{duan2015unified}
Duan, Y., Pei, M., Wang, Y., Yang, M., Qin, I., Jia, Y.: A unified
  probabilistic framework for real-time depth map fusion. J. Inf. Sci. Eng.
  \textbf{31}(4),  1309--1327 (2015)

\bibitem{evangelidis2015fusion}
Evangelidis, G.D., Hansard, M., Horaud, R.: Fusion of range and stereo data for
  high-resolution scene-modeling. IEEE transactions on pattern analysis and
  machine intelligence  \textbf{37}(11),  2178--2192 (2015)

\bibitem{girardeau2016cloudcompare}
Girardeau-Montaut, D.: Cloudcompare. France: EDF R\&D Telecom ParisTech  (2016)

\bibitem{golodetz2018collaborative}
Golodetz, S., Cavallari, T., Lord, N.A., Prisacariu, V.A., Murray, D.W., Torr,
  P.H.: Collaborative large-scale dense 3d reconstruction with online
  inter-agent pose optimisation. IEEE transactions on visualization and
  computer graphics  \textbf{24}(11),  2895--2905 (2018)

\bibitem{gu20203d}
Gu, P., Zhou, F., Yu, D., Wan, F., Wang, W., Yu, B.: A 3d reconstruction method
  using multisensor fusion in large-scale indoor scenes. Complexity
  \textbf{2020} (2020)

\bibitem{handa2014benchmark}
Handa, A., Whelan, T., McDonald, J., Davison, A.J.: A benchmark for rgb-d
  visual odometry, 3d reconstruction and slam. In: 2014 IEEE international
  conference on Robotics and automation (ICRA). pp. 1524--1531. IEEE (2014)

\bibitem{hirschmuller2007stereo}
Hirschmuller, H.: Stereo processing by semiglobal matching and mutual
  information. IEEE Transactions on pattern analysis and machine intelligence
  \textbf{30}(2),  328--341 (2007)

\bibitem{huang2021di}
Huang, J., Huang, S.S., Song, H., Hu, S.M.: Di-fusion: Online implicit 3d
  reconstruction with deep priors. In: Proceedings of the IEEE/CVF Conference
  on Computer Vision and Pattern Recognition. pp. 8932--8941 (2021)

\bibitem{izadi2011kinectfusion}
Izadi, S., Kim, D., Hilliges, O., Molyneaux, D., Newcombe, R., Kohli, P.,
  Shotton, J., Hodges, S., Freeman, D., Davison, A., et~al.: Kinectfusion:
  real-time 3d reconstruction and interaction using a moving depth camera. In:
  Proceedings of the 24th annual ACM symposium on User interface software and
  technology. pp. 559--568. ACM (2011)

\bibitem{Kahler2015infiniTAM}
K{\"{a}}hler, O., Prisacariu, V.A., Ren, C.Y., Sun, X., Torr, P.H.S., Murray,
  D.W.: Very high frame rate volumetric integration of depth images on mobile
  devices. {IEEE} Trans. Vis. Comput. Graph.  \textbf{21}(11),  1241--1250
  (2015). \doi{10.1109/TVCG.2015.2459891},
  \url{https://doi.org/10.1109/TVCG.2015.2459891}

\bibitem{kazhdan2013screened}
Kazhdan, M., Hoppe, H.: Screened poisson surface reconstruction. ACM
  Transactions on Graphics (ToG)  \textbf{32}(3),  1--13 (2013)

\bibitem{kim2009multi}
Kim, Y.M., Theobalt, C., Diebel, J., Kosecka, J., Miscusik, B., Thrun, S.:
  Multi-view image and tof sensor fusion for dense 3d reconstruction. In: 2009
  IEEE 12th international conference on computer vision workshops, ICCV
  workshops. pp. 1542--1549. IEEE (2009)

\bibitem{knapitsch2017tanks}
Knapitsch, A., Park, J., Zhou, Q.Y., Koltun, V.: Tanks and temples:
  Benchmarking large-scale scene reconstruction. ACM Transactions on Graphics
  (ToG)  \textbf{36}(4),  1--13 (2017)

\bibitem{lefloch2015anisotropic}
Lefloch, D., Weyrich, T., Kolb, A.: Anisotropic point-based fusion. In: 2015
  18th International Conference on Information Fusion (Fusion). pp. 2121--2128.
  IEEE (2015)

\bibitem{lorensen1987marching}
Lorensen, W.E., Cline, H.E.: Marching cubes: A high resolution 3d surface
  construction algorithm. ACM siggraph computer graphics  \textbf{21}(4),
  163--169 (1987)

\bibitem{van2008visualizing}
Van~der Maaten, L., Hinton, G.: Visualizing data using t-sne. Journal of
  machine learning research  \textbf{9}(11) (2008)

\bibitem{maddern2016real}
Maddern, W., Newman, P.: Real-time probabilistic fusion of sparse 3d lidar and
  dense stereo. In: 2016 IEEE/RSJ International Conference on Intelligent
  Robots and Systems (IROS). pp. 2181--2188. IEEE (2016)

\bibitem{habitat19iccv}
{Manolis Savva*}, {Abhishek Kadian*}, {Oleksandr Maksymets*}, Zhao, Y.,
  Wijmans, E., Jain, B., Straub, J., Liu, J., Koltun, V., Malik, J., Parikh,
  D., Batra, D.: Habitat: {A} {P}latform for {E}mbodied {AI} {R}esearch. In:
  Proceedings of the IEEE/CVF International Conference on Computer Vision
  (ICCV) (2019)

\bibitem{marin2016reliable}
Marin, G., Zanuttigh, P., Mattoccia, S.: Reliable fusion of tof and stereo
  depth driven by confidence measures. In: European Conference on Computer
  Vision. pp. 386--401. Springer (2016)

\bibitem{martins2018fusion}
Martins, D., Van~Hecke, K., De~Croon, G.: Fusion of stereo and still monocular
  depth estimates in a self-supervised learning context. In: 2018 IEEE
  International Conference on Robotics and Automation (ICRA). pp. 849--856.
  IEEE (2018)

\bibitem{murez2020atlas}
Murez, Z., van As, T., Bartolozzi, J., Sinha, A., Badrinarayanan, V.,
  Rabinovich, A.: Atlas: End-to-end 3d scene reconstruction from posed images.
  In: Computer Vision--ECCV 2020: 16th European Conference, Glasgow, UK, August
  23--28, 2020, Proceedings, Part VII 16. pp. 414--431. Springer (2020)

\bibitem{newcombe2011kinectfusion}
Newcombe, R.A., Izadi, S., Hilliges, O., Molyneaux, D., Kim, D., Davison, A.J.,
  Kohli, P., Shotton, J., Hodges, S., Fitzgibbon, A.W.: Kinectfusion: Real-time
  dense surface mapping and tracking. In: ISMAR. vol.~11, pp. 127--136 (2011)

\bibitem{newcombe2011dtam}
Newcombe, R.A., Lovegrove, S.J., Davison, A.J.: Dtam: Dense tracking and
  mapping in real-time. In: ICCV (2011)

\bibitem{voxel_hashing}
Nießner, M., Zollhöfer, M., Izadi, S., Stamminger, M.: Real-time 3d
  reconstruction at scale using voxel hashing. ACM Transactions on Graphics
  (TOG)  \textbf{32} (11 2013). \doi{10.1145/2508363.2508374}

\bibitem{Oleynikova2017voxblox}
Oleynikova, H., Taylor, Z., Fehr, M., Siegwart, R., Nieto, J.I.: Voxblox:
  Incremental 3d euclidean signed distance fields for on-board {MAV} planning.
  In: 2017 {IEEE/RSJ} International Conference on Intelligent Robots and
  Systems, {IROS} 2017, Vancouver, BC, Canada, September 24-28, 2017. pp.
  1366--1373. {IEEE} (2017). \doi{10.1109/IROS.2017.8202315},
  \url{https://doi.org/10.1109/IROS.2017.8202315}

\bibitem{park2018high}
Park, K., Kim, S., Sohn, K.: High-precision depth estimation with the 3d lidar
  and stereo fusion. In: 2018 IEEE International Conference on Robotics and
  Automation (ICRA). pp. 2156--2163. IEEE (2018)

\bibitem{patil2020don}
Patil, V., Van~Gansbeke, W., Dai, D., Van~Gool, L.: Don’t forget the past:
  Recurrent depth estimation from monocular video. IEEE Robotics and Automation
  Letters  \textbf{5}(4),  6813--6820 (2020)

\bibitem{poggi2016deep}
Poggi, M., Mattoccia, S.: Deep stereo fusion: combining multiple disparity
  hypotheses with deep-learning. In: 2016 Fourth International Conference on 3D
  Vision (3DV). pp. 138--147. IEEE (2016)

\bibitem{pu2019sdf}
Pu, C., Song, R., Tylecek, R., Li, N., Fisher, R.B.: Sdf-man: Semi-supervised
  disparity fusion with multi-scale adversarial networks. Remote Sensing
  \textbf{11}(5), ~487 (2019)

\bibitem{qiu2019deepLidar}
Qiu, J., Cui, Z., Zhang, Y., Zhang, X., Liu, S., Zeng, B., Pollefeys, M.:
  Deeplidar: Deep surface normal guided depth prediction for outdoor scene from
  sparse lidar data and single color image. In: {IEEE} Conference on Computer
  Vision and Pattern Recognition, {CVPR} 2019, Long Beach, CA, USA, June 16-20,
  2019. pp. 3313--3322. Computer Vision Foundation / {IEEE} (2019).
  \doi{10.1109/CVPR.2019.00343},
  \url{http://openaccess.thecvf.com/content\_CVPR\_2019/html/Qiu\_DeepLiDAR\_Deep\_Surface\_Normal\_Guided\_Depth\_Prediction\_for\_Outdoor\_Scene\_CVPR\_2019\_paper.html}

\bibitem{rozumnyi2019learned}
Rozumnyi, D., Cherabier, I., Pollefeys, M., Oswald, M.R.: Learned semantic
  multi-sensor depth map fusion. In: Internatioal Conference on Computer Vision
  Workshop (ICCVW), Workshop on 3D Reconstruction in the Wild, 2019. Seoul,
  South Korea (2019)

\bibitem{schonberger2016pixelwise}
Sch{\"o}nberger, J.L., Zheng, E., Frahm, J.M., Pollefeys, M.: Pixelwise view
  selection for unstructured multi-view stereo. In: European Conference on
  Computer Vision. pp. 501--518. Springer (2016)

\bibitem{schops2019bad}
Schops, T., Sattler, T., Pollefeys, M.: {BAD SLAM}: Bundle adjusted direct
  {RGB-D} {SLAM}. In: CVPR (2019)

\bibitem{6751517}
{Steinbrucker}, F., {Kerl}, C., {Cremers}, D., {Sturm}, J.: Large-scale
  multi-resolution surface reconstruction from rgb-d sequences. In: 2013 IEEE
  International Conference on Computer Vision. pp. 3264--3271 (2013)

\bibitem{straub2019replica}
Straub, J., Whelan, T., Ma, L., Chen, Y., Wijmans, E., Green, S., Engel, J.J.,
  Mur-Artal, R., Ren, C., Verma, S., et~al.: The replica dataset: A digital
  replica of indoor spaces. arXiv preprint arXiv:1906.05797  (2019)

\bibitem{sucar2021iMap}
Sucar, E., Liu, S., Ortiz, J., Davison, A.: {iMAP}: Implicit mapping and
  positioning in real-time. In: Proceedings of the IEEE International
  Conference on Computer Vision (2021)

\bibitem{sun2021neuralrecon}
Sun, J., Xie, Y., Chen, L., Zhou, X., Bao, H.: Neuralrecon: Real-time coherent
  3d reconstruction from monocular video. In: Proceedings of the IEEE/CVF
  Conference on Computer Vision and Pattern Recognition. pp. 15598--15607
  (2021)

\bibitem{van2012sensor}
Van~Baar, J., Beardsley, P., Pollefeys, M., Gross, M.: Sensor fusion for depth
  estimation, including tof and thermal sensors. In: 2012 Second International
  Conference on 3D Imaging, Modeling, Processing, Visualization \&
  Transmission. pp. 472--478. IEEE (2012)

\bibitem{wasenmuller2016corbs}
Wasenm{\"u}ller, O., Meyer, M., Stricker, D.: Corbs: Comprehensive rgb-d
  benchmark for slam using kinect v2. In: 2016 IEEE Winter Conference on
  Applications of Computer Vision (WACV). pp.~1--7. IEEE (2016)

\bibitem{Weder2020RoutedFusionLR}
Weder, S., Sch{\"o}nberger, J.L., Pollefeys, M., Oswald, M.R.: Routedfusion:
  Learning real-time depth map fusion. ArXiv  \textbf{abs/2001.04388} (2020)

\bibitem{weder2021neuralfusion}
Weder, S., Schonberger, J.L., Pollefeys, M., Oswald, M.R.: Neuralfusion: Online
  depth fusion in latent space. In: Proceedings of the IEEE/CVF Conference on
  Computer Vision and Pattern Recognition. pp. 3162--3172 (2021)

\bibitem{yan2021continual}
Yan, Z., Tian, Y., Shi, X., Guo, P., Wang, P., Zha, H.: Continual neural
  mapping: Learning an implicit scene representation from sequential
  observations. In: Proceedings of the IEEE/CVF International Conference on
  Computer Vision (ICCV). pp. 15782--15792 (October 2021)

\bibitem{yang2020noise}
Yang, S., Li, B., Cao, Y.P., Fu, H., Lai, Y.K., Kobbelt, L., Hu, S.M.:
  Noise-resilient reconstruction of panoramas and 3d scenes using robot-mounted
  unsynchronized commodity rgb-d cameras. ACM Transactions on Graphics (TOG)
  \textbf{39}(5),  1--15 (2020)

\bibitem{yang2019heterofusion}
Yang, S., Li, B., Liu, M., Lai, Y.K., Kobbelt, L., Hu, S.M.: Heterofusion:
  Dense scene reconstruction integrating multi-sensors. IEEE transactions on
  visualization and computer graphics  \textbf{26}(11),  3217--3230 (2019)

\bibitem{zhou2013dense}
Zhou, Q.Y., Koltun, V.: Dense scene reconstruction with points of interest. ACM
  Transactions on Graphics (ToG)  \textbf{32}(4), ~1--8 (2013)

\bibitem{zhou2014simultaneous}
Zhou, Q.Y., Koltun, V.: Simultaneous localization and calibration:
  Self-calibration of consumer depth cameras. In: Proceedings of the IEEE
  Conference on Computer Vision and Pattern Recognition. pp. 454--460 (2014)

\bibitem{zhou2013elastic}
Zhou, Q.Y., Miller, S., Koltun, V.: Elastic fragments for dense scene
  reconstruction. In: Proceedings of the IEEE International Conference on
  Computer Vision. pp. 473--480 (2013)

\bibitem{zhu2022nice}
Zhu, Z., Peng, S., Larsson, V., Xu, W., Bao, H., Cui, Z., Oswald, M.R.,
  Pollefeys, M.: Nice-slam: Neural implicit scalable encoding for slam. In:
  Proceedings of the IEEE/CVF Conference on Computer Vision and Pattern
  Recognition. pp. 12786--12796 (2022)

\bibitem{zollhofer2018state}
Zollh{\"o}fer, M., Stotko, P., G{\"o}rlitz, A., Theobalt, C., Nie{\ss}ner, M.,
  Klein, R., Kolb, A.: State of the art on 3d reconstruction with rgb-d
  cameras. In: Computer graphics forum. vol.~37, pp. 625--652. Wiley Online
  Library (2018)

\end{thebibliography}
\end{document}